\documentclass[10pt,twocolumn,letterpaper]{article}

\usepackage{cvpr}
\usepackage{times}
\usepackage{epsfig}
\usepackage{graphicx}
\usepackage{amsmath}
\usepackage{amssymb}
\usepackage{subfigure}
\usepackage{pifont}
\usepackage{comment}
\graphicspath{{img/}}

\usepackage[breaklinks=true,bookmarks=false]{hyperref}

\cvprfinalcopy 

\def\cvprPaperID{****} 

\begin{document}

\title{Computational Decomposition of Style for Controllable and Enhanced Style Transfer}

\author{Minchao Li \quad \quad Shikui Tu \quad \quad Lei Xu\\
Shanghai Jiao Tong University\\
{\tt\small \{marshal\underline{ }lmc,tushikui,leixu\}@sjtu.edu.cn}
}

\maketitle

\begin{abstract}
Neural style transfer has been demonstrated to be powerful in creating artistic image with help of Convolutional Neural Networks (CNN). However, there is still lack of computational analysis of perceptual components of the artistic style. Different from some early attempts which studied the style by some pre-processing or post-processing techniques, we investigate the characteristics of the style systematically based on feature map produced by CNN. First, we computationally decompose the style into basic elements using not only spectrum based methods including Fast Fourier Transform (FFT), Discrete Cosine Transform (DCT) but also latent variable models such Principal Component Analysis (PCA), Independent Component Analysis (ICA). Then, the decomposition of style induces various ways of controlling the style elements which could be embedded as modules in state-of-the-art style transfer algorithms. Such decomposition of style brings several advantages. It enables the computational coding of different artistic styles by our style basis with similar styles clustering together, and thus it facilitates the mixing or intervention of styles based on the style basis from more than one styles so that compound style or new style could be generated to produce styled images. Experiments demonstrate the effectiveness of our method on not only painting style transfer but also sketch style transfer which indicates possible applications on picture-to-sketch problems.
\end{abstract}


\section{Introduction}
Painting art, like Vincent van Gogh's ``The Starry Night'', have attracted people for many years. It is one of the most popular art forms for creative expression of the conceptual intention of the practitioner. Since 1990's, researches have been made by computer scientists on the artistic work, in order to understand art from the view of computer or to turn a camera photo into an artistic image automatically. One early attempt is Non-photorealistic rendering (NPR)\cite{NPR}, an area of computer graphics, which focuses on enabling artistic styles such as oil painting and drawing for digital images. However, NPR is usually limited to specific styles and hard to generalize to produce styled images for any other artistic styles.

One significant advancement was made by Gatys \etal in 2015 \cite{GatysNeuralStyle}, called neural style transfer, which could separate the representations of the image content and style learned by deep CNN and then recombine the image content from one and the image style from another to obtain styled images. During this neural style transfer process, fantastic stylized images were produced with the appearance similar to a given real artistic work, such as Vincent van Gogh's ``The Starrry Night''. The success of the style transfer indicates that artistic styles are computable and are able to be migrated from one image to another. Thus, we could learn to draw like some artists apparently without being trained for years.

Following Gatys \etal 's pioneering work, a lot of efforts have been made to improve or extend the neural style transfer algorithm.\cite{Yin} considered the semantic content and introduced the semantic style transfer network.\cite{MRF} combined the discriminatively trained CNN with the classical Markov Random Field (MRF) based texture synthesis for better mesostructure preservation in synthesized images. Semantic annotations were introduced by \cite{Champ} to achieve semantic transfer. To imporve the efficiency, \cite{Johnson} as well as \cite{Ulyanov} introduced a fast neural style transfer method, which is a feed-forward network to deal with a large set of images per training. With help of an adversarial training network, results were further improved in \cite{Adver}. For a systematic review on neural style transfer, please refer to \cite{review}.

The success of recent progress on style transfer relies on the separable representation learned by deep CNN, in which the layers of convolutional filters automatically learns low-level or abstract representations in a more expressive feature space than the raw pixel-based images. However, it is still challenging to use CNN representations for style transfer due to their uncontrollable behavior as a black-box, and thus it is still difficult to select appropriate composition of styles (e.g., textures, colors, strokes) from images due to the risk of incorporation of unpredictable or incorrect patterns. In this paper, we take a further step on the separated representations of image content and styles. We aim at a computational understanding of the artistic styles, and decompose them into basis elements that are easy to be selected and combined to obtain enhanced and controllable style transfer. Specifically, we propose two types of decomposition methods, i.e., spectrum based methods featured by Fast Fourier Transform (FFT), Discrete Cosine Transform (DCT), and latent variable models such as Principal Component Analysis (PCA), Independent Component Analysis (ICA). Then, we suggest methods of combination of styles by intervention and mixing. The computational decomposition of styles could be embedded as a module to state-of-the-art neural transfer algorithms. Experiments demonstrate the effectiveness of style decomposition in style transfer. We also demonstrate that controlling the style bases enables us to transfer the Chinese landscape painting very well and to transfer the sketch style for a task similar to picture-to-sketch \cite{2017sketch,2018sketch}.

\section{Related Work}
Style transfer generates a styled image having similar semantic content as the content image and similar style as the style image. Conventional style transfer is realized by patch-based texture synthesis methods \cite{texture,fastTexture} where style is approximated as texture. Given a texture image, patch-based texture synthesis methods can automatically generate new image with the same texture. However, arbitary style images are quite different from textures images \cite{fastTexture}, since patches of arbitary style images from different regions are usually distinguished while patches of texture images from different regions are always similar, which limits the functional ability of patch-based texture synthesis method in style transfer. Moreover, control of the texture transferred by varying the patch size (shown in Figure 2 of \cite{texture}) is limited due to the duplicated texture patterns in the texture image.

The neural style transfer algorithm proposed by Gatys \etal \cite{GatysNeuralStyle} is a milestone in style transfer referring to his previous research \cite{gatysTexture} which pioneers to take advantage of pre-trained CNN on imageNet \cite{imagenet}. Rather than using previous texture synthesis methods which are implemented directly on the pixels of raw images, \cite{gatysTexture} uses the feature map of the image which proves to preserve better semantic infomation of the image. Content similarity is measured by comparing the feature map while style similarity is measured by comparing the Gram matrix of the feature map. Algorithm proposed in \cite{GatysNeuralStyle} starts with a noise image and finally converges to the styled image by iterative learning. The loss function $\mathcal{L}$ is composed of the content loss $\mathcal{L}_{content}$ and the style loss $\mathcal{L}_{style}$. $\mathcal{L}_{content}$ is measured by the square error between the feature map of certain layer () while $\mathcal{L}_{style}$ is measured by the square error between the Gram matrix $G_l$ of the feature maps from some certain layers. Notate $h_l, w_l, c_l$ as the height, width and the channel number of the feature map $\mathcal{F}$ in layer $l$ and $e_l$ as the weight of layer $l$ contributing to the style loss $\mathcal{L}_{style}$. $\mathcal{F}_l^{pred}$, $\mathcal{F}_l^{content}$ and $\mathcal{F}_l^{style}$ denote the feature map of the styled image, content image and style image correspondingly, where $\mathcal{F}_l$ is treated as 2-dimensional data ($\mathcal{F}_l \in \mathcal{R}^{(h_l w_l) \times c_l}$).

\begin{align}
\mathcal{L} = \alpha \mathcal{L}_{content} + \beta \mathcal{L}_{style}
\label{totalLoss}
\end{align}

\begin{align}
\mathcal{L}_{content} = \frac{1}{2}(\mathcal{F}_l^{pred} - \mathcal{F}_l^{content}) ^ 2
\end{align}

\begin{align}
G_l = \mathcal{F}_l^T \times \mathcal{F}_l, G_l \in \mathcal{R}^{c_l \times c_l}
\end{align}

\begin{align}
\mathcal{L}_{style} = \sum_{l} e_l \frac{1}{4h_l^2w_l^2c_l^2}(G_l^{pred} - G_l^{style}) ^ 2
\end{align}

\begin{figure*}[htb]
\centering
\includegraphics[width=\linewidth]{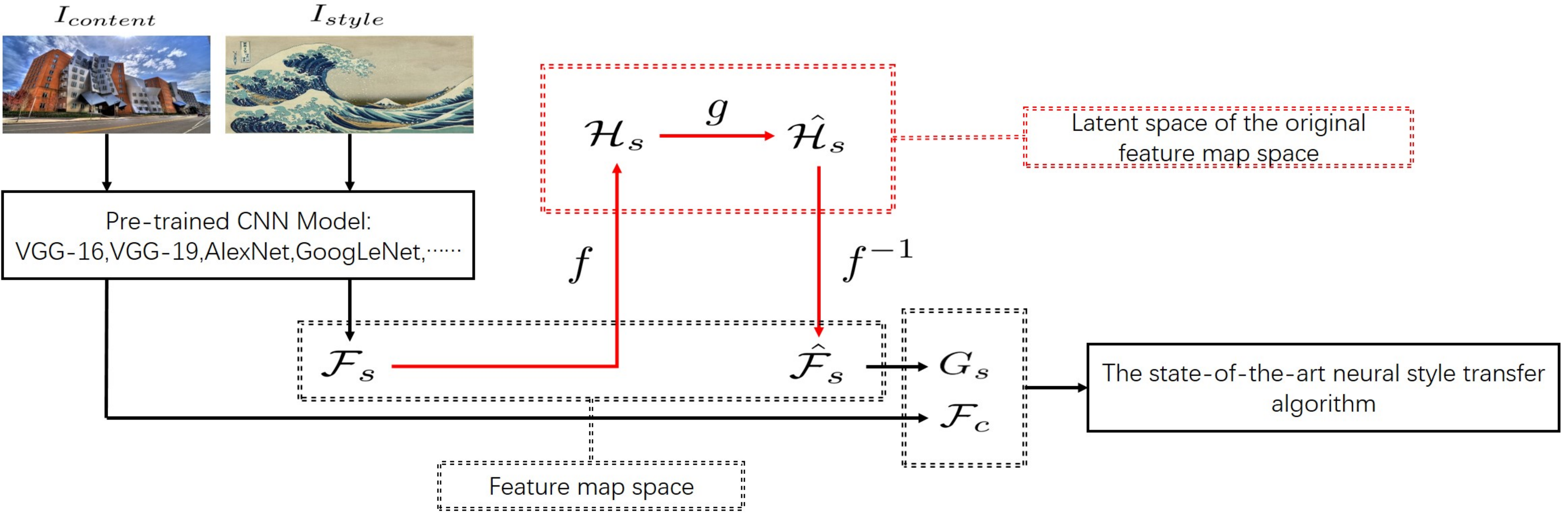}
\caption{An overview of our method is indicated by red, where the dotted red rectangle represents the latent space expanded by the style basis, $f$ denotes computational decomposition of the style feature map $F_s$, $g$ denotes mixing or intervention within the latent space. The red part works as a computational module embedded in Gatys’ or other neural style transfer algorithms.}
\label{method}
\end{figure*}

Gatys \etal then proposed the methods of spatial control, color control and scale control for neural style transfer in \cite{ControlFactor}. Spatial control of neural style transfer can transfer style of specific regions of the style image via guided feature maps. Given the binary spatial guidance channels for both content and style image, one approach to generate the guided feature map is to multiply the guidance channel with the feature map element-wise while another approach is to concatnate the guidance channel with the feature map. Spatial control of neural style transfer serves as an effective method for semnatic control in neural style transfer \cite{Champ,stable}. Color control is realized with help of YUV color space , by which a styled image is first generated using \cite{GatysNeuralStyle}. The luminance channel of content image is replaced with that of the styled image which manage to preserve the color of content image \cite{YUV}. Besides, \cite{ControlFactor} referred to histogram matching methods \cite{histogram} which serve as the second approach to preserve the color of the content image. Although both approaches are feasible for color control of style transfer, we cannot control the color of style image transferred with arbitrary degree which makes the control of color binary-like (1: the styled image using \cite{GatysNeuralStyle} with all color of the style image transferred; 0: the styled image using color control in \cite{ControlFactor} with no color of the style image transferred). Moreover, \cite{ControlFactor} proposed feasible method of mixing detailed texture of one style image $I_s^{1}$ (like stroke) with course texture of another style image $I_s^{2}$ (like color) as scale control which first generates a new style image $I_s^{new}$ with $I_s^{1},I_s^{2}$ as content image and style image respectively using Gram matrix from lower layers in CNN. It can be noticed that the scale levels of the style depends on different layers in CNN which represents different abstract levels. However, since the number of layers in CNN is finite (for VGG19 at most 19 layers), the scale of style can only be controlled with finite degree.

The limitation of control over neural style transfer proposed by pre-processing and post-processing methods in \cite{ControlFactor} derives from the lack of computational analysis of the artistic style which is the foundation of continuous control for neural style transfer. Inspired by spatial control in \cite{ControlFactor} that operations on the feature map could affect the style transferred, we implement different approaches to analyze the feature map and succeed in computational decomposition of the style via projecting feature map into latent space that is expanded by style basis, like color and stroke. Since every point in the well-organized latent space can be decoded back to one style, the control of style basis can be continuous. Meanwhile, our work facilitates the mixing or intervention of styles based on the style basis from more than one styles so that compound style or new style could be generated which enhances the diversity of styled images.

\section{Methods}\label{sec:methods}

Thanks to the powerful representation learning by deep CNN, the separation of content and style enables style transfer from an artistic painting to a natural image \cite{GatysNeuralStyle}. However, it is still challenging to use CNN representations for style transfer due to their uncontrollable behavior as a black-box, and thus it is still difficult to select appropriate composition of styles (e.g., textures, colors, strokes) from images due to the risk of incorporation of unpredictable or incorrect patterns. In the following, we propose to decompose the feature map of the style image into style basis in a latent space, in which it becomes easy to mix or intervene style bases of different styles to generate compound styles or new styles which are then projected back to the feature map space. Such decomposition process enables us to continuously control the composition of the style basis and enhance the diversity of the synthesized styled image. Please refer to Figure \ref{method} for an overview of our method, which could be implemented as a computational module in Gatys' or other neural style transfer algorithms.

Given the content image $I_{content}$ and style image $I_{style}$, we decompose the style by function $f$ from the feature map $\mathcal{F}_s$ of the style image to $\mathcal{H}_s$ in the latent space which is consisted of style basis $S_i$. We can mix or intervene the style basis via function $g$ which is operated on style basis to generate the desired style coded by $\hat{\mathcal{H}_s}$. Using the inverse function $f^{-1}$, $\hat{\mathcal{H}}_s$ is projected back to the feature map space to get $\hat{F}_s$, which replace the original $\mathcal{F}_s$ for style transfer. Our method can serve as embedded module for the state-of-the-art neural style transfer algorithms, as shown in Figure \ref{method} by red.

It can be noted that the module can be regarded as a general transformation from original style feature map $\mathcal{F}_s$ to new style feature map $\hat{\mathcal{F}_s}$. If we let $\hat{\mathcal{F}_s} = \mathcal{F}_s$, our method degenerates back to traditional neural style transfer \cite{GatysNeuralStyle}.

Next, we will introduce two types of decomposition function $f$ and also suggest some control functions $g$. Since the transformation of the feature map is only done on the feature map of the style image, we simply notate $\mathcal{F}_s$ as $\mathcal{F}$ the denote the feature map of the style image and $\mathcal{H}_s$ as $\mathcal{H}$ in the rest of the paper. We notate $h$ and $w$ as the height and width of each channel in the feature map.

\begin{table*}[htbp]
\centering
\begin{tabular}{|c|c|c|}
\hline
Method & Decomposition function $f$ & Projection back \\
\hline
2-d FFT & $\mathcal{H}(u,v) = \frac{1}{hw} \sum^{h-1}_{x=0} \sum^{w-1}_{y=0} \mathcal{F}(x,y)e^{-2(\frac{ux}{h} + \frac{vy}{w})\pi i}, \mathcal{F} \in \mathcal{R}^{h \times w \times c}$ & inverse 2-d FFT\\
\hline 
2-d DCT & $\mathcal{H}(u,v) = c(u)c(v) \sum^{h-1}_{x=0} \sum^{w-1}_{y=0} \mathcal{F}(x,y)cos[\frac{(x+0.5)\pi}{h}u]cos[\frac{(y+0.5)\pi}{w}v]$ & inverse 2-d DCT\\
 & $\mathcal{F} \in \mathcal{R}^{h \times w \times c}, c(u) = \sqrt{\frac{1}{N}},u = 0$ and $c(u) = \sqrt{\frac{2}{N}},u \neq 0$ & \\
\hline
PCA & $\mathcal{F} = UDV^T, U = [v_1,\dots,v_{hw}], \mathcal{H} = U \times \mathcal{F}, \mathcal{F} \in \mathcal{R}^{(hw) \times c}, U \in \mathcal{R}^{(hw) \times (hw)}$ & $\hat{\mathcal{F}} = U^T \times \hat{\mathcal{H}}$\\
\hline
ICA & $ [S, A] = fastICA(\mathcal{F}), \mathcal{H} = S, \mathcal{F} \in \mathcal{R}^{(hw) \times c}, S \in \mathcal{R}^{c \times (hw)}, A \in \mathcal{R}^{c \times c}$ & $\hat{\mathcal{F}} = (A \times \hat{\mathcal{H}})^T$\\
\hline
\end{tabular}
\centering
\caption{The mathematical details of $f$ and $f^{-1}$}
\label{math}
\end{table*}

\subsection{Decomposed by spectrum transforms}

We adopt 2-dimensional Fast Fourier Transform (FFT) and 2-dimensional Discrete Cosine Transform (DCT) as the decomposition function with details given in Table \ref{math}. Both methods are implemented in channel level of $\mathcal{F}$ where each channel is treated as 2-dimensional data like a gray image. 

Through the transform by 2-d FFT and 2-d DCT, the style feature map was decomposed as frequencies in the spectrum space where the style is coded by frequency that forms style bases. We will see that some style bases, such as stroke and color, actually correspond to different level of frequencies. With help of decomposition, similar styles are quantified to be close to each other as a cluster in the spectrum space, and it is easy to combine the existing styles to generate compound styles or new styles $\hat{\mathcal{H}}$ by appropriately varying the style codes. $\hat{\mathcal{H}}$ is then projected back to the feature map space via the inverse function of 2-d FFT and 2-d DCT shown in Table \ref{math}.

\subsection{Decomposed by latent variable models}

We consider another type of decomposition  by latent variable models, such as Principal Component Analysis (PCA) or Independent Component Analysis (ICA), which decompose the input signal into uncorrelated or independent components. Details are referred to Table \ref{math}, where each channel of the feature map $\mathcal{F}$ is vectorized as one input vector.

\begin{itemize}
\item \textbf{Principal Component Analysis (PCA)}: 

We implement PCA from the perspective of matrix factorization. The eigenvectors are computed via Singular Value Decomposition (SVD). Then, the style is coded as linear combination of orthogonal eigenvectors, which could be regarded as style bases. By varying the combination of eigenvectors, compound styles or new styles are generated and then projected back to feature map space via the inverse of the matrix of the eigenvectors. 

\item \textbf{Independent Component Analysis (ICA)}: 

We implement ICA via the fastICA algorithm \cite{fastICA}, so that we decompose the style feature map into statistically independent components, which could be regarded as the style bases. Similar to PCA, we could control the combination of independent components to obtain compound styles or new styles, and then project them back to the feature map space.

\end{itemize}
\subsection{Control function $g$}

The control function $g$ in Figure \ref{method} defines style operations in the latent space expanded by the decomposed style basis. Instead of operating directly on the feature map space, such operations within the latent space bring several advantages. First, after decomposition, style bases are of least redundancy or independent to each other,  operations on them are easier to control; Second, the latent space could be made as a low dimensional manifold against noise, by focusing on several key frequencies for the spectrum or principal components in terms of maximum data variation; Third, continuous operations, such as linear mixing, intervention, and interpolation, are possible, and thus the diversity of the output style is enhanced, and even new styles could be sampled from the latent space. Fourth, multiple styles are able to be better mixed and transferred simultaneously.

Let $S_{i}^{(n)}, i \in \mathbb{Z}$ denote the i-th style basis of n-th style image. Notate $\{S_{i}^{(n)} | i \in I\}, I \subset \mathbb{Z}$ as $S_{I}^{(n)}$.

\begin{itemize}
\item \textbf{Single style basis}: Project the latent space on one of the style basis $S_j$. That is $S_i = 0$ if $i \neq j$
\item \textbf{Intervention}: Reduce or amplify the effect of one style basis $S_j$ by multiplying $I$ while keeping other style bases unchanged. That is $S_i = I * S_i$ if $i = j$
\item \textbf{Mixing}: Combine the style bases of $n$ styles. That is $S = \{S_I^{(1)}, S_J^{(2)}, \dots, S_K^{(n)}\}$
\end{itemize}

\section{Experiments}
We demonstrate the performance of our method using the fast neural style transfer algorithm \cite{fast-algorithm,Johnson,Ulyanov}. We take the feature map `relu4\underline{ }1' from pre-trained VGG-19 model \cite{VGG} as input to our style decomposition method because we try every single activation layer in VGG-19 and find that `relu4\underline{ }1' is more suitable for style transfer.

\subsection{Inferiority of feature map and necessity of latent space}
Here, we demonstrate that it is not suitable for the style control function $g$ to be applied on the feature map space directly because feature map space is possibly formed by a complicated mixture of style bases. To check whether the basis of feature map $\mathcal{F}$ can form the style bases, we first experimented on the channels of $\mathcal{F}$, then the pixels of $\mathcal{F}$.

\begin{figure}[htb]
\centering
\subfigure[]{
\begin{minipage}[t]{\linewidth}
\centering
\includegraphics[width=\linewidth]{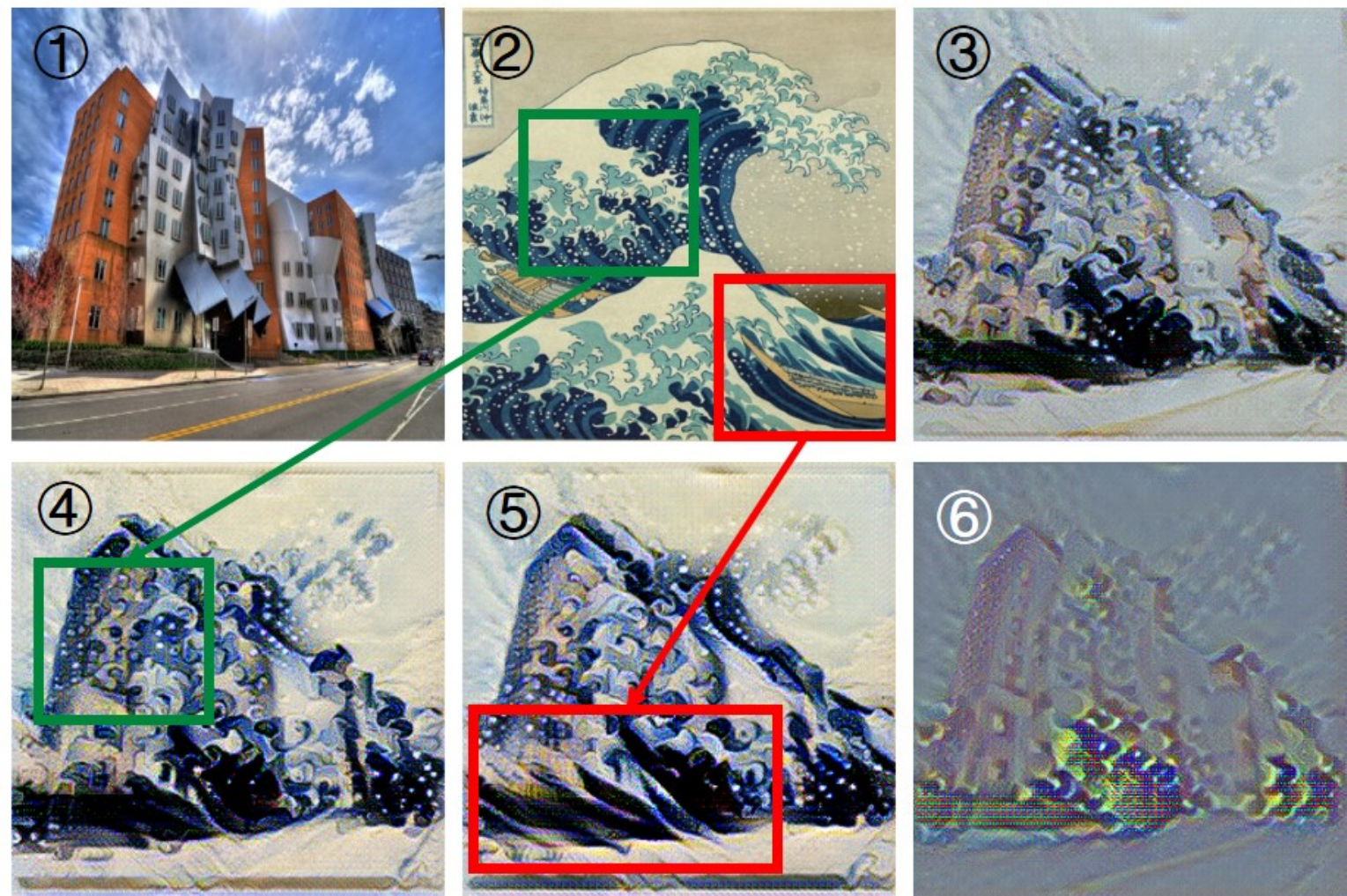}
\end{minipage}%
}%
\centering
\caption{(a)\ding{172} content image (Stata Center); \ding{173} style image (``The Great Wave off Kanagawa'' by Katsushika Hokusai); \ding{174} styled image by traditional neural style transfer \cite{GatysNeuralStyle}; \ding{175}\ding{176}\ding{177} are the results of implementing control function directly on the feature map $\mathcal{F}$. Specifically, we amplify some pixles of $\mathcal{F}$ which generate \ding{175}\ding{176} and preserve a subset of channels of $\mathcal{F}$ which generate \ding{177}. }
\label{inferior}
\end{figure}

\subsubsection{Channels of $\mathcal{F}$}
Assume style is encoded in space $\mathcal{H} = \{S_{1},S_{2},\dots,S_{n}\}$ which is expanded by style basis $S_{i}$. The superiority of space $\mathcal{H}$ should result in that the intuitive similarity of style basis $S_{i}$ conforms to the cluster of $S_{i}$ in space $\mathcal{H}$ under Euclidean distance.

Based on the above assumption, we generate the subset $C$ of channels of $\mathcal{F}$ that could possibly represent color basis with semi-supervised method using style images in Figure \ref{manifold}(a-c). It can be noticed that Chinese paintings and pen sketches (Figure \ref{manifold}(a,c)) share the same color style while oil painting (Figure \ref{manifold}(b)) has an exclusive one. We iteratively find the largest channel set $C_{max}$ (384 channels included) whose clustering result out of K-means \cite{kmeans} conforms to the following \textbf{clustering standard for color basis:}

\begin{itemize}
\item No cluster contains both \textbf{oil painting} and \textbf{Chinese painting or pen sketch}.
\item One cluster contains only one or two points, since K-means is not adpative to the cluster number and the cluster number is set as $3$.
\end{itemize}

However, if we only use $C_{max}$ to transfer style, the styled image (Figure \ref{inferior}(a)\ding{177}) isn't well stylized and doesn't indicate any color style of the style image (Figure \ref{inferior}\ding{173}), which probably indicates that the channels of $\mathcal{F}$ are not suitable to form independent style basis.

In comparison, not only does the clustering result of the color basis in spectrum space (defined in Section \ref{sec:decompose}) using K-means conform to the above clustering standard, but the styled image using single color basis (Figure \ref{decompose}(c)) also works well and meets our intuitive standard for color, which indicates that with help of proper decompostion functions, the superiority of latent space $\mathcal{H}$ is reachable.

\subsubsection{Pixels of $\mathcal{F}$}
We give intervention $I=2.0$ to certain region of each channel of $\mathcal{F}$ to see if any intuitive style basis is amplified. The styled images are shown in Figure \ref{inferior}(a)\ding{175}\ding{176}. Rectangles in style image (Figure \ref{inferior}(a)\ding{173}) are the intervened regions corespondingly. Compared to the styled image using \cite{GatysNeuralStyle} (Figure \ref{inferior}(a)\ding{174}), when the region of small waves in style image is intervened (green rectangle in the style image), the effect of small blue circles in the styled image are amplyfied (green rectangle in the styled image) while when the region of large waves in style image is intervened (red rectangle in the style image), the effect of long blue curves in the styled image are amplyfied (red rectangle in the styled image). Actually, implementing control function $g$ on the pixels of the channels of $\mathcal{F}$ is quite similar to the methods proposed for spatial control of neural style transfer \cite{ControlFactor} which controls style transfer via a spatially guided feature map defined by a binary or real-valued mask on a region of the feature map. Yet it fails to computationally decompose the style basis.

\subsection{Transfer by single style basis} \label{sec:decompose}
\begin{figure}[tb]
\centering
\subfigure[]{
\begin{minipage}[t]{0.25\linewidth}
\centering
\includegraphics[width=\linewidth]{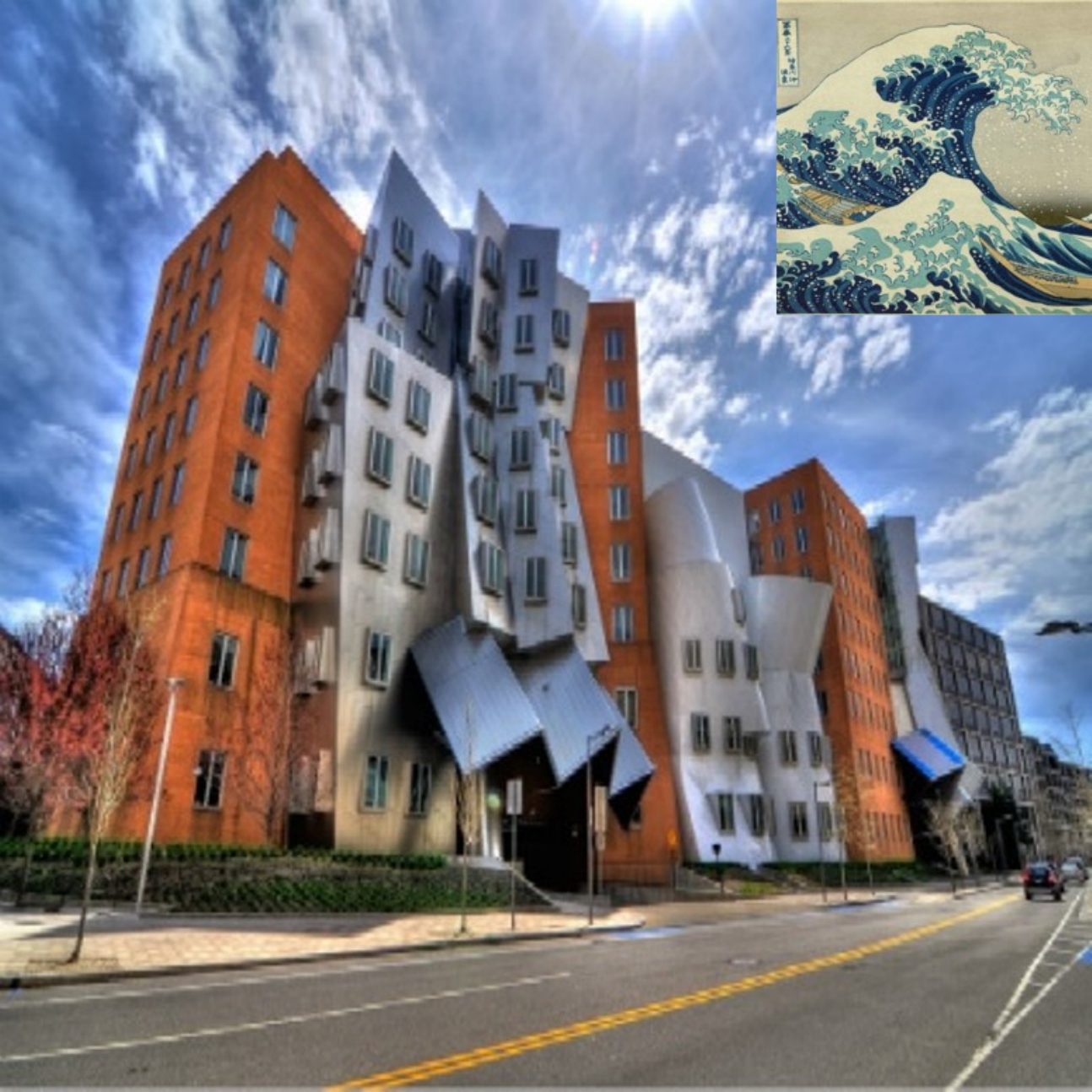}
\end{minipage}%
}%
\subfigure[]{
\begin{minipage}[t]{0.25\linewidth}
\centering
\includegraphics[width=\linewidth]{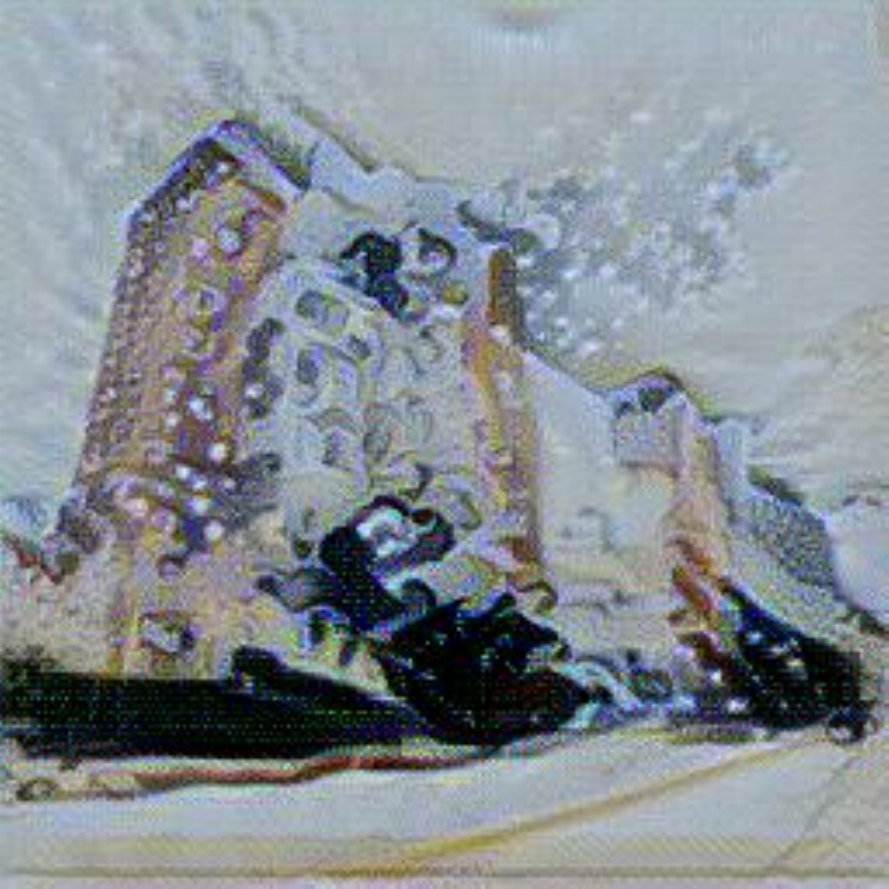}
\end{minipage}%
}%
\subfigure[]{
\begin{minipage}[t]{0.25\linewidth}
\centering
\includegraphics[width=\linewidth]{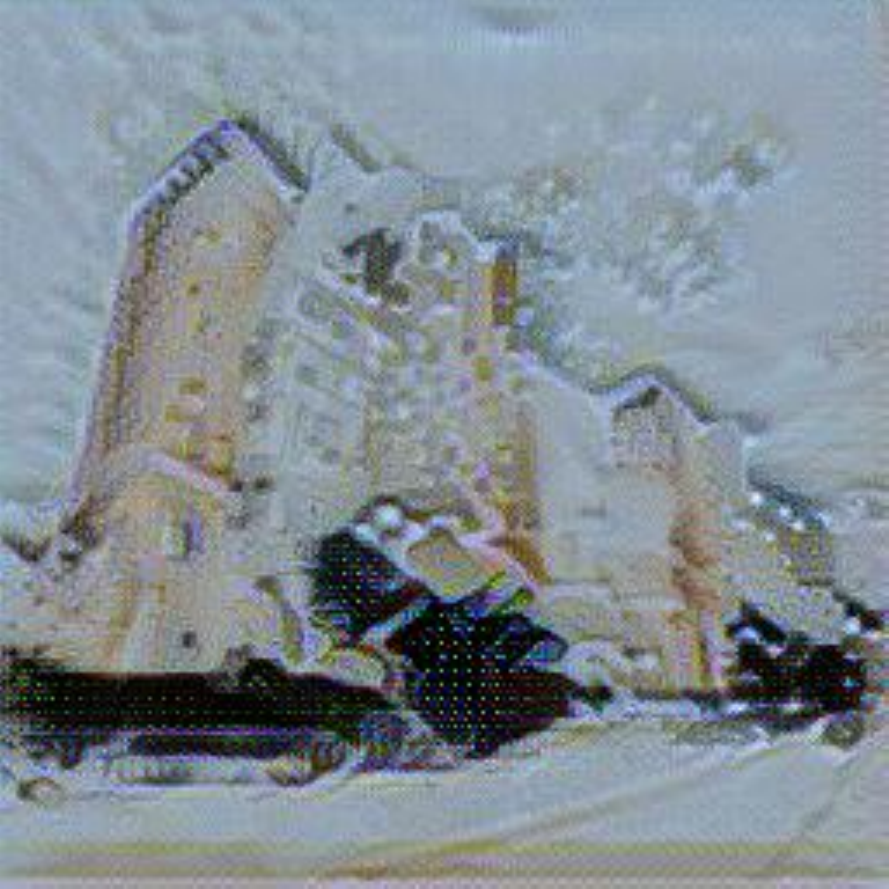}
\end{minipage}%
}%
\subfigure[]{
\begin{minipage}[t]{0.25\linewidth}
\centering
\includegraphics[width=\linewidth]{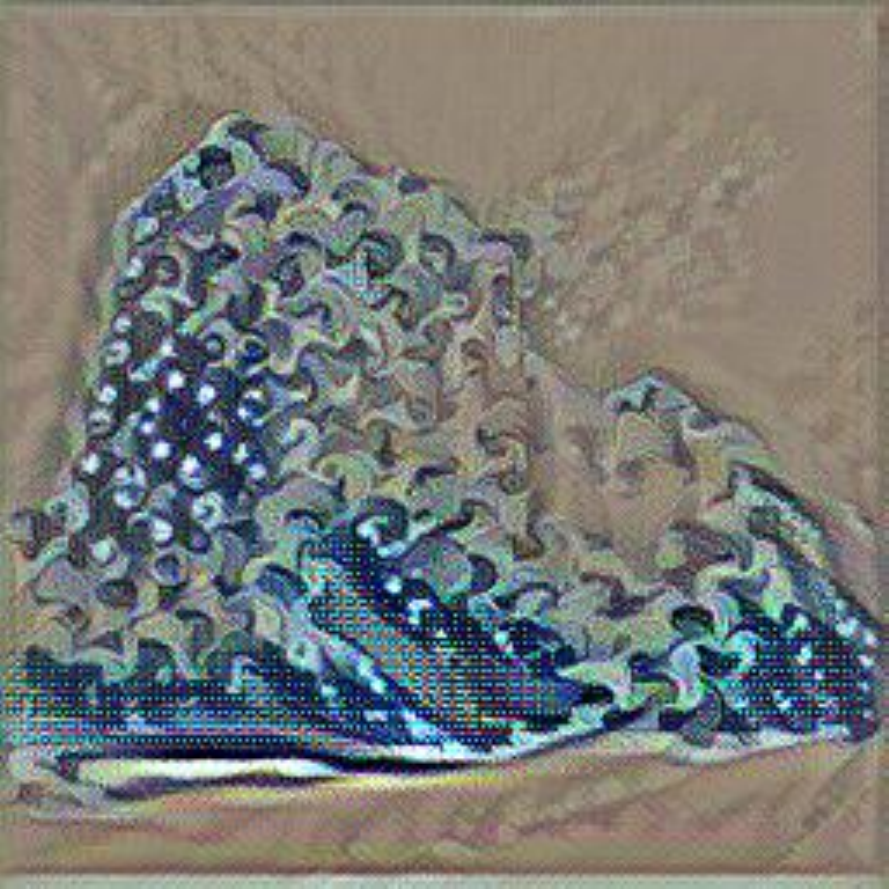}
\end{minipage}%
}%
\vfill
\subfigure[]{
\begin{minipage}[t]{0.25\linewidth}
\centering
\includegraphics[width=\linewidth]{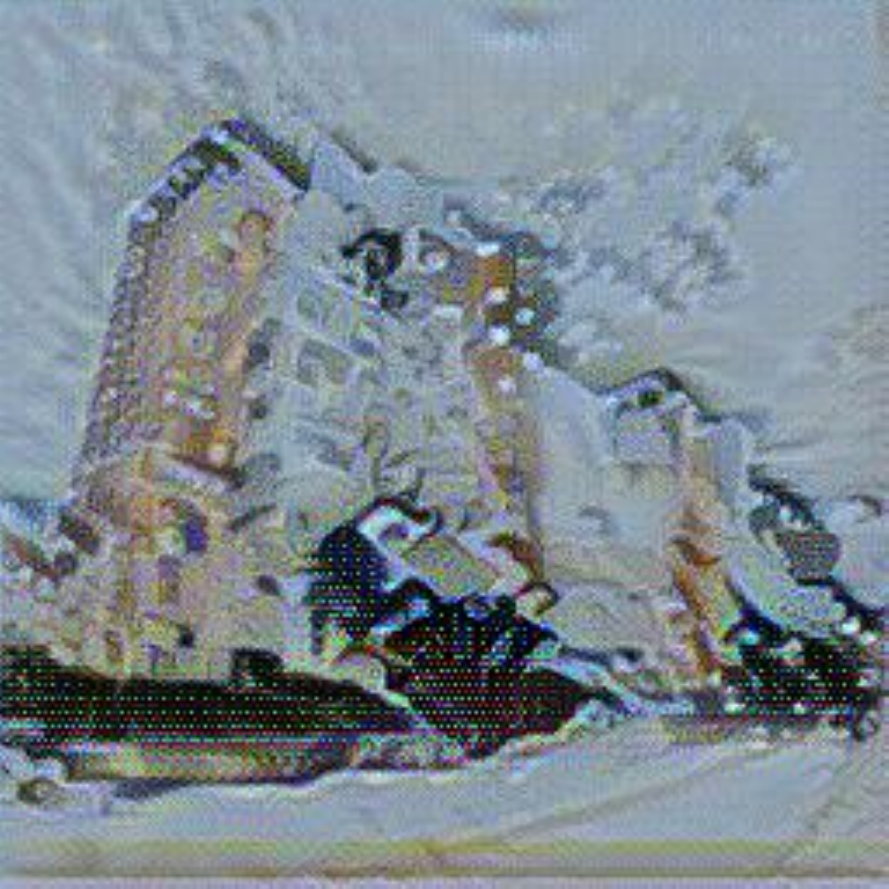}
\end{minipage}%
}%
\subfigure[]{
\begin{minipage}[t]{0.25\linewidth}
\centering
\includegraphics[width=\linewidth]{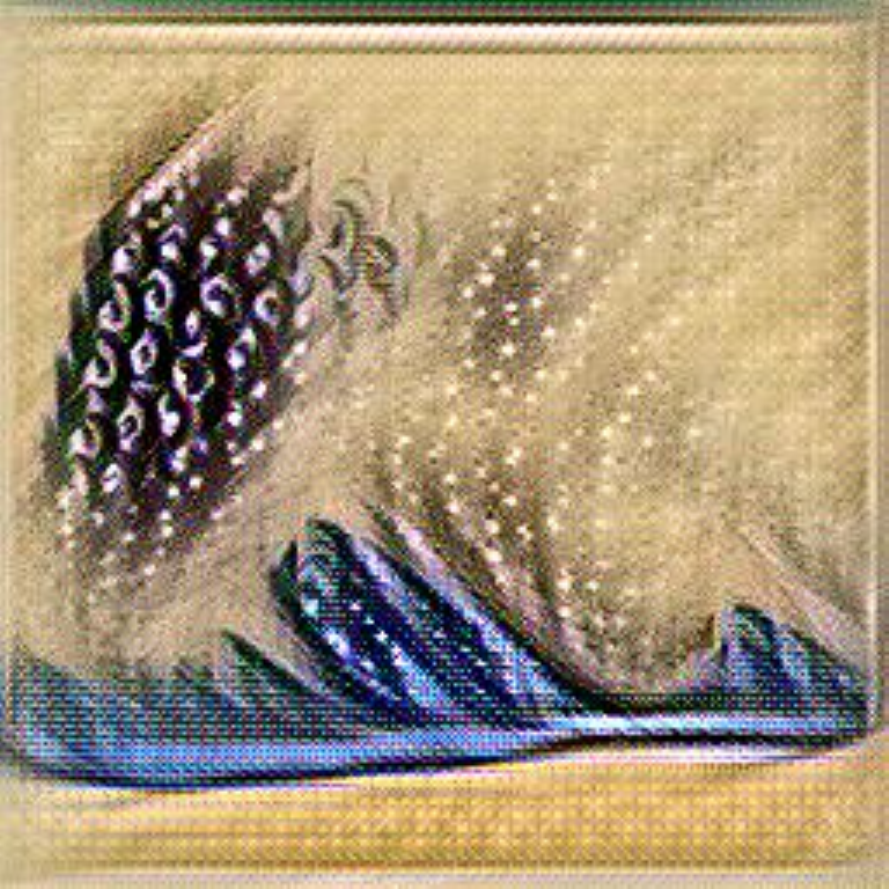}
\end{minipage}%
}%
\subfigure[]{
\begin{minipage}[t]{0.25\linewidth}
\centering
\includegraphics[width=\linewidth]{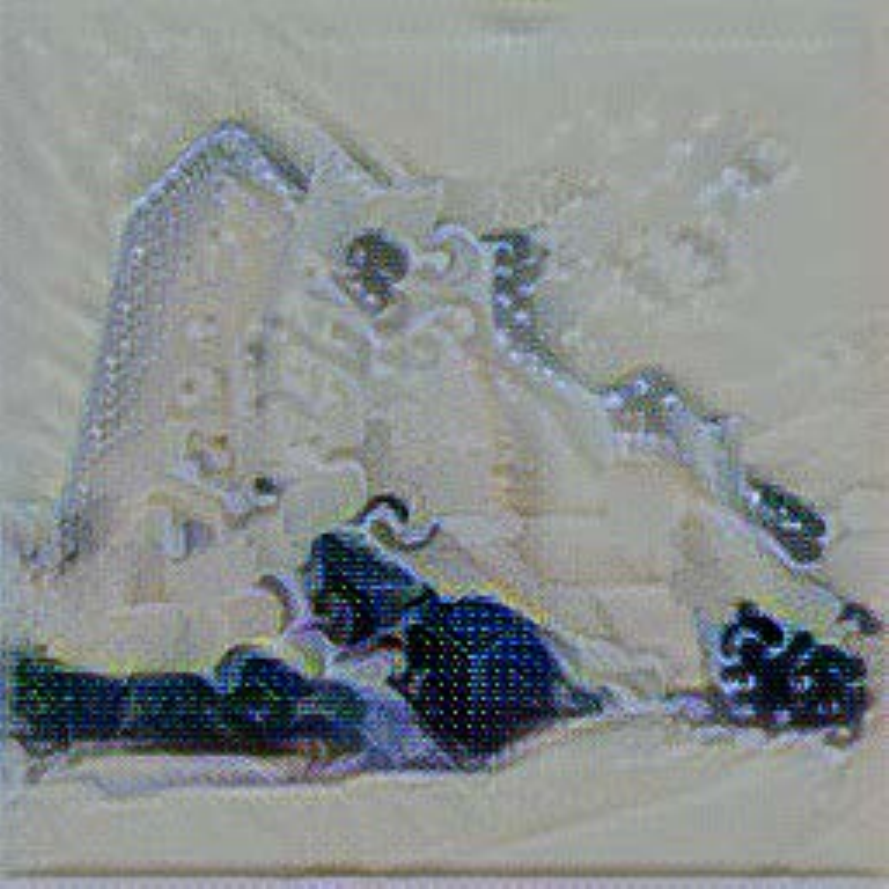}
\end{minipage}%
}%
\subfigure[]{
\begin{minipage}[t]{0.25\linewidth}
\centering
\includegraphics[width=\linewidth]{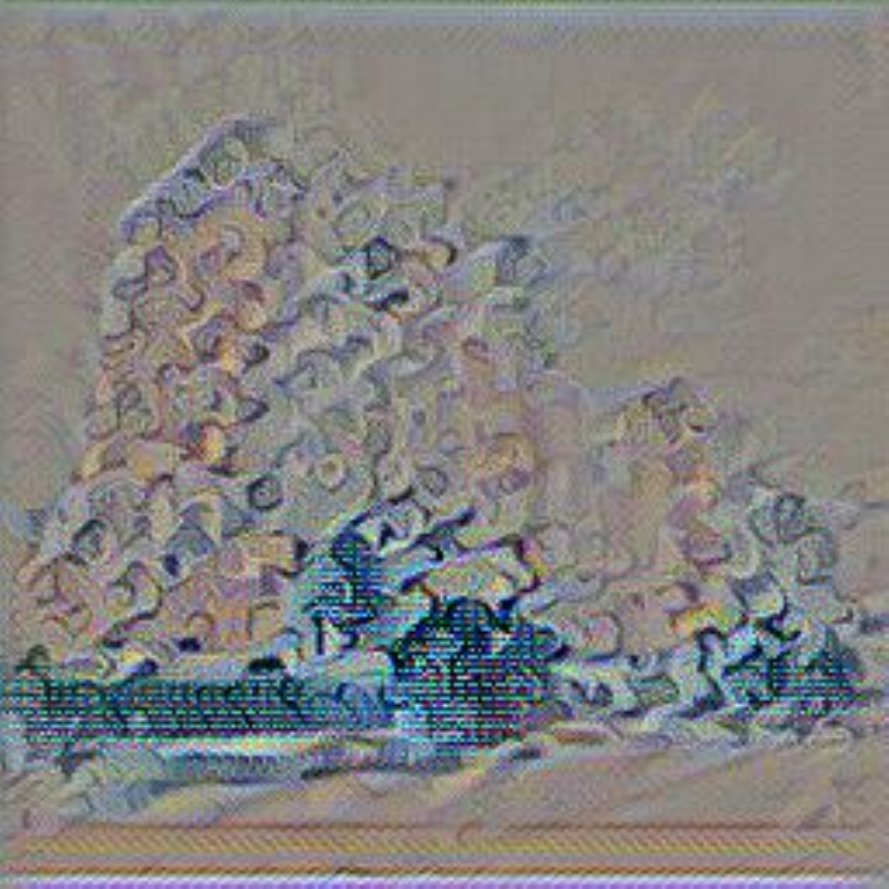}
\end{minipage}%
}%
\centering
\caption{(a) the original content and style images; (b) styled image by traditional neural style transfer \cite{GatysNeuralStyle}; (c-h) results of preserving one style basis by different methods. Specifically, (c-d) FFT;  (e-f) PCA; (g-h) ICA where (c,e,g) aim to transfer the color of style image while (d,f,h) aim to transfer the stroke of style image.}
\label{decompose}
\end{figure}

To check whether $\mathcal{H}$ is consisted of style bases, we transfer style with single style basis preserved. We conduct the experiment on $\mathcal{H}$ generated by different decomposition functions, including FFT, DCT, PCA as well as ICA, with details mentioned in Section \ref{sec:methods} and results shown in Figure \ref{decompose}. 

We preserve the DC component only and the rest frequency components only in the spectrum space generated by FFT respectively with results shown in Figure \ref{decompose}(c)(d). Figure \ref{decompose}(c) preserves the color of style while Figure \ref{decompose}(d) preserves the wave-like stroke, which indicates that FFT is feasible for style decomposition. The result of DCT is quite similar to that of FFT, with DC component representing color and the rest representing stroke.

Besides to visual eveluation shown in Figure \ref{decompose}(c)(d), we analyze the spectrum space by projecting the style bases into low dimensional space, which can analytically demonstrate the effectiveness and robustness of spectrum based methods. Given the spectrum space of $\mathcal{F}$, we vectorize the DC component as well as the rest frequency components  as color vector $v_{color}$ and stroke vector $v_{stroke}$, ($v_{color} \in \mathcal{R}^{1 \times c}$, $v_{stroke} \in \mathcal{R}^{1 \times ((hw - 1)c)}$). Via Isomap \cite{Isomap}, we project $v_{color}$ and $v_{stroke}$ to $u_{color}, u_{stroke} \in \mathcal{R}$ which forms X-axis and Y-axis of the 2-dimensional plane for visualization where every style is encoded as a point. We experiment on 3 artistic styles, including Chinese painting, oil painting and pen sketch, and each style contains 10 pictures which is shown in Figure \ref{manifold}(a-c). Chinese paintings and pen sketches share similar color style which is sharply distinguished with oil paintings' while the stroke of three artistic styles are quite different from each other. Thus, as in shown in Figure \ref{manifold}(d), Chinese paintings and pen sketches are close to each other and both stay away from oil paintings in X-axis which represents color while three styles are respectively separable in Y-axis which represents stroke, which completely satisfies our analysis of the three artistic styles. When we apply to large scale of style images (Figure \ref{manifold}(e)), X-axis represents the linear transition from dull-colored to rich-colored. However, we fail to conclude any notable linear transition for Y-axis from the 2-dimensional visualization probably because it is hard to describe the style of stroke (boldface,length,curvity,etc.) using only one dimension. Yet, clustering by K-means on the original spectrum still conforms to the true label. In summary, spectrum based methods do work for large scale.

Unlike spectrum based methods, the bases of latent space via PCA or ICA are either uncorrelated components or independent signals. Via PCA, the most principle component (Figure \ref{decompose}(e)) fails to separate color and stroke well while the rest components (Figure \ref{decompose}(f)) fail to represent any style basis, which indicates PCA isn't a suitable method for style decomposition.

The results of ICA (Figure \ref{decompose}(g)(h)) are as good as the results of FFT but show significant differences. The color basis and stroke basis are formed by the following method. We sum up each column of mixing matrix $A$ where $A_{i,j}$ denotes the contribution of j-th independent signal to the i-th channel in $\mathcal{F}$ to get $A^{sum} \in \mathcal{R}^{c}$ where $A^{sum}_{j}$ denotes the overall contribution of the j-th independent signal to $\mathcal{F}$. Sort $A^{sum}$ in acsent order to get $arg \in \mathcal{R}^{c}$, where $arg_{j}$ denotes the index of signal which ranges $j$ in $A_{sum}$. The stroke basis is formed with $S_{arg_{i}}, i \in [0,n-1] \cup [c-n,c-1]$ while the color basis is formed with the rest independent signals. Using ICA, the color basis (Figure \ref{decompose}(g)) is more murky than Figure \ref{decompose}(c) while the stroke basis (Figure \ref{decompose}(h)) retains the profile of curves with less stroke color preserved compared to Figure \ref{decompose}(d), which indicates both ICA and spectrum based methods work for style decomposition, but generates different results.

\begin{figure}[tb]
\centering
\subfigure[]{
\begin{minipage}[t]{0.174\linewidth}
\centering
\includegraphics[width=\linewidth]{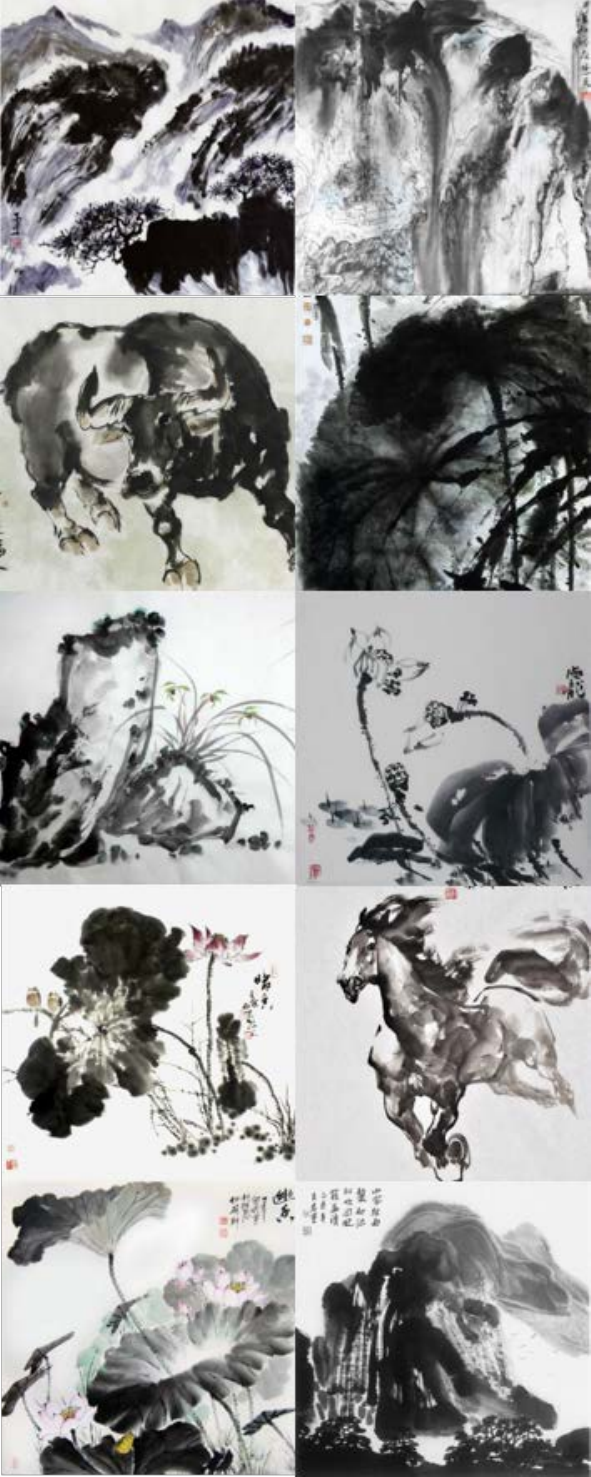}
\end{minipage}%
}%
\subfigure[]{
\begin{minipage}[t]{0.174\linewidth}
\centering
\includegraphics[width=\linewidth]{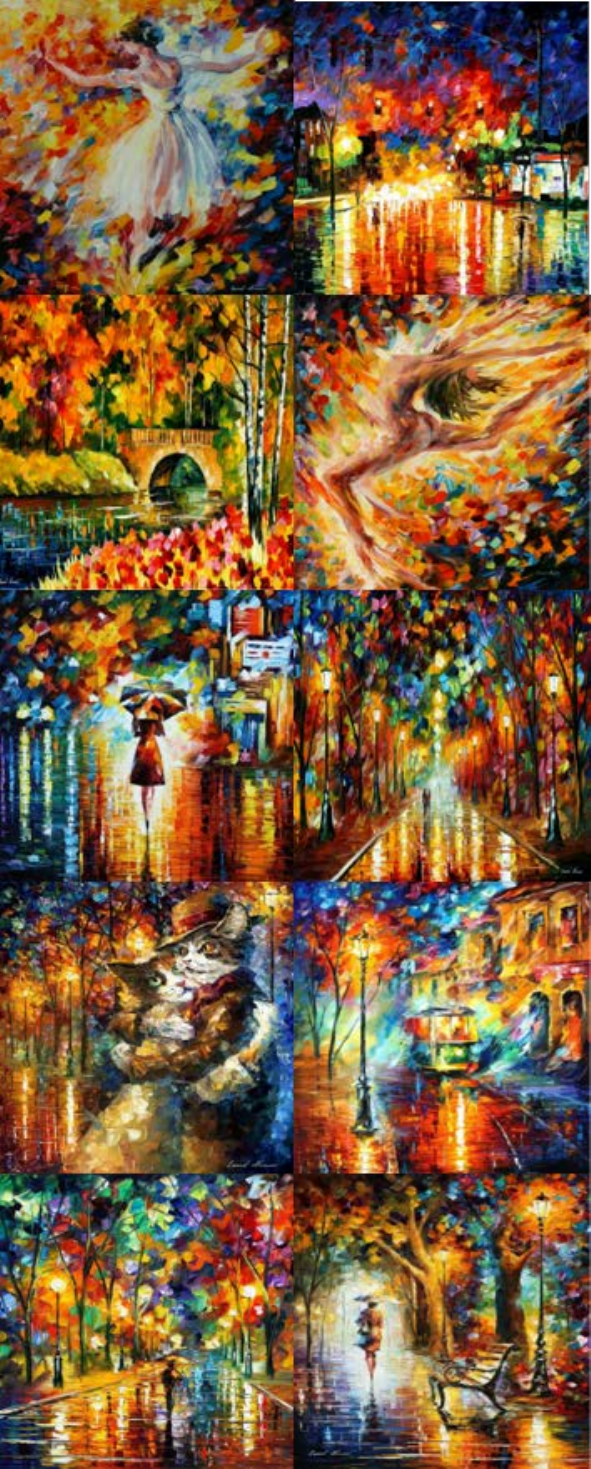}
\end{minipage}%
}%
\subfigure[]{
\begin{minipage}[t]{0.174\linewidth}
\centering
\includegraphics[width=\linewidth]{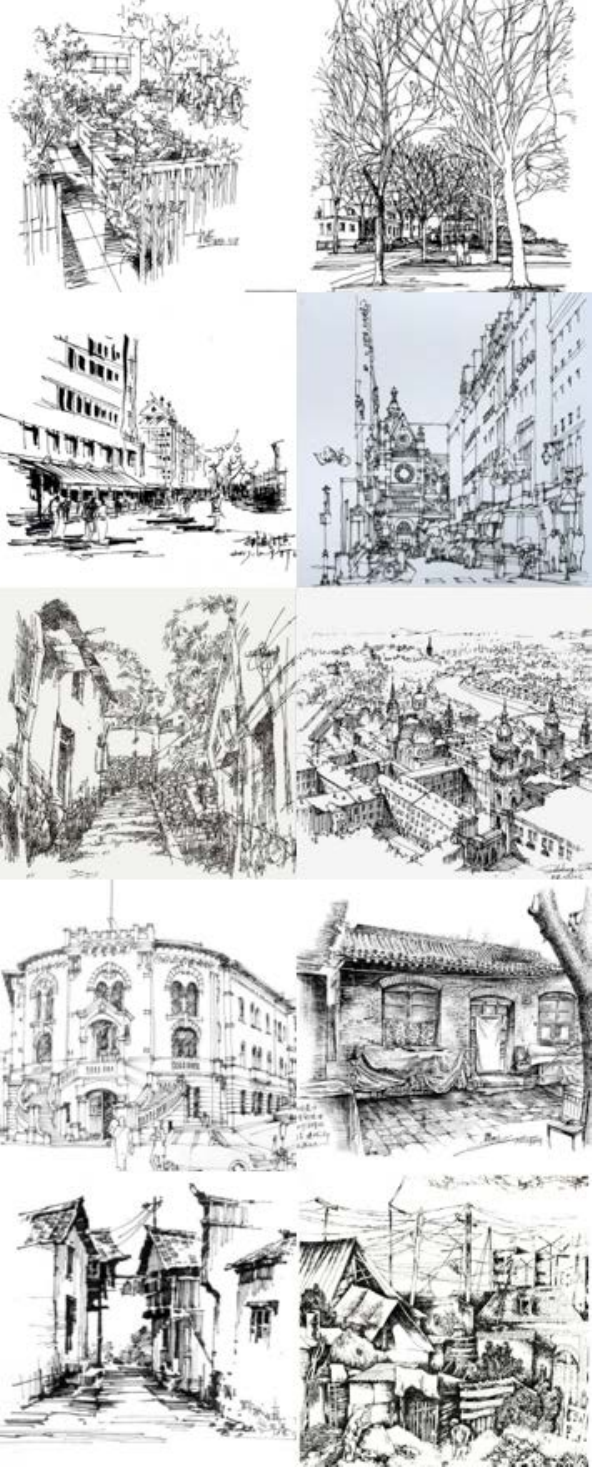}
\end{minipage}%
}%
\subfigure[]{
\begin{minipage}[t]{0.48\linewidth}
\centering
\includegraphics[width=\linewidth]{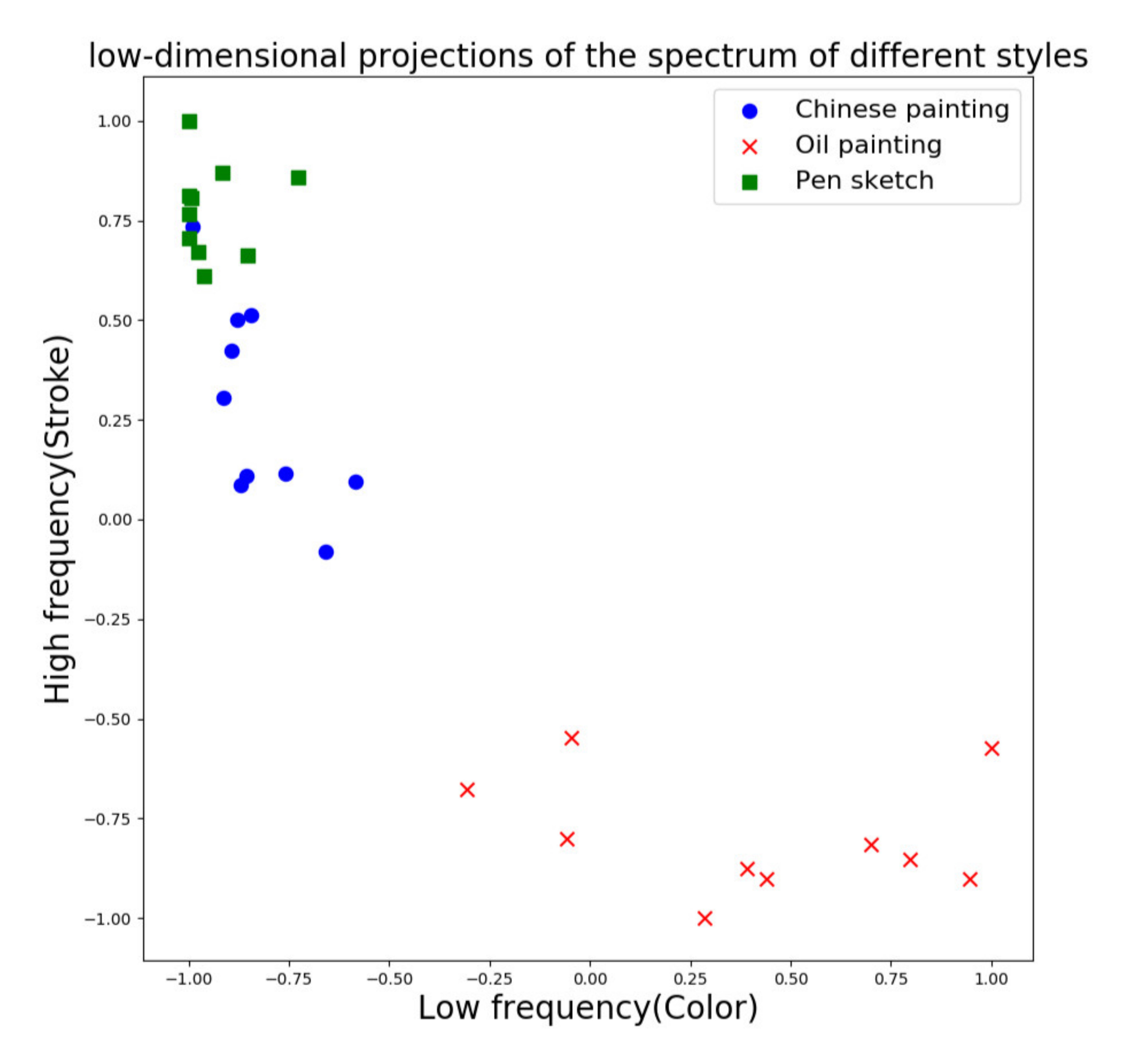}
\end{minipage}%
}%
\vfill
\subfigure[]{
\begin{minipage}[t]{\linewidth}
\centering
\includegraphics[width=\linewidth]{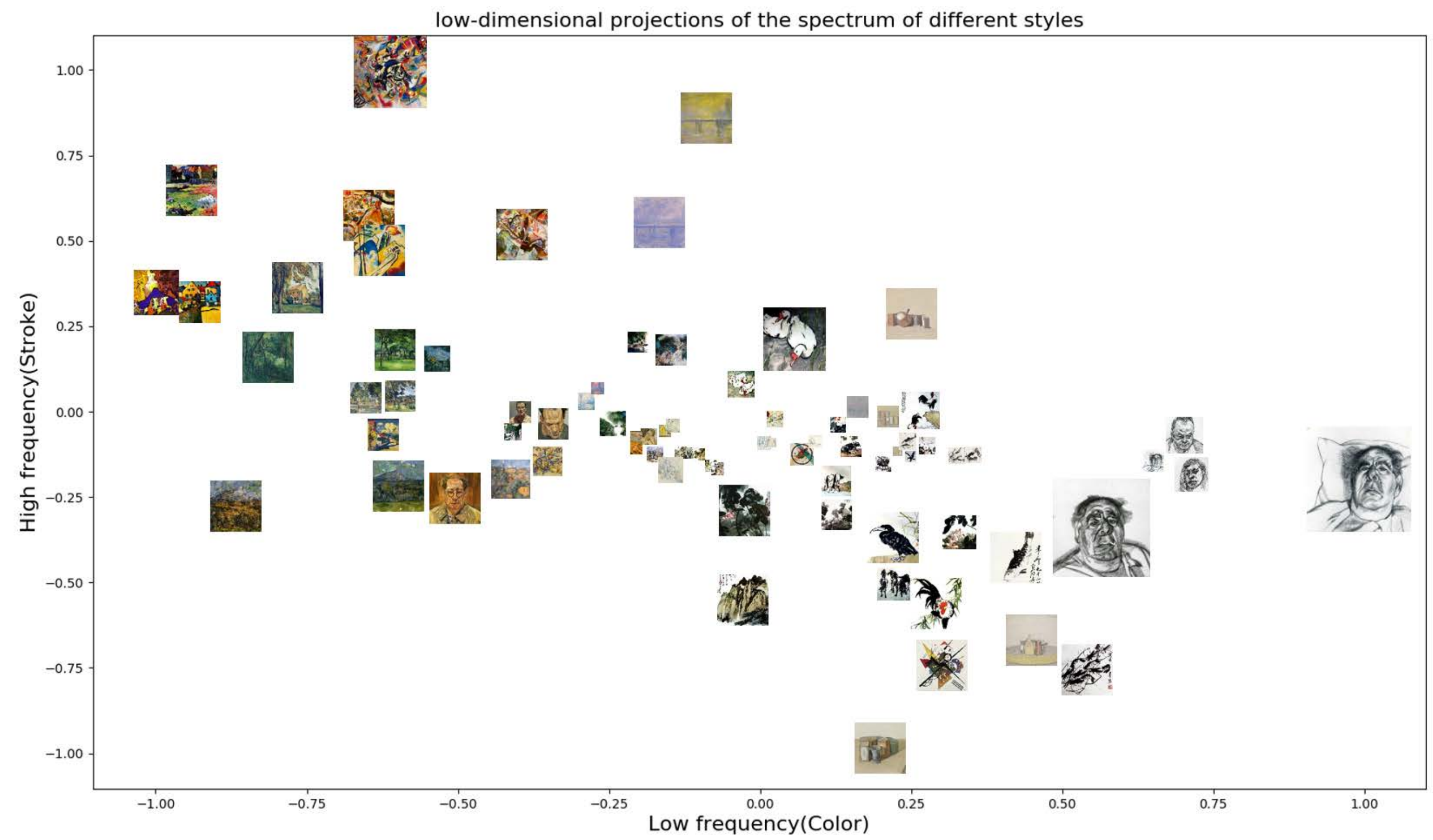}
\end{minipage}%
}%
\centering
\caption{(a) Chinese paintings; (b) Oil paintings (by Leonid Afremov); (c) Pen sketches; (d) low-dimensional projections of the spectrum of style(a-c) via Isomap; (e) low-dimensional projections of the spectrum of large scale of style images via Isomap \cite{Isomap}. The size of each image shown above does not indicate any other information, but is set to prevent the overlap of the images only.}
\label{manifold}
\end{figure}

\begin{figure*}[htb]
\centering
\subfigure[]{
\begin{minipage}[t]{0.286\linewidth}
\centering
\includegraphics[width=\linewidth]{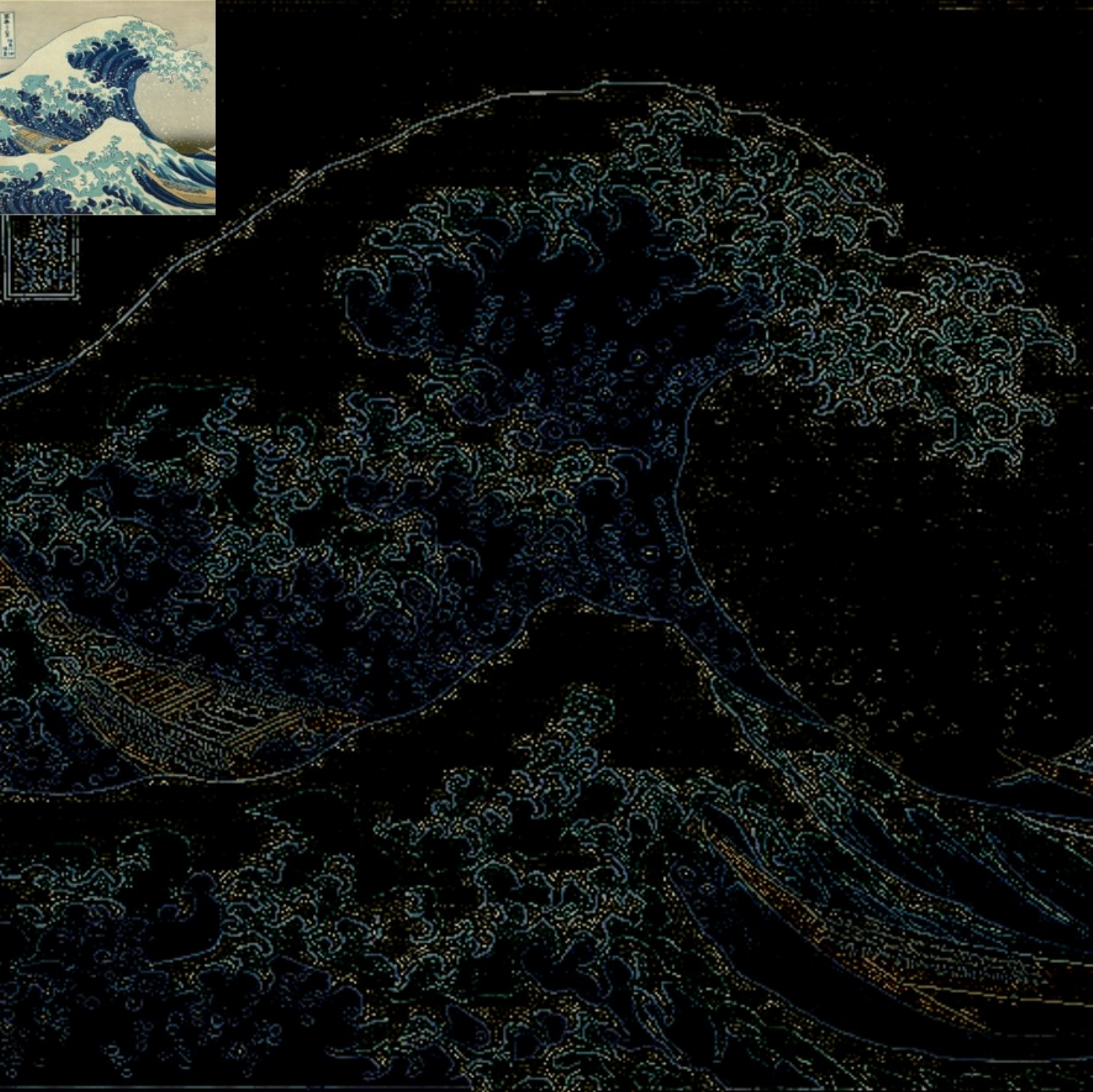}
\end{minipage}%
}%
\subfigure[]{
\begin{minipage}[t]{0.357\linewidth}
\centering
\includegraphics[width=\linewidth]{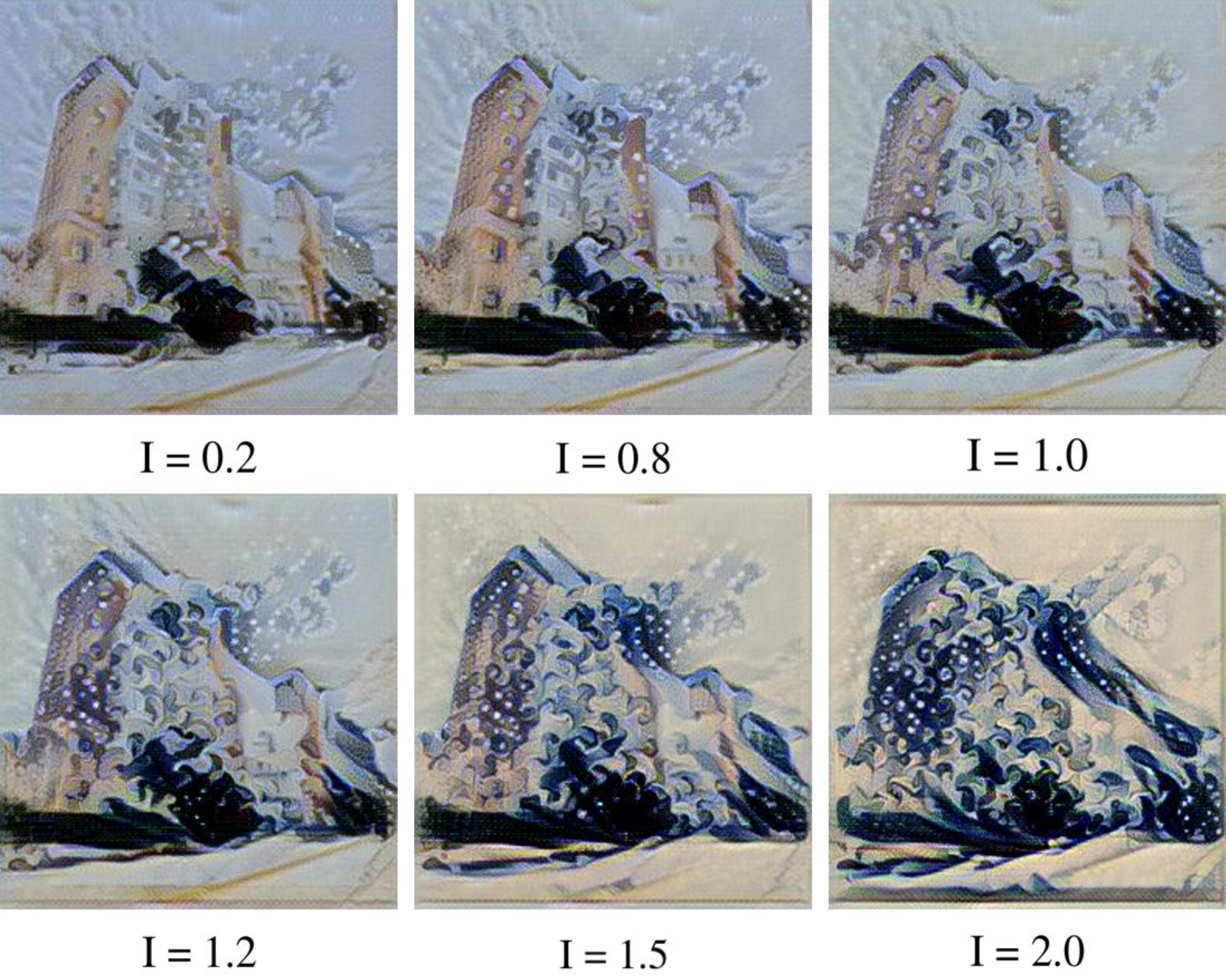}
\end{minipage}%
}%
\subfigure[]{
\begin{minipage}[t]{0.357\linewidth}
\centering
\includegraphics[width=\linewidth]{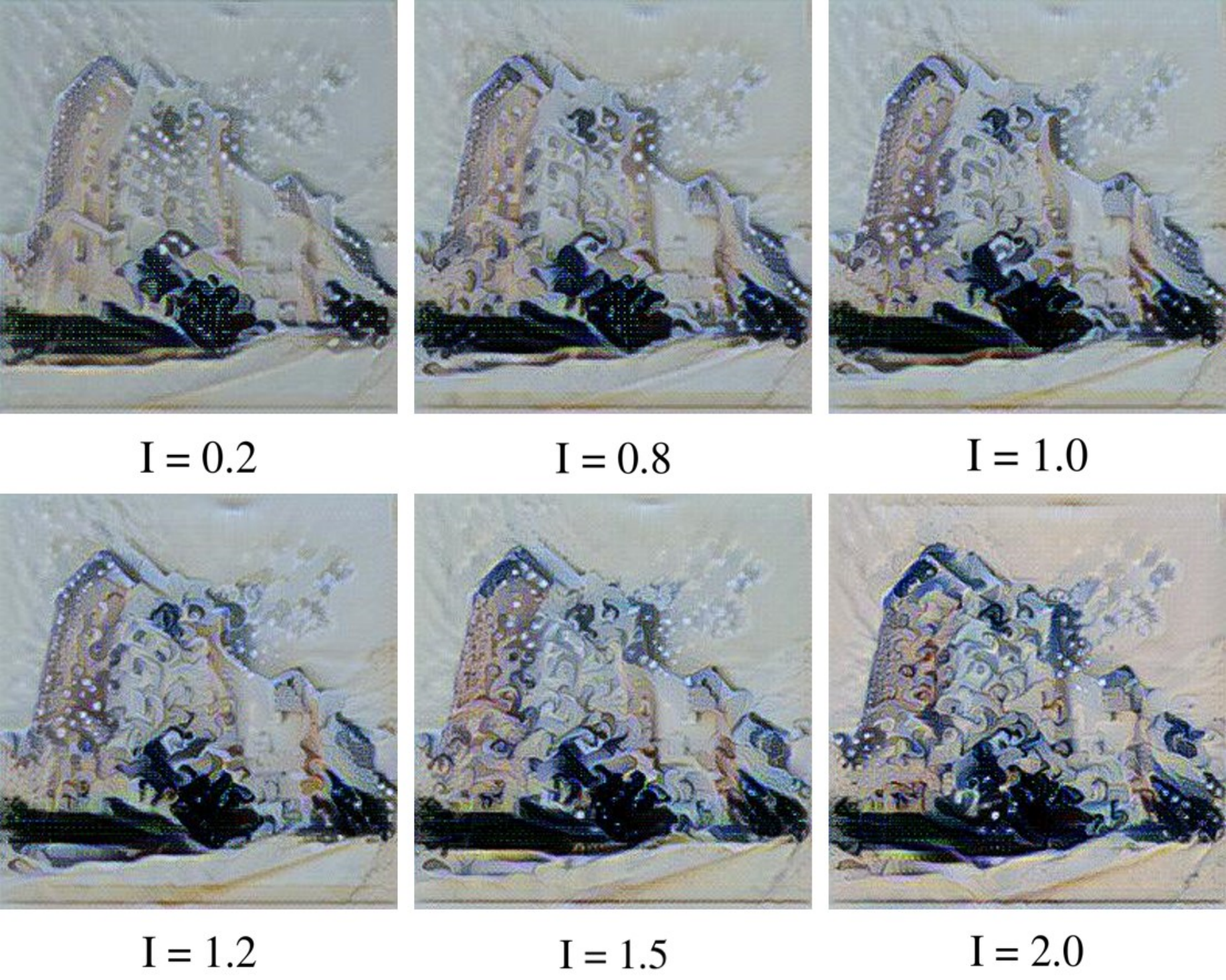}
\end{minipage}%
}%
\vfill
\subfigure[]{
\begin{minipage}[t]{0.286\linewidth}
\centering
\includegraphics[width=\linewidth]{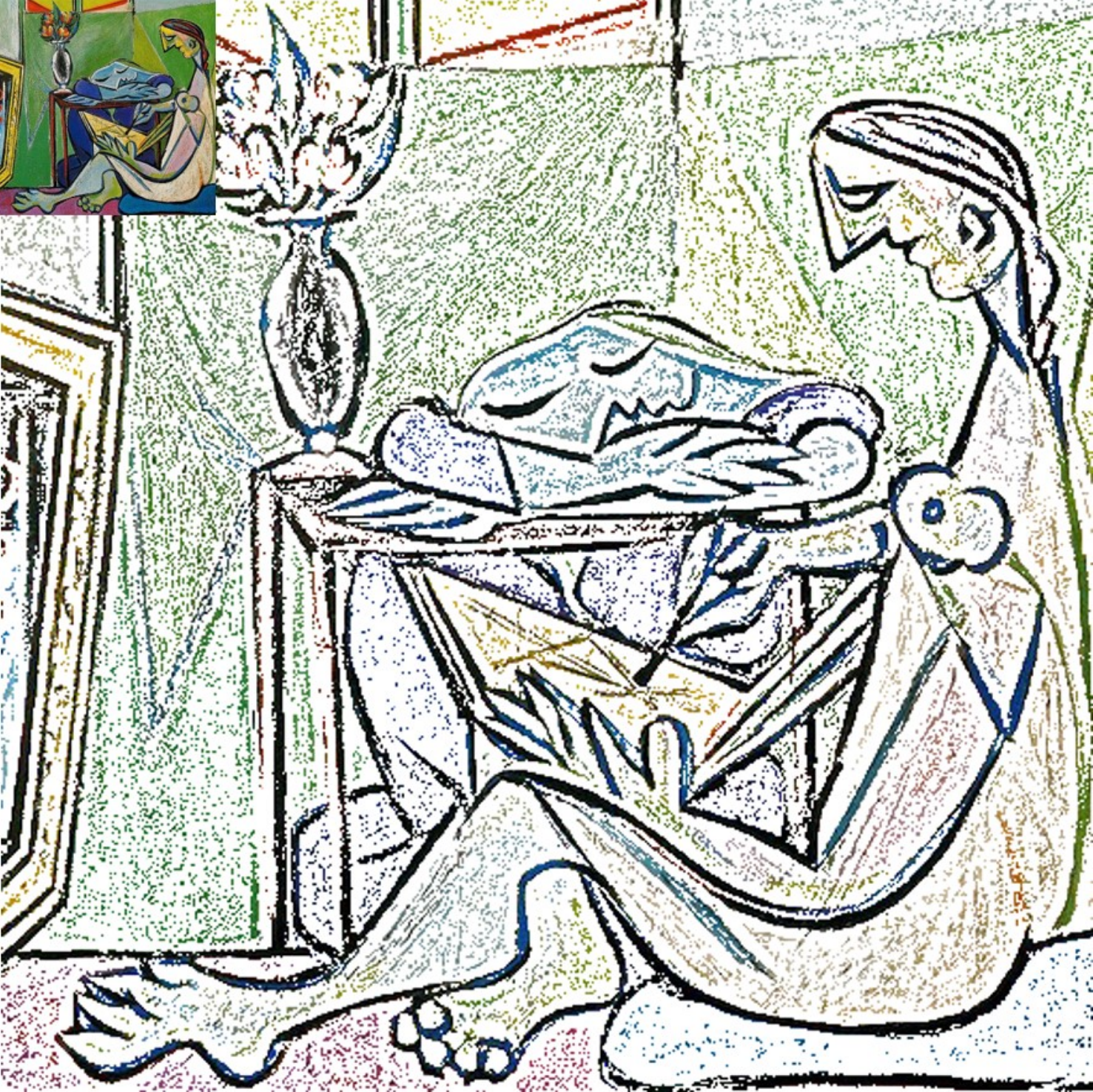}
\end{minipage}%
}%
\subfigure[]{
\begin{minipage}[t]{0.357\linewidth}
\centering
\includegraphics[width=\linewidth]{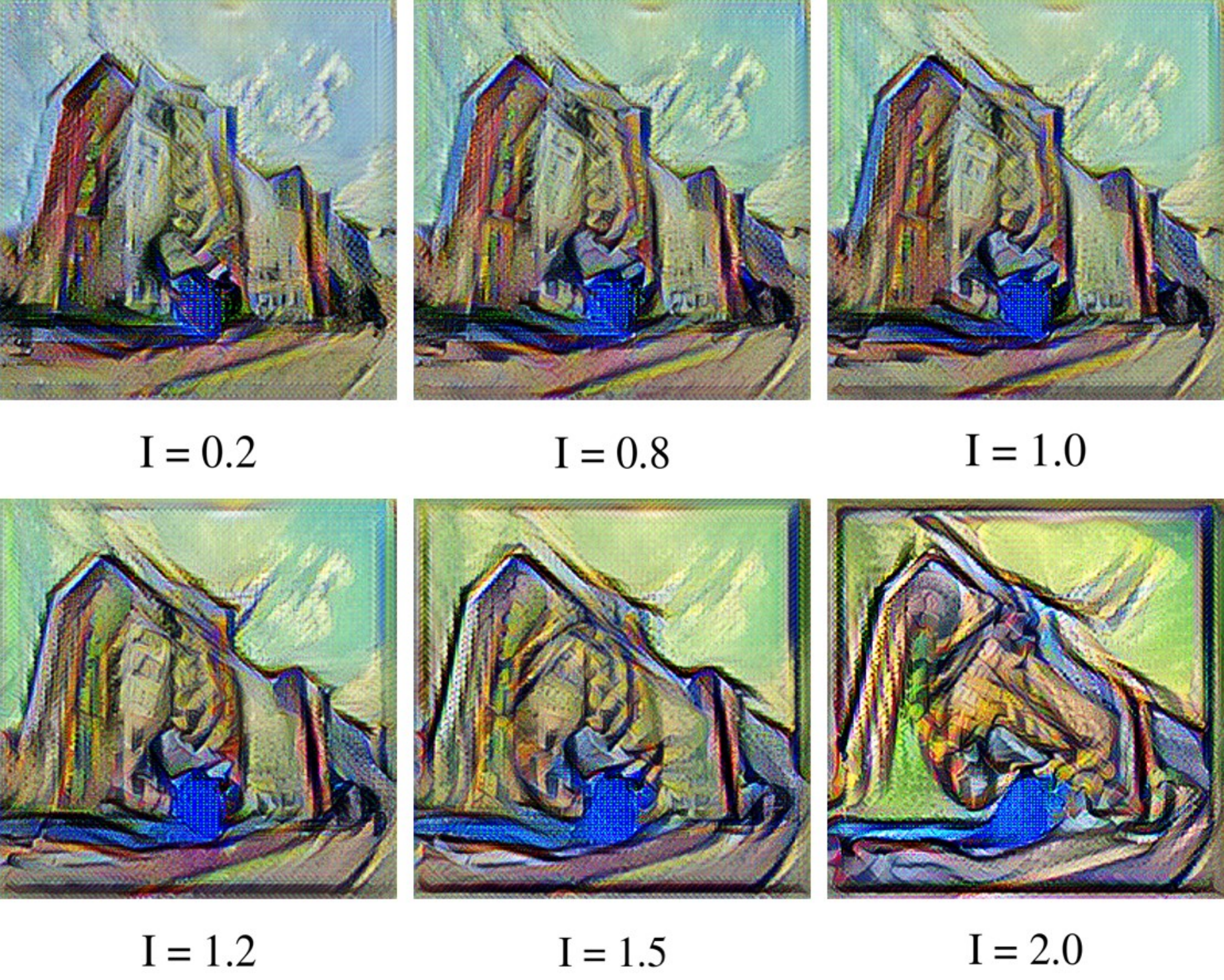}
\end{minipage}%
}%
\subfigure[]{
\begin{minipage}[t]{0.357\linewidth}
\centering
\includegraphics[width=\linewidth]{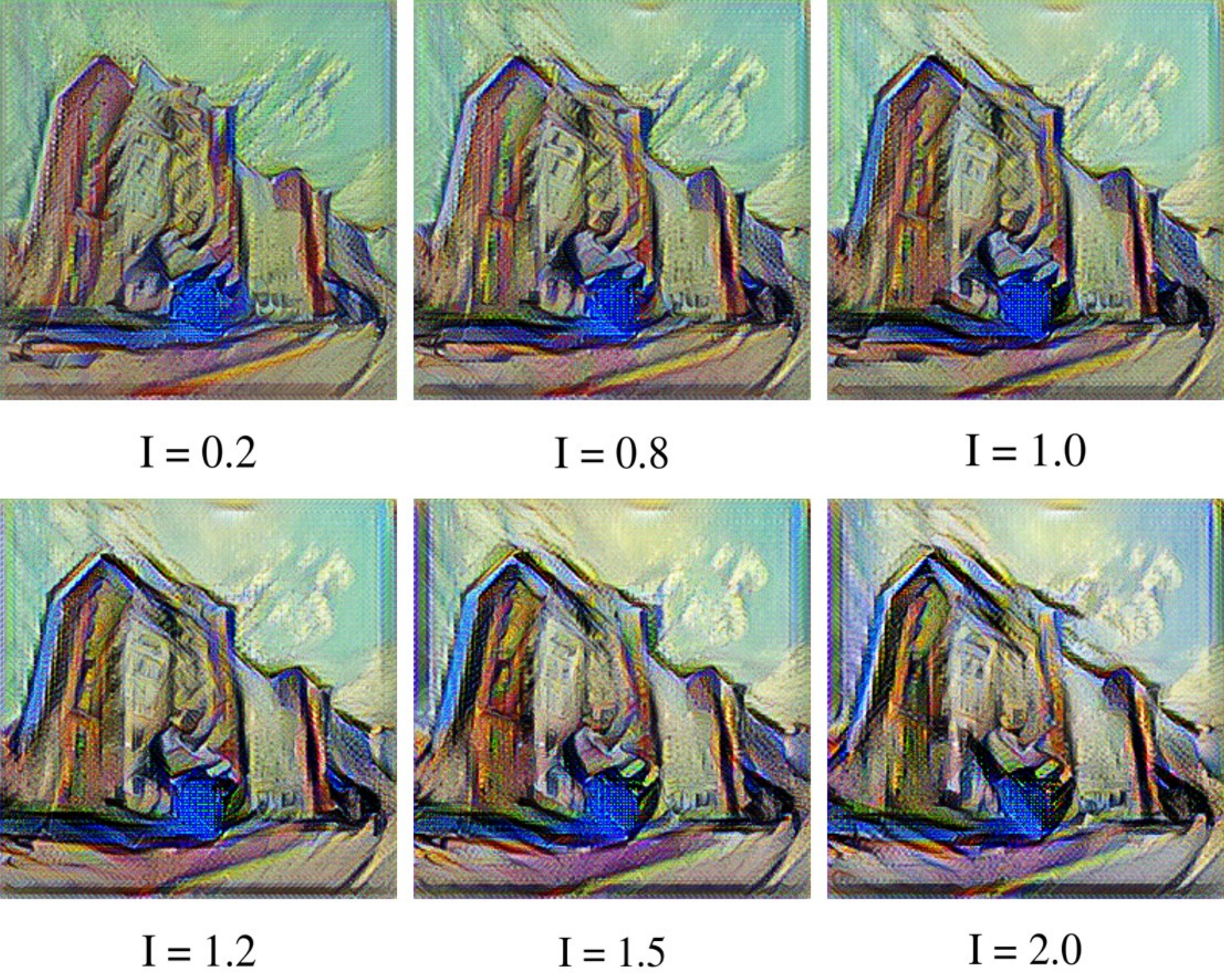}
\end{minipage}%
}%
\centering
\caption{(a) the stroke of the style image `wave'; (d) the stroke of the style image `aMuse'(``A muse'' by Pablo Picasso); (b,c,e,f) the results of giving different intervention $I$ to the stroke basis using different methods. Specifically, (b,e) spectrum based method; (c.f) ICA.}
\label{spectrumIntervention}
\end{figure*}

\subsection{Transfer by intervention}\label{sec:intervention}
We give intervention to the stroke basis via control function $g$ to demonstrate the controllable diversified styles and distinguish the difference in stroke basis between spectrum based methods and ICA. We experimented on various styles and we here demonstrate two of them (dur to space limitation), `wave' and `aMuse', to indicate the robustness of our experiment. The strokes of `wave' (Figure \ref{spectrumIntervention}(a)) are curves with light and dark blue while the strokes of `aMuse' (Figure \ref{spectrumIntervention}(d)) are black bold lines and coarse powder-like dots in green, blue, yellow, etc. We further divide the concept of stroke into stroke color and stroke profile (which is character like curve, bold line and coarse powder-like dot) to better illustrate the difference between two methods.

Intervention impacts both stroke color and stroke profile using spectrum based methods. With intervention increasing, we see more exaggarated curves with darker blue in Figure \ref{spectrumIntervention}(b) and more black bold lines as well as greener and yellower oil-painting-like sky in Figure \ref{spectrumIntervention}(e).

Compared to spectrum based method, there is slight difference in color between results using different intervention. Intervention only impacts stroke profile using ICA. With intervention increasing, only the curvity of curves is amplified (Figure \ref{spectrumIntervention}(c)) and only the margin of profile and the grainy of color become more obvious (Figure \ref{spectrumIntervention}(f)).

However different two methods are, the stroke effect of style can be reduced or amplified within our control, which greatly enhances the diversity of styled image.

\subsection{Transfer by mixing}

Current style mixing method, interpolation, cannot mix the style bases of different styles because styles are integrally mixed however interpolation weights are modified (Figure \ref{mixing}(g-i)), which limits the diversity of style mixing. Based on the success spectrum based methods and ICA in style decomposition, we experiment to mix the stroke of `wave' with the color of `aMuse' to check whether such newly compound artistic style can be transferred to the styled image.

Specifically, for ICA, we not only need to replace the color basis of `wave' with that of `aMuse' but also should replace the rows of mixing matrix $A$ corresponding to the exchanged signals. Both spectrum based methods (Figure \ref{mixing}(d-f)) and ICA (Figure \ref{mixing}(j-l)) work well in mixing style bases of different styles and the difference conforms to the conclusion given in Section \ref{sec:intervention}. Moreover, we can intervenve the style basis when mixing, which further enhances the diversity of style mixing.

\begin{figure*}[ht]
\centering
\subfigure[content and style]{
\begin{minipage}[t]{0.165\linewidth}
\centering
\includegraphics[width=\linewidth]{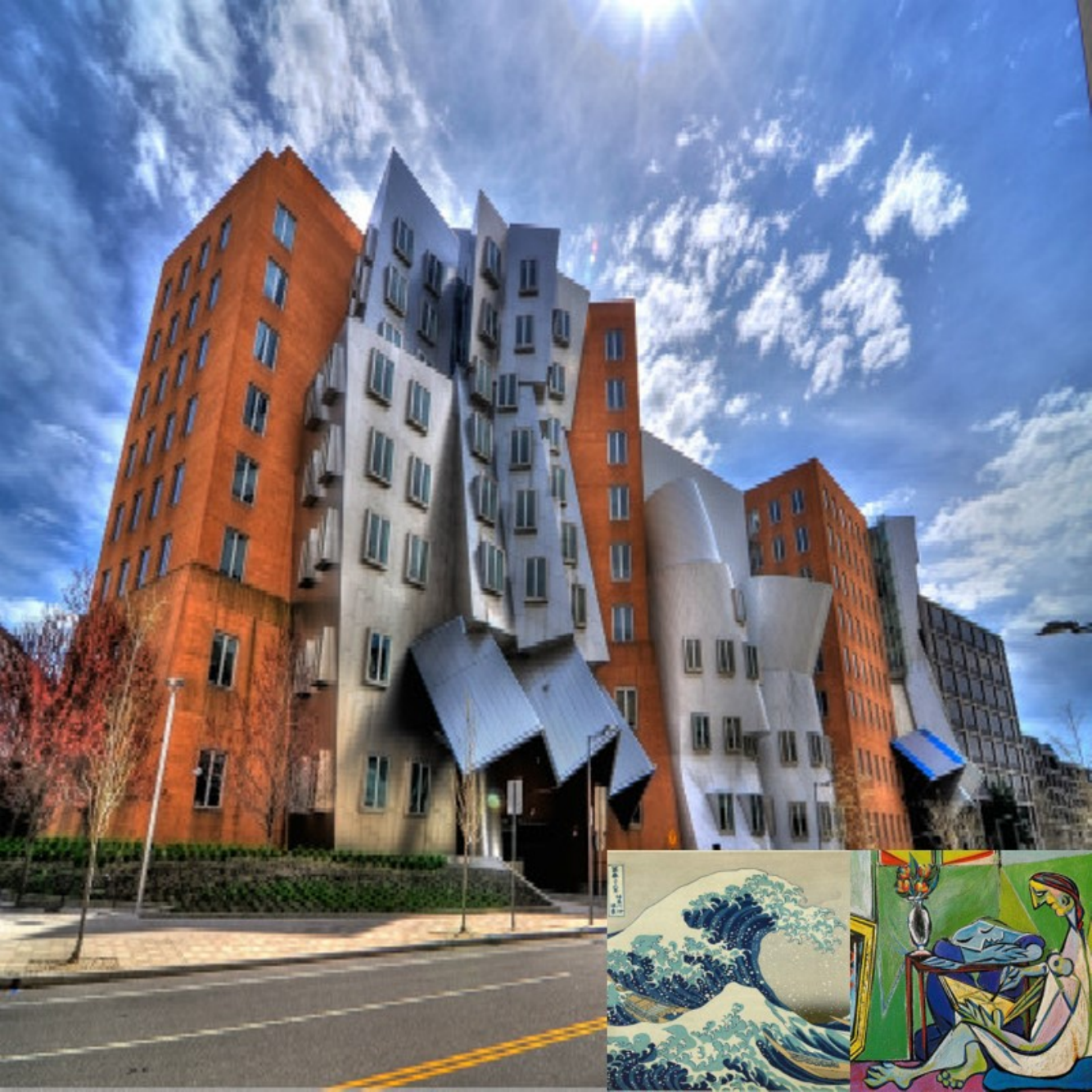}
\end{minipage}%
}%
\subfigure[wave style]{
\begin{minipage}[t]{0.165\linewidth}
\centering
\includegraphics[width=\linewidth]{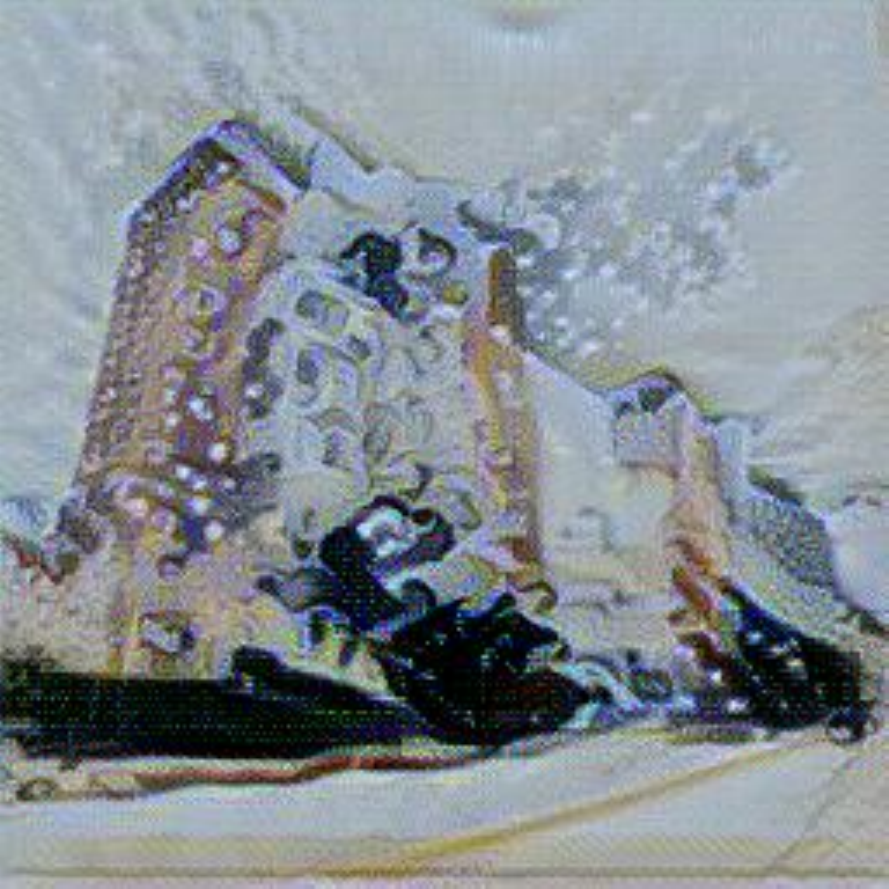}
\end{minipage}%
}%
\subfigure[aMuse style]{
\begin{minipage}[t]{0.165\linewidth}
\centering
\includegraphics[width=\linewidth]{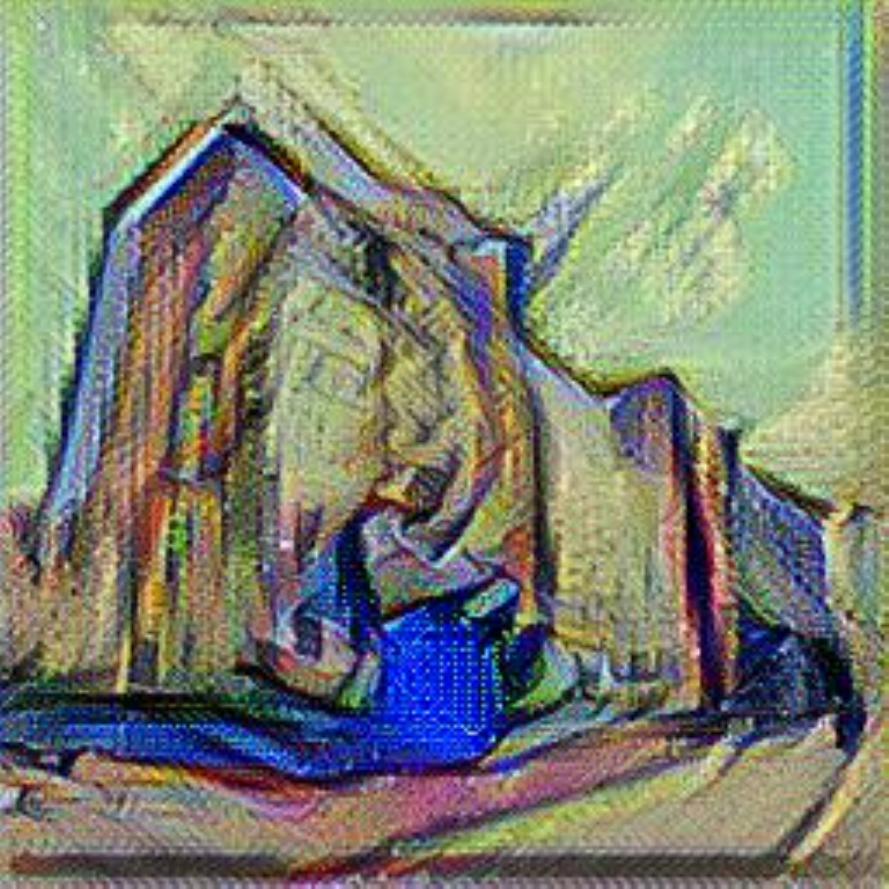}
\end{minipage}%
}%
\subfigure[I = 1.0]{
\begin{minipage}[t]{0.165\linewidth}
\centering
\includegraphics[width=\linewidth]{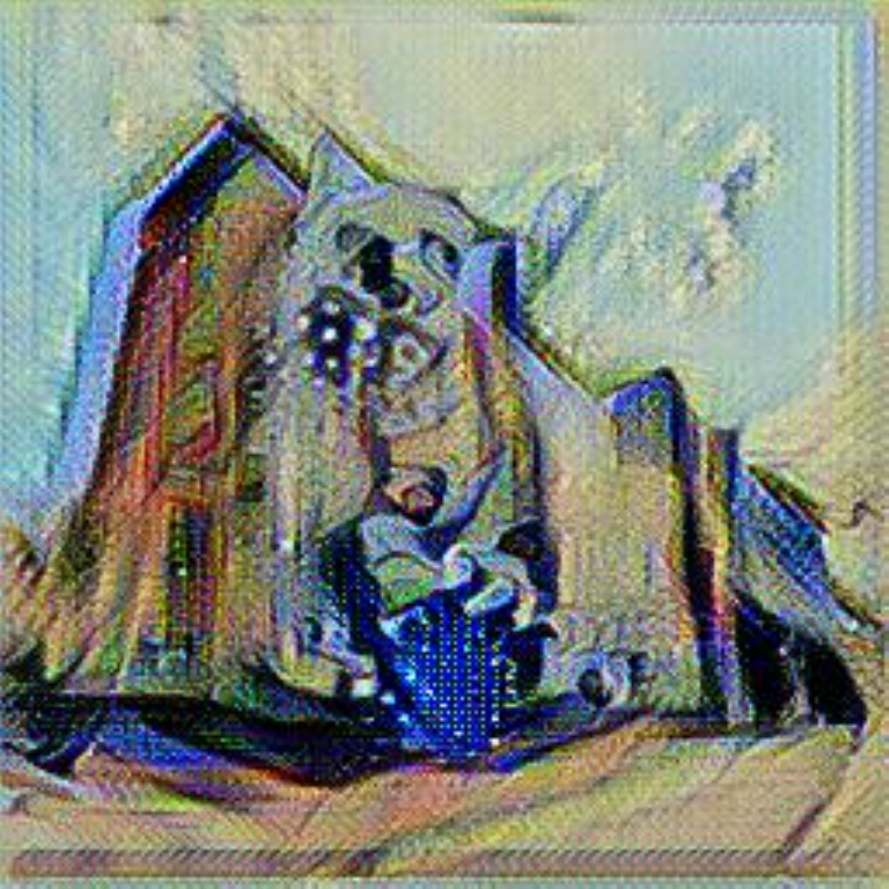}
\end{minipage}%
}%
\subfigure[I = 1.5]{
\begin{minipage}[t]{0.165\linewidth}
\centering
\includegraphics[width=\linewidth]{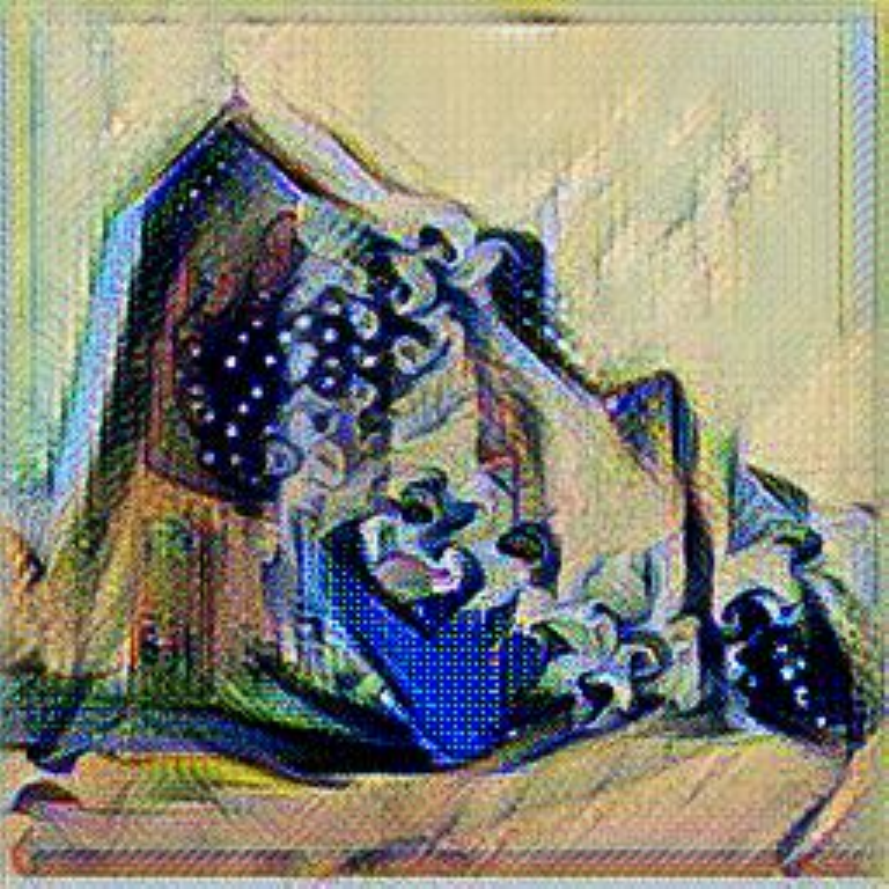}
\end{minipage}%
}%
\subfigure[I = 2.0]{
\begin{minipage}[t]{0.165\linewidth}
\centering
\includegraphics[width=\linewidth]{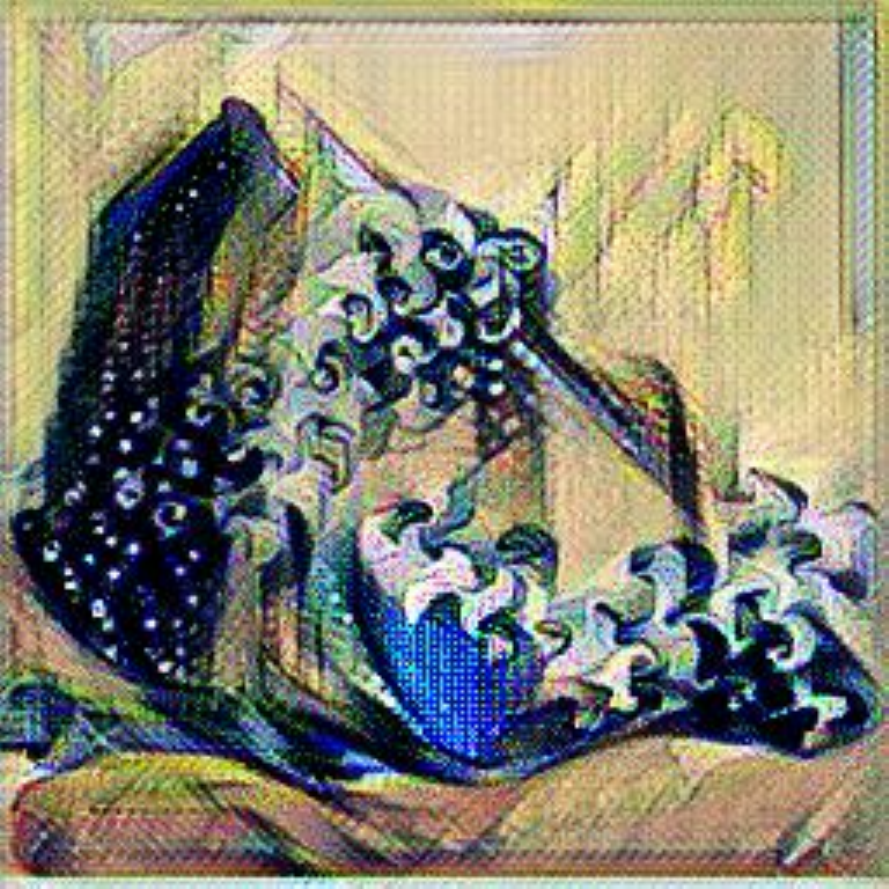}
\end{minipage}%
}%
\vfill
\subfigure[I1 = 0.3, I2 = 0.7]{
\begin{minipage}[t]{0.165\linewidth}
\centering
\includegraphics[width=\linewidth]{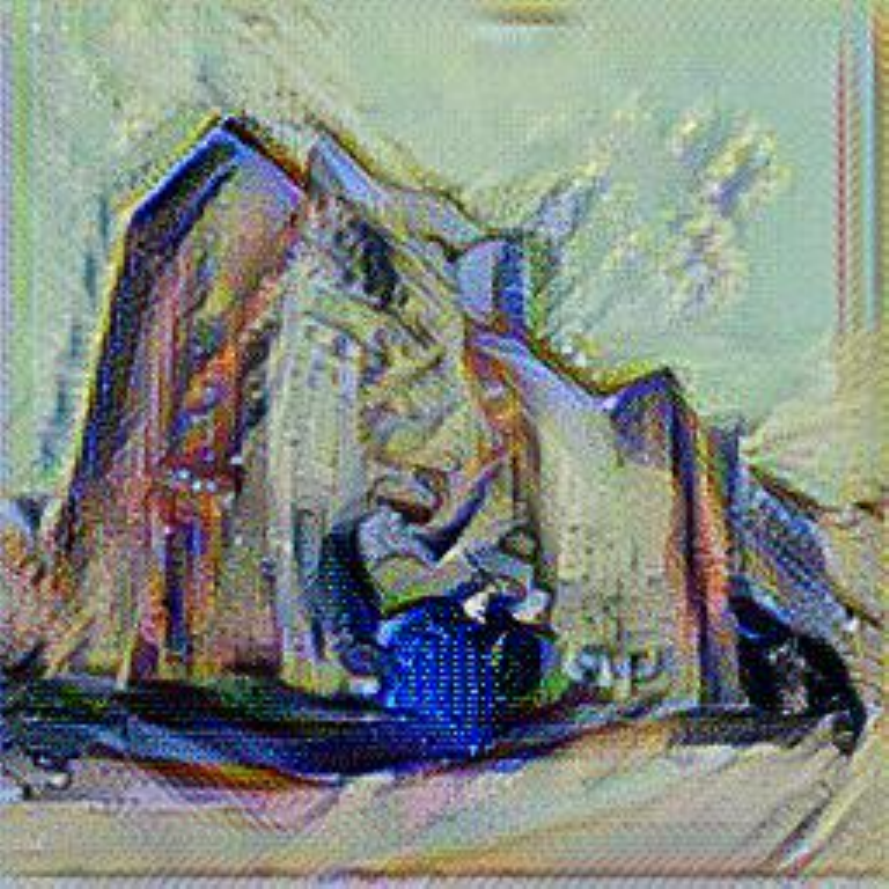}
\end{minipage}%
}%
\subfigure[I1 = 0.5, I2 = 0.5]{
\begin{minipage}[t]{0.165\linewidth}
\centering
\includegraphics[width=\linewidth]{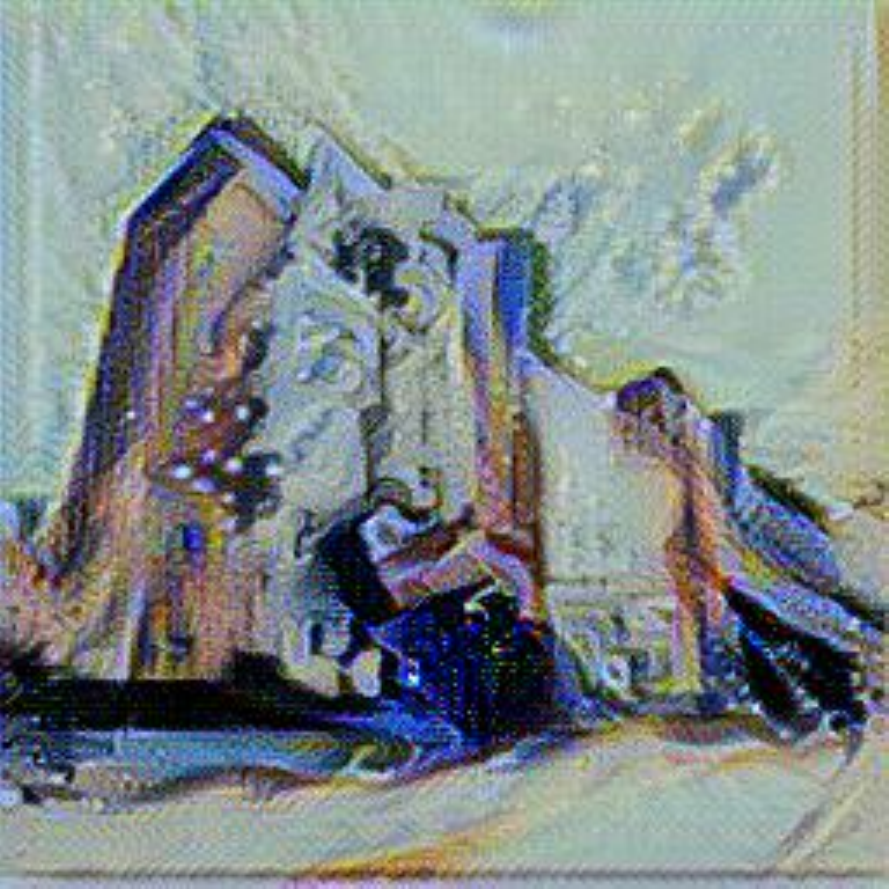}
\end{minipage}%
}%
\subfigure[I1 = 0.7, I2 = 0.3]{
\begin{minipage}[t]{0.165\linewidth}
\centering
\includegraphics[width=\linewidth]{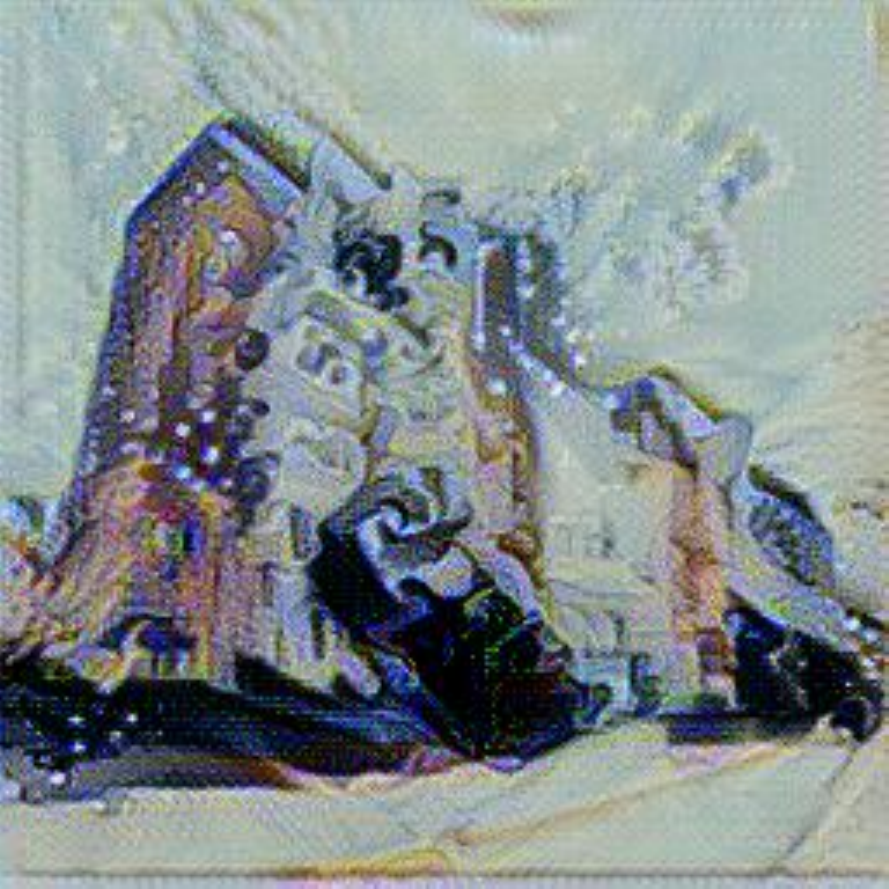}
\end{minipage}%
}%
\subfigure[I = 1.0]{
\begin{minipage}[t]{0.165\linewidth}
\centering
\includegraphics[width=\linewidth]{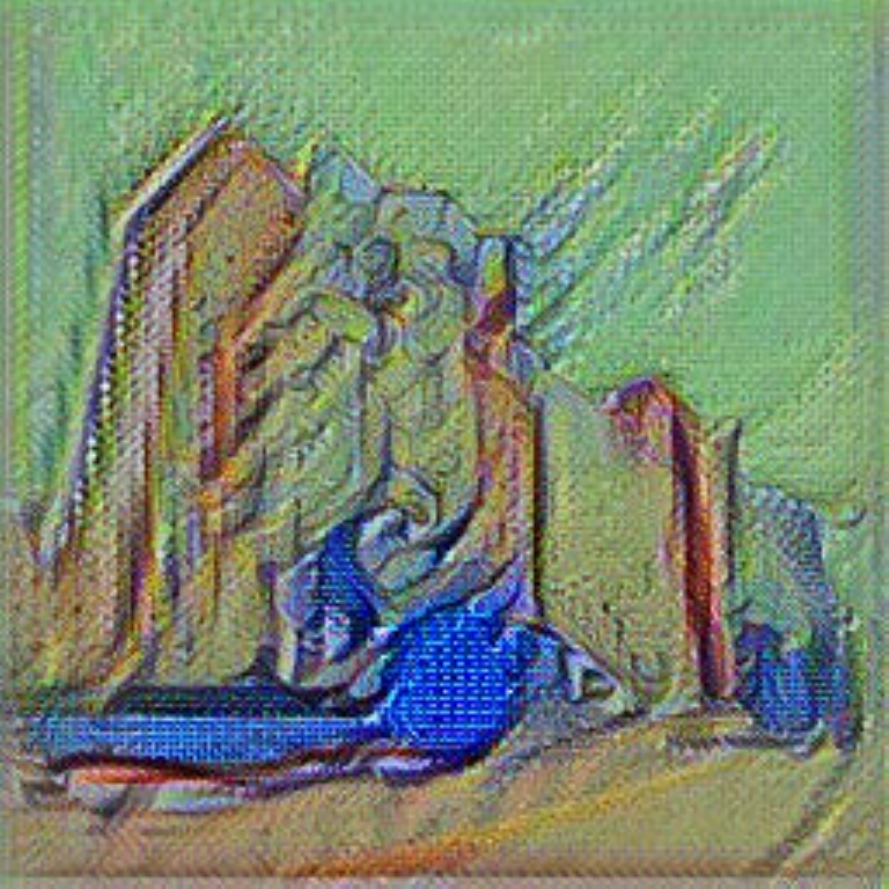}
\end{minipage}%
}%
\subfigure[I = 2.0]{
\begin{minipage}[t]{0.165\linewidth}
\centering
\includegraphics[width=\linewidth]{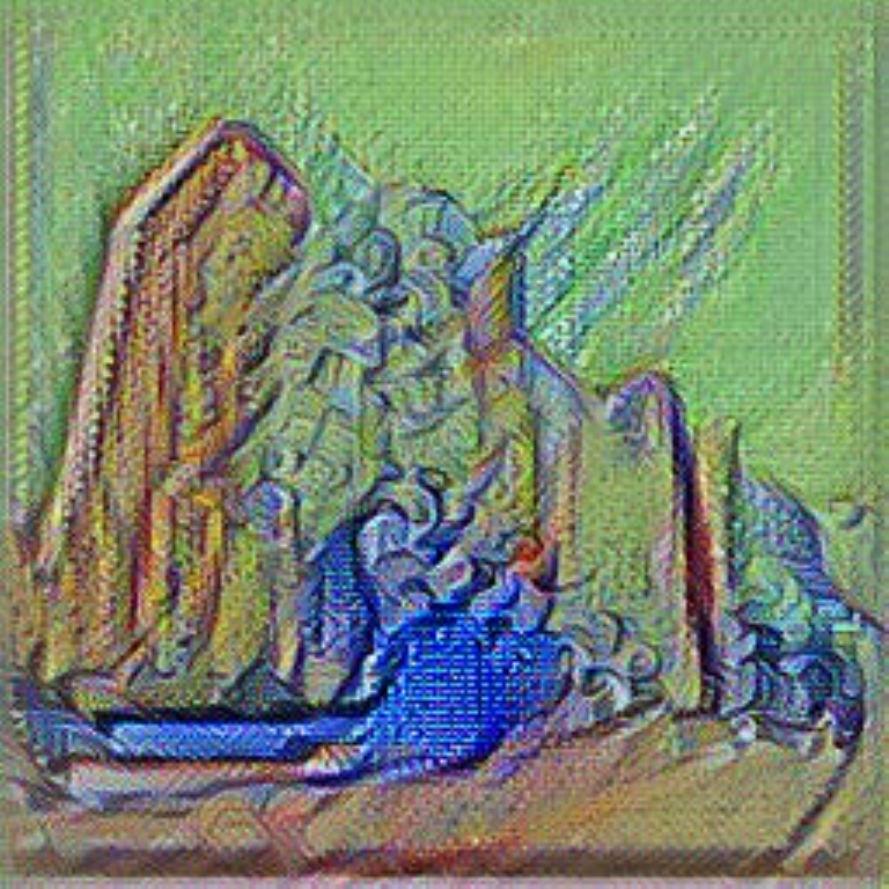}
\end{minipage}%
}%
\subfigure[I = 3.0]{
\begin{minipage}[t]{0.165\linewidth}
\centering
\includegraphics[width=\linewidth]{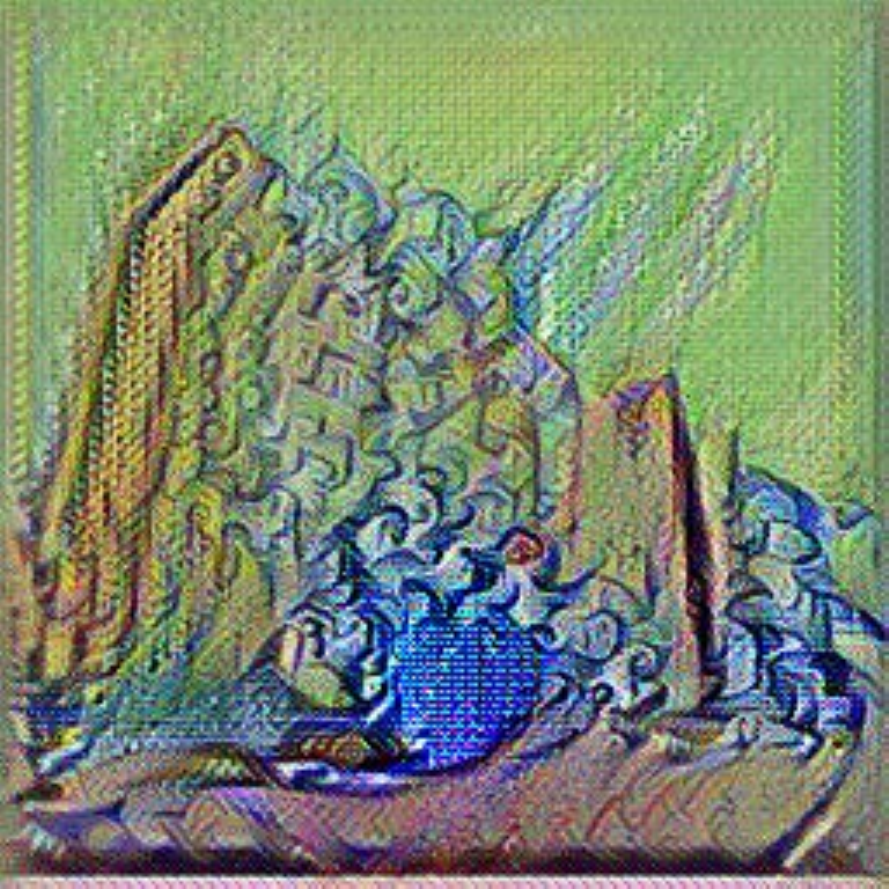}
\end{minipage}%
}%
\centering
\caption{(a) the content image and two style images; (b)(c) styled image of single style using traditional methods \cite{GatysNeuralStyle}; (g-i) interpolation mixing where I1 and I2 are the weights of `wave' and `aMuse' in interpolation; (d-f,j-l) results of mixing the color of `aMuse' and the stroke of `wave' where I is the intervention to the stroke of `wave'. Specifically, (d-f) use FFT; (j-l) use ICA.}
\label{mixing}
\end{figure*}

\begin{figure}[t]
\centering
\subfigure[]{
\begin{minipage}[t]{0.25\linewidth}
\centering
\includegraphics[width=\linewidth]{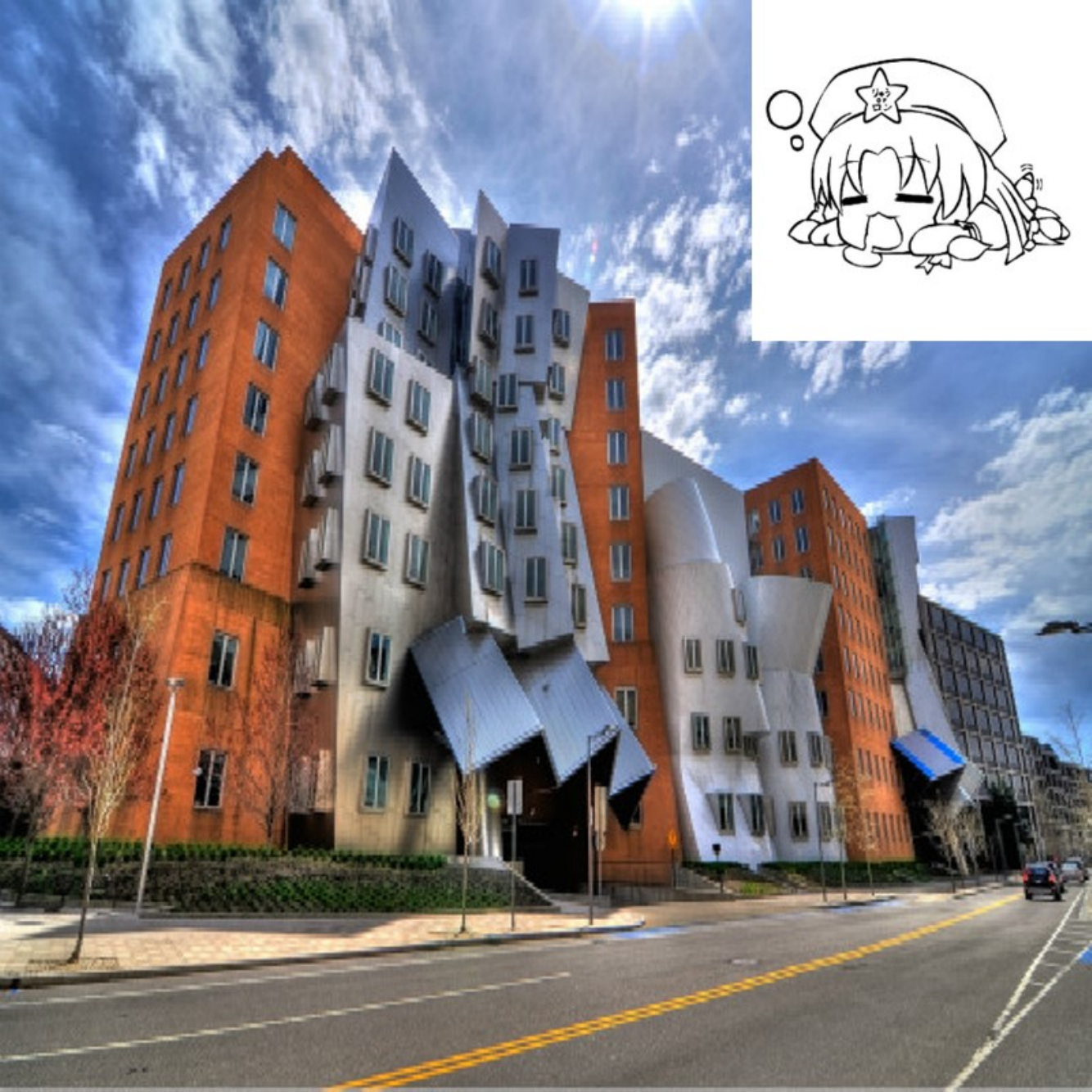}
\end{minipage}%
}%
\subfigure[]{
\begin{minipage}[t]{0.25\linewidth}
\centering
\includegraphics[width=\linewidth]{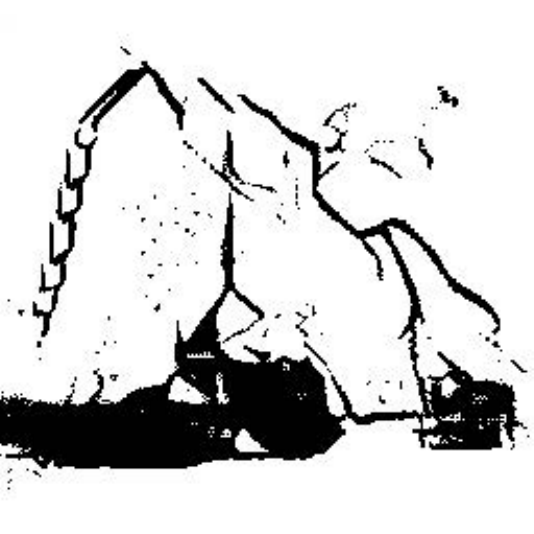}
\end{minipage}%
}%
\subfigure[]{
\begin{minipage}[t]{0.25\linewidth}
\centering
\includegraphics[width=\linewidth]{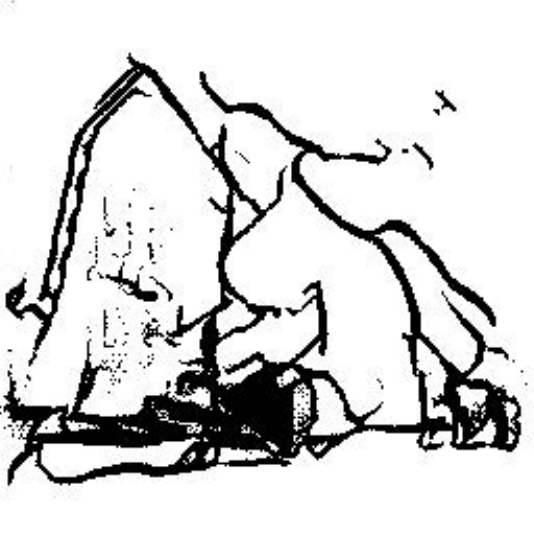}
\end{minipage}%
}%
\subfigure[]{
\begin{minipage}[t]{0.25\linewidth}
\centering
\includegraphics[width=\linewidth]{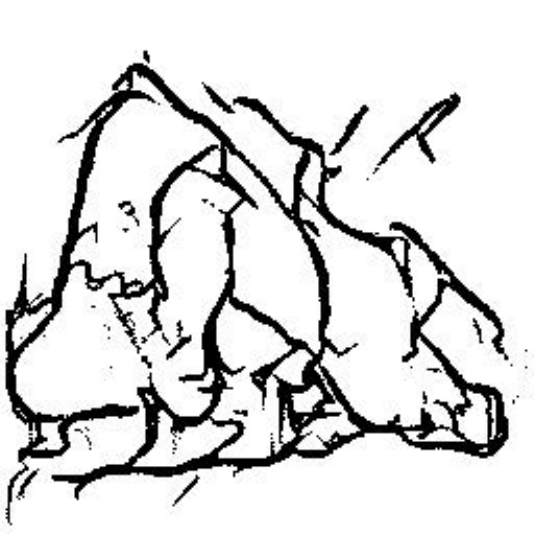}
\end{minipage}%
}%
\centering
\caption{Picture-to-sketch using style transfer and binarization. (a) content image and style image; (b-d) styled images. From (b) to (d), the number of stroke increases as more details of the content image are restored.}
\label{sketch}
\end{figure}

\begin{figure}[t]
\centering
\subfigure[]{
\begin{minipage}[t]{0.25\linewidth}
\centering
\includegraphics[width=\linewidth]{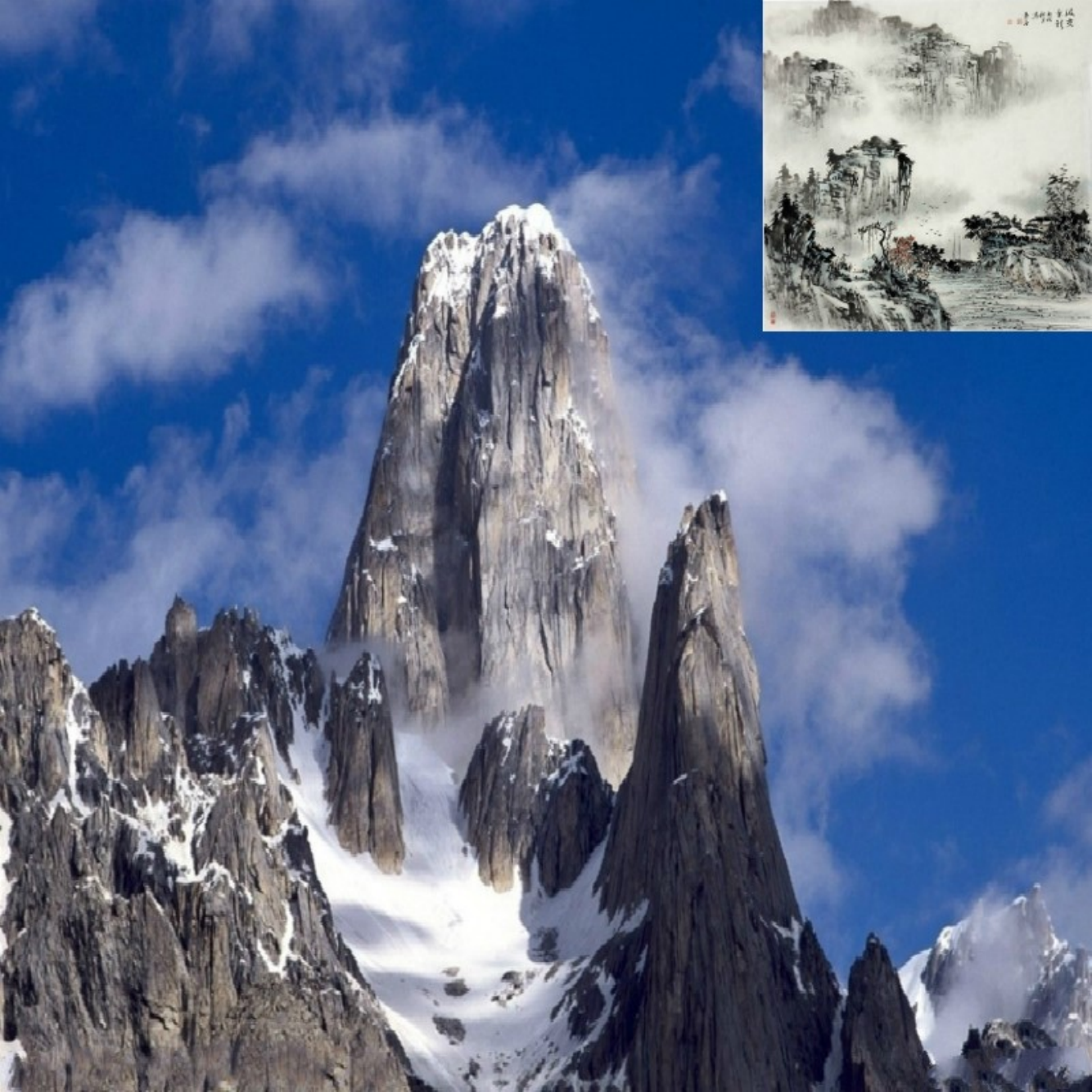}
\end{minipage}%
}%
\subfigure[]{
\begin{minipage}[t]{0.25\linewidth}
\centering
\includegraphics[width=\linewidth]{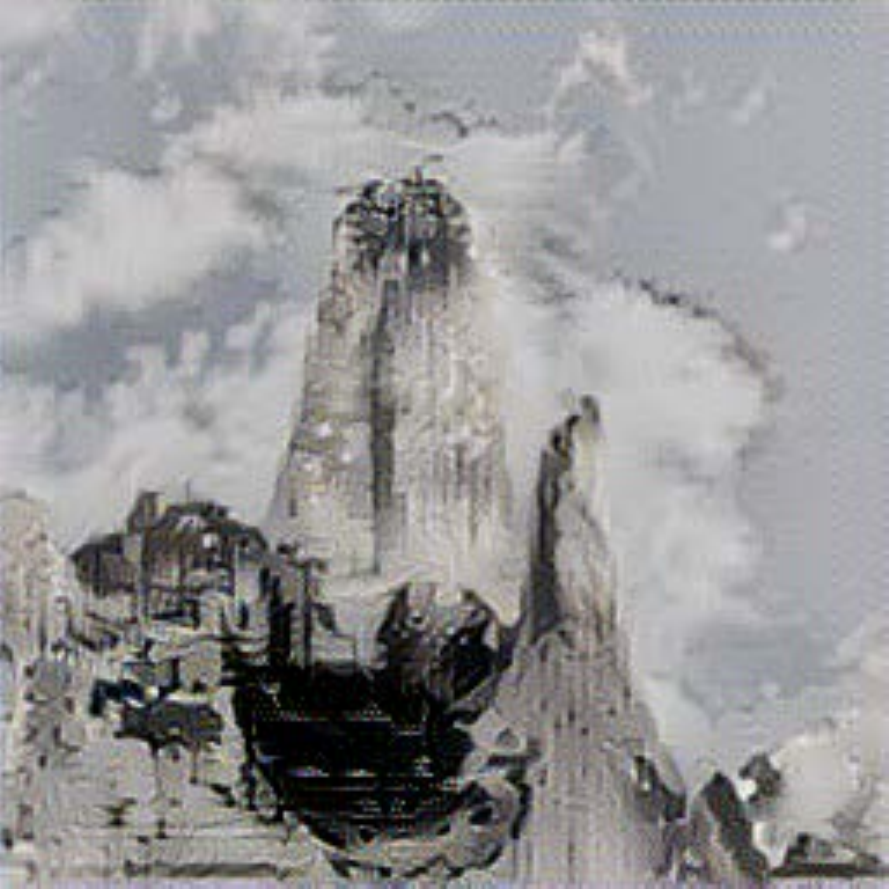}
\end{minipage}%
}%
\subfigure[]{
\begin{minipage}[t]{0.25\linewidth}
\centering
\includegraphics[width=\linewidth]{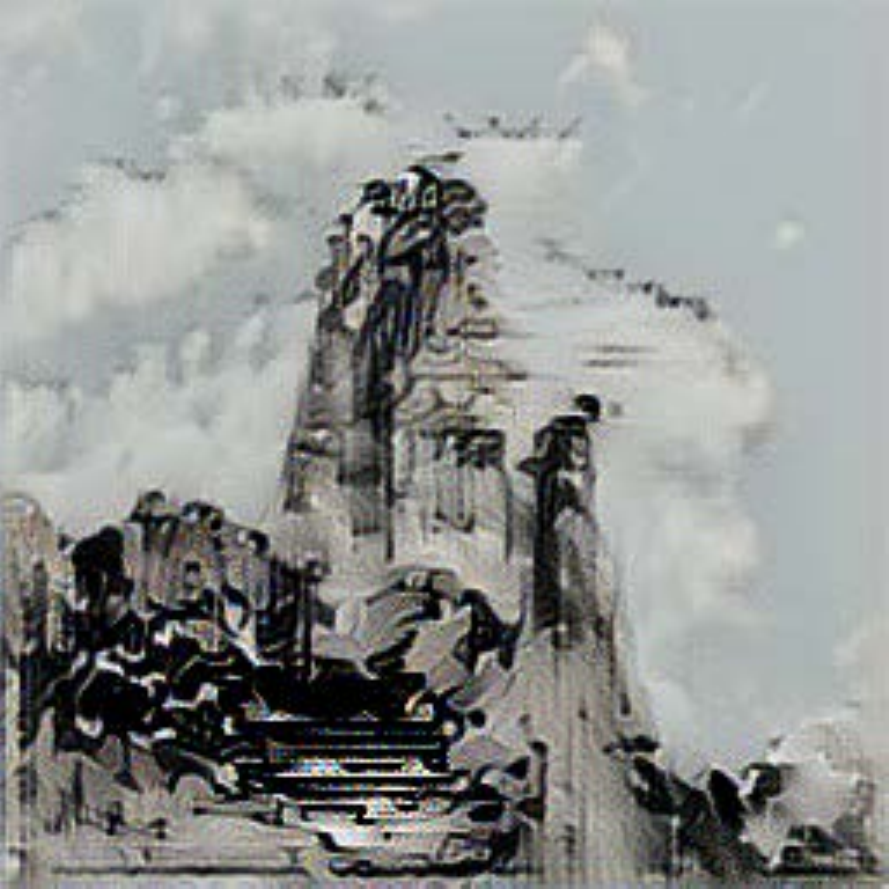}
\end{minipage}%
}%
\subfigure[]{
\begin{minipage}[t]{0.25\linewidth}
\centering
\includegraphics[width=\linewidth]{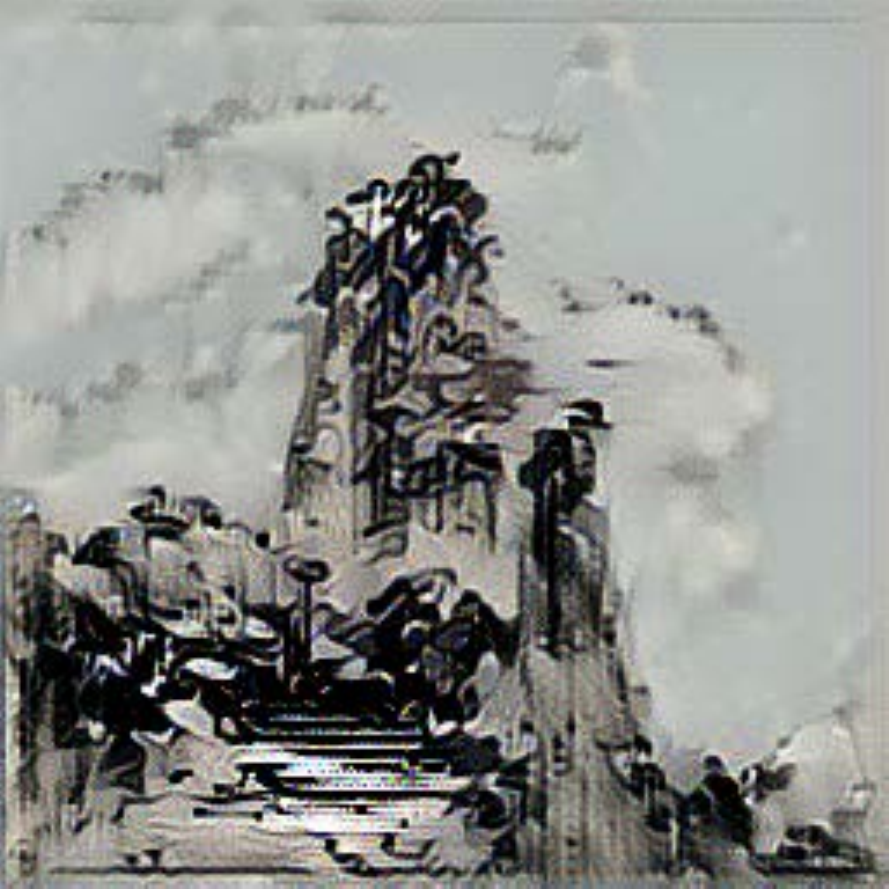}
\end{minipage}%
}%
\centering
\caption{Neural style transfer of Chinese painting with stroke controlled. (a) content image and style image (by Zaixin Miao); (b-d) styled images. From (b) to (d), the strokes are getting more detailed which gradually turns freehand style into finebrush style.}
\label{chinese}
\end{figure}


\subsection{Sketch style transfer}

Picture-to-sketch problem challenges how computer can understand and represent the concept of objects both abstractly and semantically. State-of-the-art methods \cite{2017sketch,2018sketch} use variance model of genarative adversary network (GAN) via both supervised and un-supervised methods. One obstacle mentioned by \cite{2018sketch} is that using supervised learning only may result in unstablity due to the noise in the dataset which is caused by variant sketch styles for the same data sample. Controllable neural style transfer proposed by us tackles the above obstacle because inconsistent styles are no more burdens, but can in turn increase the style diversity of output images. Moreover, as is shown in  Figure \ref{sketch}, our method can control the abstract level by reserving major semantic details and minor ones automatically. In addition, our method does not require vector sketch dataset, but as the tradeoff, we cannot generate sketch stroke by stroke like \cite{2017sketch,2018sketch}.


\subsection{Chinese painting style transfer}

Chinese painting is an exclusive artistic style which does not have plentiful color like the Western painting, but mostly represents the artistic conception by strokes. Taking advantage of effective controls over stroke via our methods, the Chinese painting styled image can be either misty-like (Figure \ref{chinese}(b)) which can be called as freehand-brush or meticulous representation (Figure \ref{chinese}(d)) which is called as fine-brush.

\section{Conclusions}
Artistic styles are made of basic elements, each with distinct characteristics and functionality. Developing such a style decomposition method facilitate the quantitative control of the styles in one or more images to be transfer to another natural image, while still keeping the basic content of natural image. In this paper, we proposed a novel computational decomposition method, and demonstrated its strengths via extensive experiments. To our best knowledge, it is the first such study, which could serve as a computational module embedded in those neural style transfer algorithms. We implemented the decomposition function by spectrum transform or latent variable models, and thus it enables us to computationally and continuously control the styles by linear mixing or intervention. Experiments showed that our method enhanced the flexibility of style mixing and the diversity of stylization. Moreover, our method could be applied in picture-to-sketch problems by transferring the sketch style, and it captures the key feature and facilitates the stylization of the Chinese painting style.

{\small
\bibliographystyle{ieee}
\bibliography{reference}
}

\newpage
\section*{Appendix}
\subsection*{A. The Stylization Effect of Every Activation Layer in VGG19}

\begin{figure*}[htb]
\centering
\subfigure[content]{
\begin{minipage}[t]{0.2\linewidth}
\centering
\includegraphics[width=\linewidth]{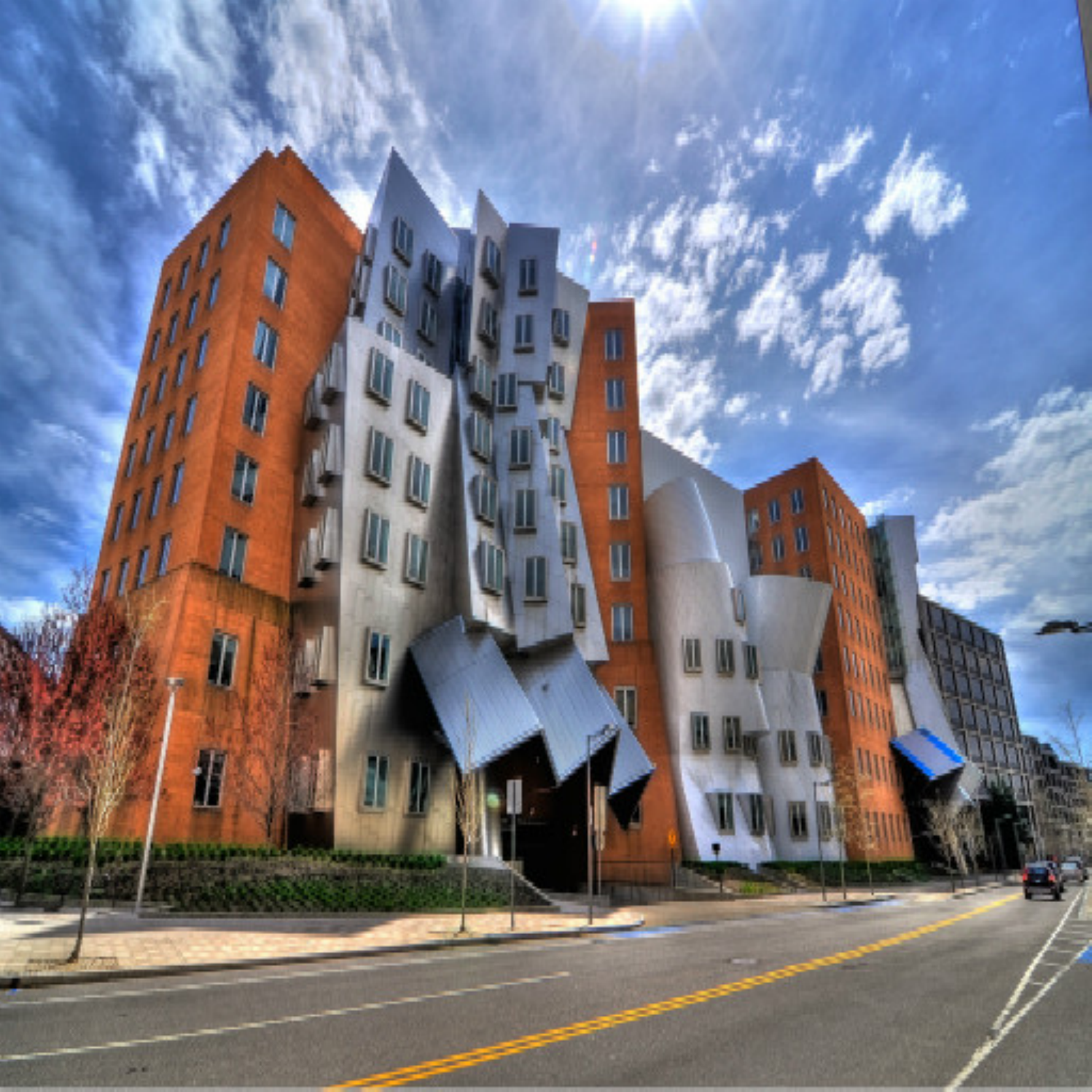}
\end{minipage}%
}%
\subfigure[style]{
\begin{minipage}[t]{0.2\linewidth}
\centering
\includegraphics[width=\linewidth]{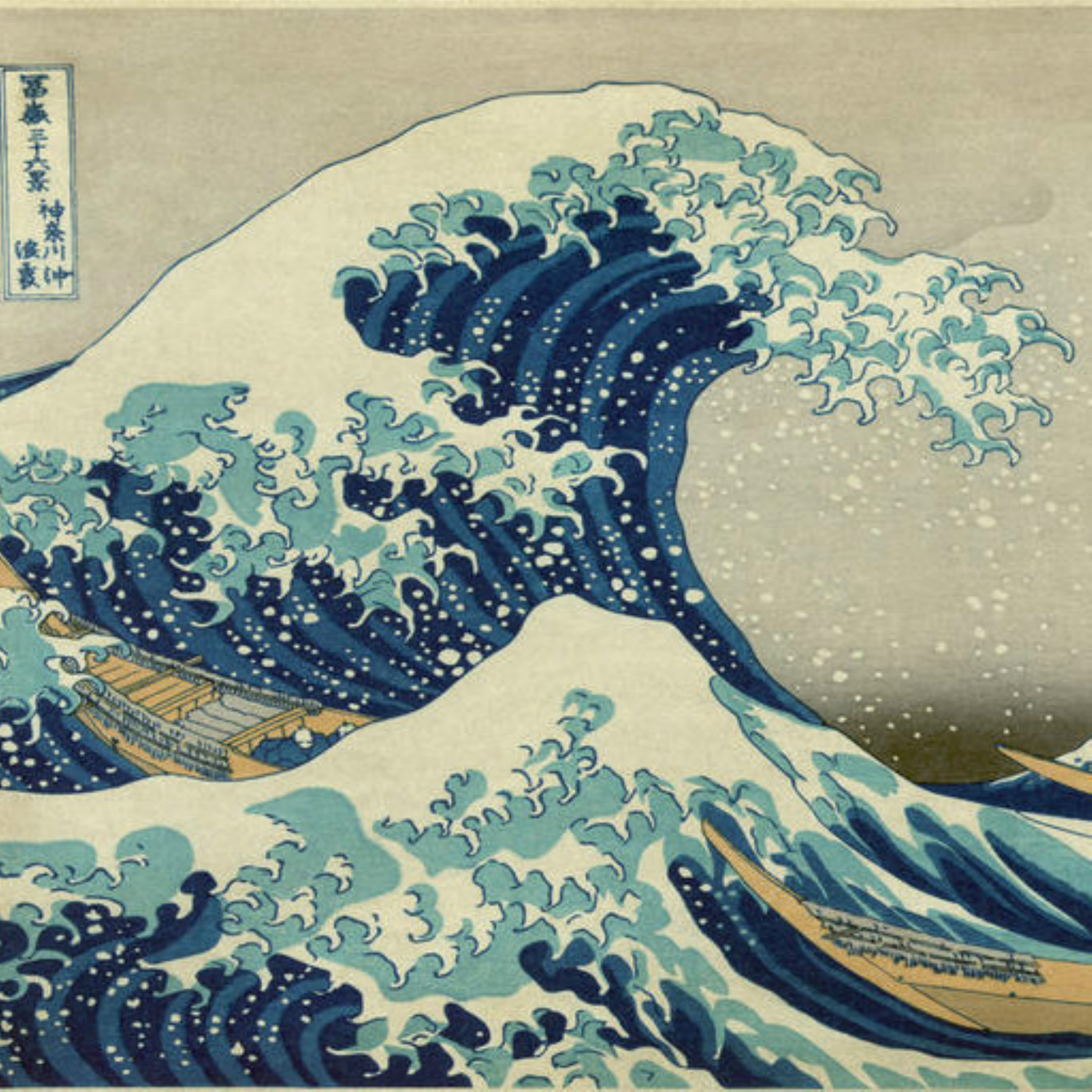}
\end{minipage}%
}%
\vfill

\subfigure[relu1-1]{
\begin{minipage}[t]{0.2\linewidth}
\centering
\includegraphics[width=\linewidth]{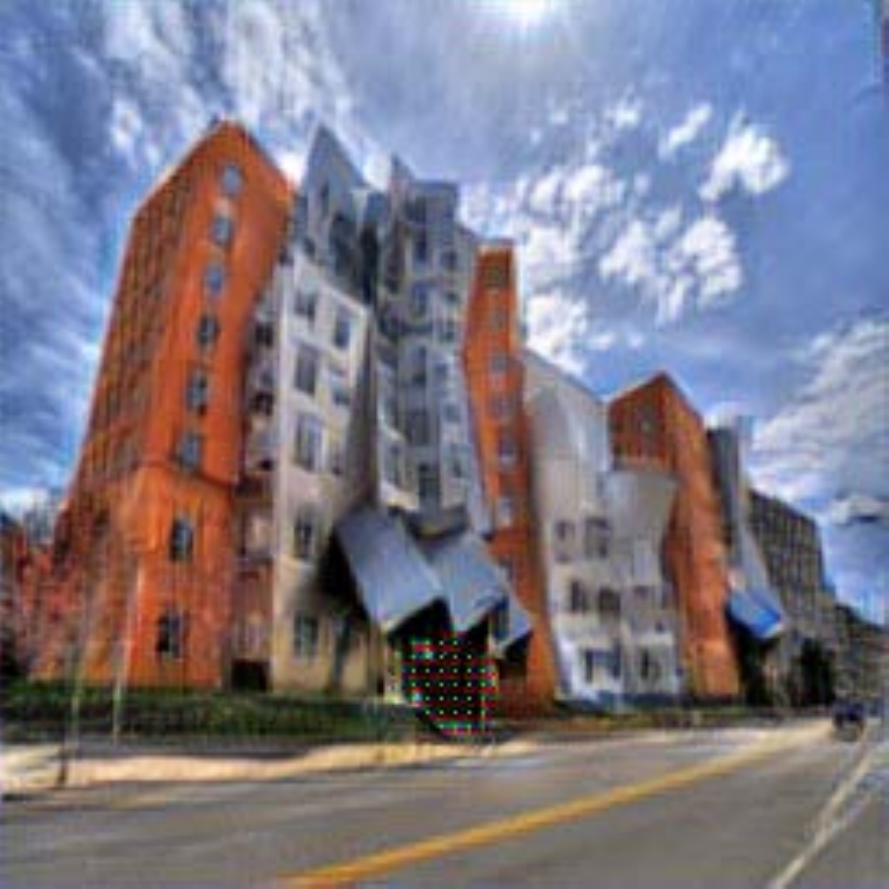}
\end{minipage}%
}%
\subfigure[relu1-2]{
\begin{minipage}[t]{0.2\linewidth}
\centering
\includegraphics[width=\linewidth]{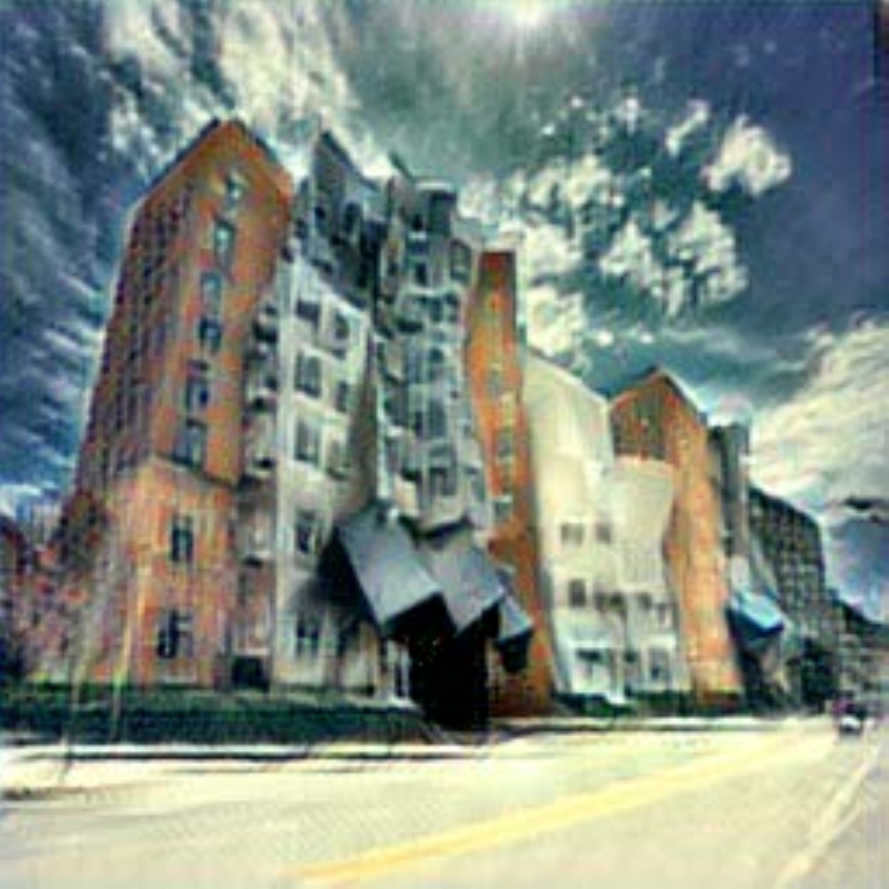}
\end{minipage}%
}%
\subfigure[relu2-1]{
\begin{minipage}[t]{0.2\linewidth}
\centering
\includegraphics[width=\linewidth]{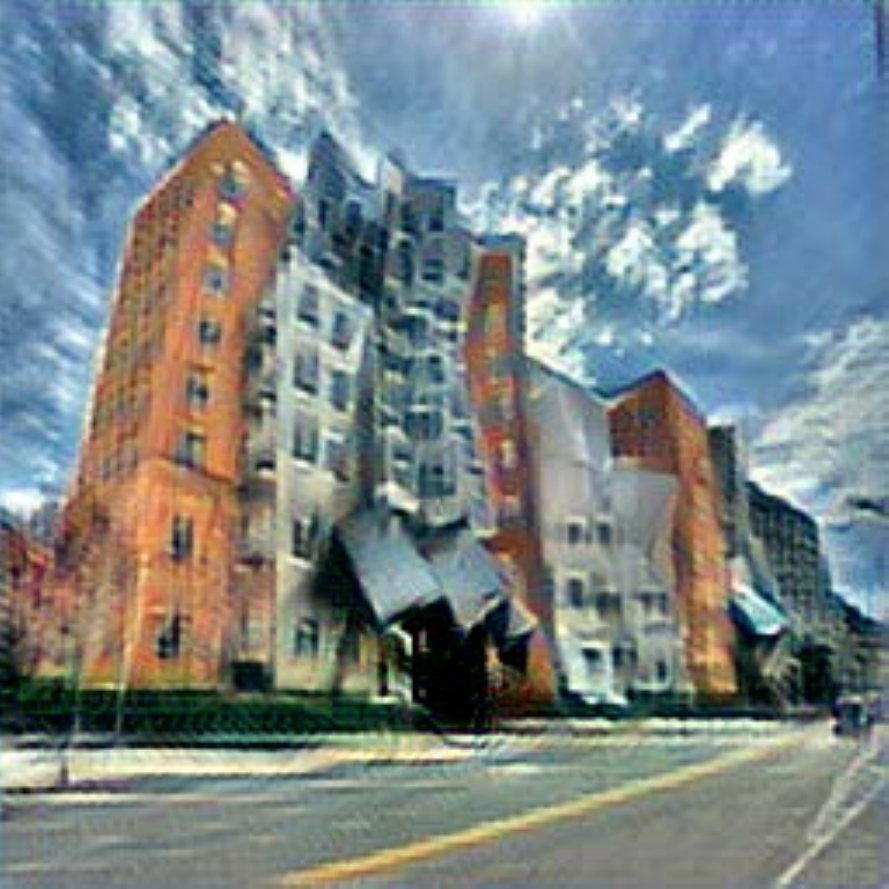}
\end{minipage}%
}%
\subfigure[relu2-2]{
\begin{minipage}[t]{0.2\linewidth}
\centering
\includegraphics[width=\linewidth]{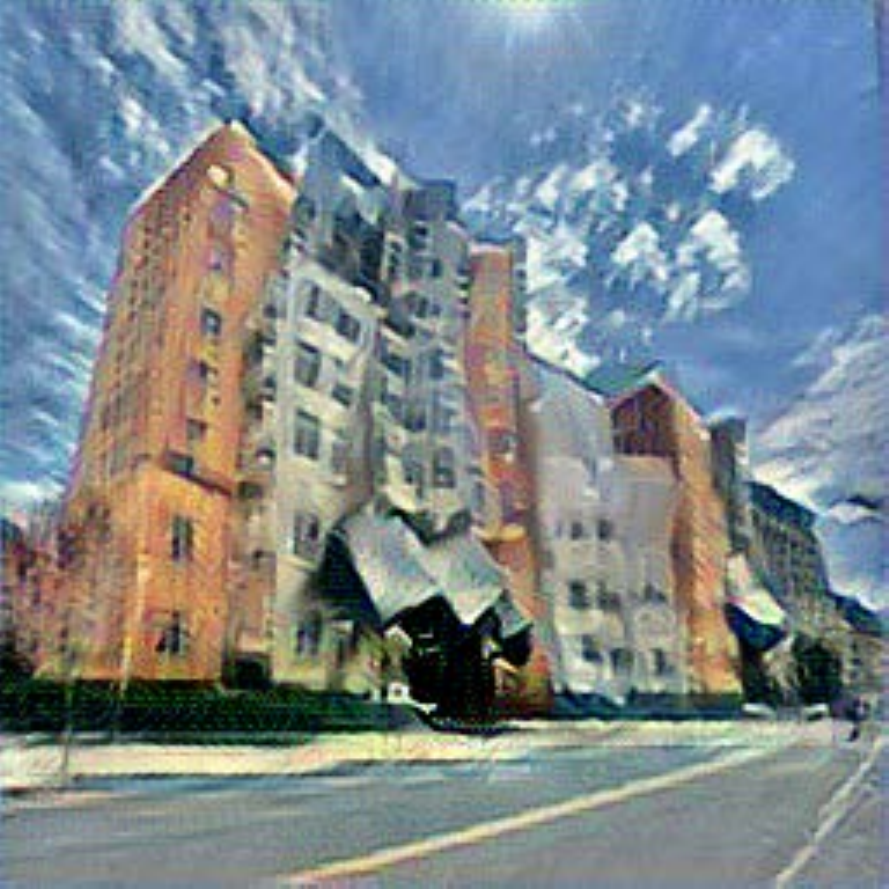}
\end{minipage}%
}%
\vfill

\subfigure[relu3-1]{
\begin{minipage}[t]{0.2\linewidth}
\centering
\includegraphics[width=\linewidth]{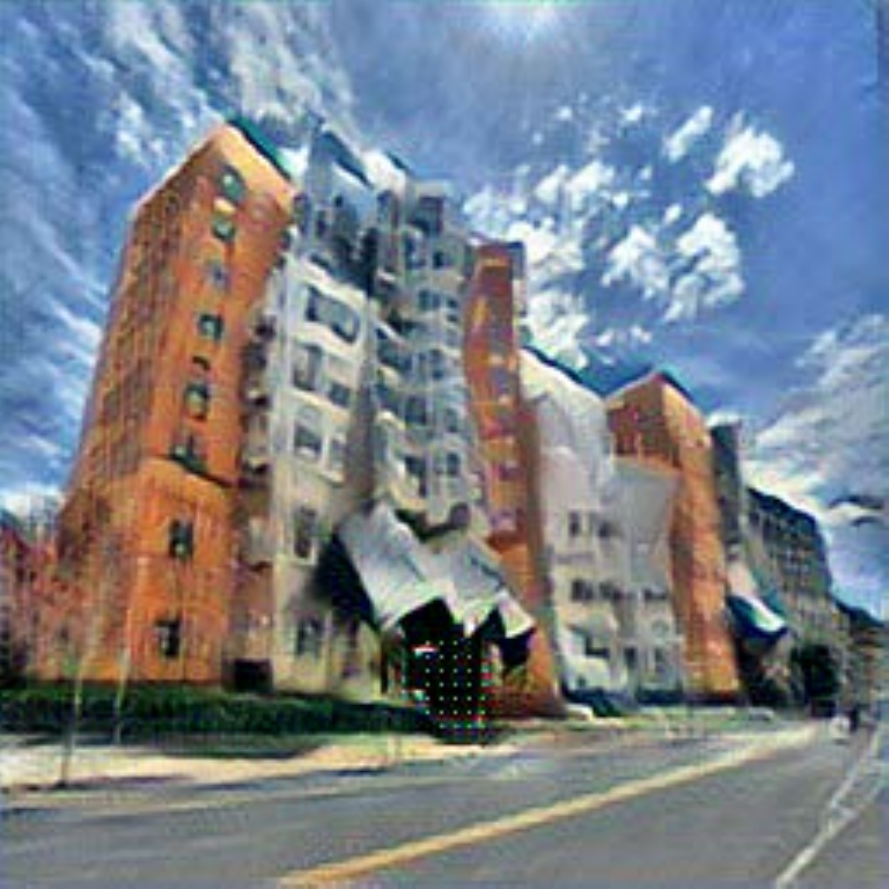}
\end{minipage}%
}%
\subfigure[relu3-2]{
\begin{minipage}[t]{0.2\linewidth}
\centering
\includegraphics[width=\linewidth]{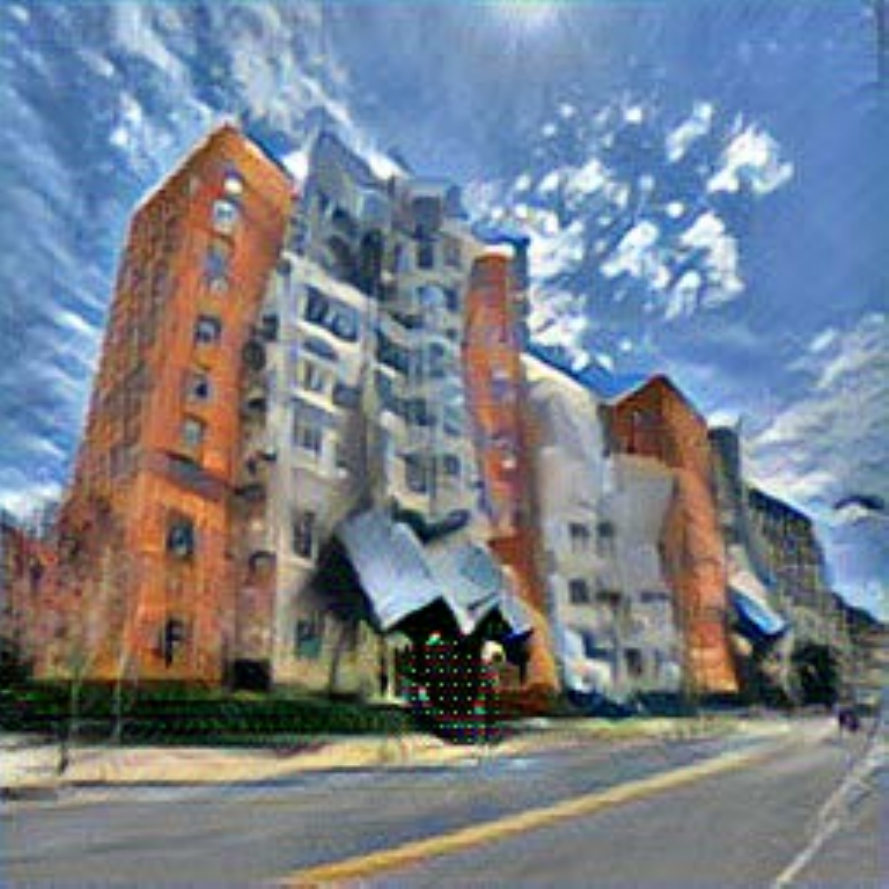}
\end{minipage}%
}%
\subfigure[relu3-3]{
\begin{minipage}[t]{0.2\linewidth}
\centering
\includegraphics[width=\linewidth]{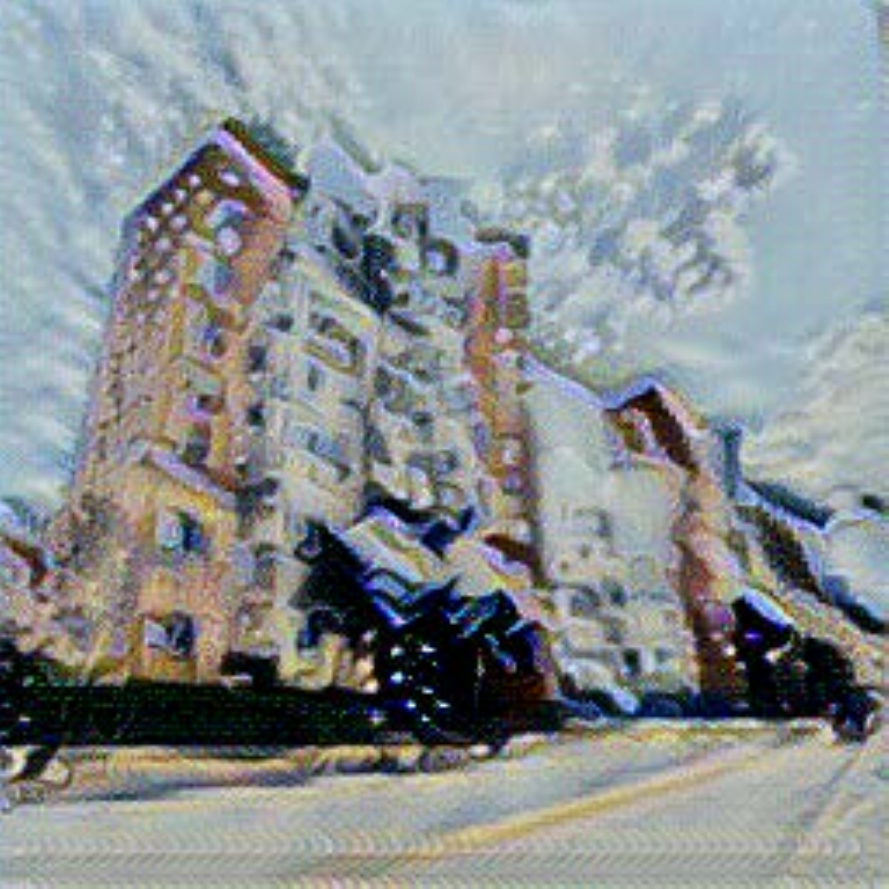}
\end{minipage}%
}%
\subfigure[relu3-4]{
\begin{minipage}[t]{0.2\linewidth}
\centering
\includegraphics[width=\linewidth]{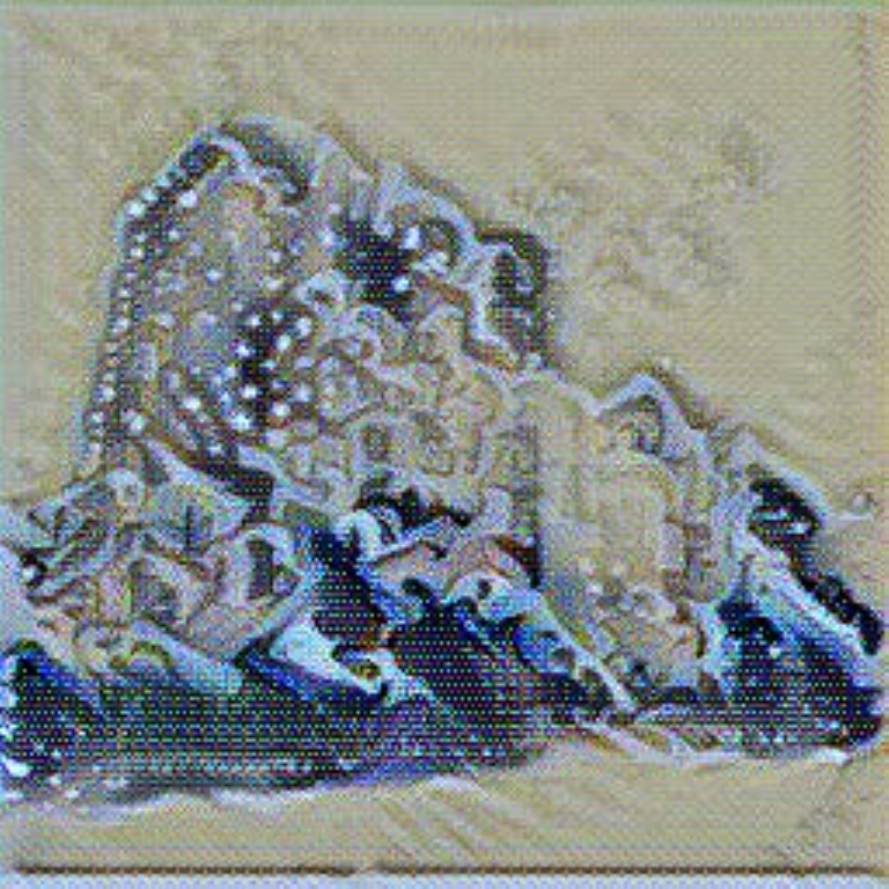}
\end{minipage}%
}%
\vfill

\subfigure[relu4-1]{
\begin{minipage}[t]{0.2\linewidth}
\centering
\includegraphics[width=\linewidth]{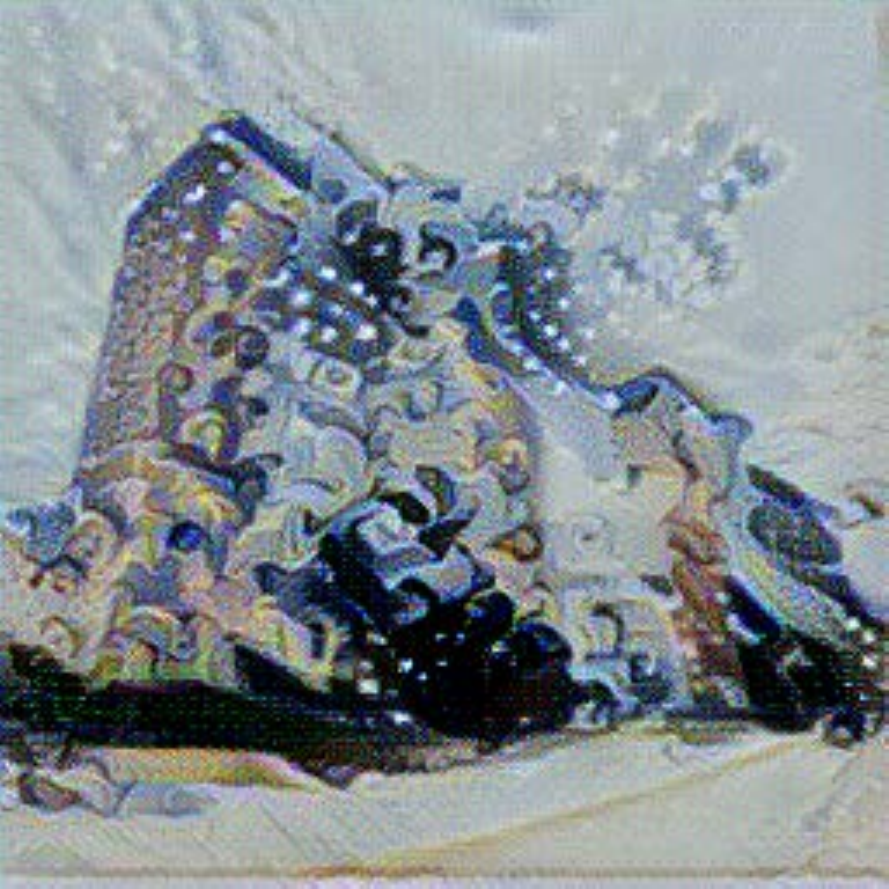}
\end{minipage}%
}%
\subfigure[relu4-2]{
\begin{minipage}[t]{0.2\linewidth}
\centering
\includegraphics[width=\linewidth]{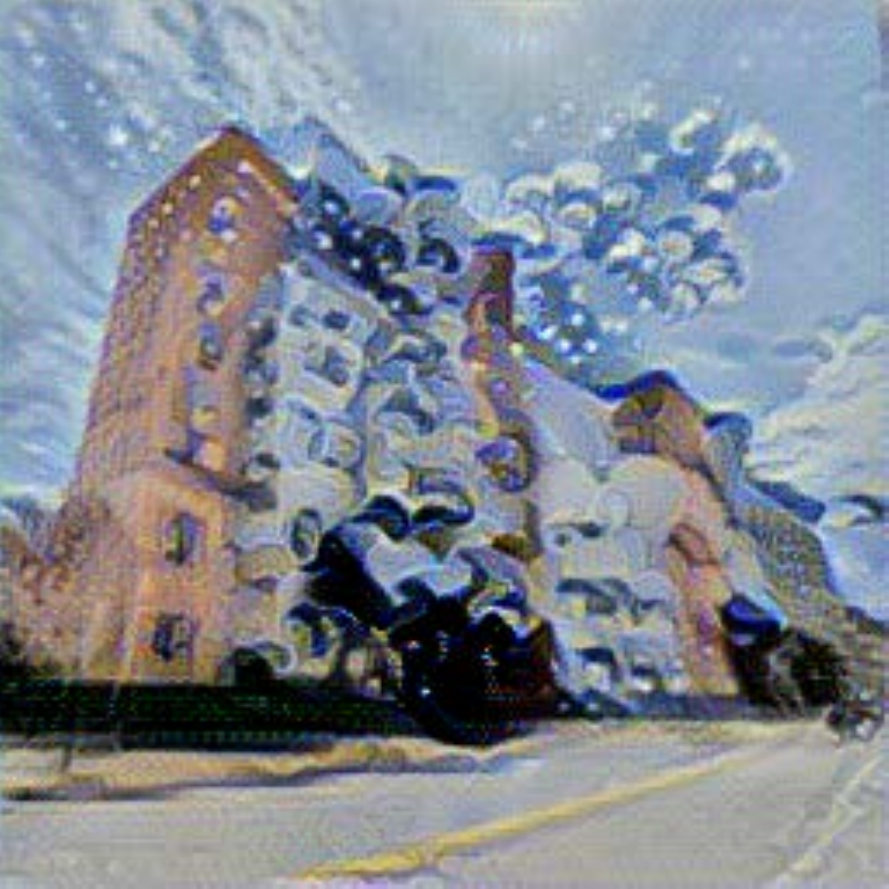}
\end{minipage}%
}%
\subfigure[relu4-3]{
\begin{minipage}[t]{0.2\linewidth}
\centering
\includegraphics[width=\linewidth]{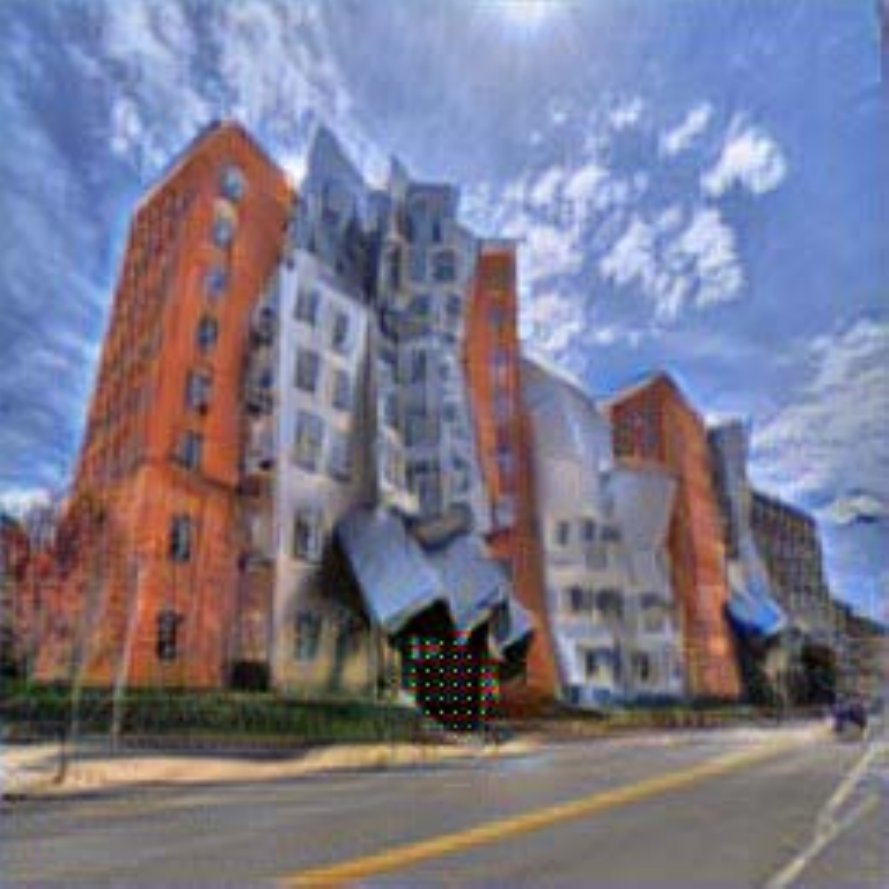}
\end{minipage}%
}%
\subfigure[relu4-4]{
\begin{minipage}[t]{0.2\linewidth}
\centering
\includegraphics[width=\linewidth]{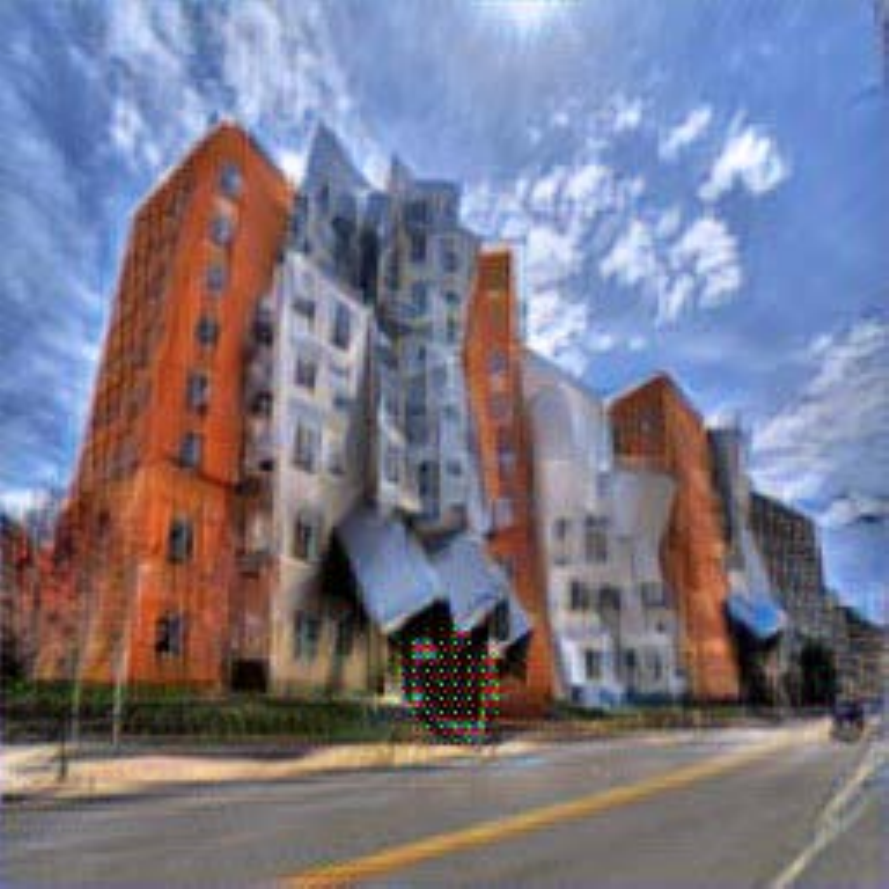}
\end{minipage}%
}%
\vfill

\subfigure[relu5-1]{
\begin{minipage}[t]{0.2\linewidth}
\centering
\includegraphics[width=\linewidth]{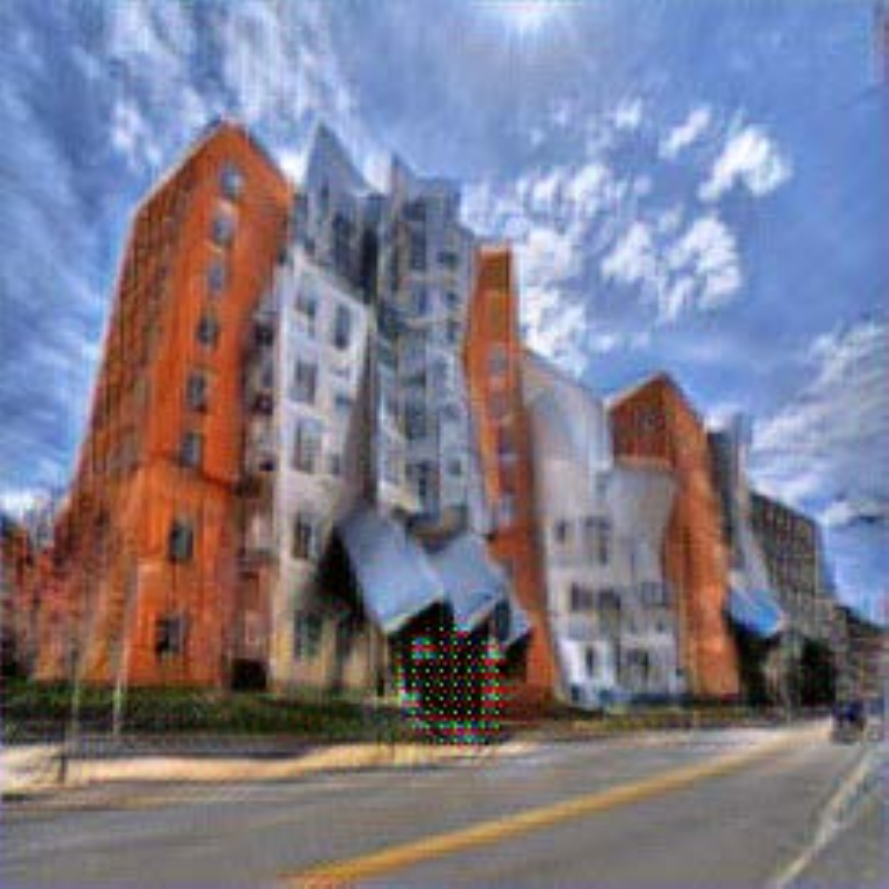}
\end{minipage}%
}%
\subfigure[relu5-2]{
\begin{minipage}[t]{0.2\linewidth}
\centering
\includegraphics[width=\linewidth]{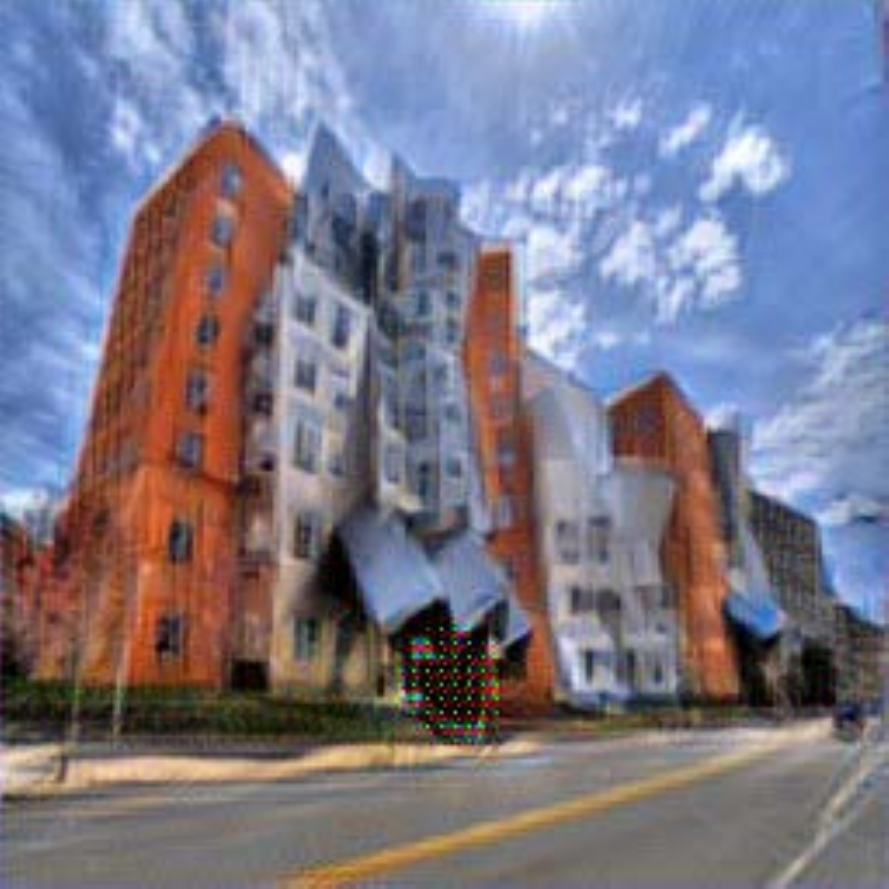}
\end{minipage}%
}%
\subfigure[relu5-3]{
\begin{minipage}[t]{0.2\linewidth}
\centering
\includegraphics[width=\linewidth]{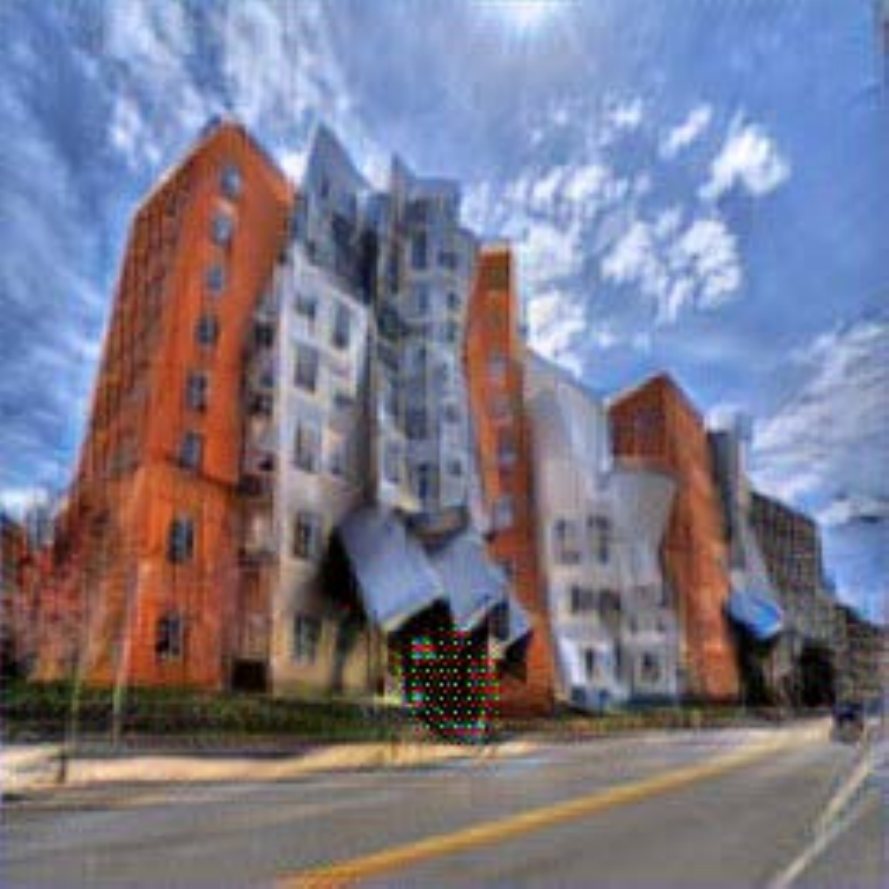}
\end{minipage}%
}%
\subfigure[relu5-4]{
\begin{minipage}[t]{0.2\linewidth}
\centering
\includegraphics[width=\linewidth]{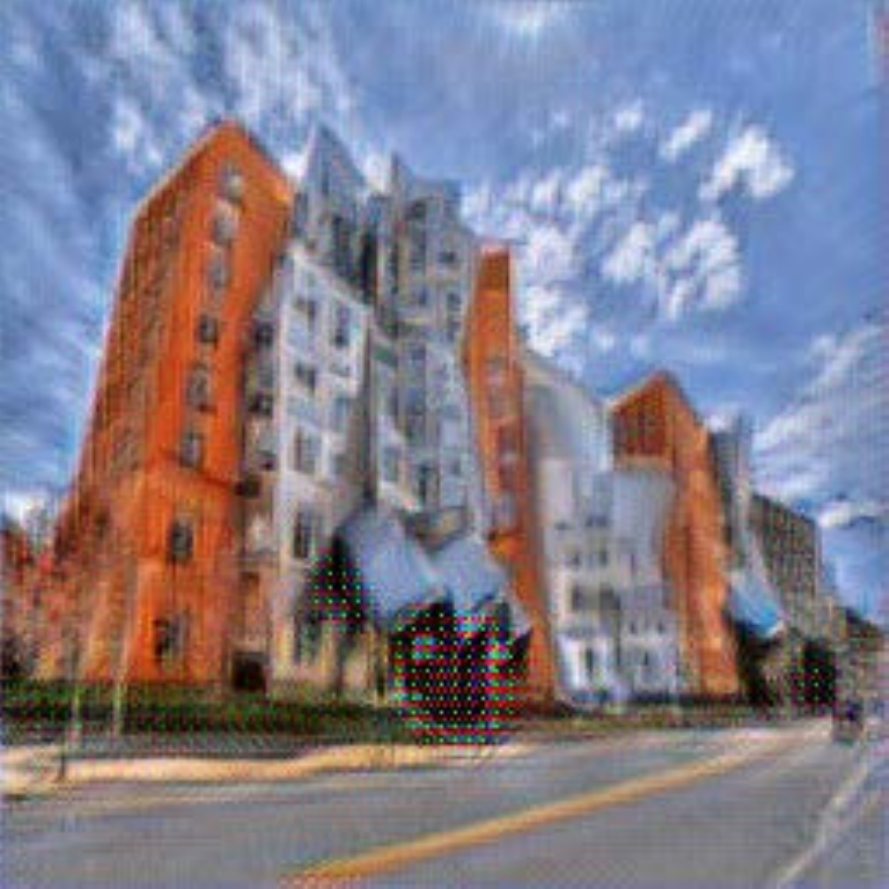}
\end{minipage}%
}%

\centering
\caption{Styled images using \cite{GatysNeuralStyle} with 500 epoches using every single activation layer from the pre-trained VGG19.}
\label{wave-layer}
\end{figure*}

\begin{figure*}[htb]
\centering
\subfigure[content]{
\begin{minipage}[t]{0.2\linewidth}
\centering
\includegraphics[width=\linewidth]{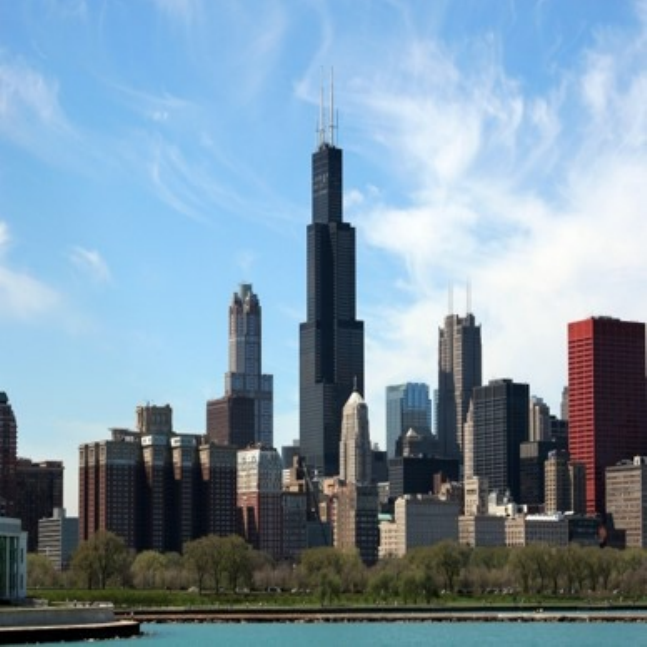}
\end{minipage}%
}%
\subfigure[style]{
\begin{minipage}[t]{0.2\linewidth}
\centering
\includegraphics[width=\linewidth]{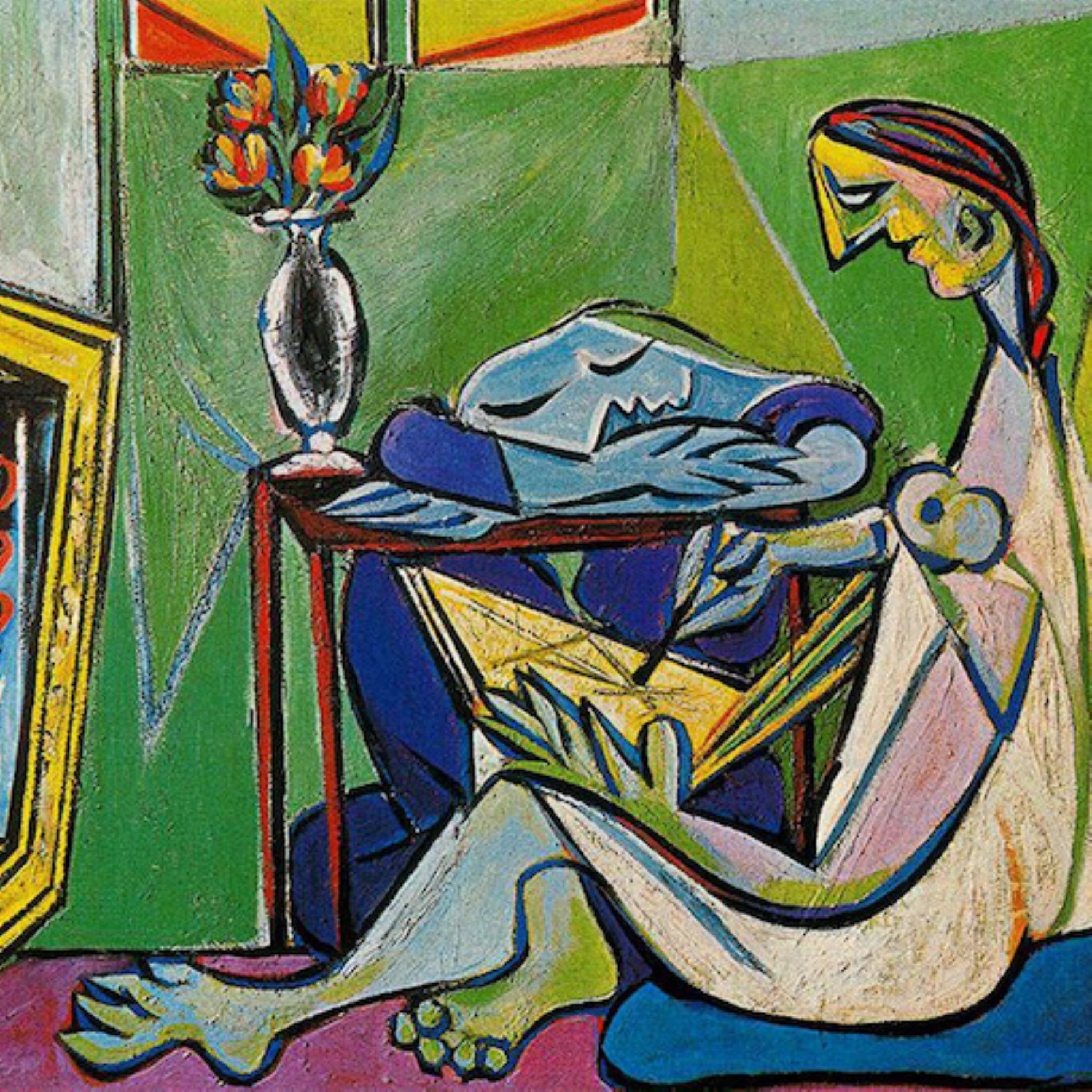}
\end{minipage}%
}%
\vfill

\subfigure[relu1-1]{
\begin{minipage}[t]{0.2\linewidth}
\centering
\includegraphics[width=\linewidth]{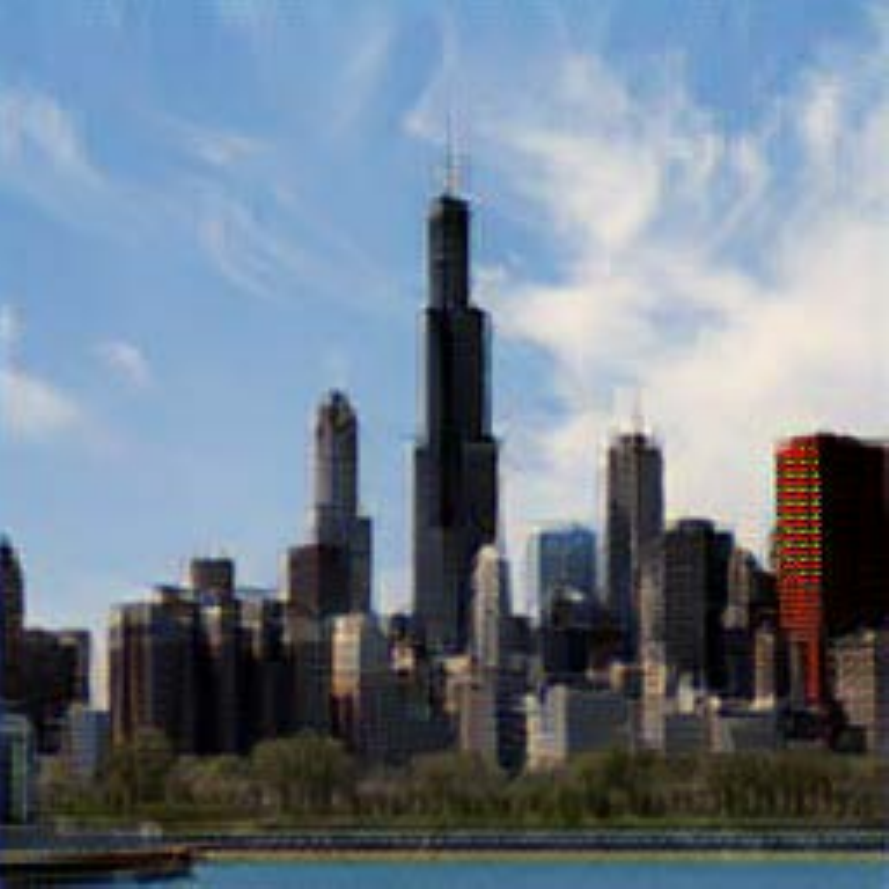}
\end{minipage}%
}%
\subfigure[relu1-2]{
\begin{minipage}[t]{0.2\linewidth}
\centering
\includegraphics[width=\linewidth]{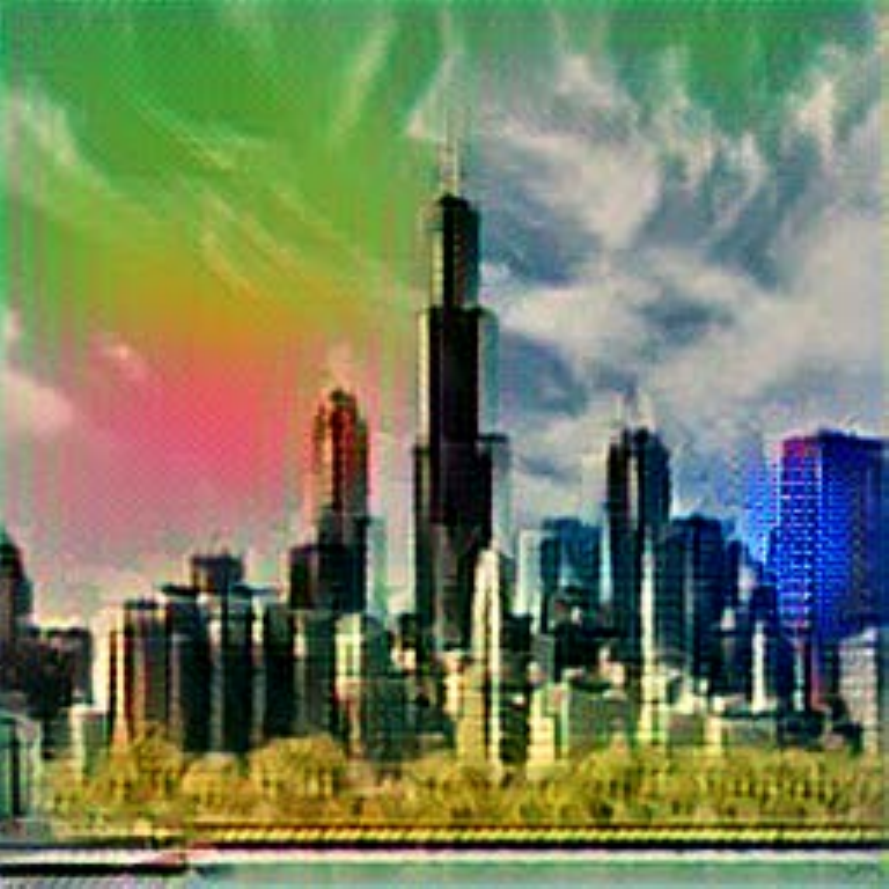}
\end{minipage}%
}%
\subfigure[relu2-1]{
\begin{minipage}[t]{0.2\linewidth}
\centering
\includegraphics[width=\linewidth]{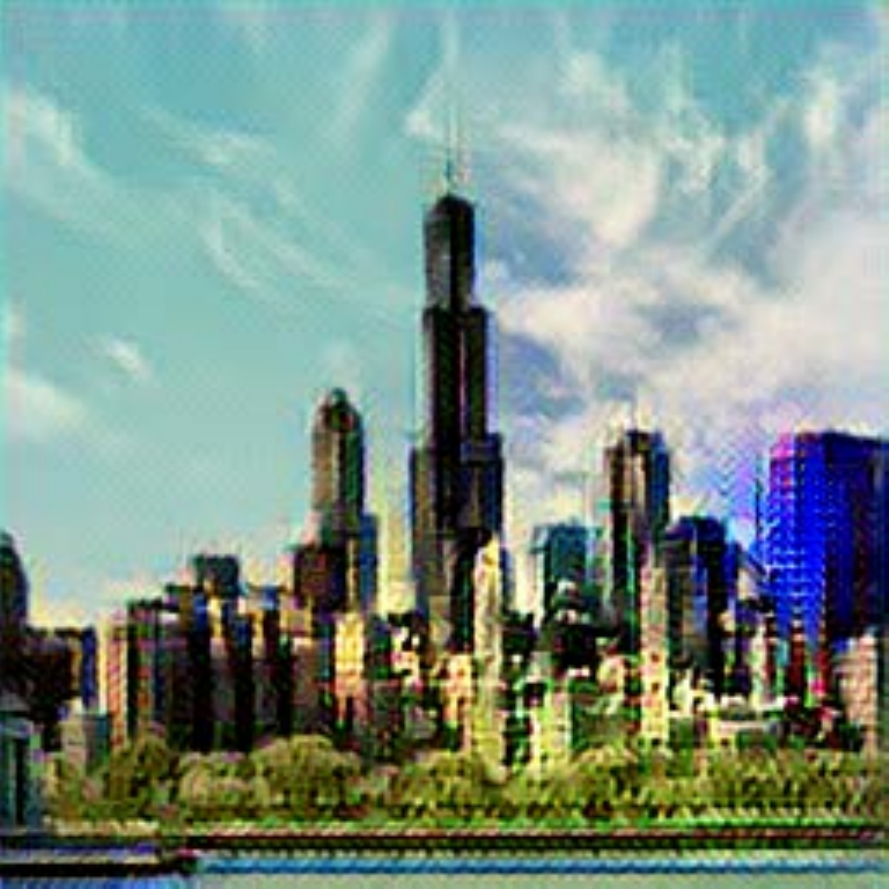}
\end{minipage}%
}%
\subfigure[relu2-2]{
\begin{minipage}[t]{0.2\linewidth}
\centering
\includegraphics[width=\linewidth]{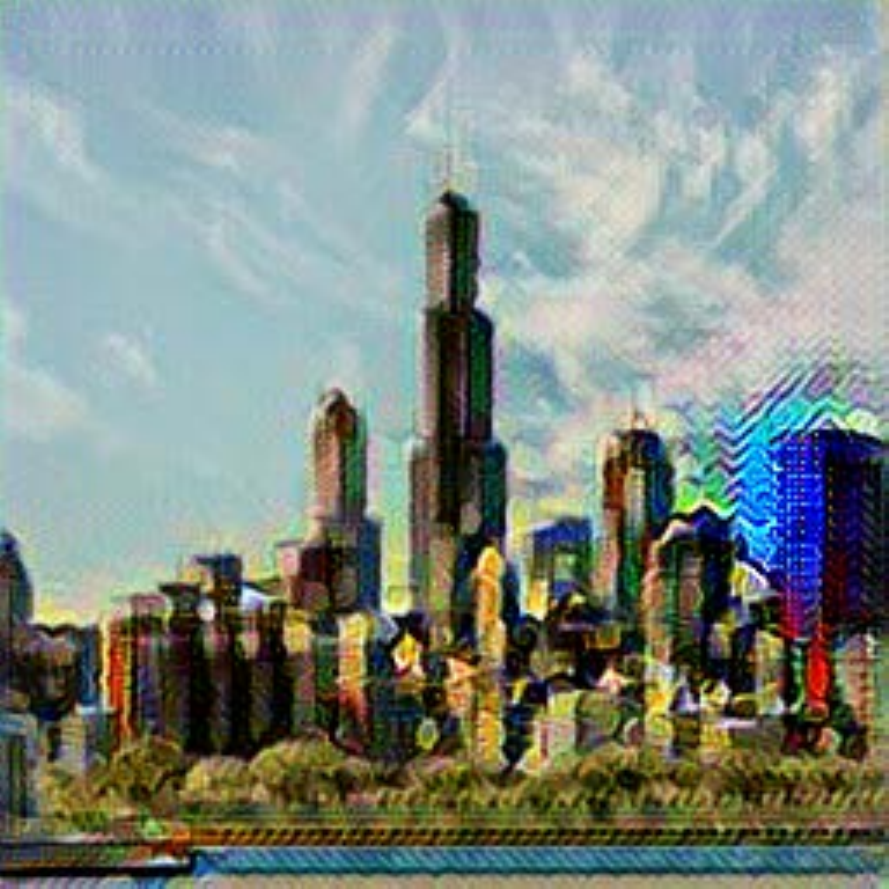}
\end{minipage}%
}%
\vfill

\subfigure[relu3-1]{
\begin{minipage}[t]{0.2\linewidth}
\centering
\includegraphics[width=\linewidth]{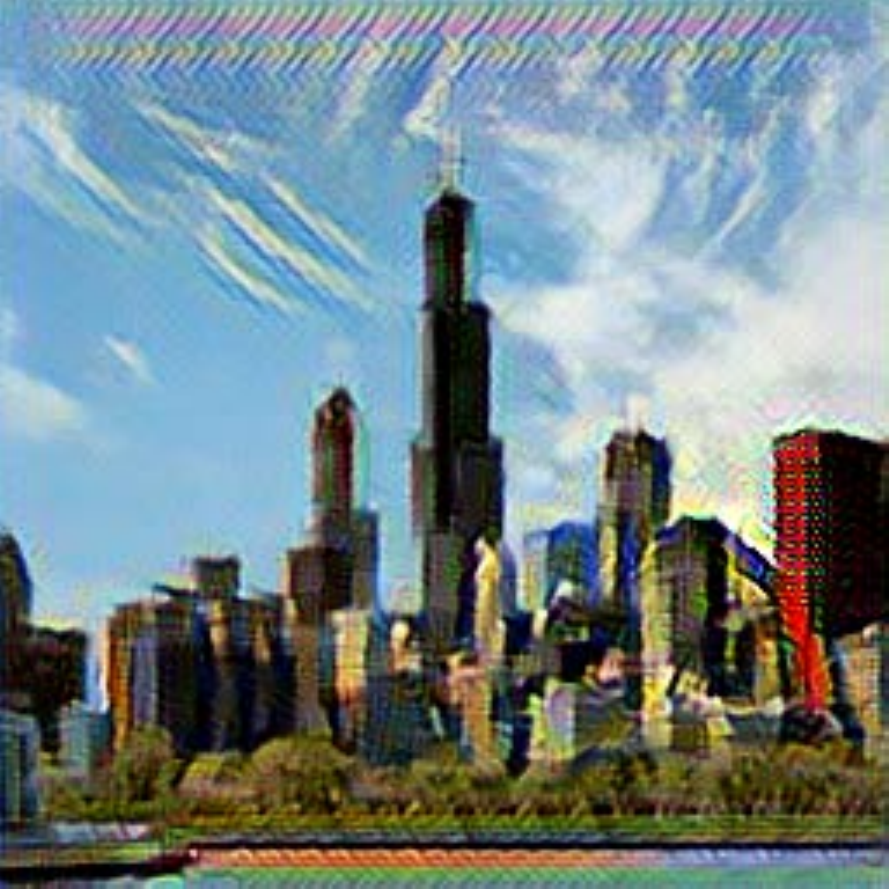}
\end{minipage}%
}%
\subfigure[relu3-2]{
\begin{minipage}[t]{0.2\linewidth}
\centering
\includegraphics[width=\linewidth]{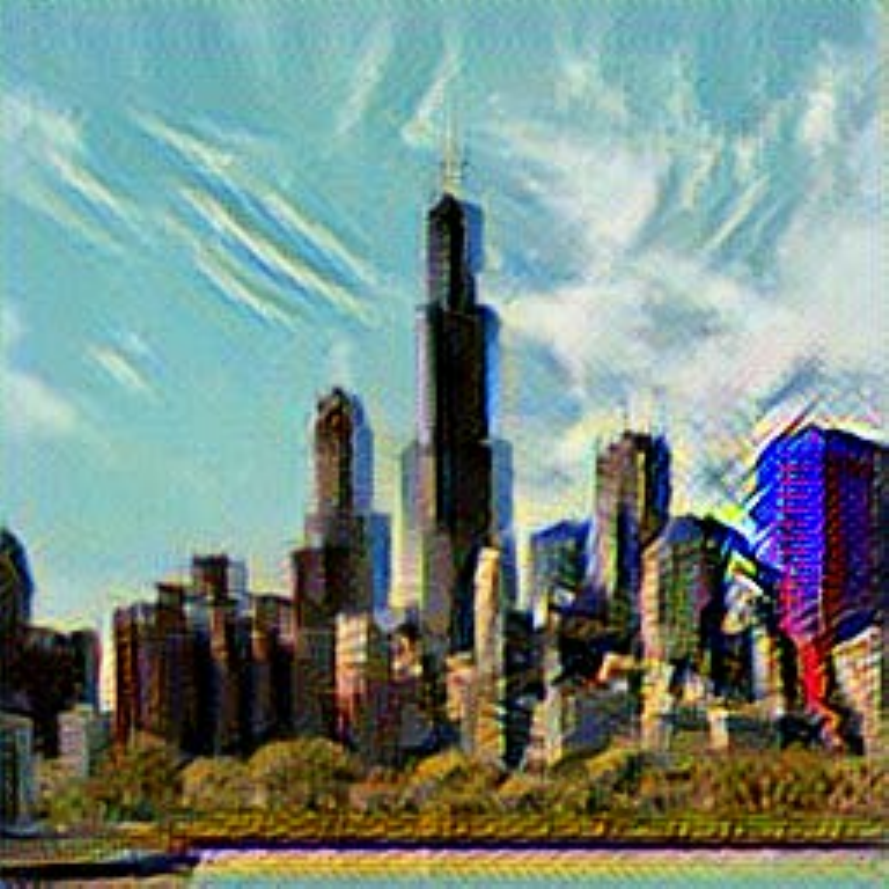}
\end{minipage}%
}%
\subfigure[relu3-3]{
\begin{minipage}[t]{0.2\linewidth}
\centering
\includegraphics[width=\linewidth]{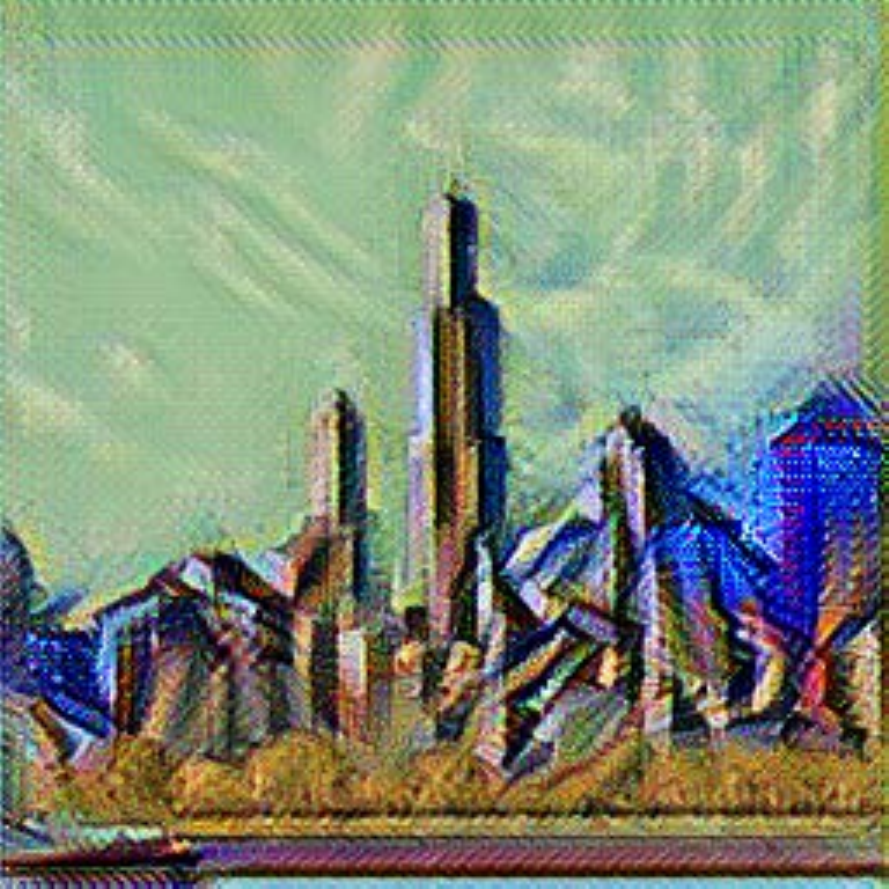}
\end{minipage}%
}%
\subfigure[relu3-4]{
\begin{minipage}[t]{0.2\linewidth}
\centering
\includegraphics[width=\linewidth]{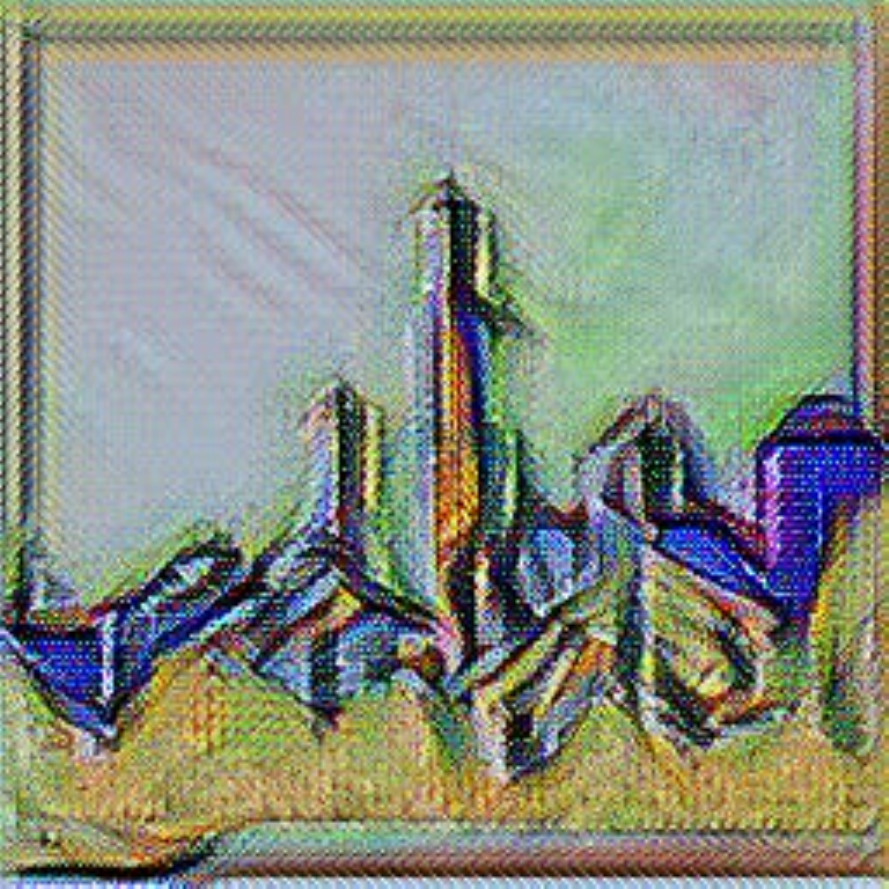}
\end{minipage}%
}%
\vfill

\subfigure[relu4-1]{
\begin{minipage}[t]{0.2\linewidth}
\centering
\includegraphics[width=\linewidth]{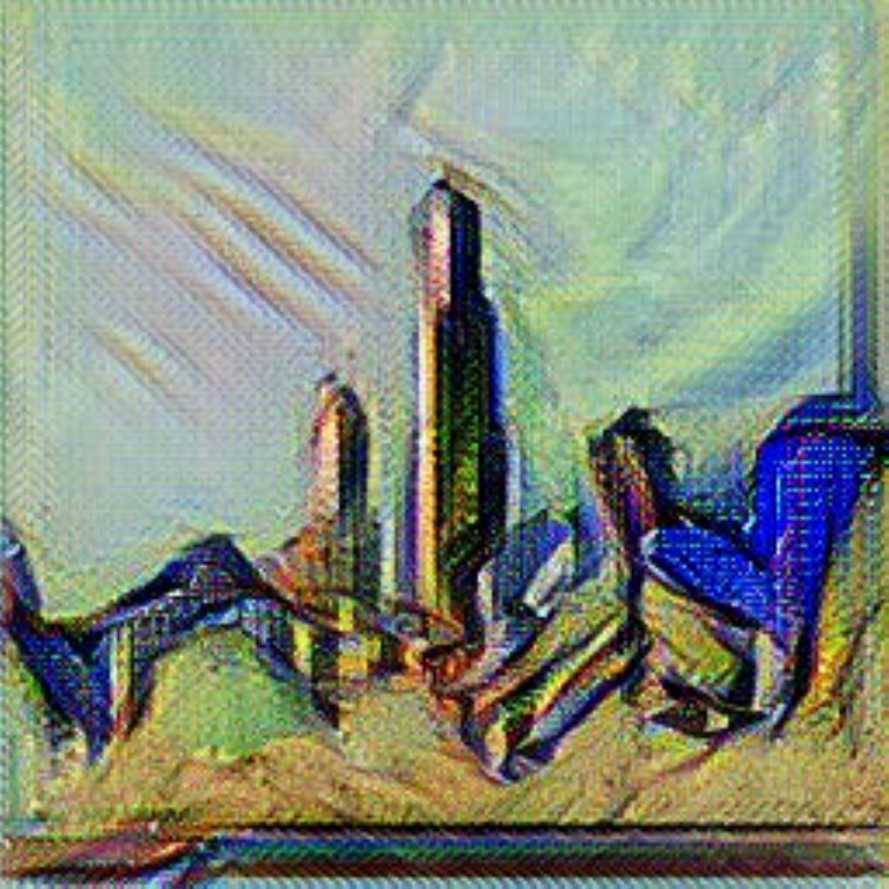}
\end{minipage}%
}%
\subfigure[relu4-2]{
\begin{minipage}[t]{0.2\linewidth}
\centering
\includegraphics[width=\linewidth]{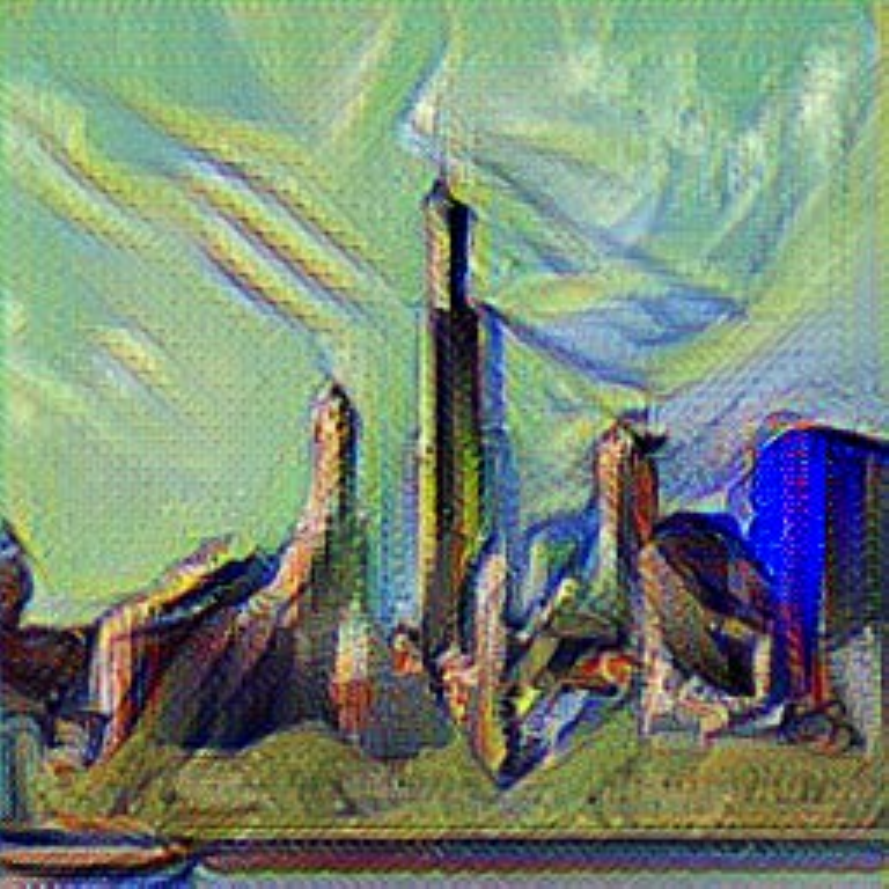}
\end{minipage}%
}%
\subfigure[relu4-3]{
\begin{minipage}[t]{0.2\linewidth}
\centering
\includegraphics[width=\linewidth]{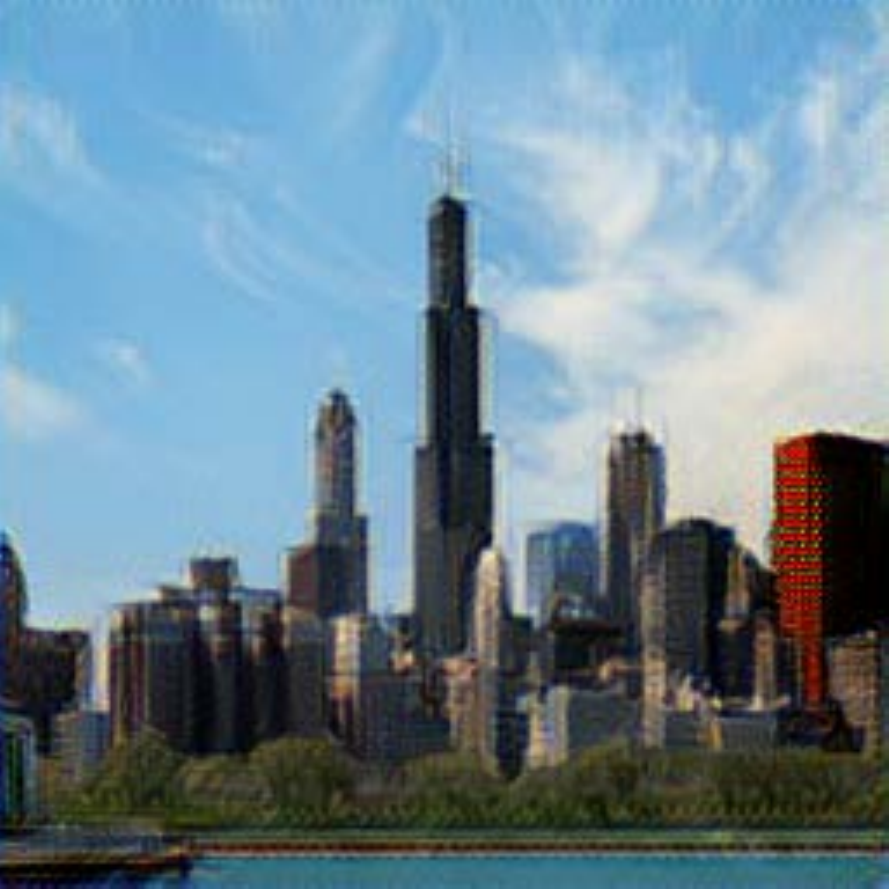}
\end{minipage}%
}%
\subfigure[relu4-4]{
\begin{minipage}[t]{0.2\linewidth}
\centering
\includegraphics[width=\linewidth]{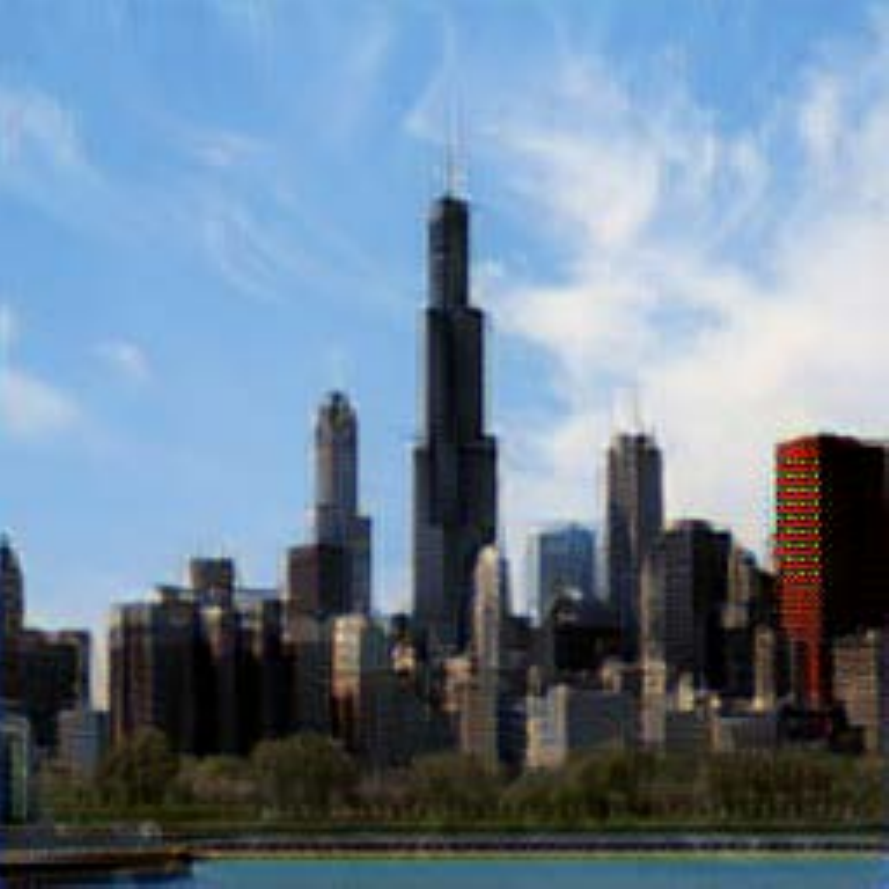}
\end{minipage}%
}%
\vfill

\subfigure[relu5-1]{
\begin{minipage}[t]{0.2\linewidth}
\centering
\includegraphics[width=\linewidth]{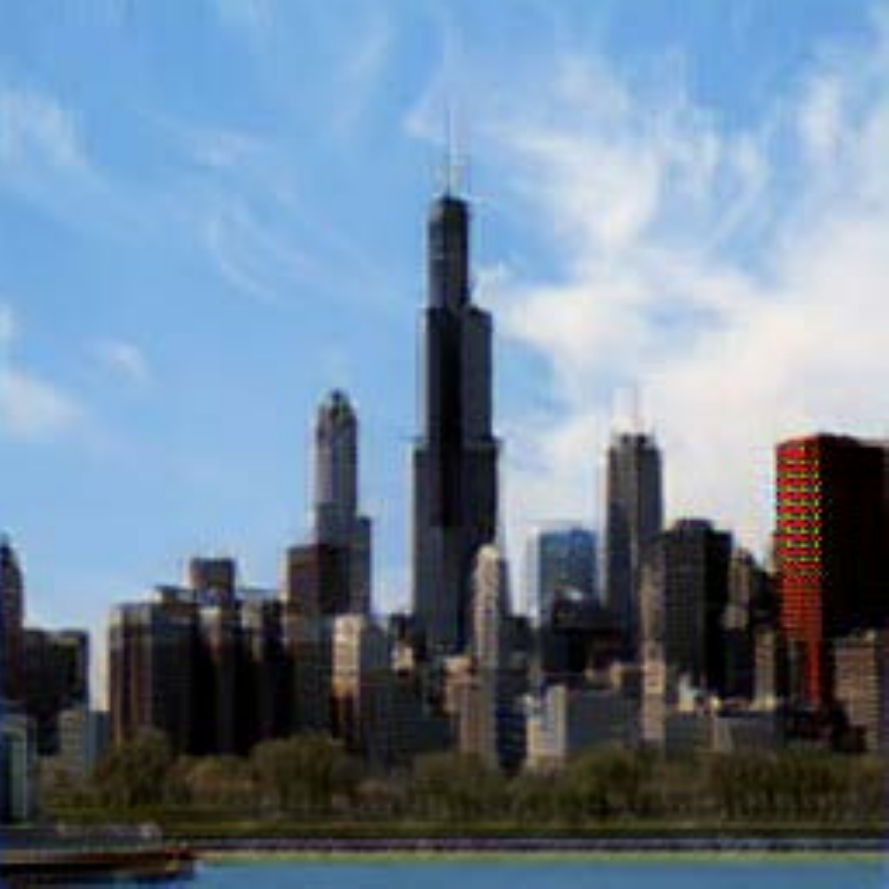}
\end{minipage}%
}%
\subfigure[relu5-2]{
\begin{minipage}[t]{0.2\linewidth}
\centering
\includegraphics[width=\linewidth]{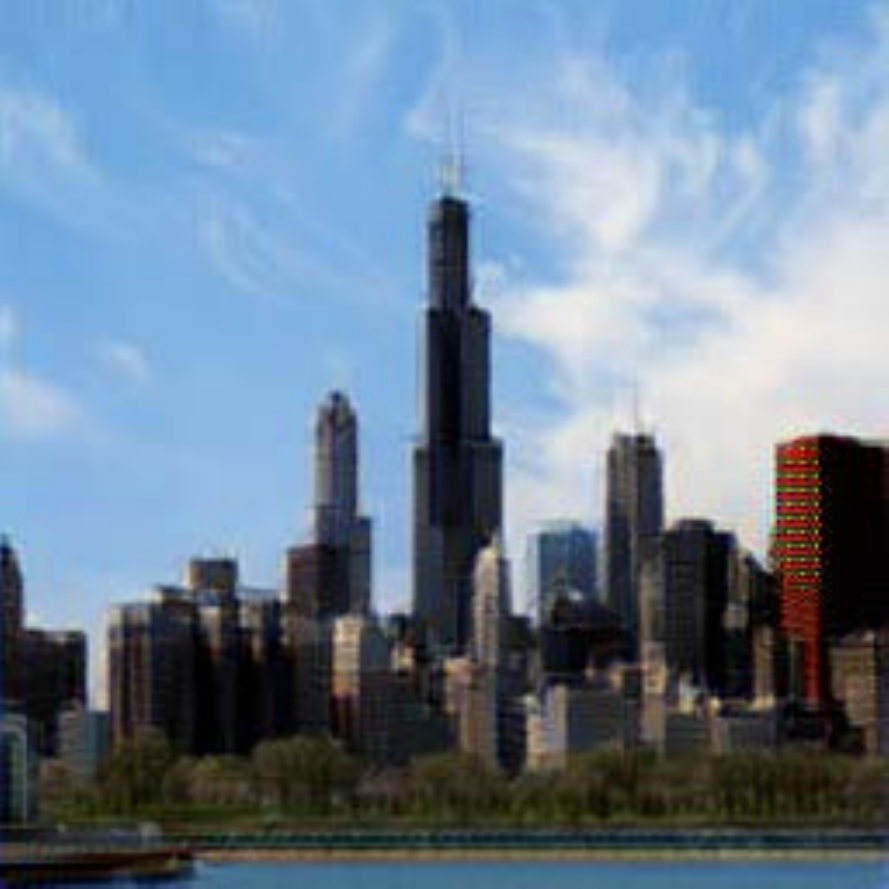}
\end{minipage}%
}%
\subfigure[relu5-3]{
\begin{minipage}[t]{0.2\linewidth}
\centering
\includegraphics[width=\linewidth]{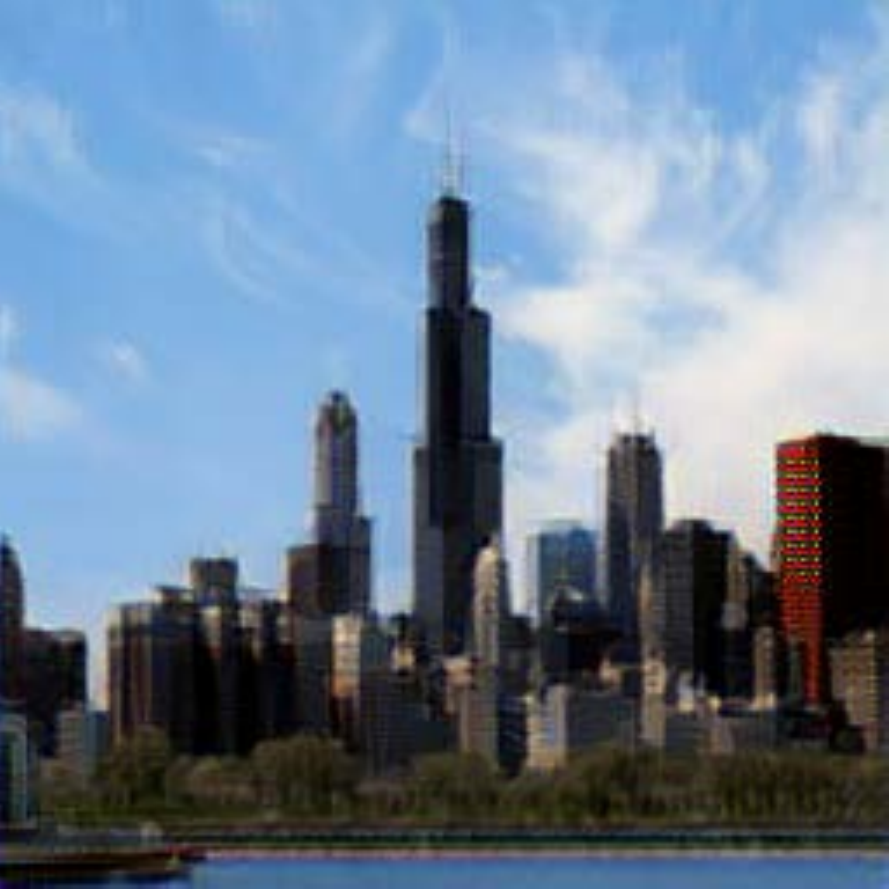}
\end{minipage}%
}%
\subfigure[relu5-4]{
\begin{minipage}[t]{0.2\linewidth}
\centering
\includegraphics[width=\linewidth]{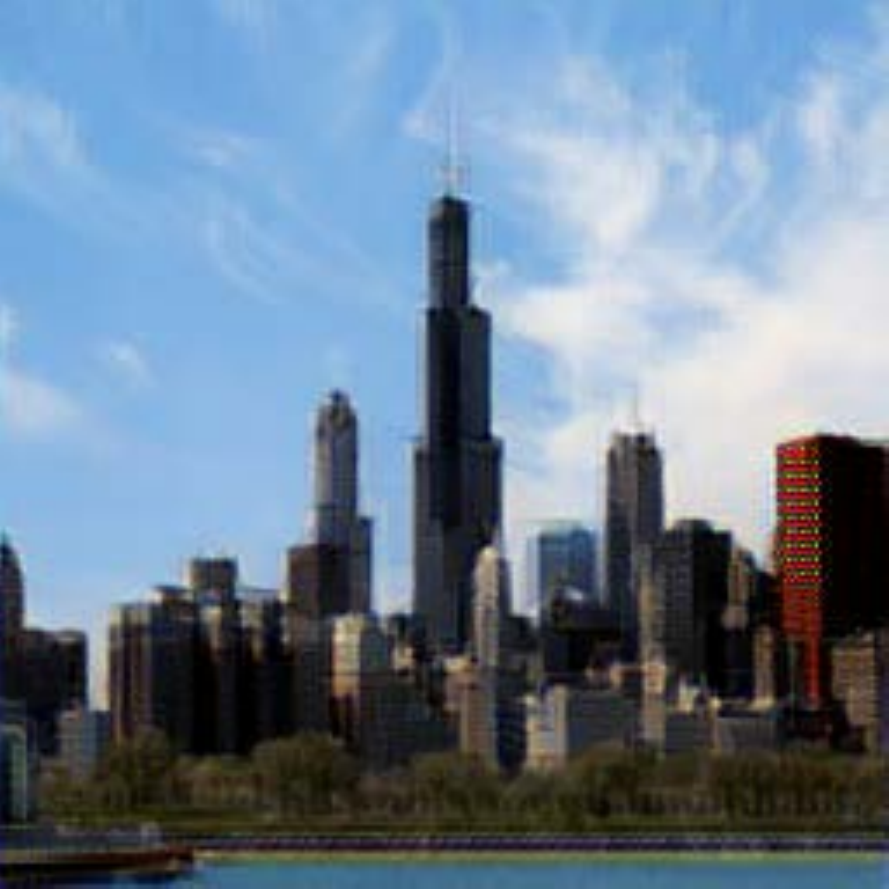}
\end{minipage}%
}%
\centering
\caption{Styled images using \cite{GatysNeuralStyle} with 500 epoches using every single activation layer from the pre-trained VGG19.}
\label{lamuse-layer}
\end{figure*}

Since different layers in VGG19 \cite{VGG} represent different abstract level, we experiment the stylization effect of every activation layer on a couple of different content and style images as are shown in Figure \ref{wave-layer} and Figure \ref{lamuse-layer}. From our experiment, it can be noticed that not every layer is effective in style transfer and among those that work, shallow layers only transfer the coarse scale of style (color) while deep layers can transfer both the coarse scale (color) and detailed scale (stroke) of style, which conforms to the result of scale control in \cite{ControlFactor}. Since `relu4\underline{ }1' performs the best in style transfer after same amount of iterations, we determine to study the feature map of `relu4\underline{ }1' in our research.

\begin{figure*}[tb]
\centering
\subfigure[relu1-1]{
\begin{minipage}[t]{0.25\linewidth}
\centering
\includegraphics[width=\linewidth]{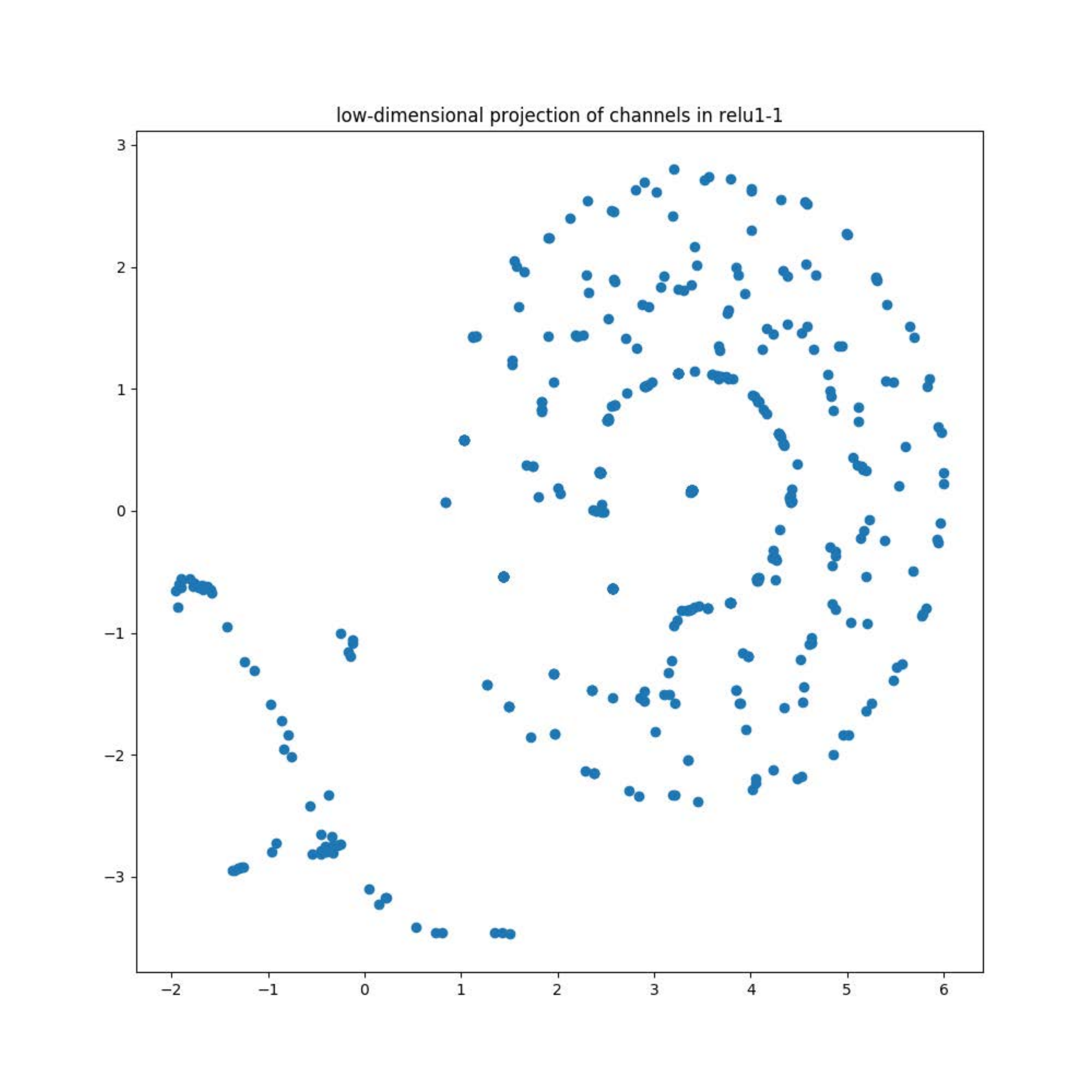}
\end{minipage}%
}%
\subfigure[relu1-2]{
\begin{minipage}[t]{0.25\linewidth}
\centering
\includegraphics[width=\linewidth]{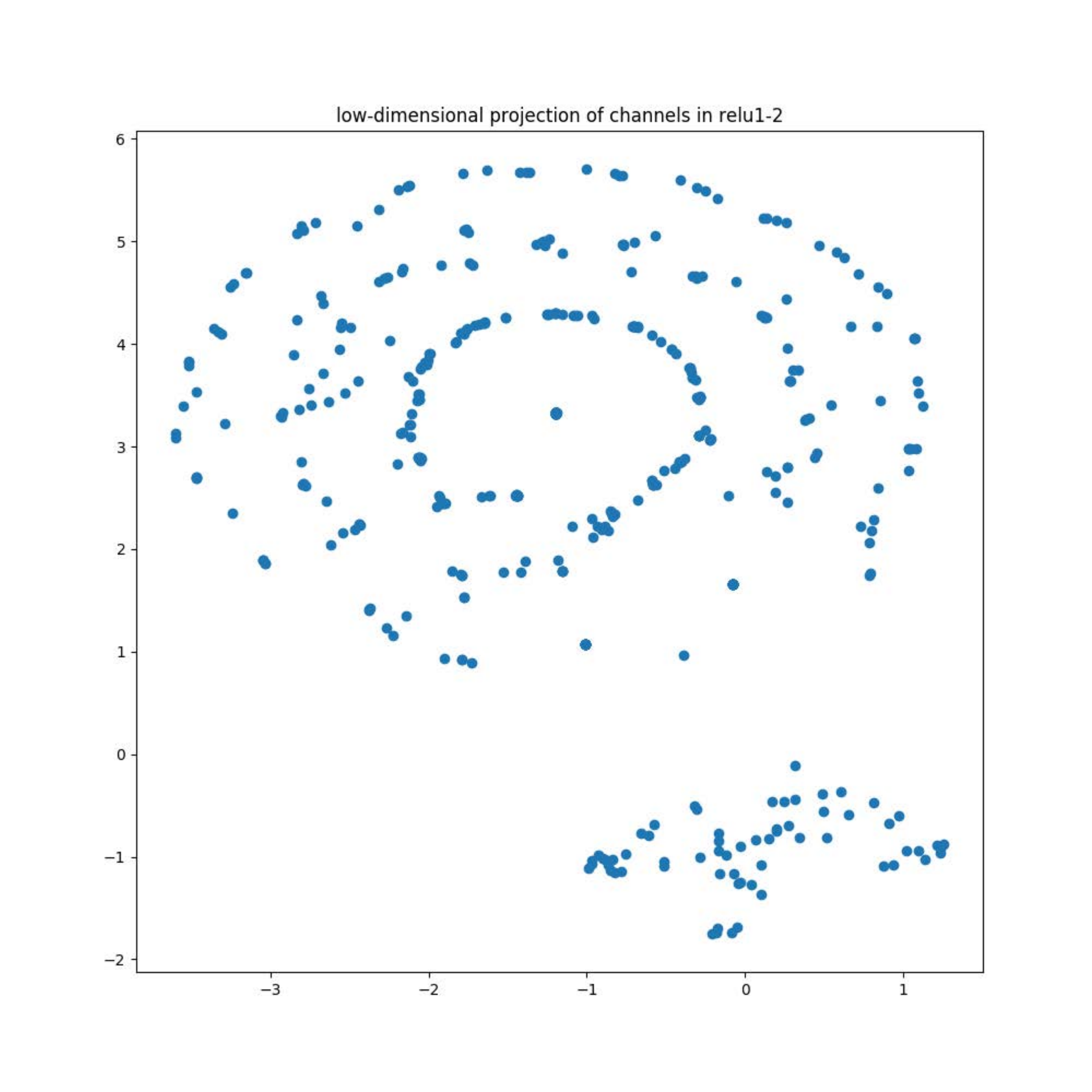}
\end{minipage}%
}%
\subfigure[relu2-1]{
\begin{minipage}[t]{0.25\linewidth}
\centering
\includegraphics[width=\linewidth]{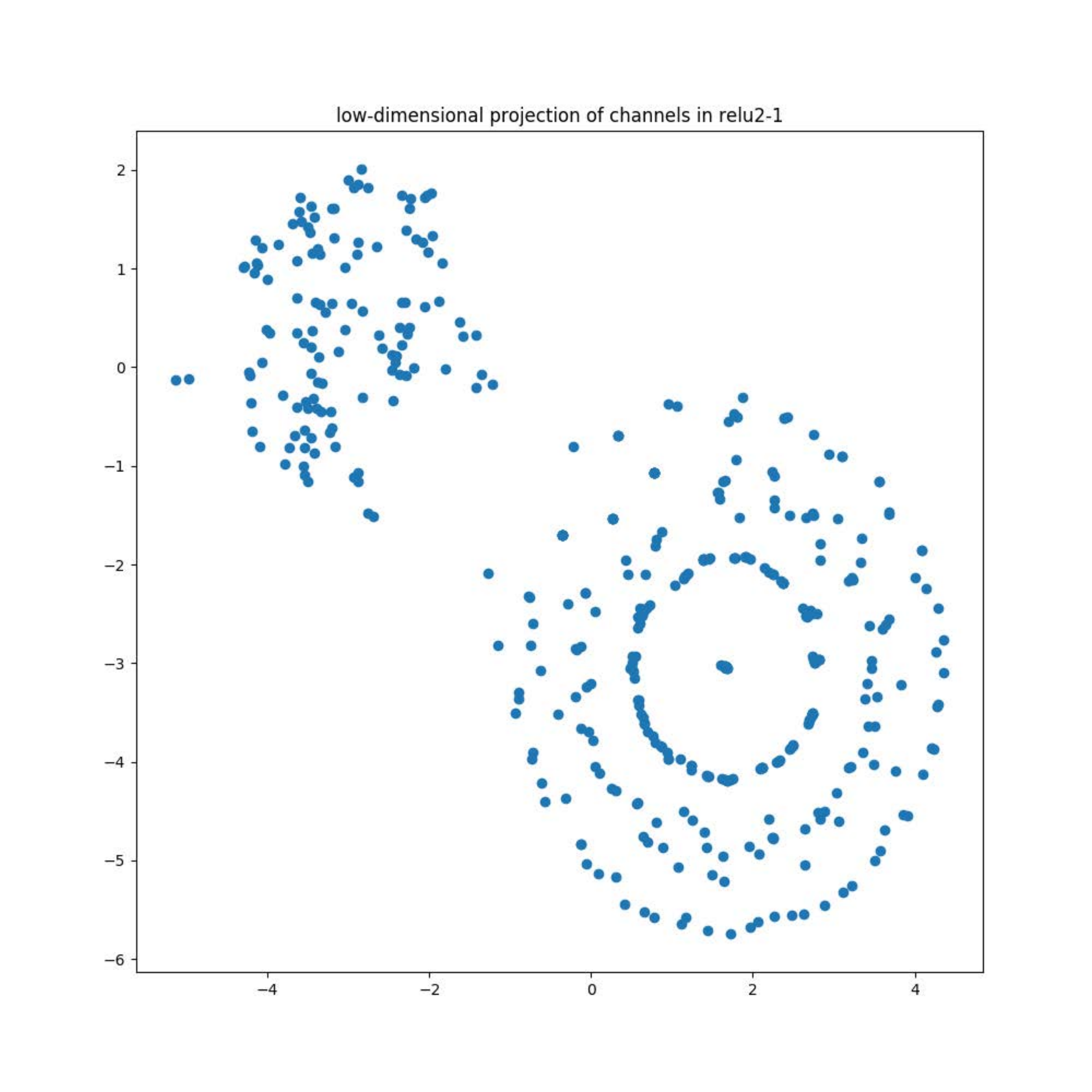}
\end{minipage}%
}%
\subfigure[relu2-2]{
\begin{minipage}[t]{0.25\linewidth}
\centering
\includegraphics[width=\linewidth]{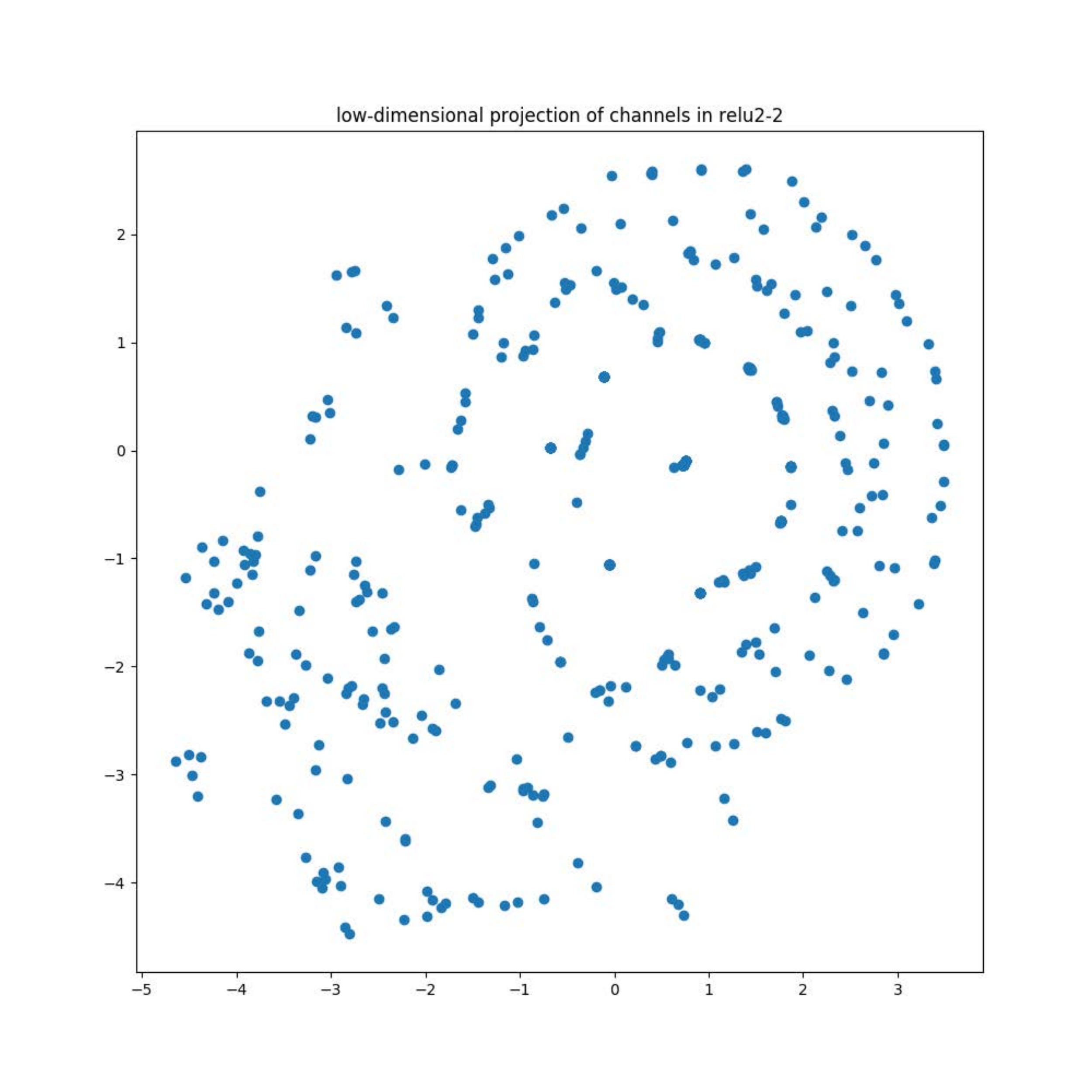}
\end{minipage}%
}%
\vfill

\subfigure[relu3-1]{
\begin{minipage}[t]{0.25\linewidth}
\centering
\includegraphics[width=\linewidth]{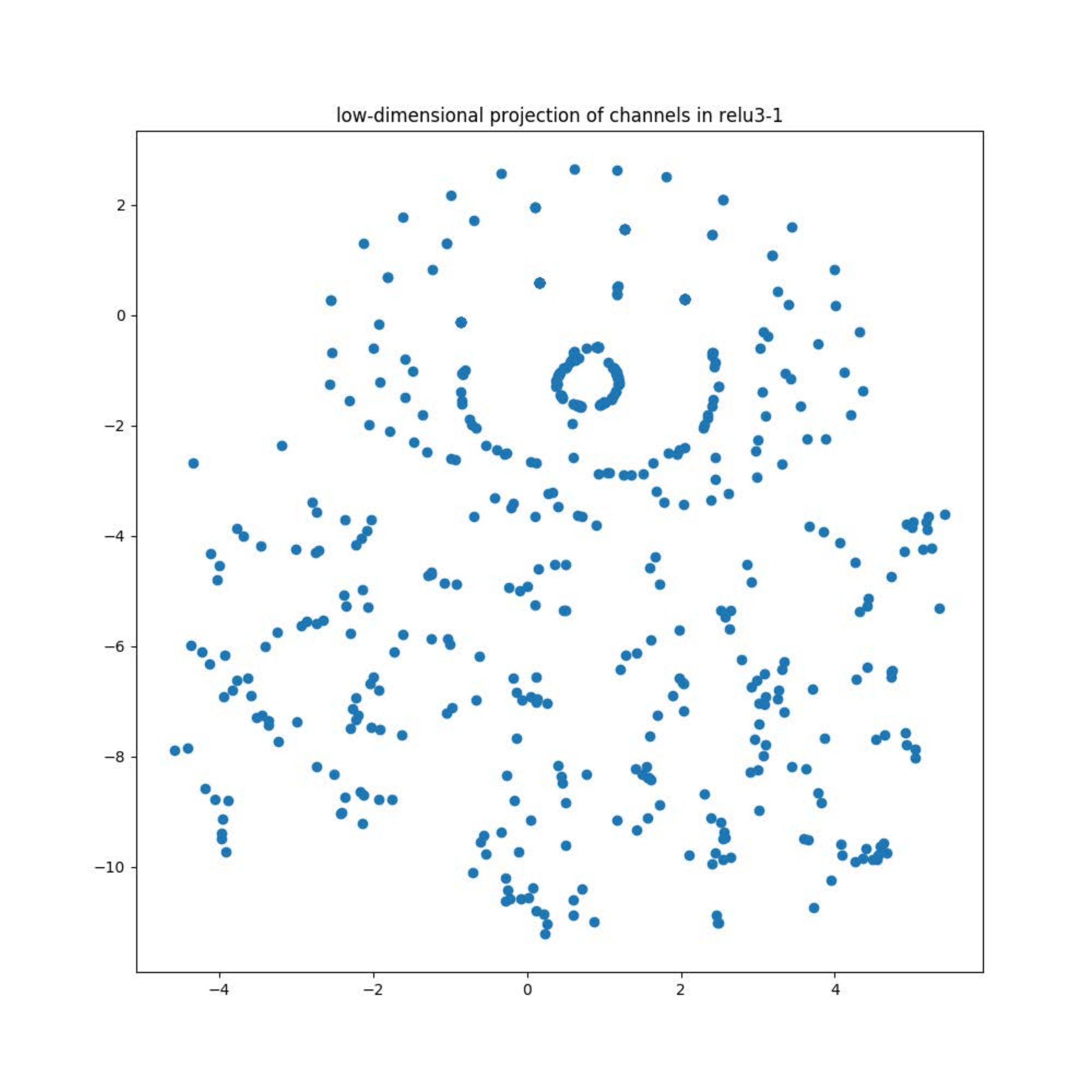}
\end{minipage}%
}%
\subfigure[relu3-2]{
\begin{minipage}[t]{0.25\linewidth}
\centering
\includegraphics[width=\linewidth]{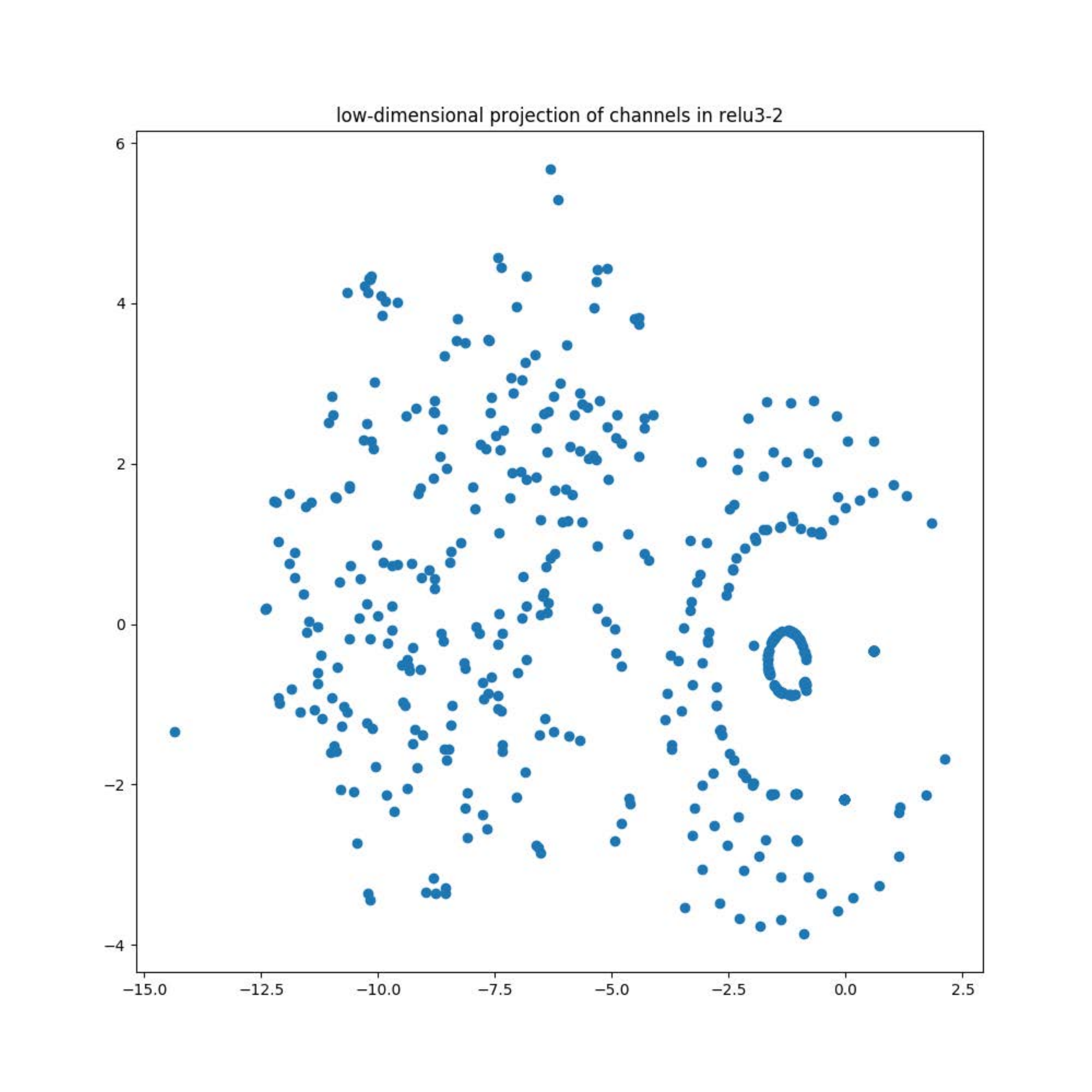}
\end{minipage}%
}%
\subfigure[relu3-3]{
\begin{minipage}[t]{0.25\linewidth}
\centering
\includegraphics[width=\linewidth]{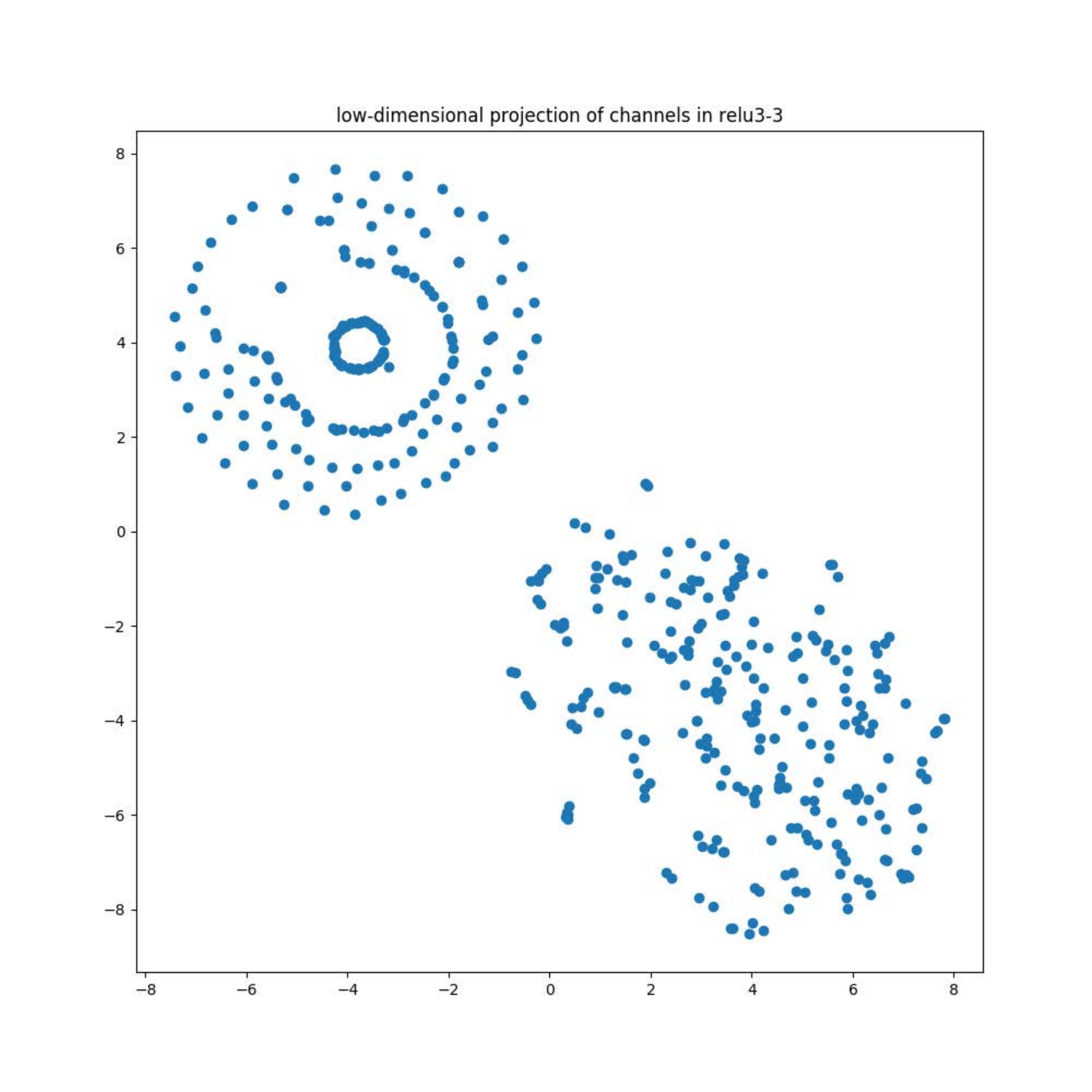}
\end{minipage}%
}%
\subfigure[relu3-4]{
\begin{minipage}[t]{0.25\linewidth}
\centering
\includegraphics[width=\linewidth]{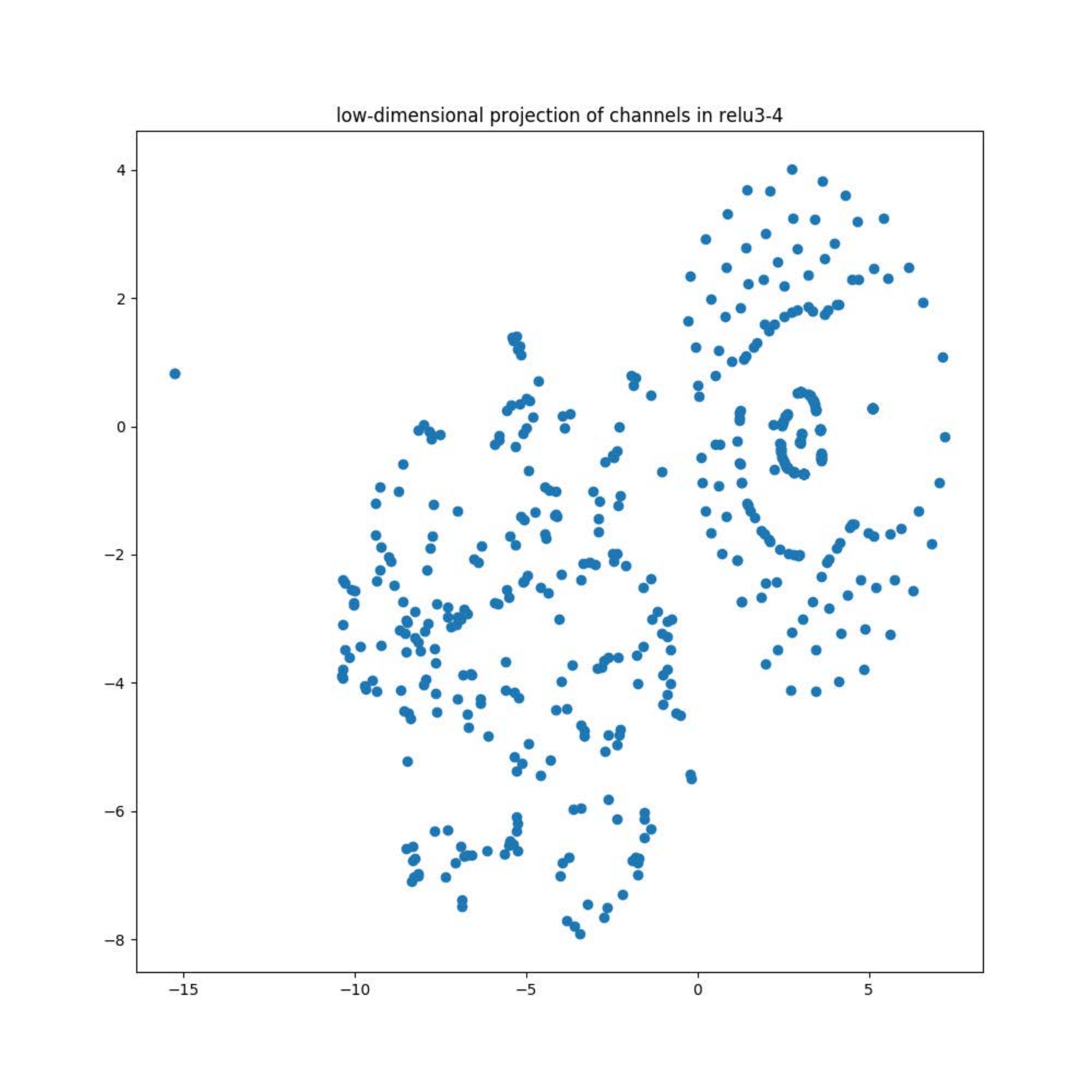}
\end{minipage}%
}%
\vfill

\subfigure[relu4-1]{
\begin{minipage}[t]{0.25\linewidth}
\centering
\includegraphics[width=\linewidth]{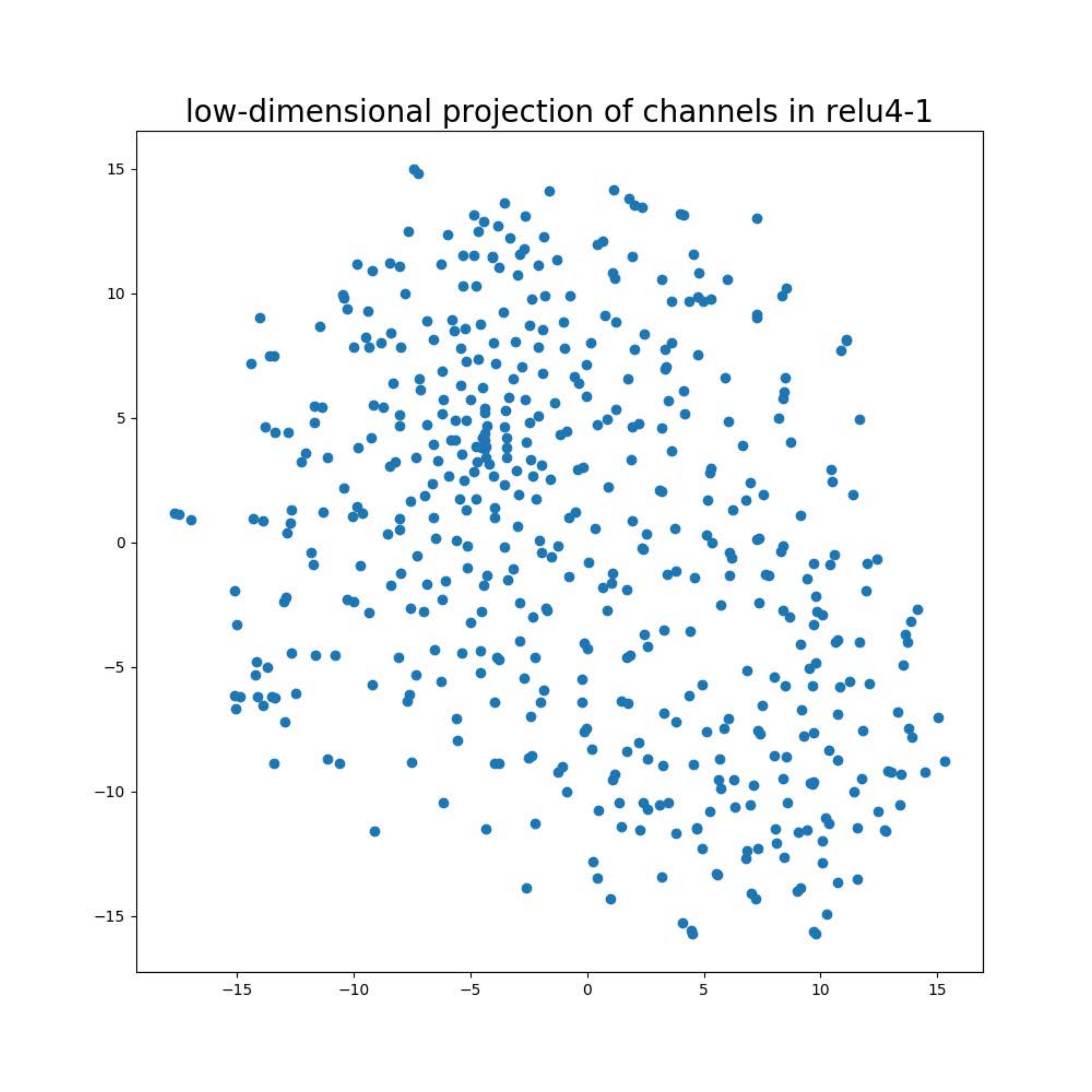}
\end{minipage}%
}%
\subfigure[relu4-2]{
\begin{minipage}[t]{0.25\linewidth}
\centering
\includegraphics[width=\linewidth]{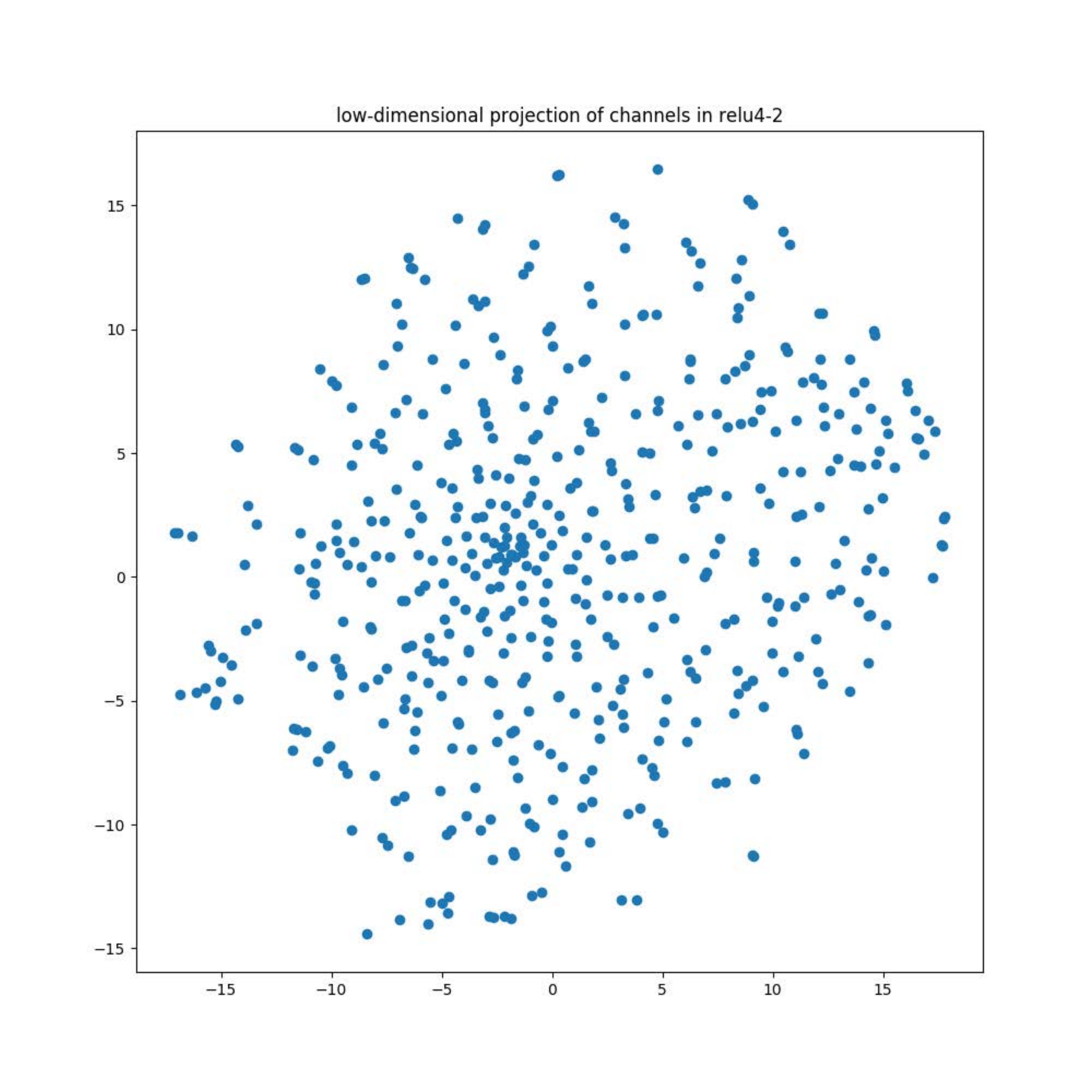}
\end{minipage}%
}%
\subfigure[relu4-3]{
\begin{minipage}[t]{0.25\linewidth}
\centering
\includegraphics[width=\linewidth]{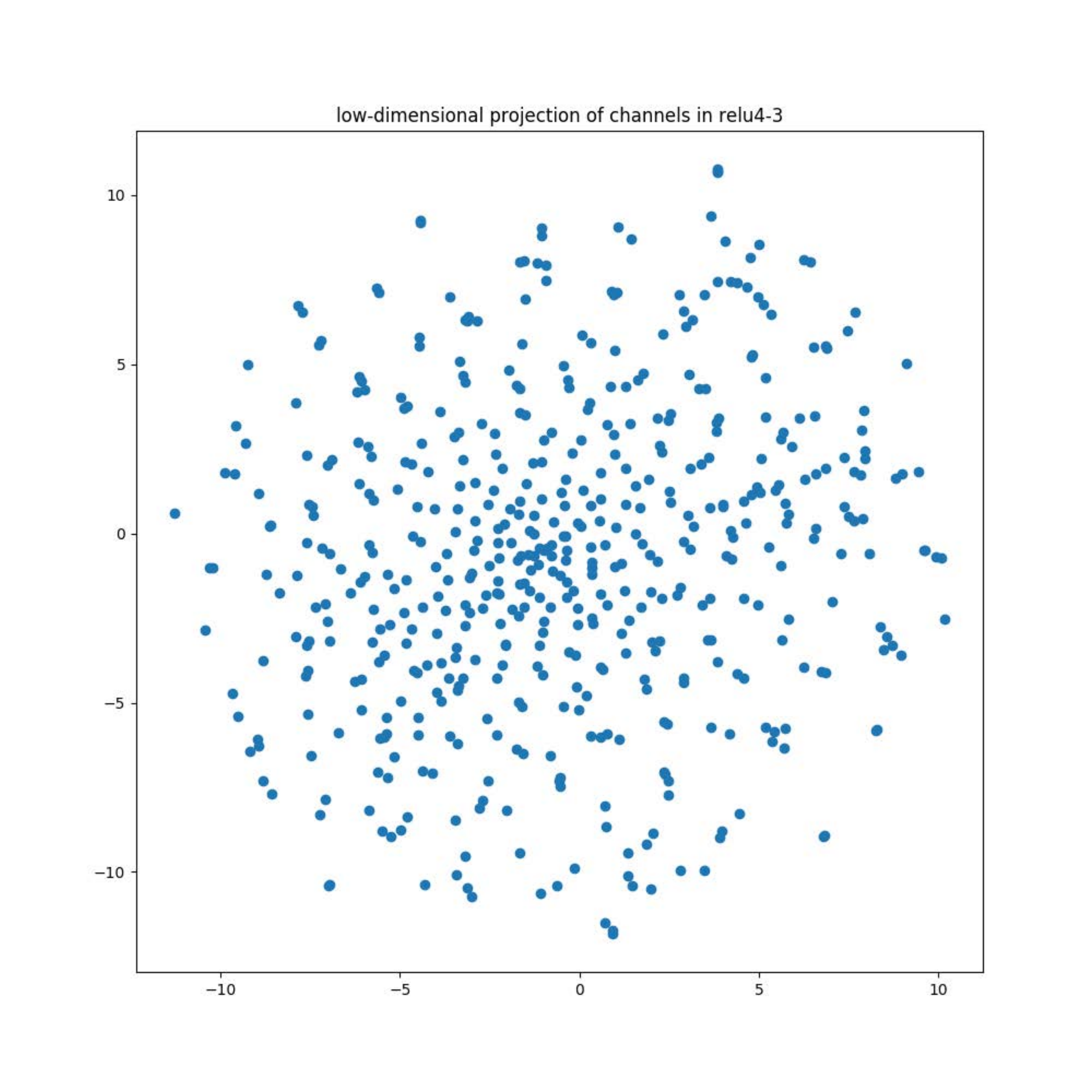}
\end{minipage}%
}%
\subfigure[relu4-4]{
\begin{minipage}[t]{0.25\linewidth}
\centering
\includegraphics[width=\linewidth]{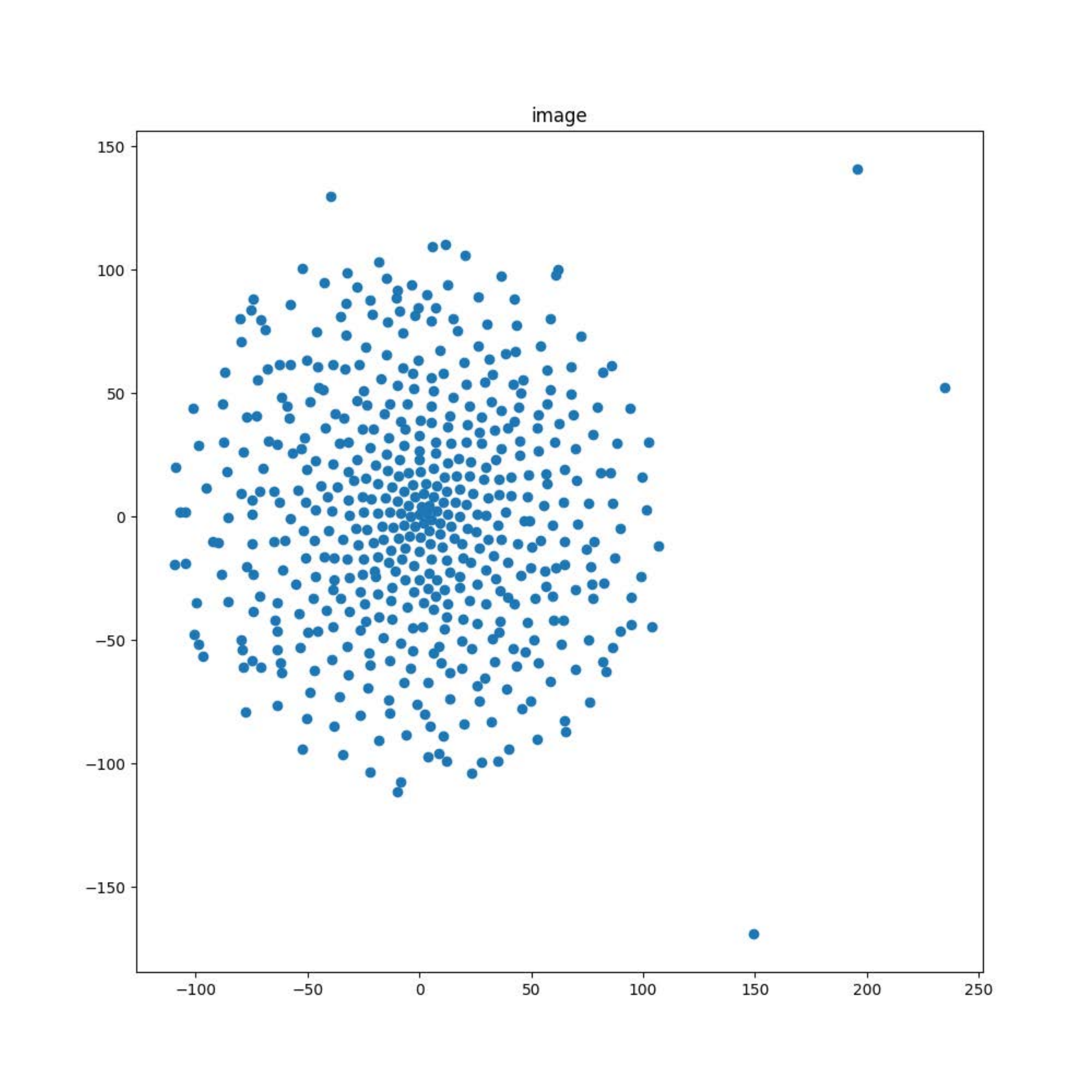}
\end{minipage}%
}%
\vfill

\subfigure[relu5-1]{
\begin{minipage}[t]{0.25\linewidth}
\centering
\includegraphics[width=\linewidth]{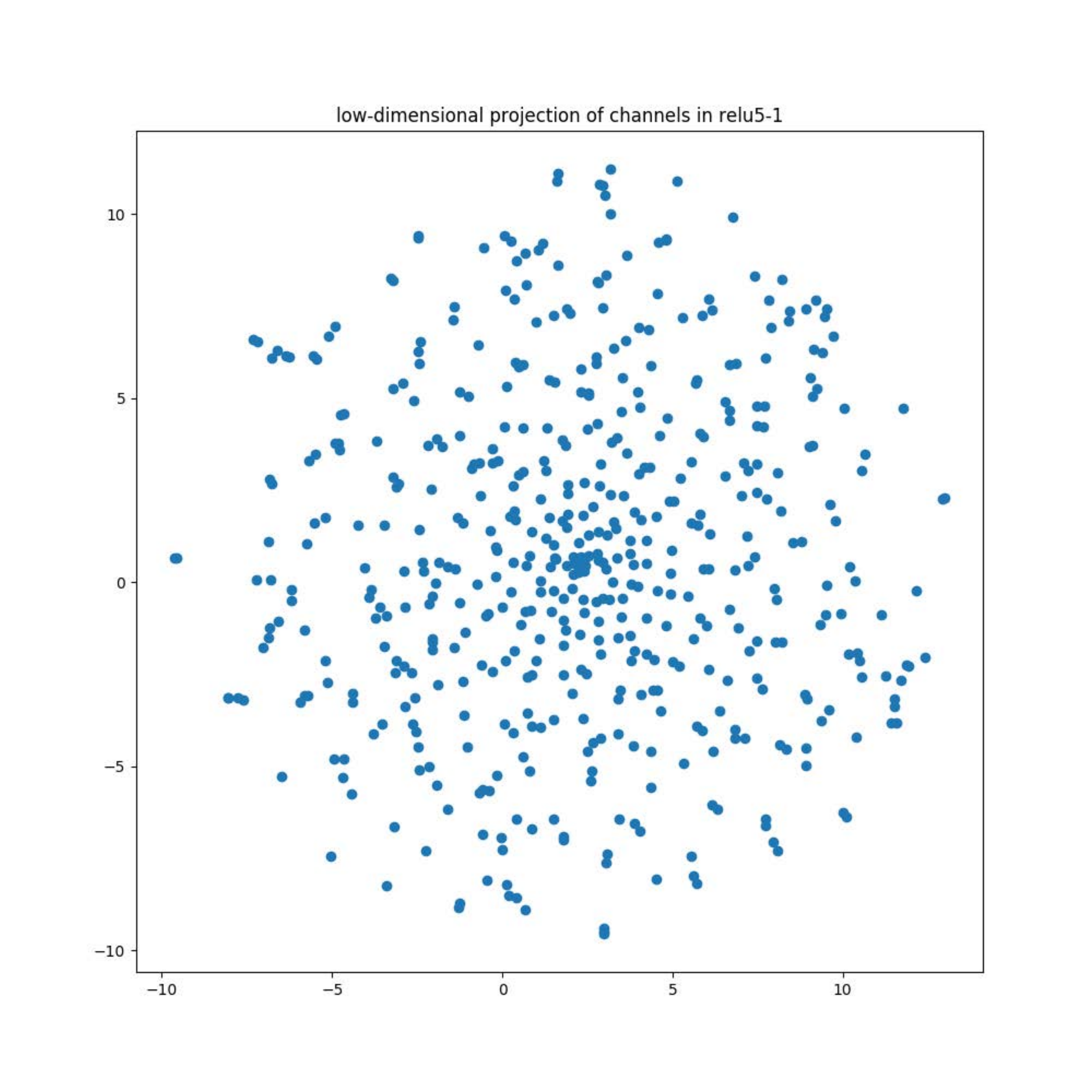}
\end{minipage}%
}%
\subfigure[relu5-2]{
\begin{minipage}[t]{0.25\linewidth}
\centering
\includegraphics[width=\linewidth]{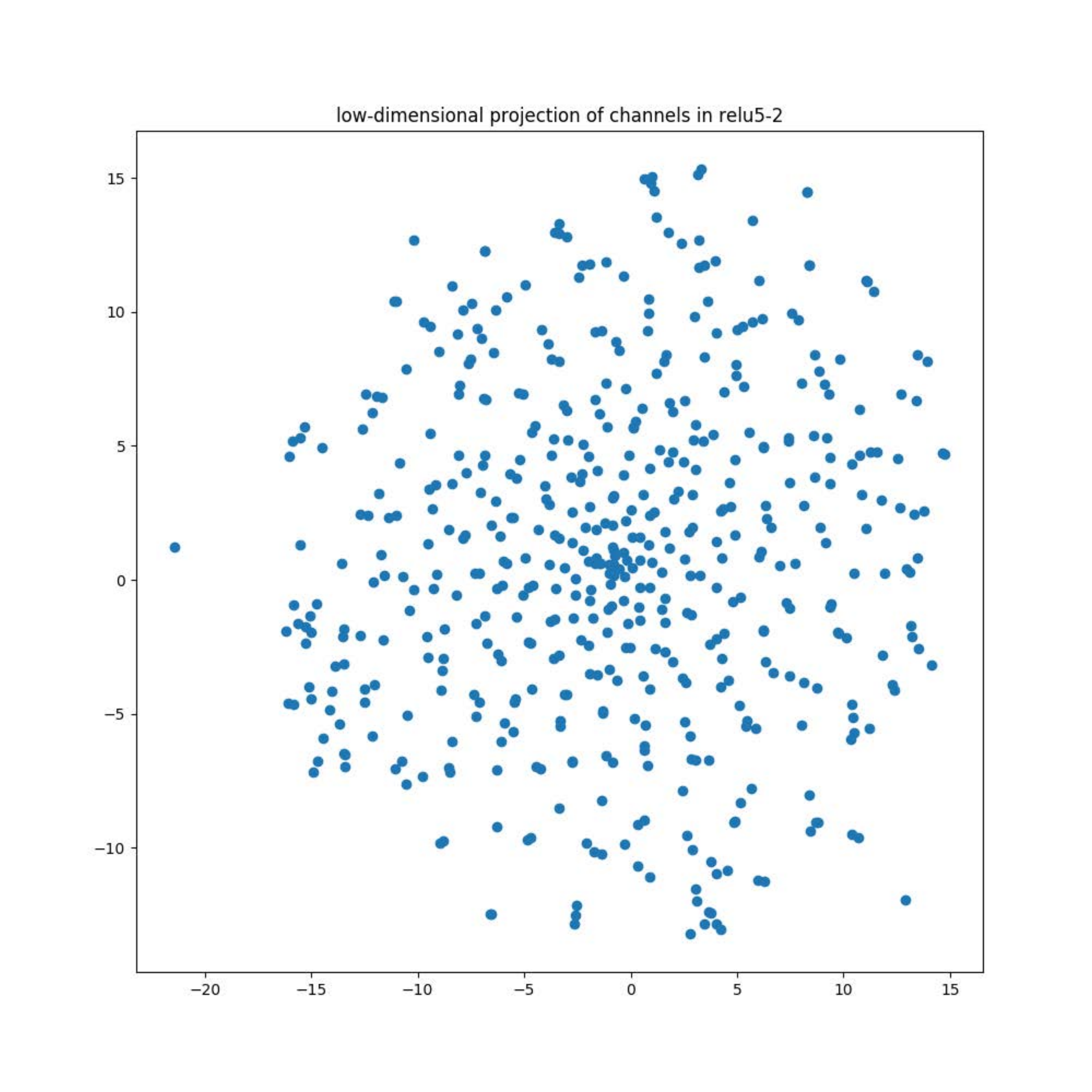}
\end{minipage}%
}%
\subfigure[relu5-3]{
\begin{minipage}[t]{0.25\linewidth}
\centering
\includegraphics[width=\linewidth]{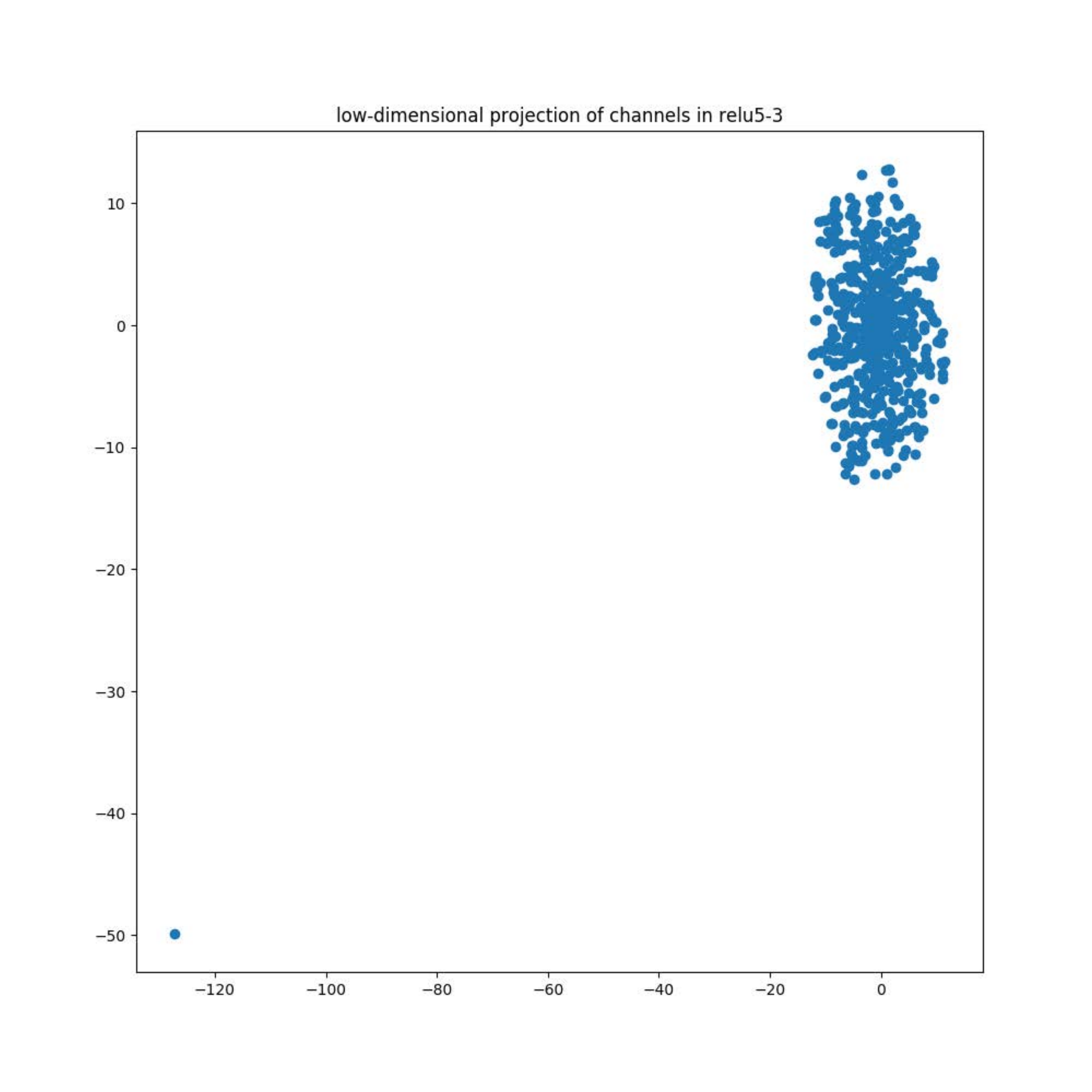}
\end{minipage}%
}%
\subfigure[relu5-4]{
\begin{minipage}[t]{0.25\linewidth}
\centering
\includegraphics[width=\linewidth]{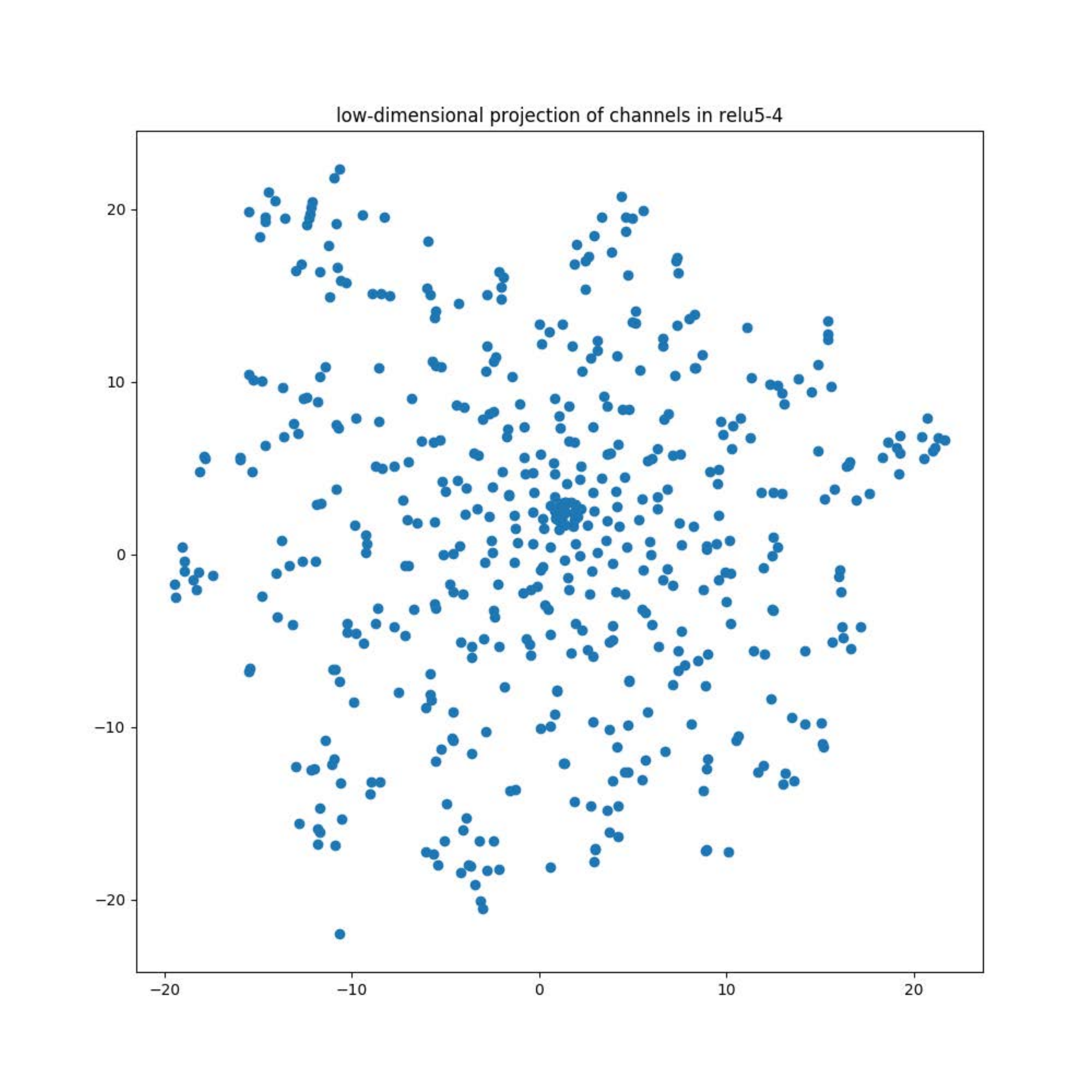}
\end{minipage}%
}%
\centering
\caption{Low dimensional projection of all channels of the feature map of style image (in Figure \ref{wave-layer}(b)) of every single layer via t-SNE \cite{tsne}.}
\label{wave-visualization}
\end{figure*}

\begin{figure*}[tb]
\centering
\subfigure[relu1-1]{
\begin{minipage}[t]{0.25\linewidth}
\centering
\includegraphics[width=\linewidth]{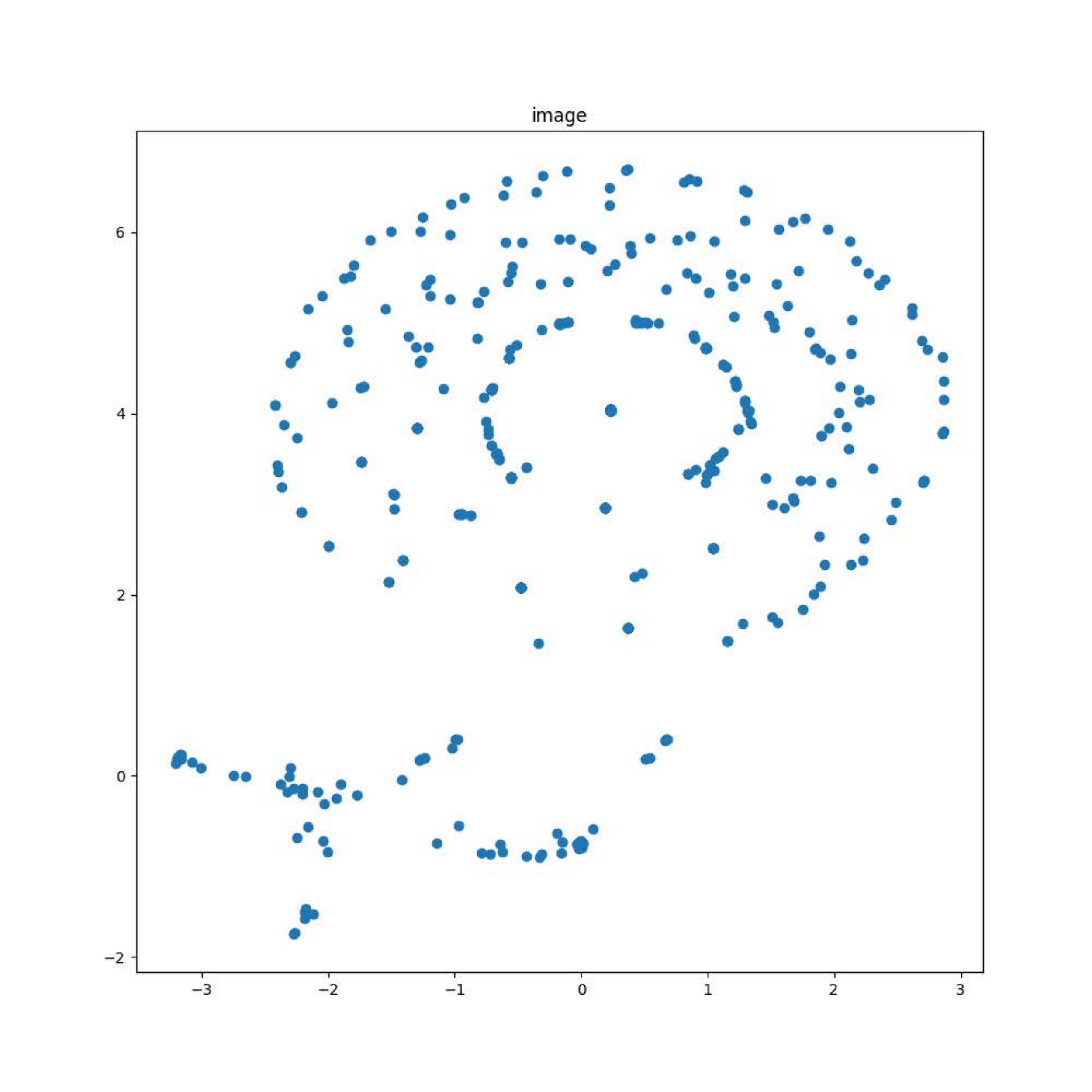}
\end{minipage}%
}%
\subfigure[relu1-2]{
\begin{minipage}[t]{0.25\linewidth}
\centering
\includegraphics[width=\linewidth]{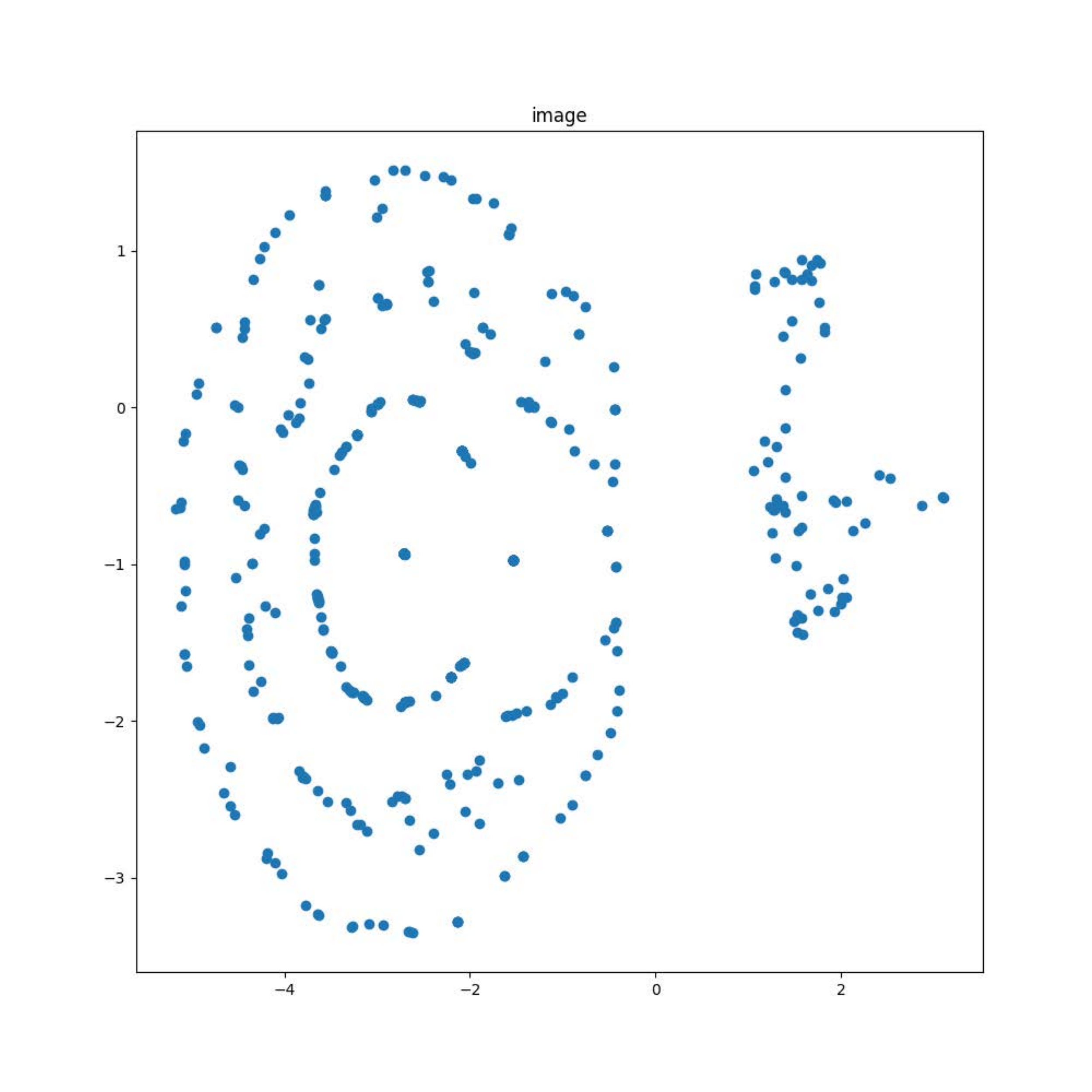}
\end{minipage}%
}%
\subfigure[relu2-1]{
\begin{minipage}[t]{0.25\linewidth}
\centering
\includegraphics[width=\linewidth]{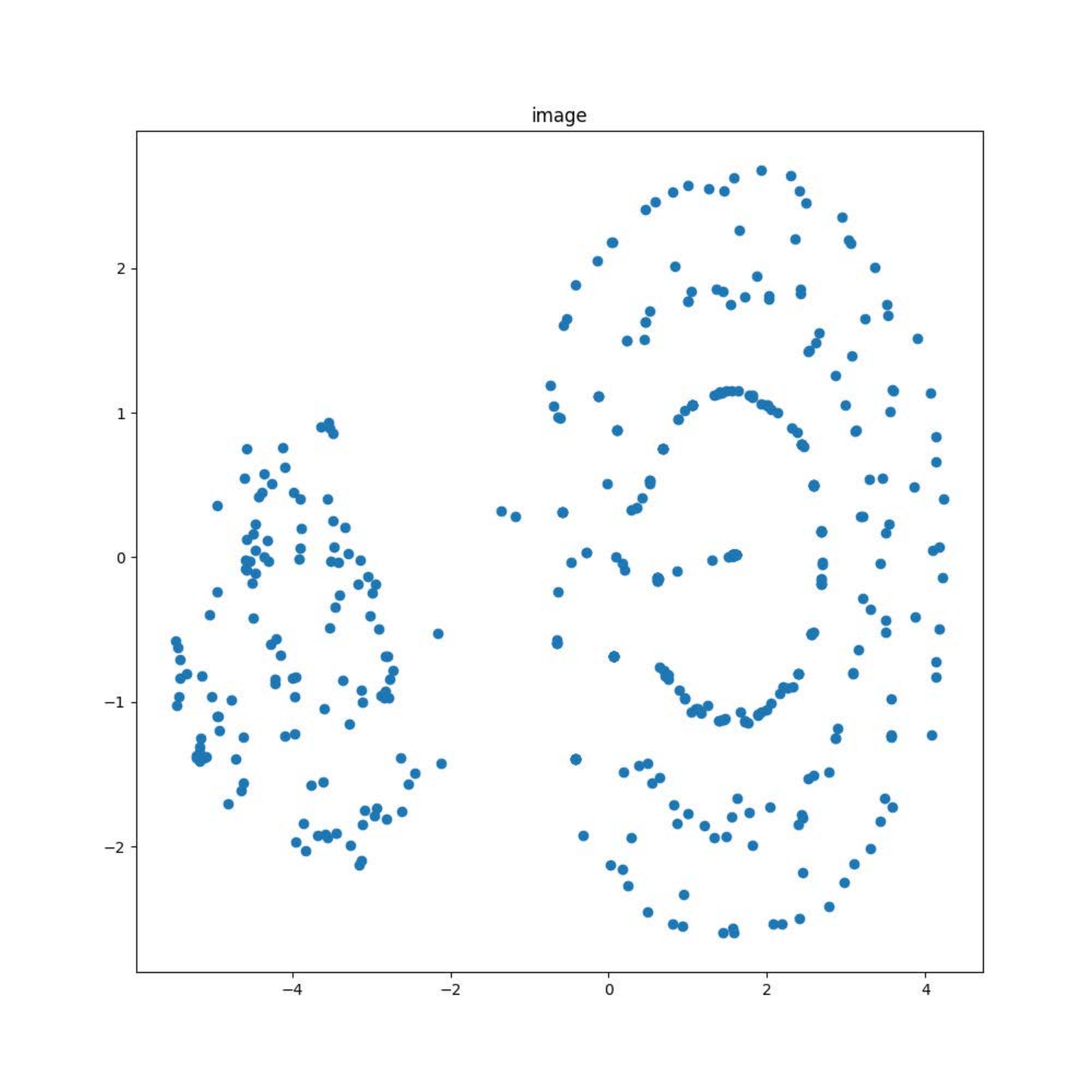}
\end{minipage}%
}%
\subfigure[relu2-2]{
\begin{minipage}[t]{0.25\linewidth}
\centering
\includegraphics[width=\linewidth]{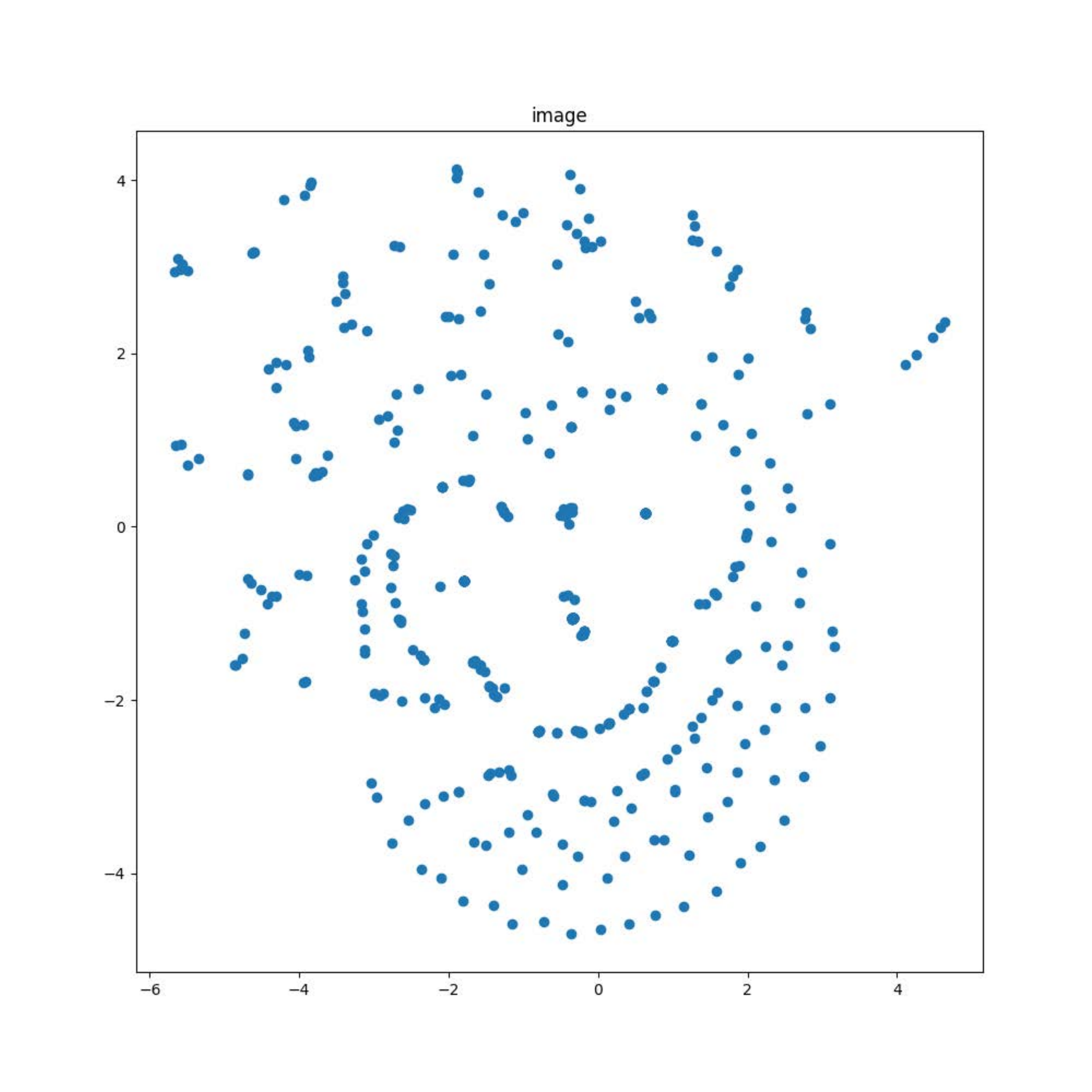}
\end{minipage}%
}%
\vfill

\subfigure[relu3-1]{
\begin{minipage}[t]{0.25\linewidth}
\centering
\includegraphics[width=\linewidth]{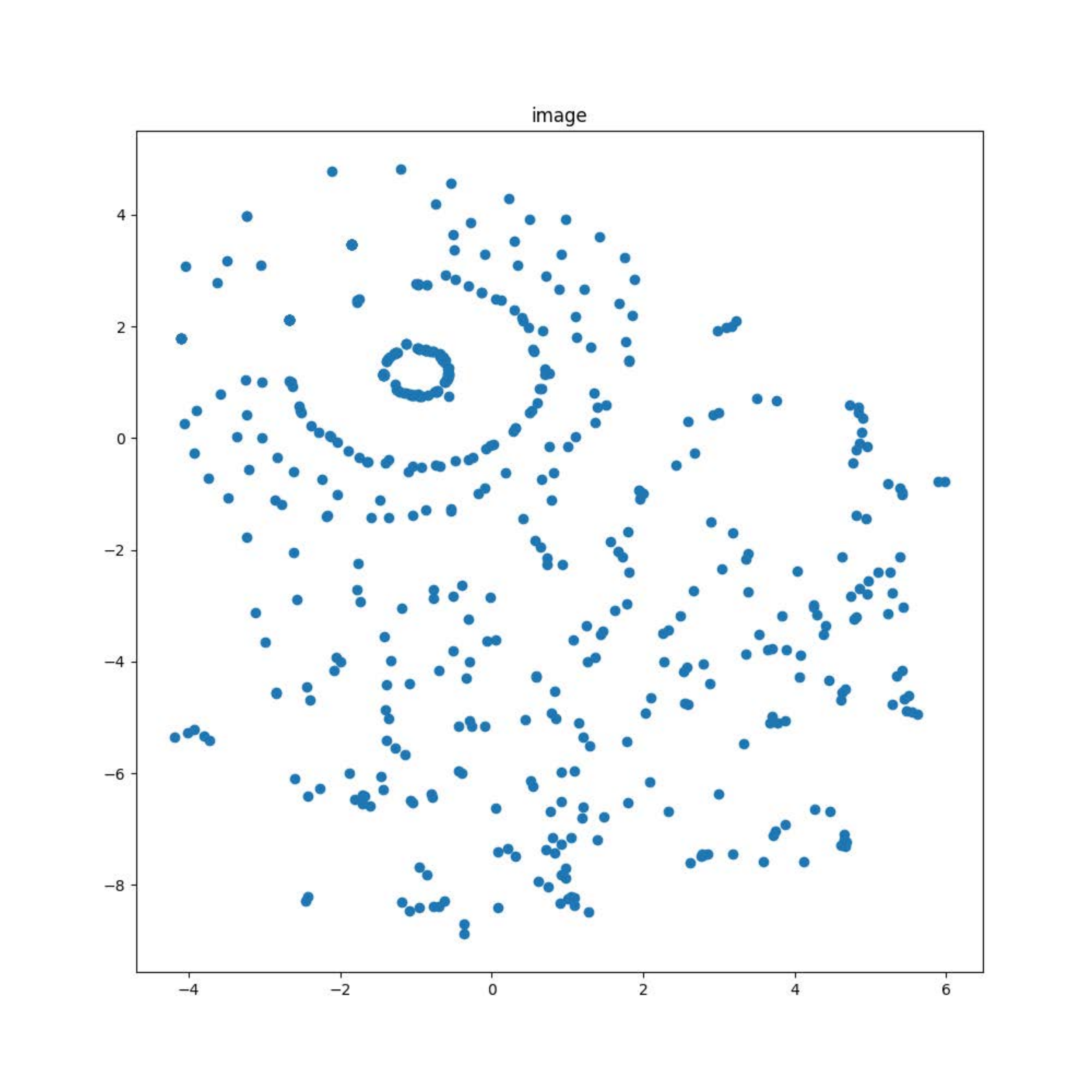}
\end{minipage}%
}%
\subfigure[relu3-2]{
\begin{minipage}[t]{0.25\linewidth}
\centering
\includegraphics[width=\linewidth]{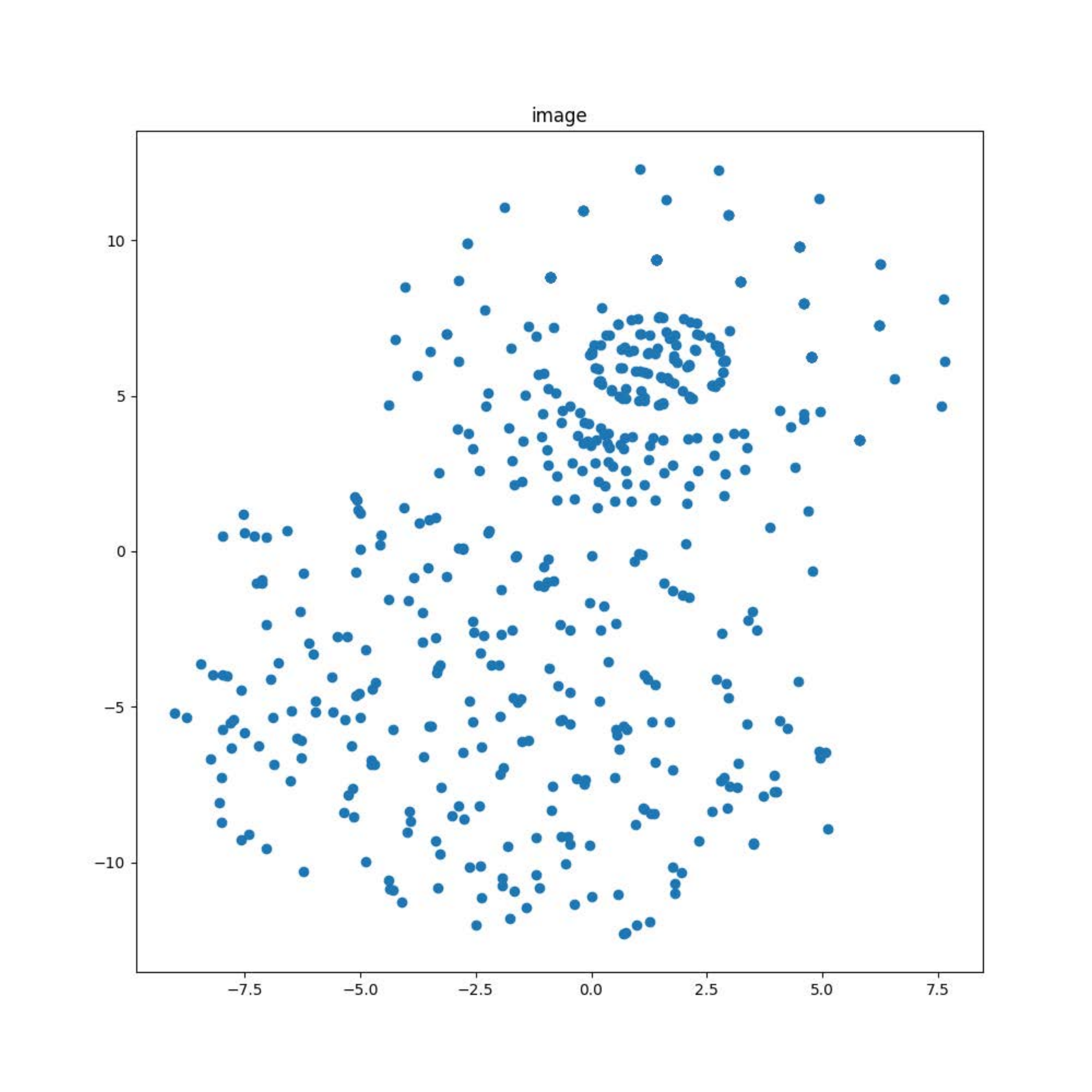}
\end{minipage}%
}%
\subfigure[relu3-3]{
\begin{minipage}[t]{0.25\linewidth}
\centering
\includegraphics[width=\linewidth]{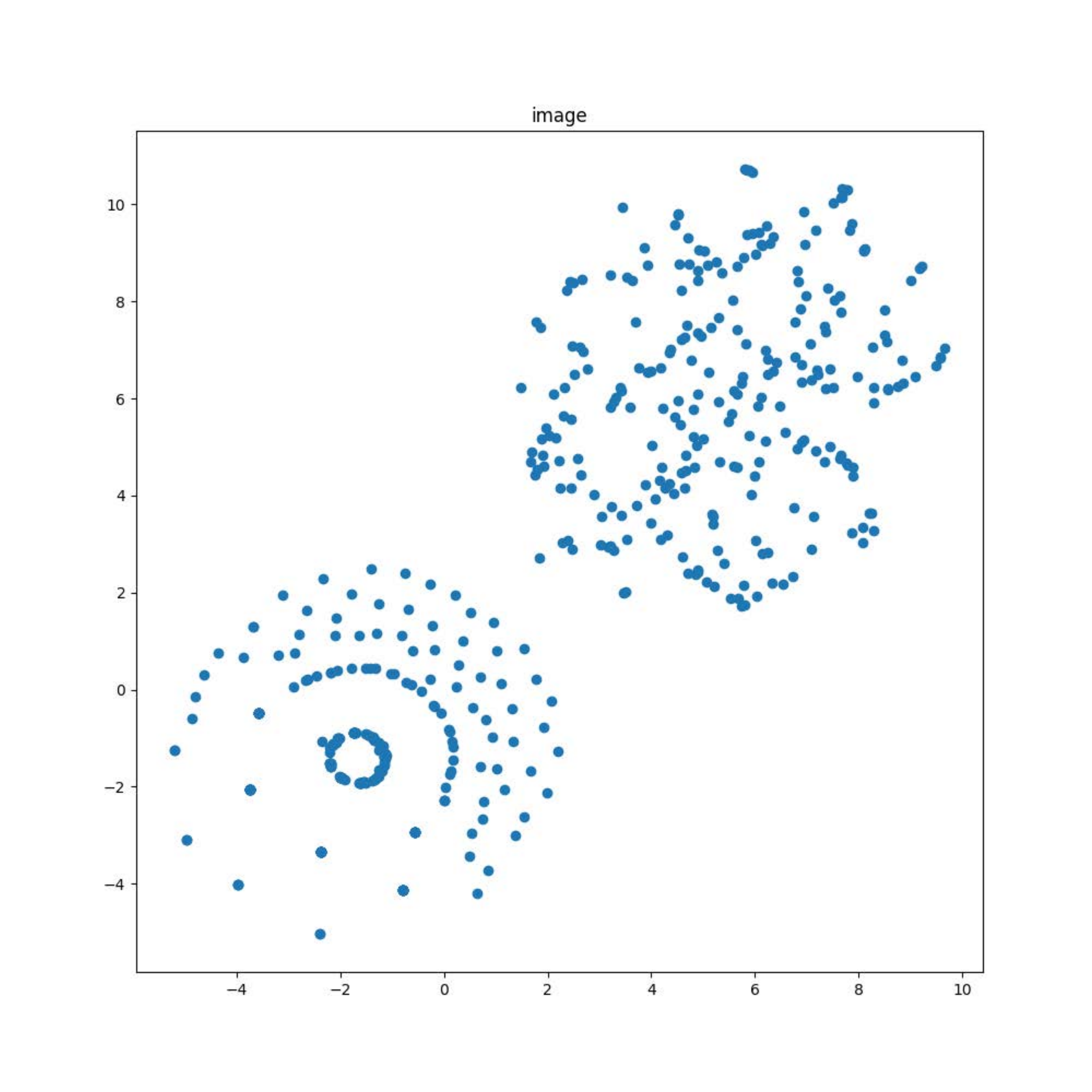}
\end{minipage}%
}%
\subfigure[relu3-4]{
\begin{minipage}[t]{0.25\linewidth}
\centering
\includegraphics[width=\linewidth]{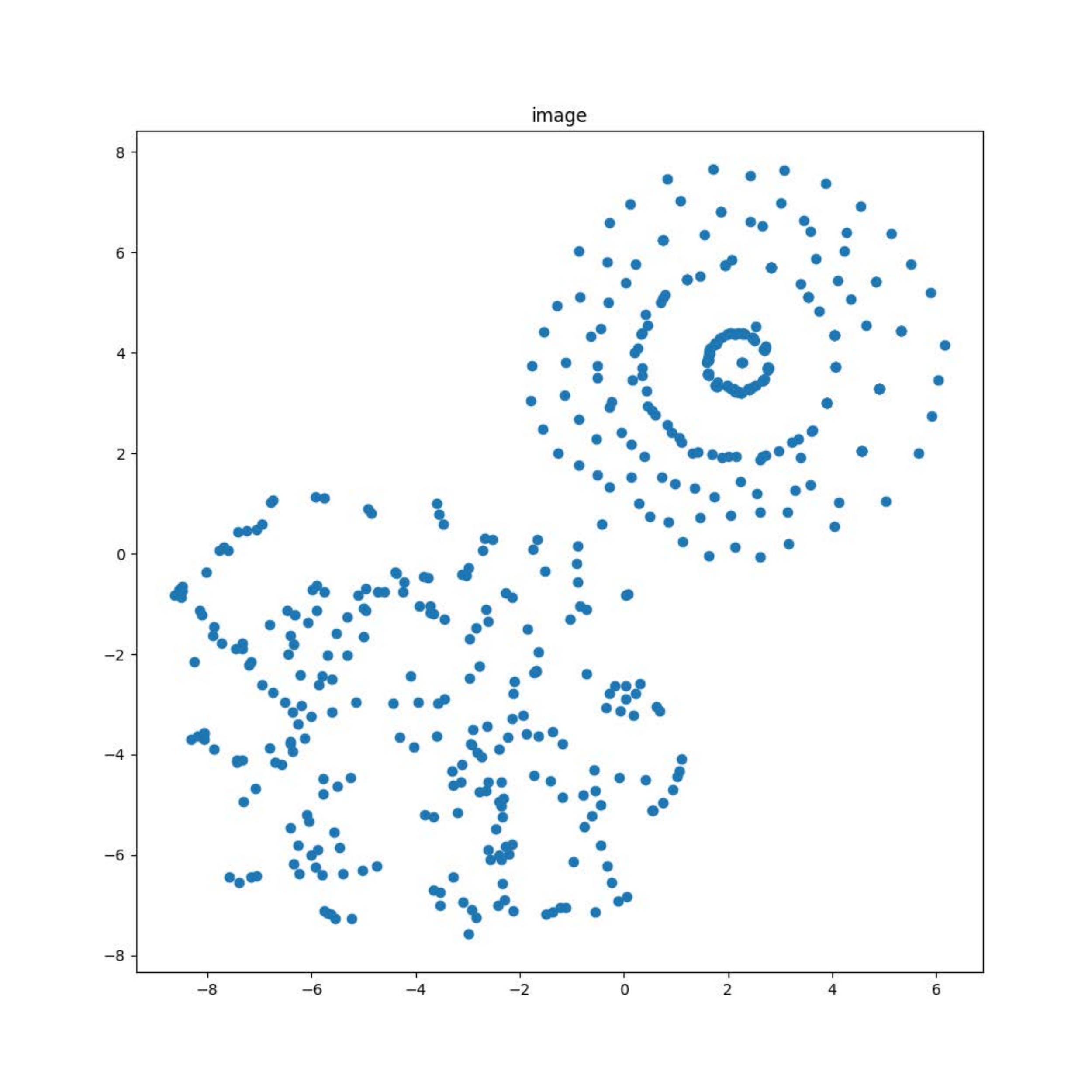}
\end{minipage}%
}%
\vfill

\subfigure[relu4-1]{
\begin{minipage}[t]{0.25\linewidth}
\centering
\includegraphics[width=\linewidth]{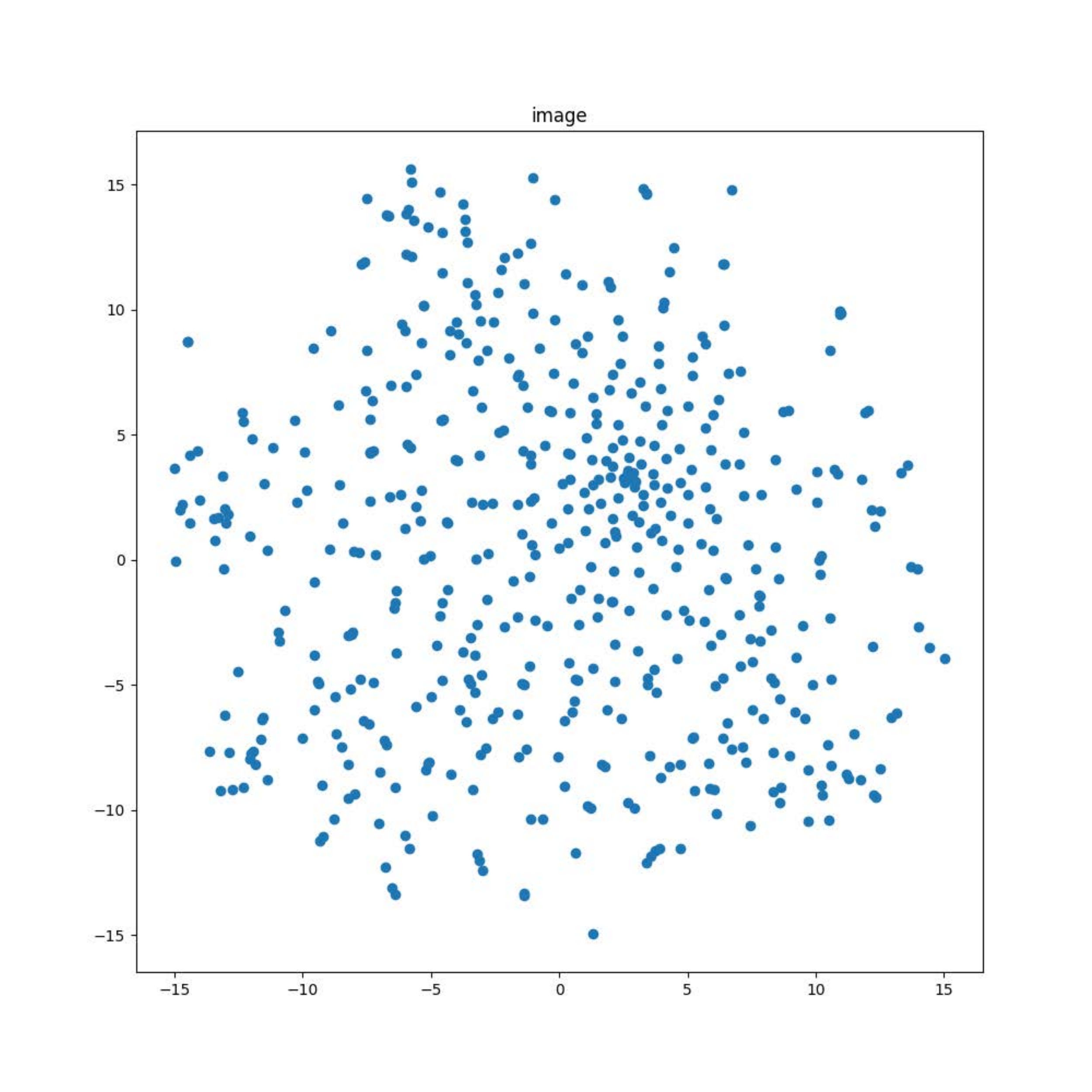}
\end{minipage}%
}%
\subfigure[relu4-2]{
\begin{minipage}[t]{0.25\linewidth}
\centering
\includegraphics[width=\linewidth]{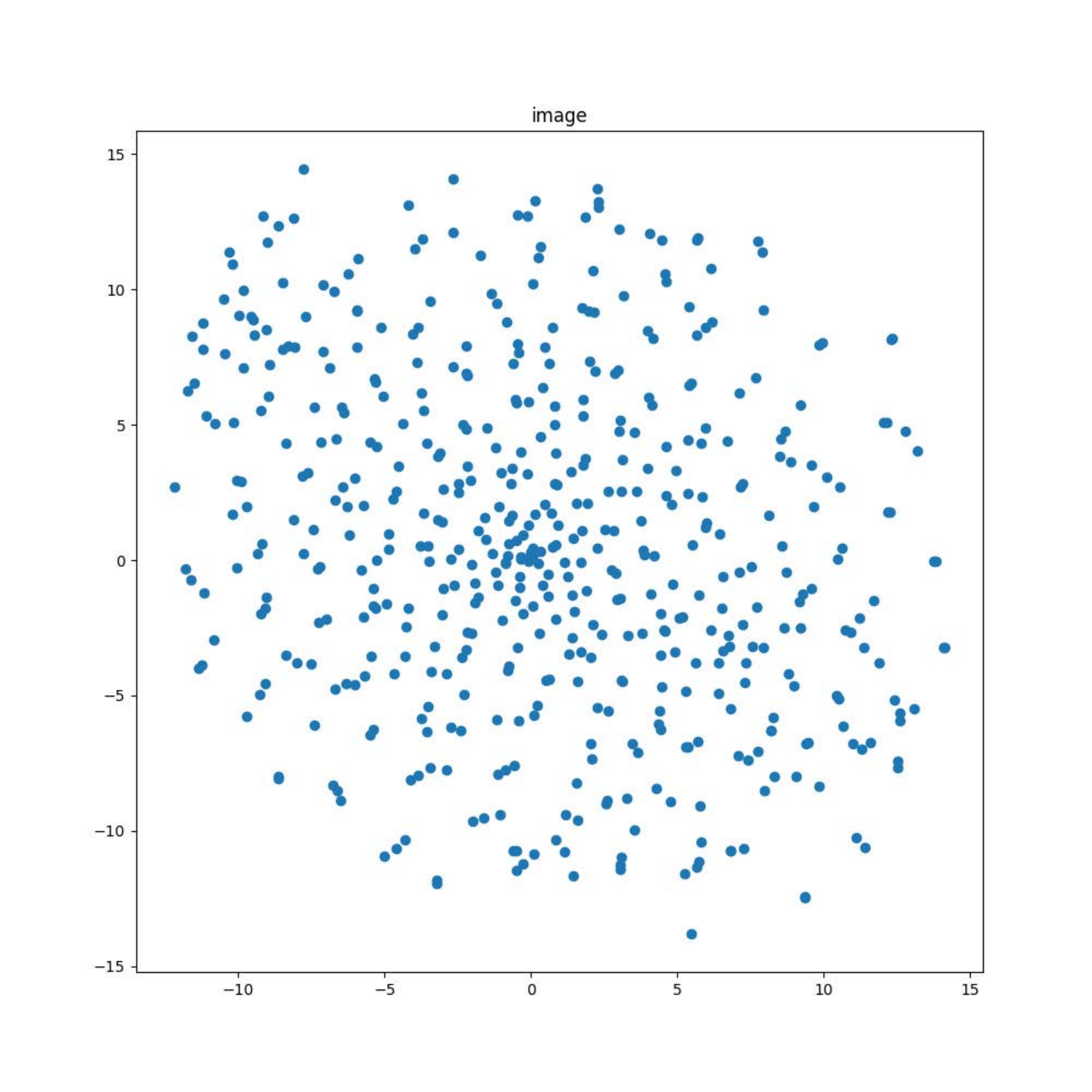}
\end{minipage}%
}%
\subfigure[relu4-3]{
\begin{minipage}[t]{0.25\linewidth}
\centering
\includegraphics[width=\linewidth]{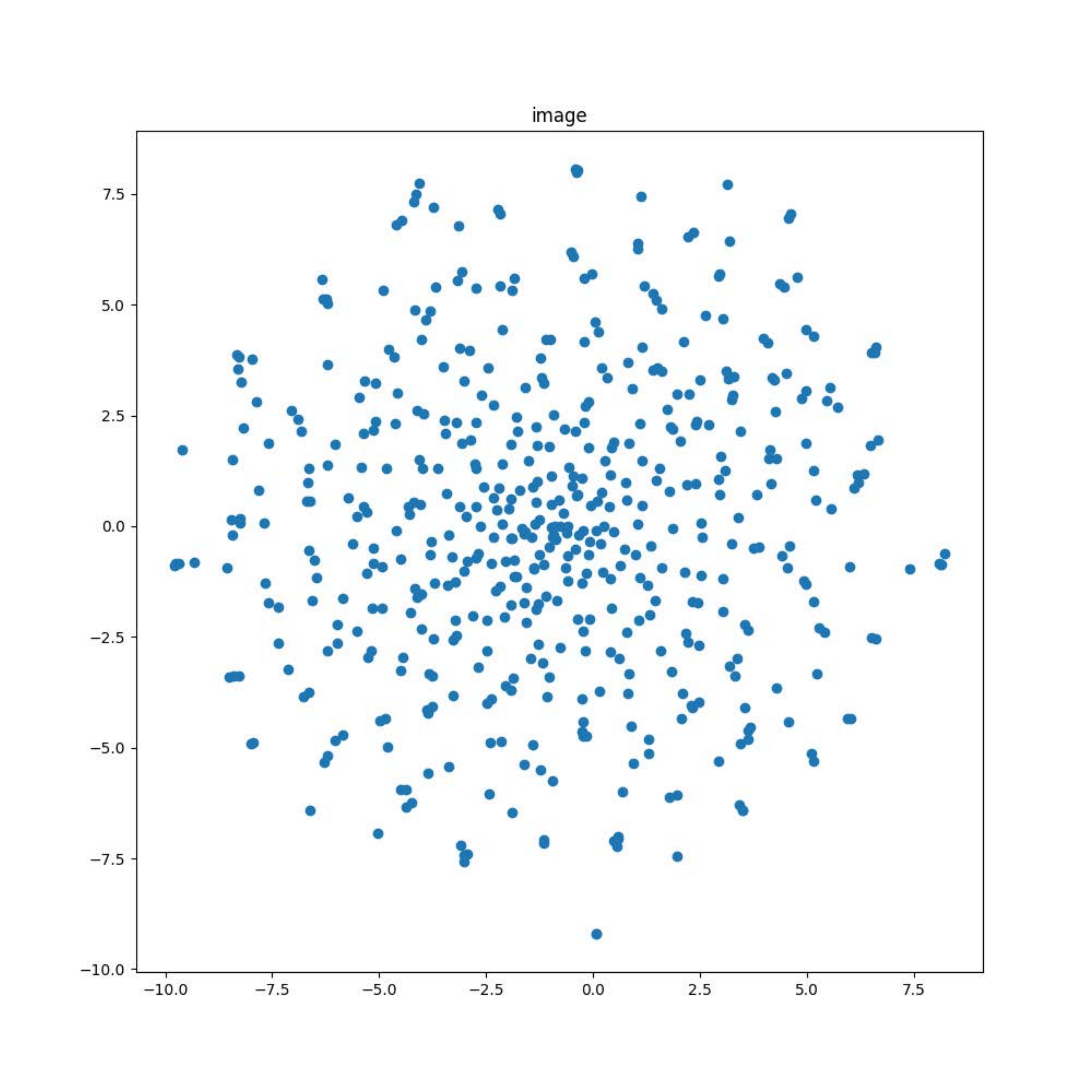}
\end{minipage}%
}%
\subfigure[relu4-4]{
\begin{minipage}[t]{0.25\linewidth}
\centering
\includegraphics[width=\linewidth]{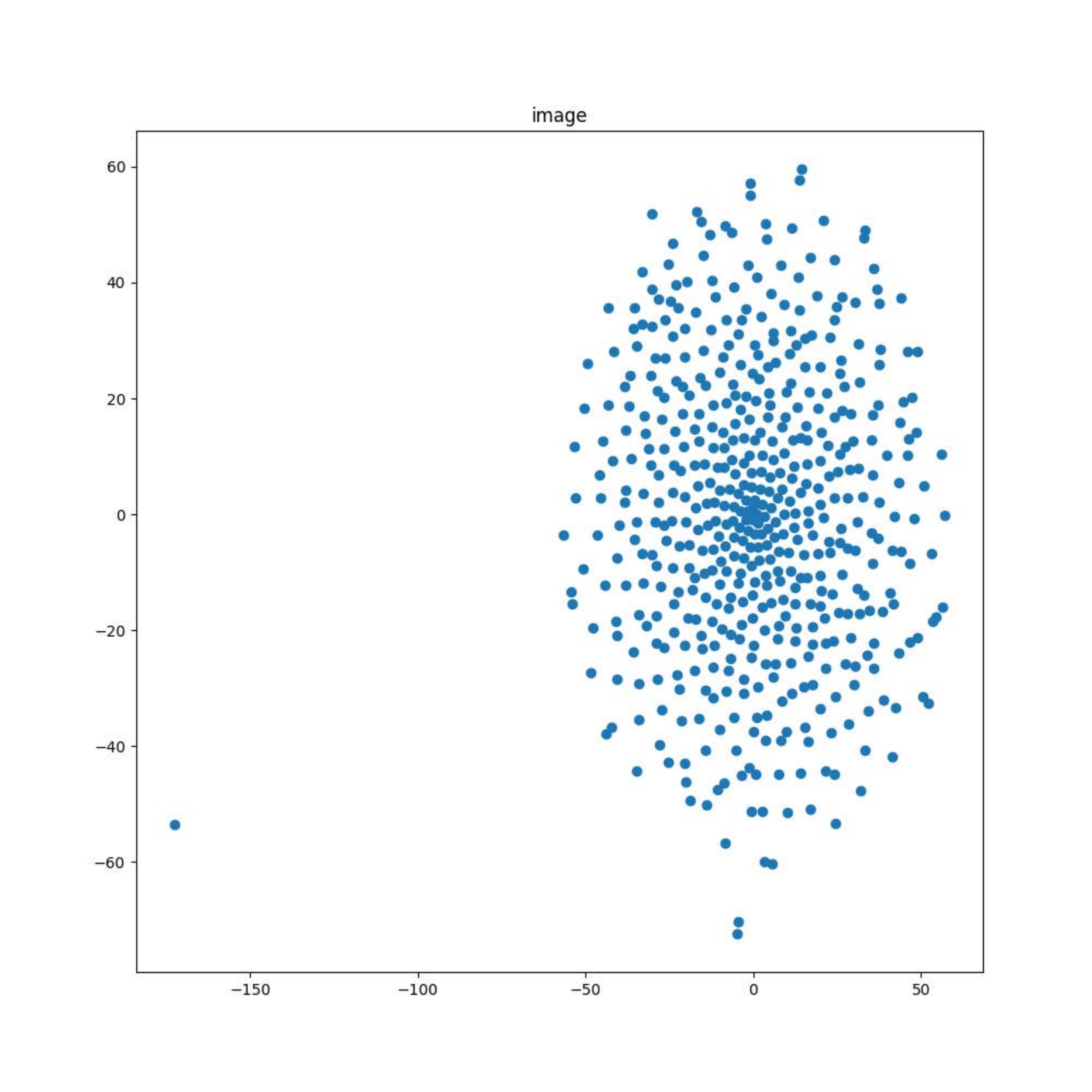}
\end{minipage}%
}%
\vfill

\subfigure[relu5-1]{
\begin{minipage}[t]{0.25\linewidth}
\centering
\includegraphics[width=\linewidth]{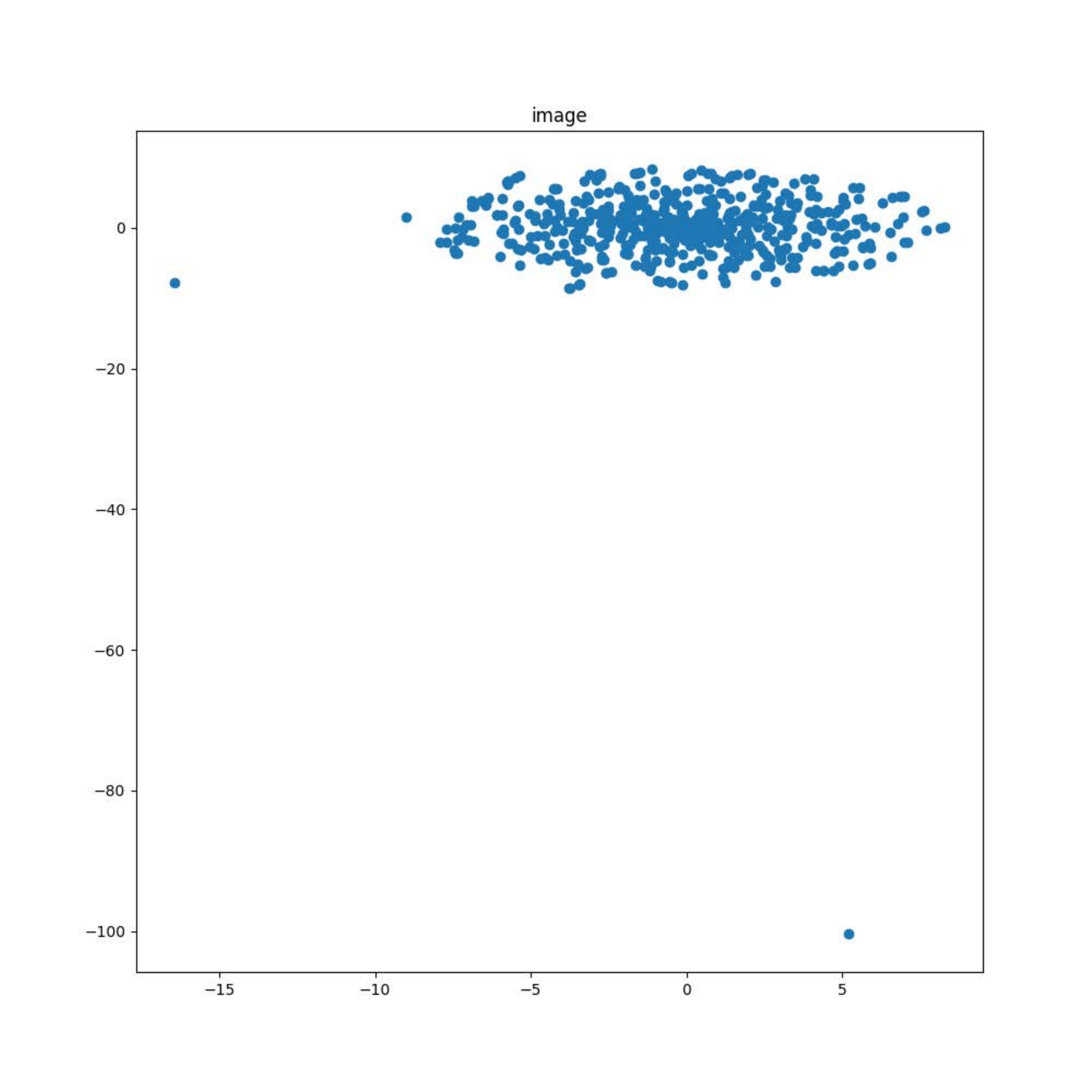}
\end{minipage}%
}%
\subfigure[relu5-2]{
\begin{minipage}[t]{0.25\linewidth}
\centering
\includegraphics[width=\linewidth]{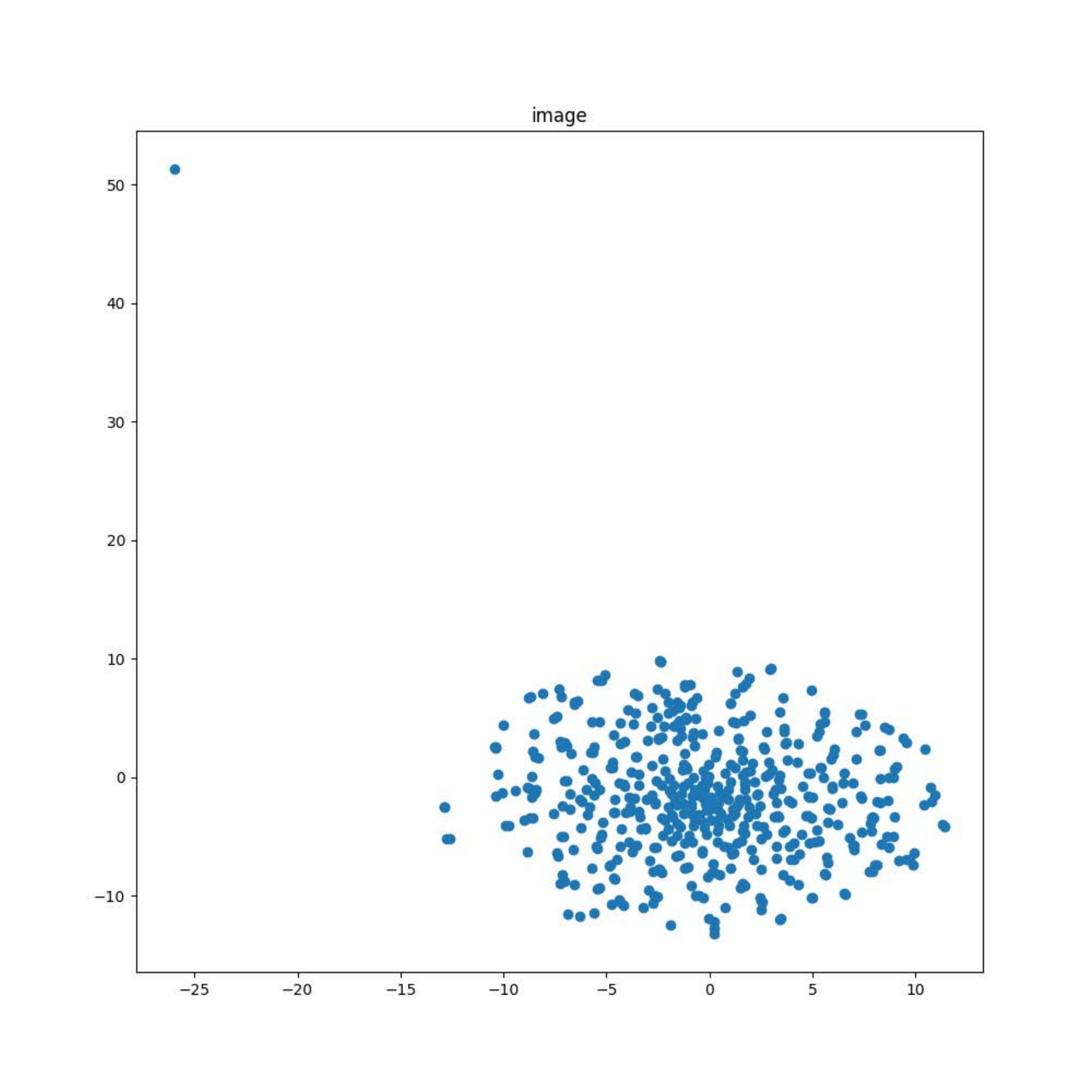}
\end{minipage}%
}%
\subfigure[relu5-3]{
\begin{minipage}[t]{0.25\linewidth}
\centering
\includegraphics[width=\linewidth]{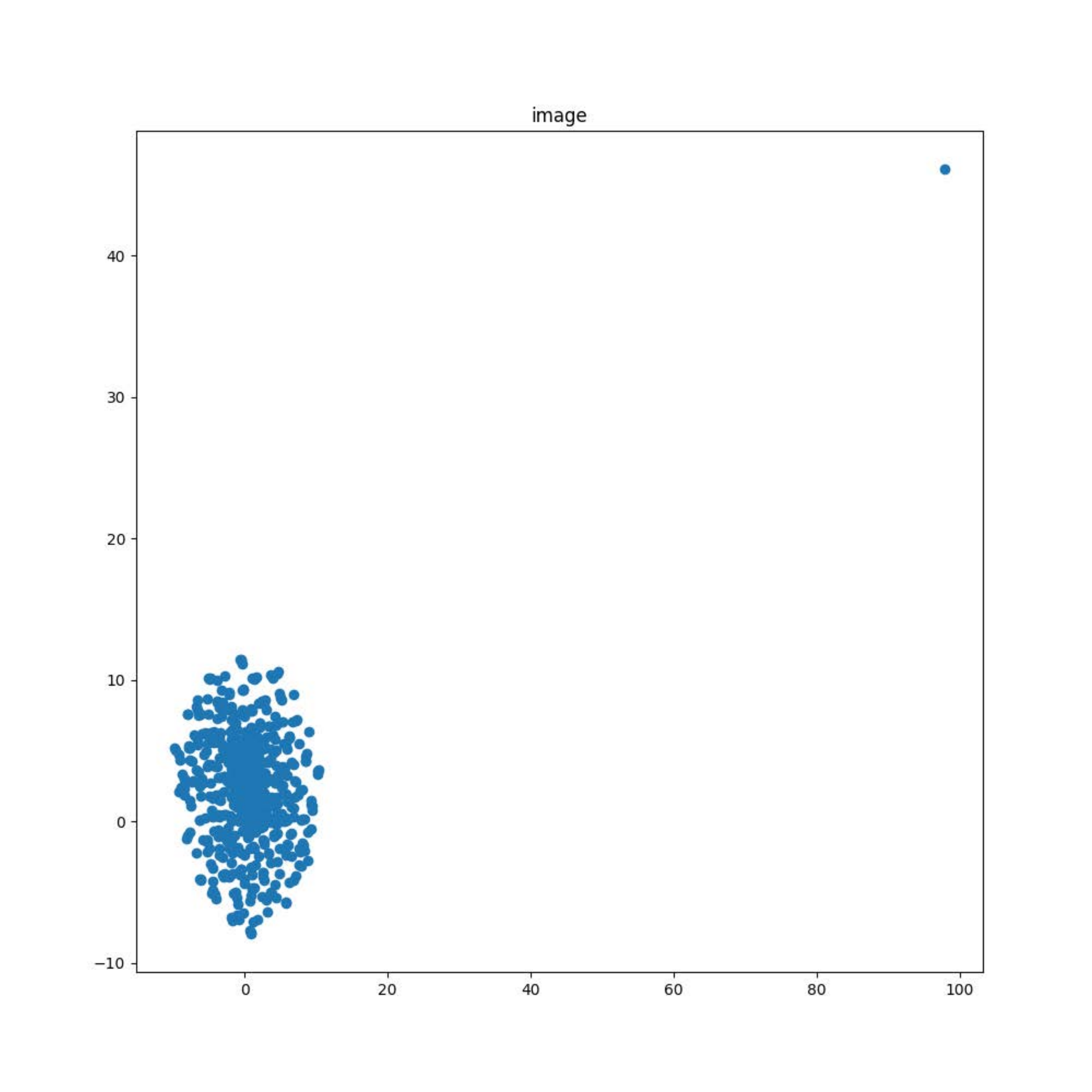}
\end{minipage}%
}%
\subfigure[relu5-4]{
\begin{minipage}[t]{0.25\linewidth}
\centering
\includegraphics[width=\linewidth]{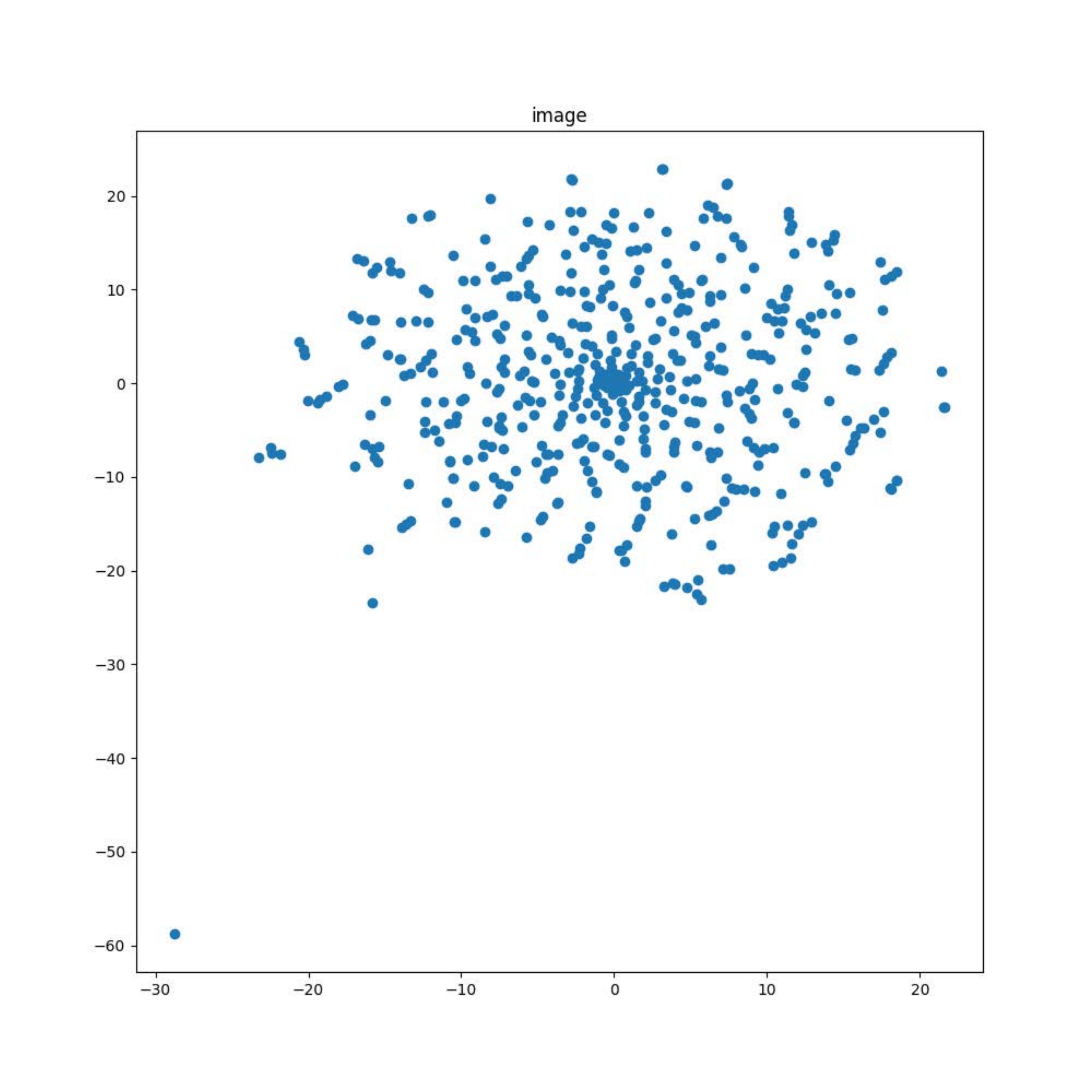}
\end{minipage}%
}%
\centering
\caption{Low dimensional projection of all channels of the feature map of style image (in Figure \ref{lamuse-layer}(b)) of every single layer via t-SNE \cite{tsne}.}
\label{lamuse-visualization}
\end{figure*}

We further visualize each channel of the feature map of the style image using t-SNE \cite{tsne}, as are shwon in Figure \ref{wave-visualization} and Figure \ref{lamuse-visualization} where the similarity of the result is quite interesting. However, we could not explain the specific pattern in the visualization result and the relationship between the similarity of the visualization results with the similarity of the stylization effects of different VGG layers yet.

\subsection*{B. The Manifold of Spectrum Based Methods}
We analyze the spectrum space by projecting the style bases via Isomap \cite{Isomap} into low dimensional space where the X-axis represents the color basis and the Y-axis represents the stroke basis, which can analytically demonstrate the effectiveness and robustness of spectrum based methods. Three artistic styles are experimented (shown in Figure \ref{manifold}(a-c)). Chinese paintings and pen sketches share similar color style which is sharply distinguished with oil paintings' while the stroke of three artistic styles are quite different from each other. Thus, as in shown in Figure \ref{manifold}(d), Chinese paintings and pen sketches are close to each other and both stay away from oil paintings in X-axis which represents color while three styles are respectively separable in Y-axis which represents stroke, which completely satisfies our analysis of the three artistic styles. 

When we apply the same method to large scale of style images (Figure \ref{manifold-2}), X-axis clearly represents the linear transition from dull-colored to rich-colored. However, we fail to conclude any notable linear transition for Y-axis from the 2-dimensional visualization probably because it is hard to describe the style of stroke (boldface,length,curvity,etc.) using only one dimension. 
\begin{figure*}[tb]
\centering
\subfigure[]{
\begin{minipage}[t]{0.174\linewidth}
\centering
\includegraphics[width=\linewidth]{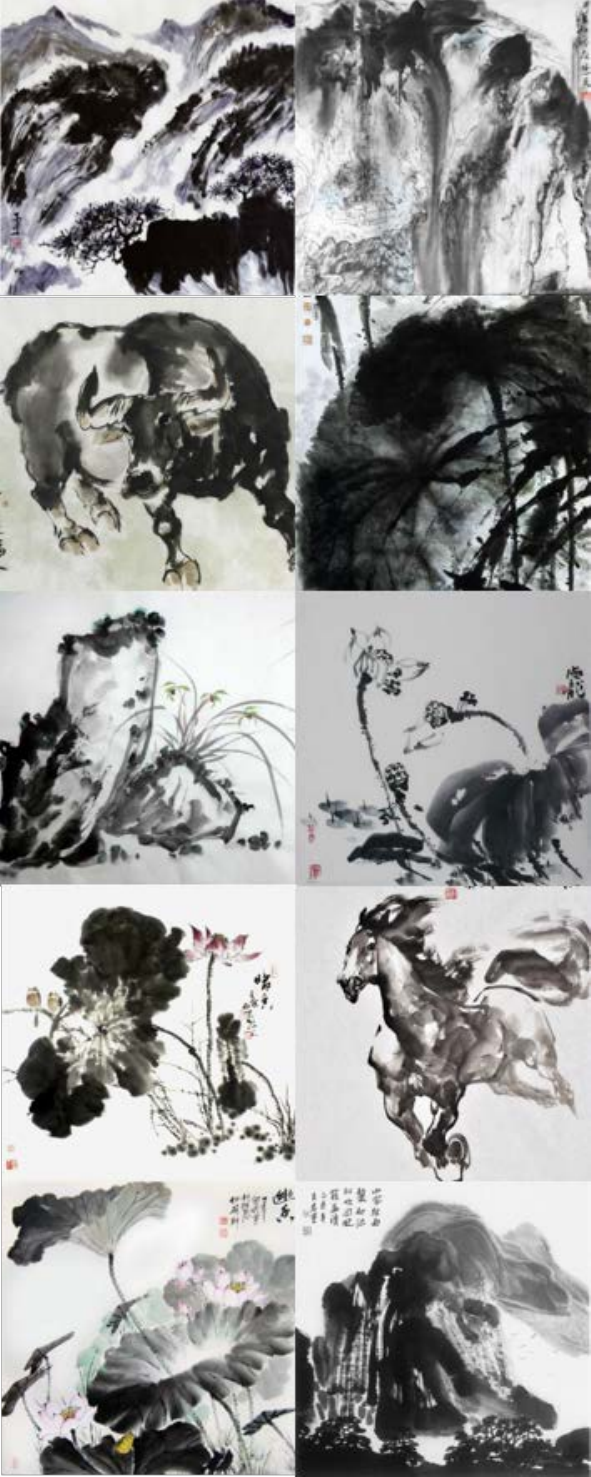}
\end{minipage}%
}%
\subfigure[]{
\begin{minipage}[t]{0.174\linewidth}
\centering
\includegraphics[width=\linewidth]{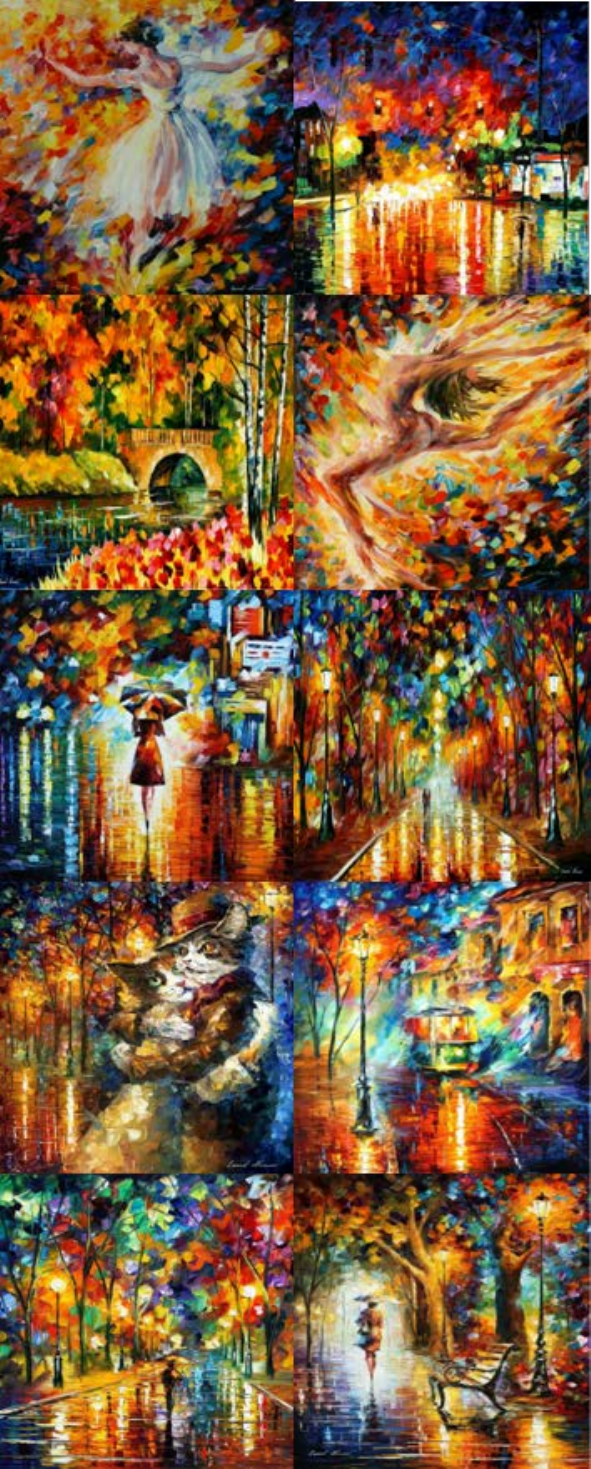}
\end{minipage}%
}%
\subfigure[]{
\begin{minipage}[t]{0.174\linewidth}
\centering
\includegraphics[width=\linewidth]{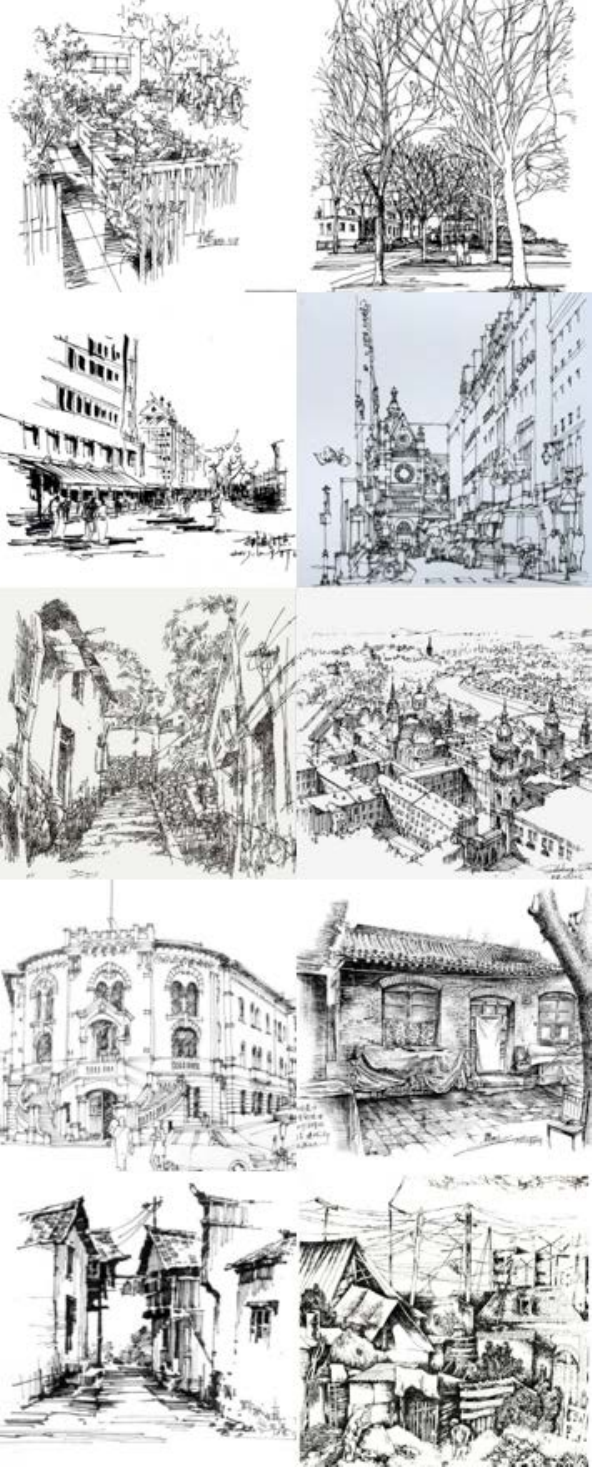}
\end{minipage}%
}%
\subfigure[]{
\begin{minipage}[t]{0.48\linewidth}
\centering
\includegraphics[width=\linewidth]{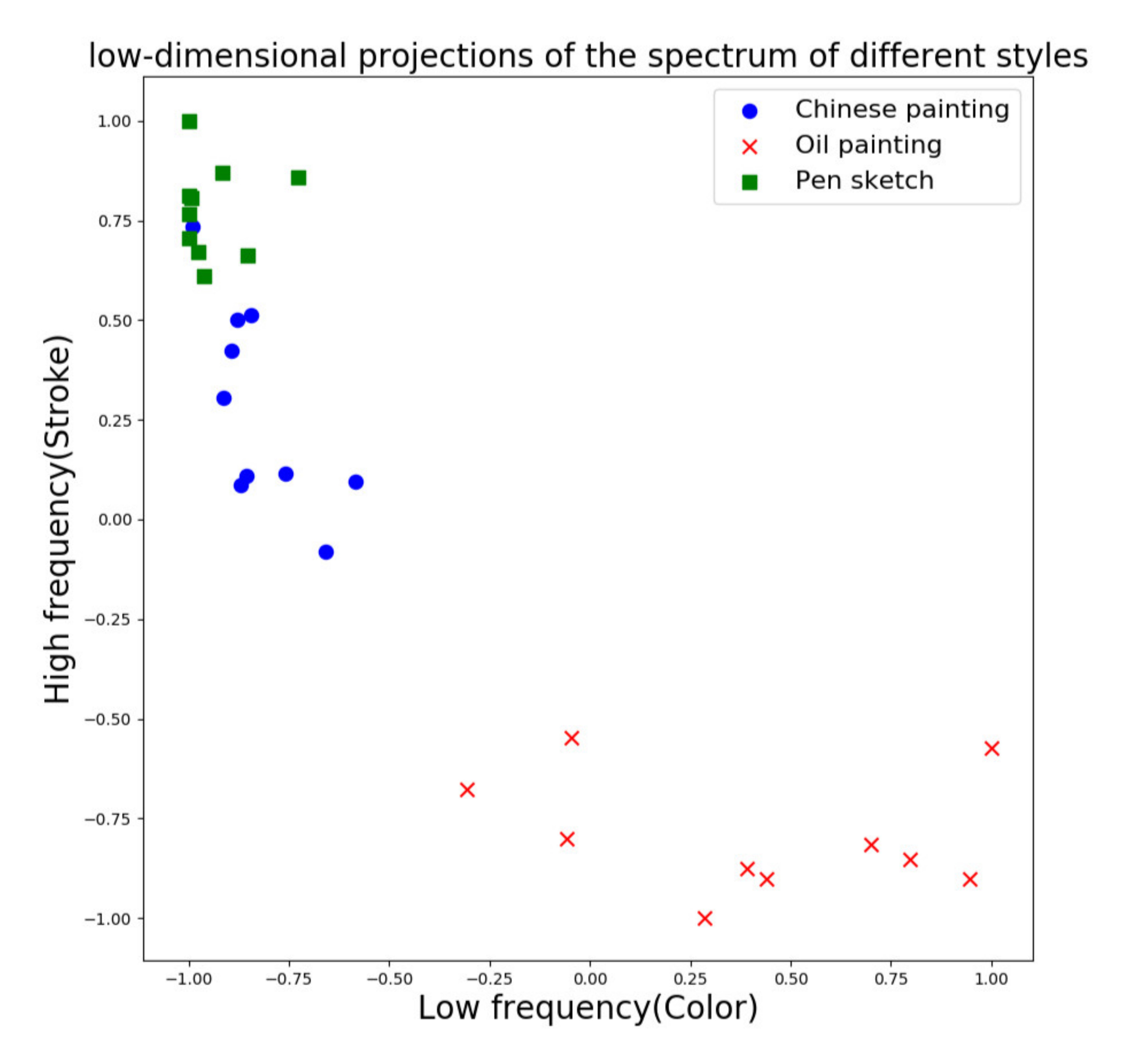}
\end{minipage}%
}%
\centering
\caption{(a) Chinese paintings; (b) Oil paintings (by Leonid Afremov); (c) Pen sketches; (d) low-dimensional projections of the spectrum of style(a-c) via Isomap \cite{Isomap}.}
\label{manifold}
\end{figure*}

\begin{figure*}[tb]
\centering
\subfigure[]{
\begin{minipage}[t]{\linewidth}
\centering
\includegraphics[width=\linewidth]{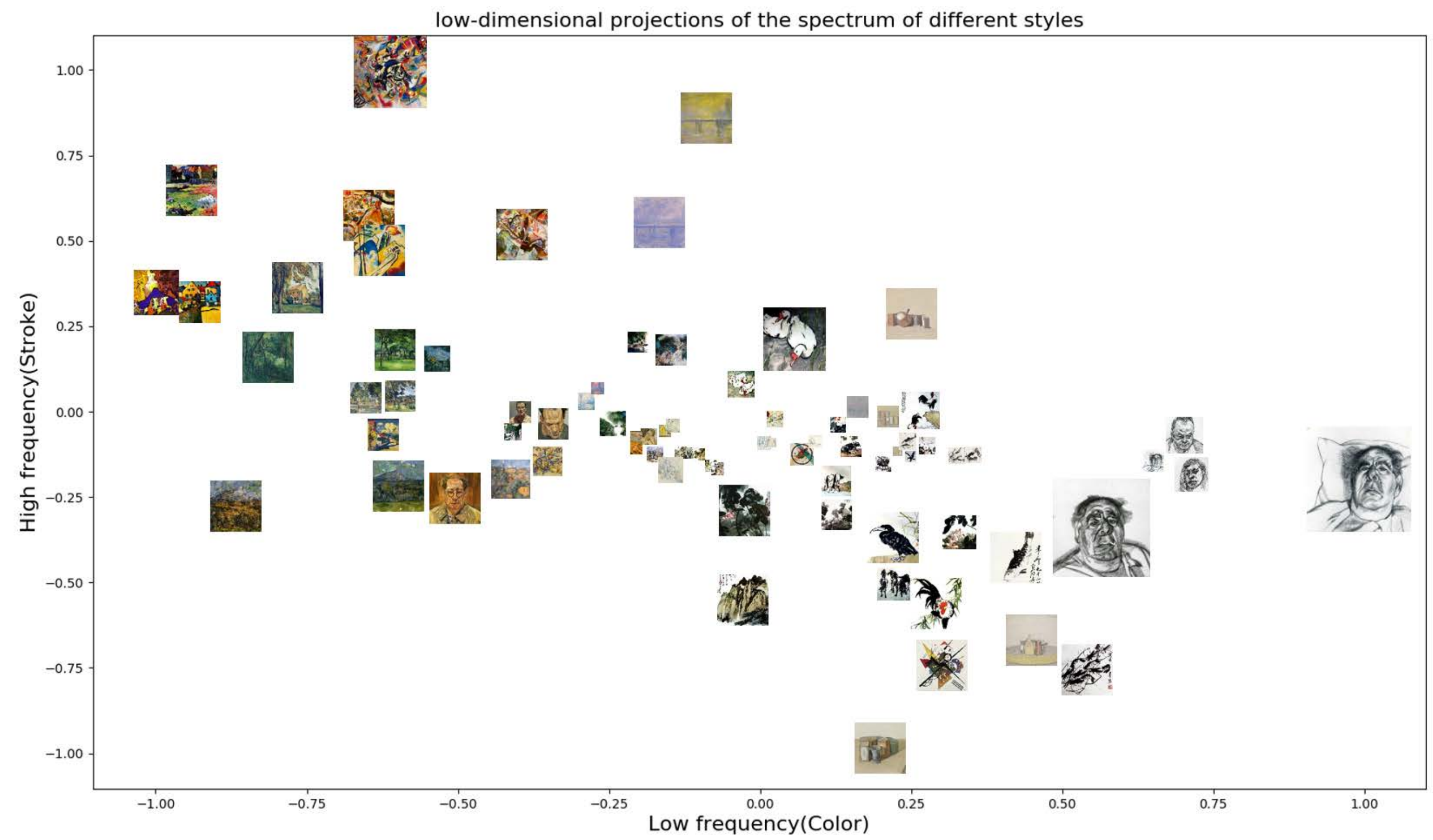}
\end{minipage}%
}%
\centering
\caption{low-dimensional projections of the spectrum of large scale of style images via Isomap \cite{Isomap}. The size of each image shown above does not indicate any other information, but is set to prevent the overlap of the images only.}
\label{manifold-2}
\end{figure*}

\subsection*{C. Stroke Intervention}
\begin{figure*}[htb]
\centering
\subfigure{
\begin{minipage}[t]{0.2\linewidth}
\centering
\includegraphics[width=\linewidth]{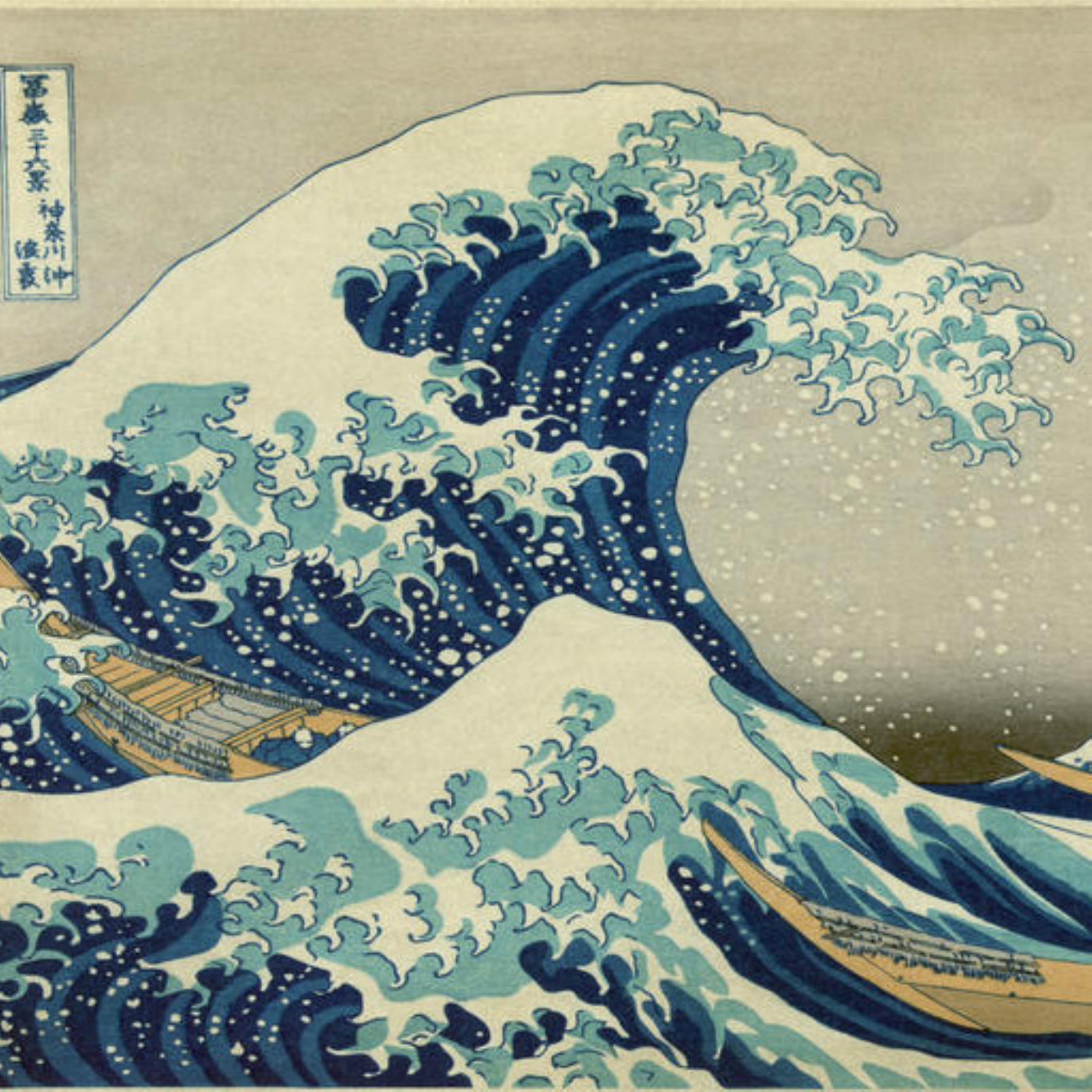}
\end{minipage}%
}%
\subfigure{
\begin{minipage}[t]{0.2\linewidth}
\centering
\includegraphics[width=\linewidth]{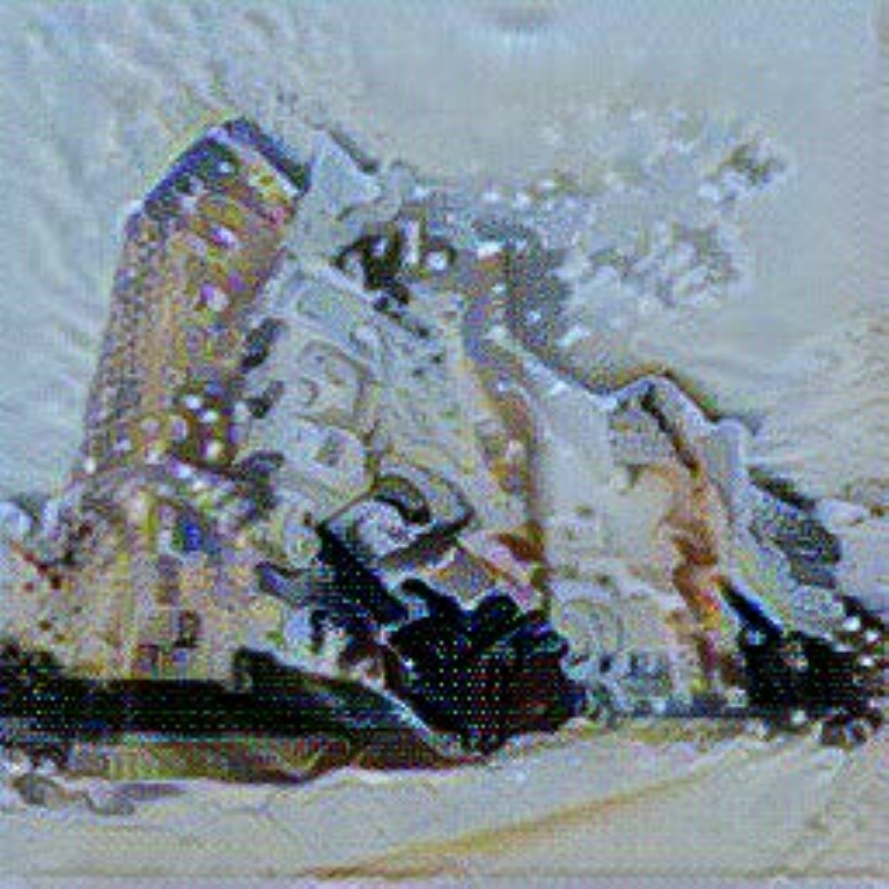}
\end{minipage}%
}%
\subfigure{
\begin{minipage}[t]{0.2\linewidth}
\centering
\includegraphics[width=\linewidth]{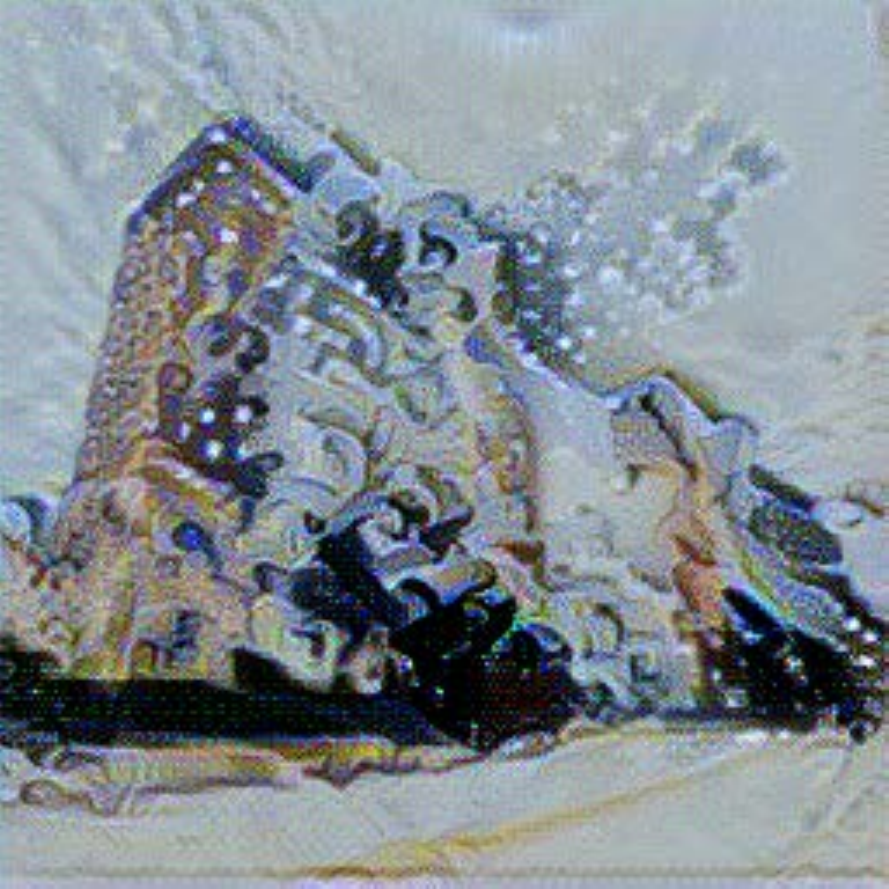}
\end{minipage}%
}%
\subfigure{
\begin{minipage}[t]{0.2\linewidth}
\centering
\includegraphics[width=\linewidth]{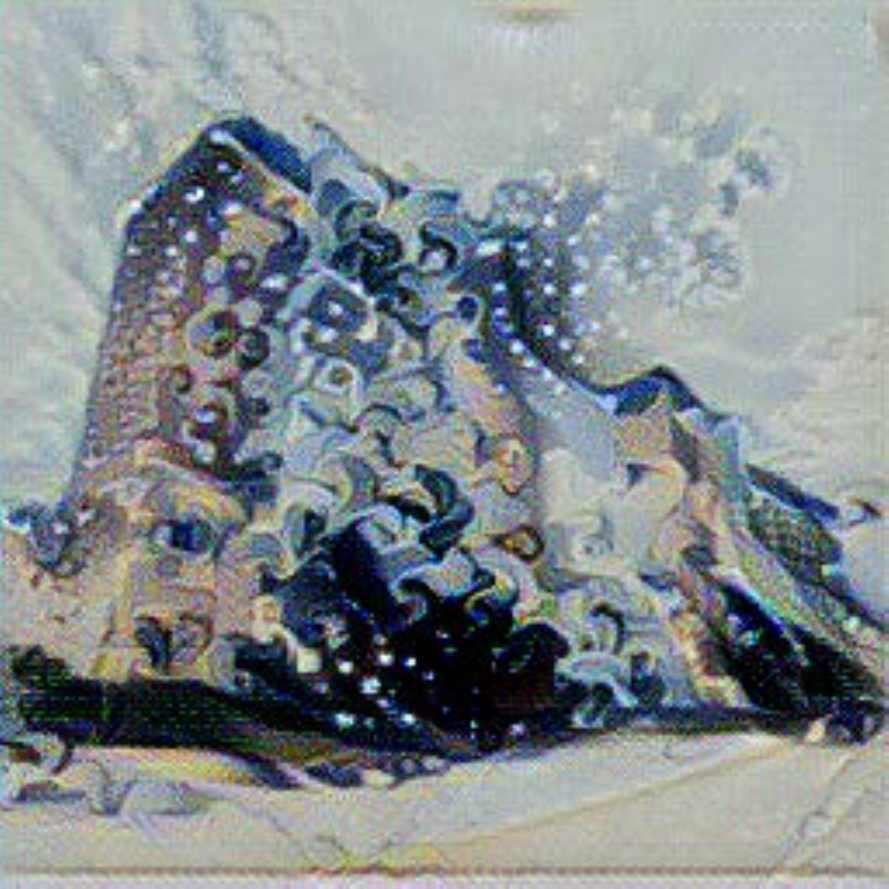}
\end{minipage}%
}%
\subfigure{
\begin{minipage}[t]{0.2\linewidth}
\centering
\includegraphics[width=\linewidth]{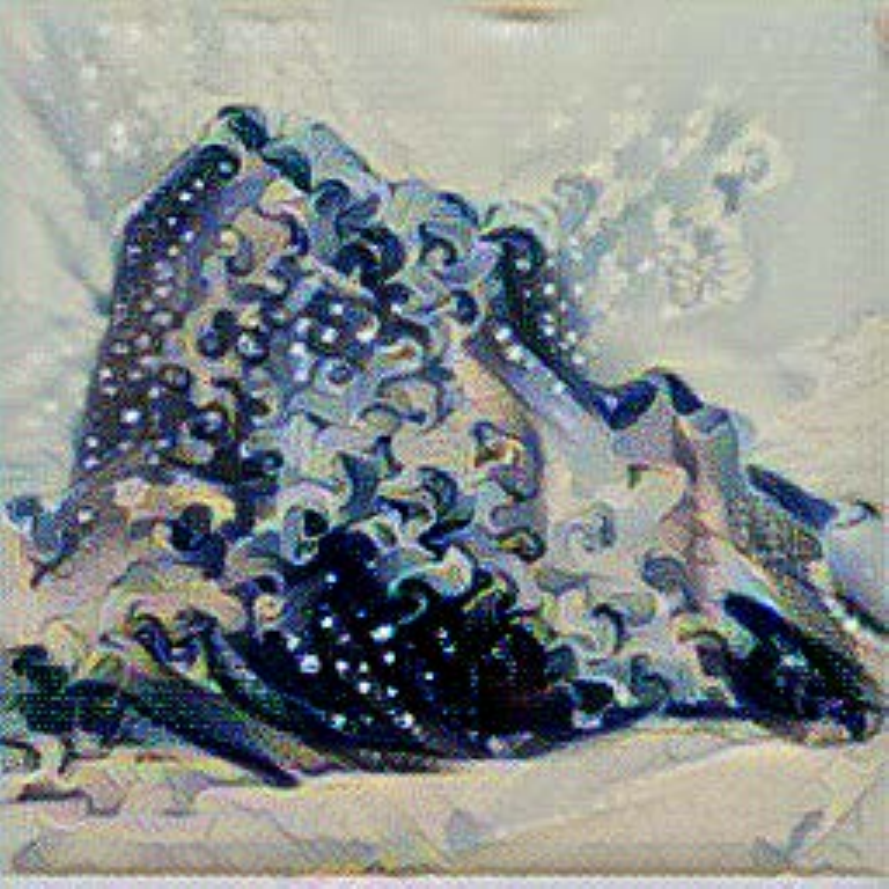}
\end{minipage}%
}%
\vfill

\subfigure{
\begin{minipage}[t]{0.2\linewidth}
\centering
\includegraphics[width=\linewidth]{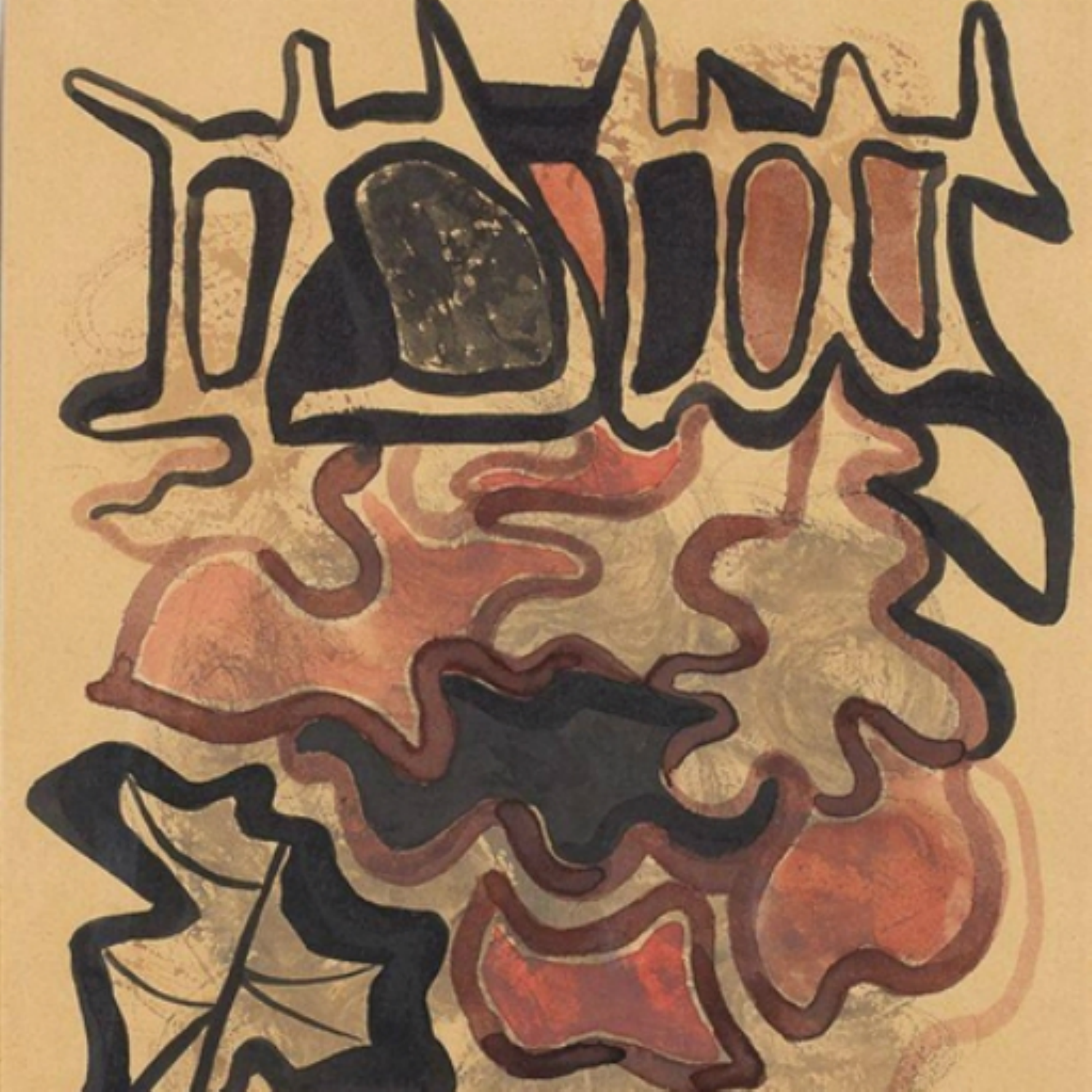}
\end{minipage}%
}%
\subfigure{
\begin{minipage}[t]{0.2\linewidth}
\centering
\includegraphics[width=\linewidth]{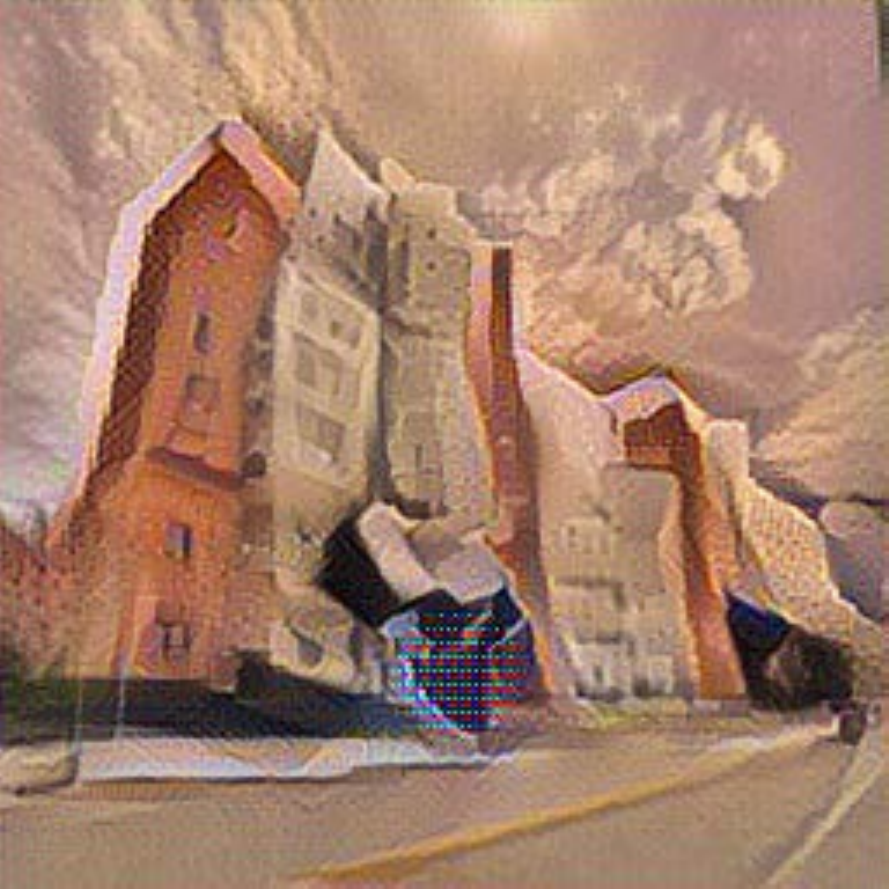}
\end{minipage}%
}%
\subfigure{
\begin{minipage}[t]{0.2\linewidth}
\centering
\includegraphics[width=\linewidth]{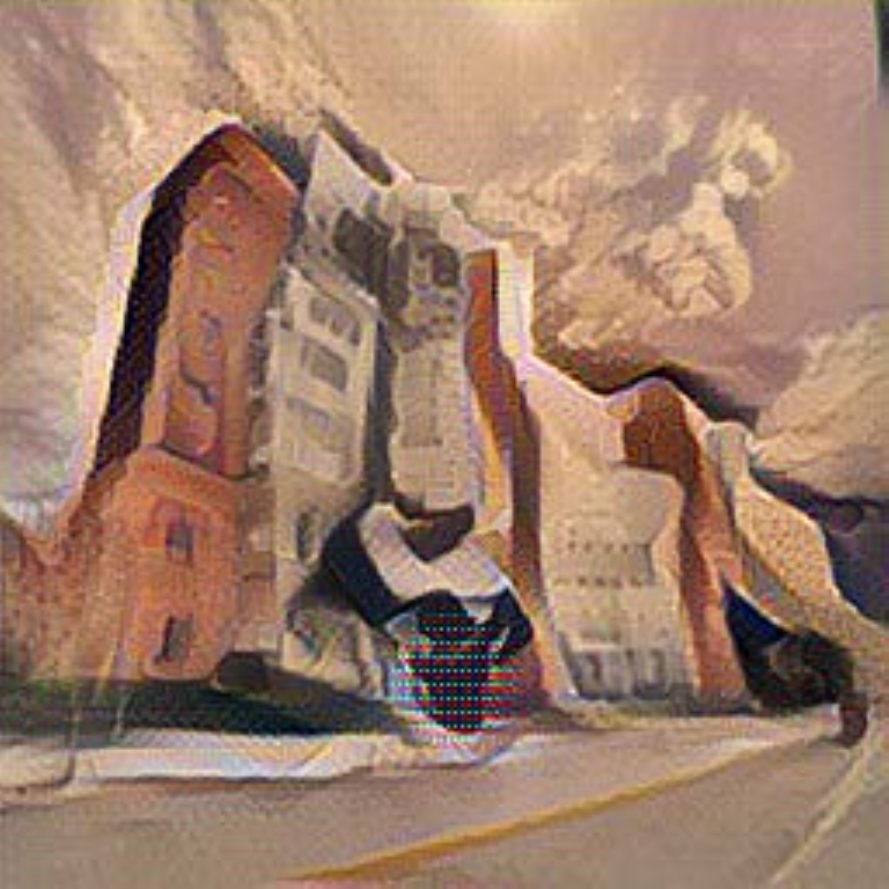}
\end{minipage}%
}%
\subfigure{
\begin{minipage}[t]{0.2\linewidth}
\centering
\includegraphics[width=\linewidth]{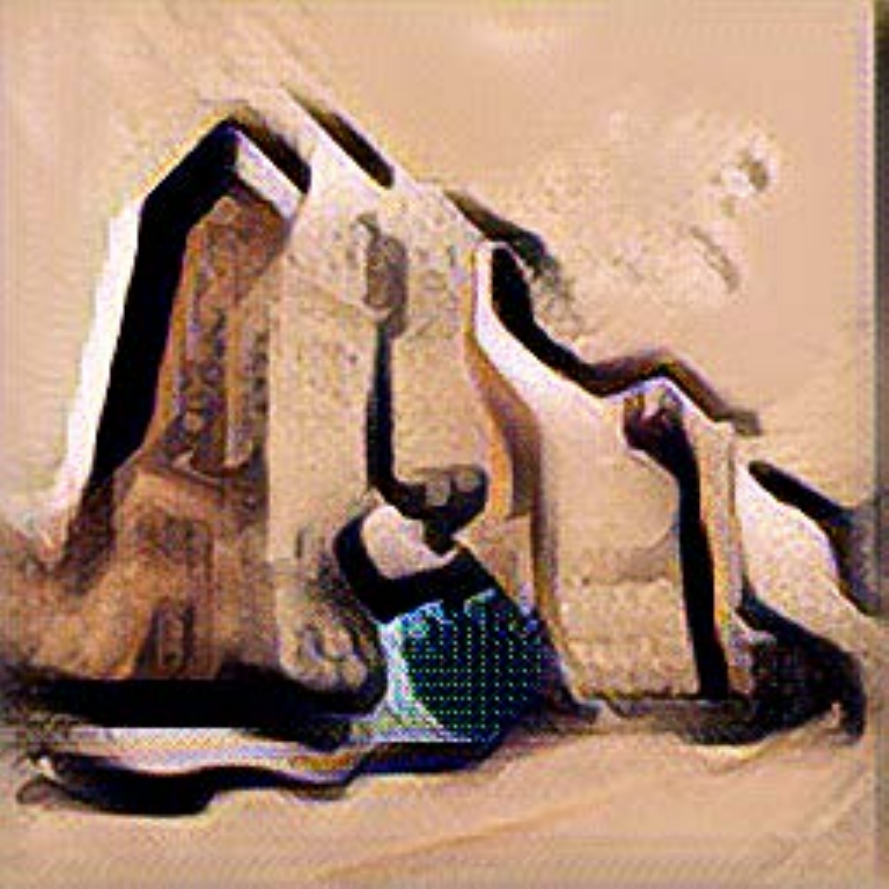}
\end{minipage}%
}%
\subfigure{
\begin{minipage}[t]{0.2\linewidth}
\centering
\includegraphics[width=\linewidth]{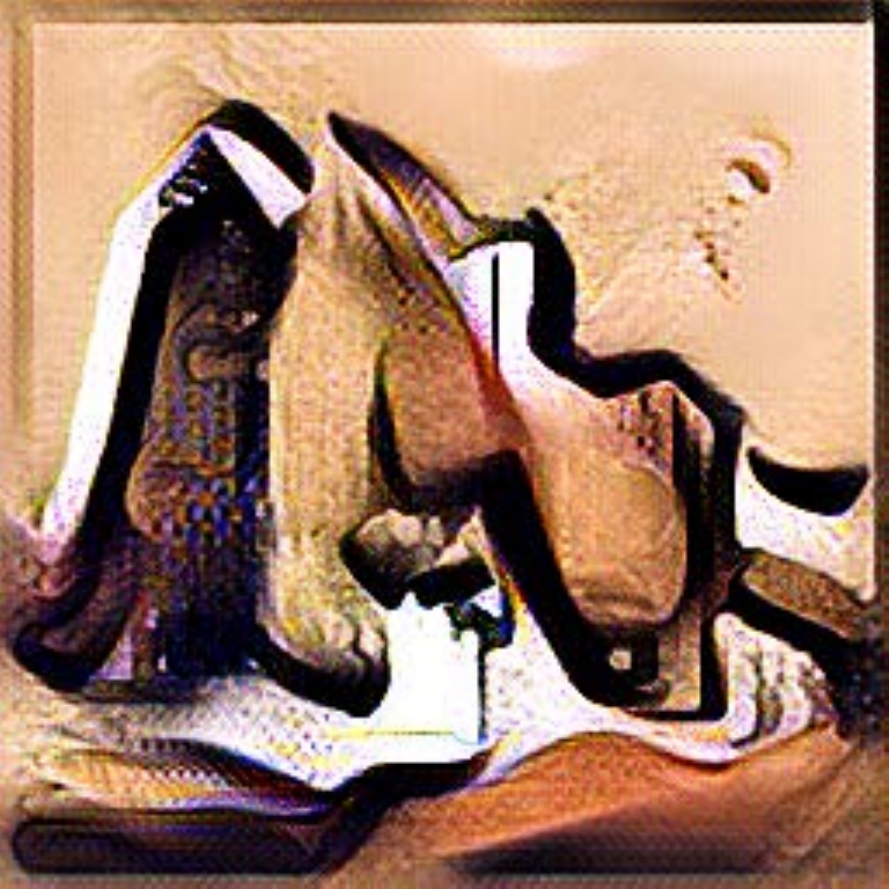}
\end{minipage}%
}%
\vfill

\subfigure{
\begin{minipage}[t]{0.2\linewidth}
\centering
\includegraphics[width=\linewidth]{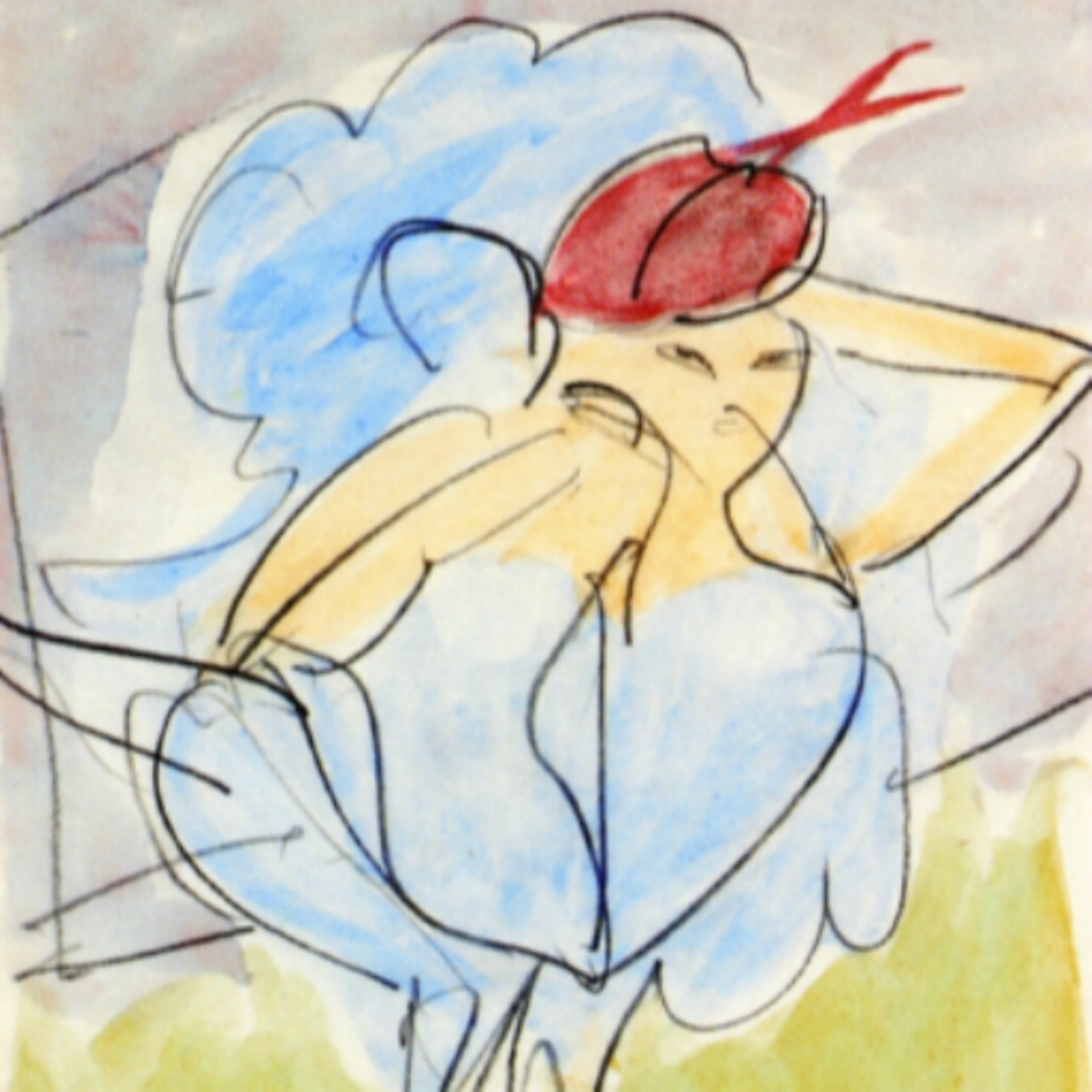}
\end{minipage}%
}%
\subfigure{
\begin{minipage}[t]{0.2\linewidth}
\centering
\includegraphics[width=\linewidth]{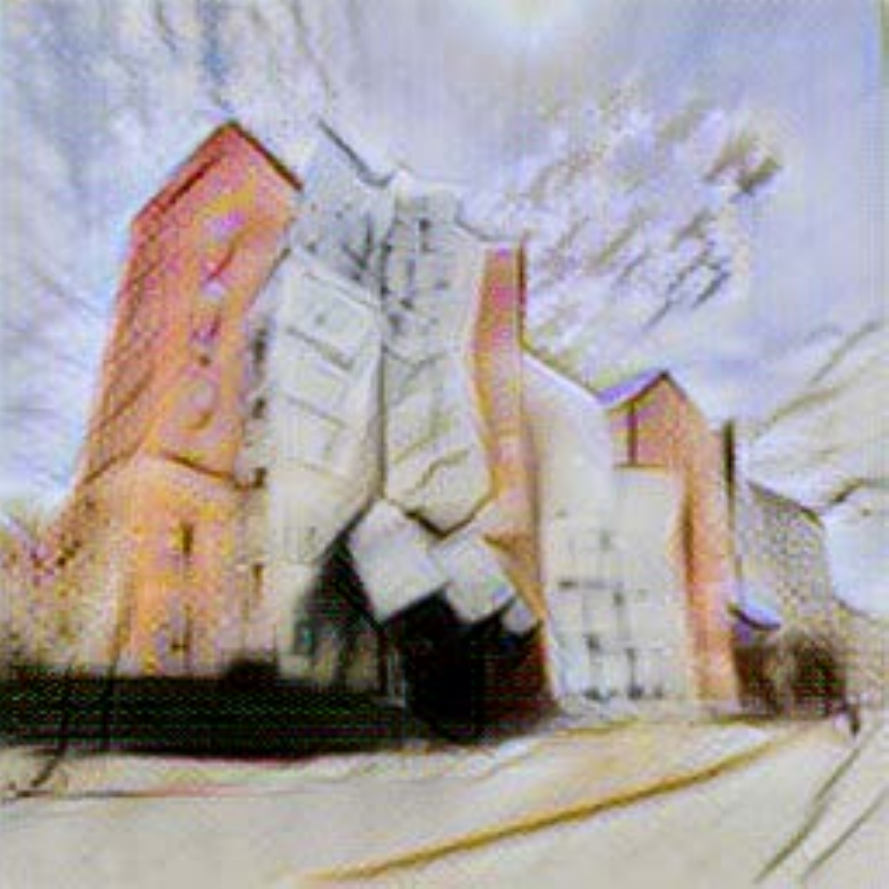}
\end{minipage}%
}%
\subfigure{
\begin{minipage}[t]{0.2\linewidth}
\centering
\includegraphics[width=\linewidth]{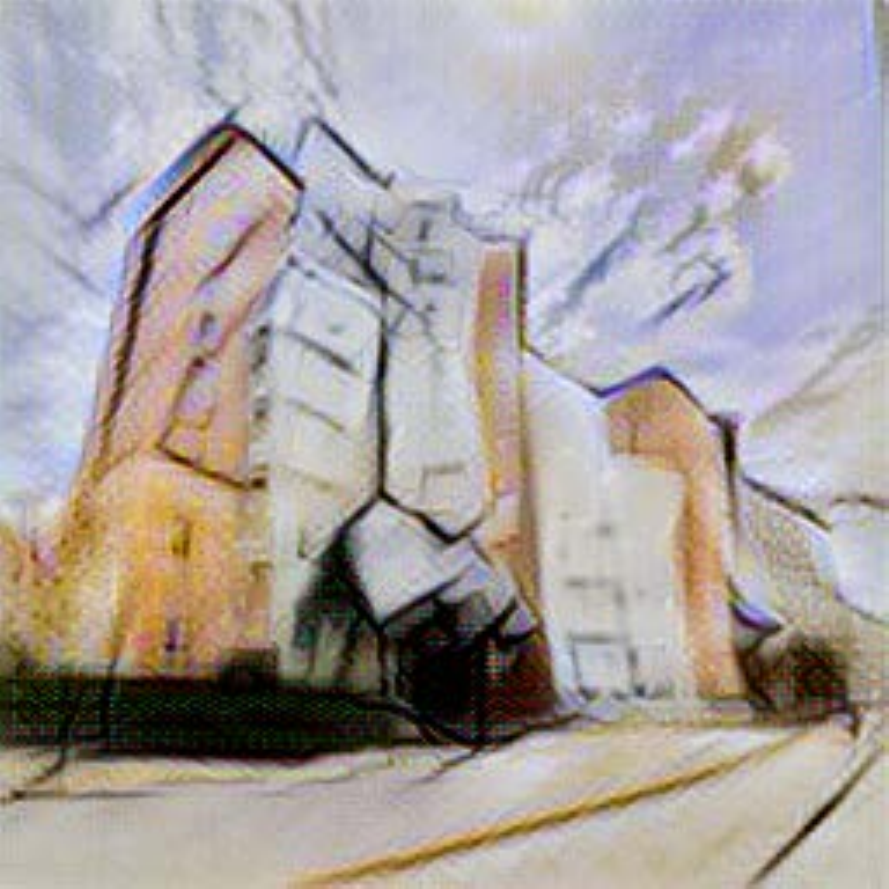}
\end{minipage}%
}%
\subfigure{
\begin{minipage}[t]{0.2\linewidth}
\centering
\includegraphics[width=\linewidth]{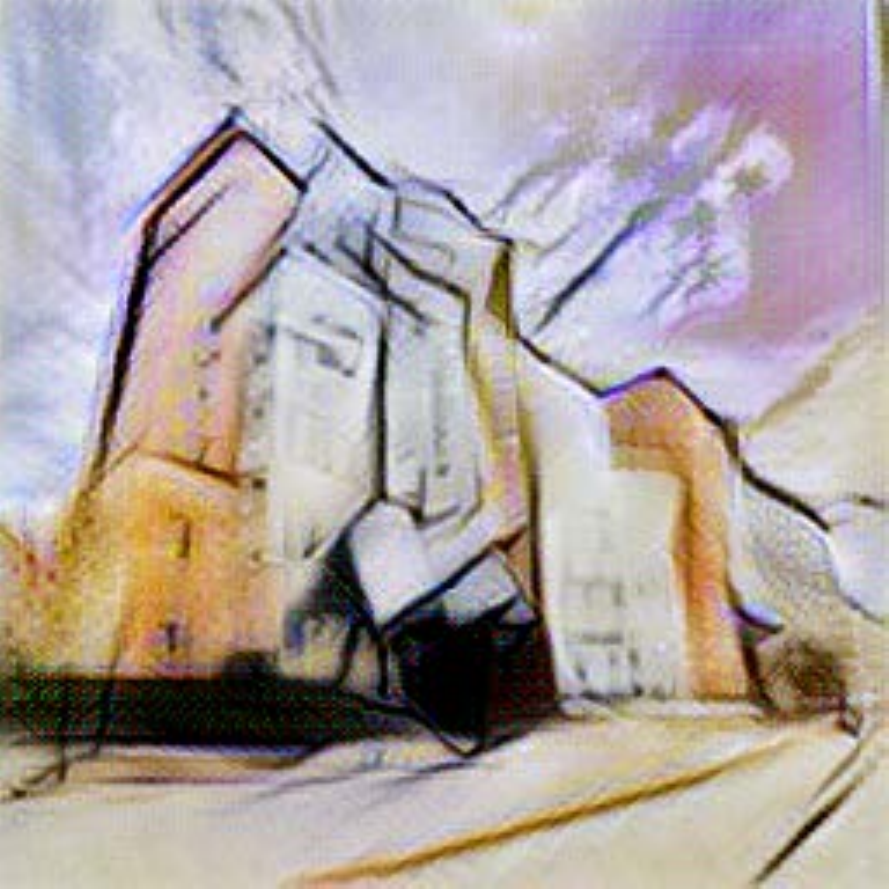}
\end{minipage}%
}%
\subfigure{
\begin{minipage}[t]{0.2\linewidth}
\centering
\includegraphics[width=\linewidth]{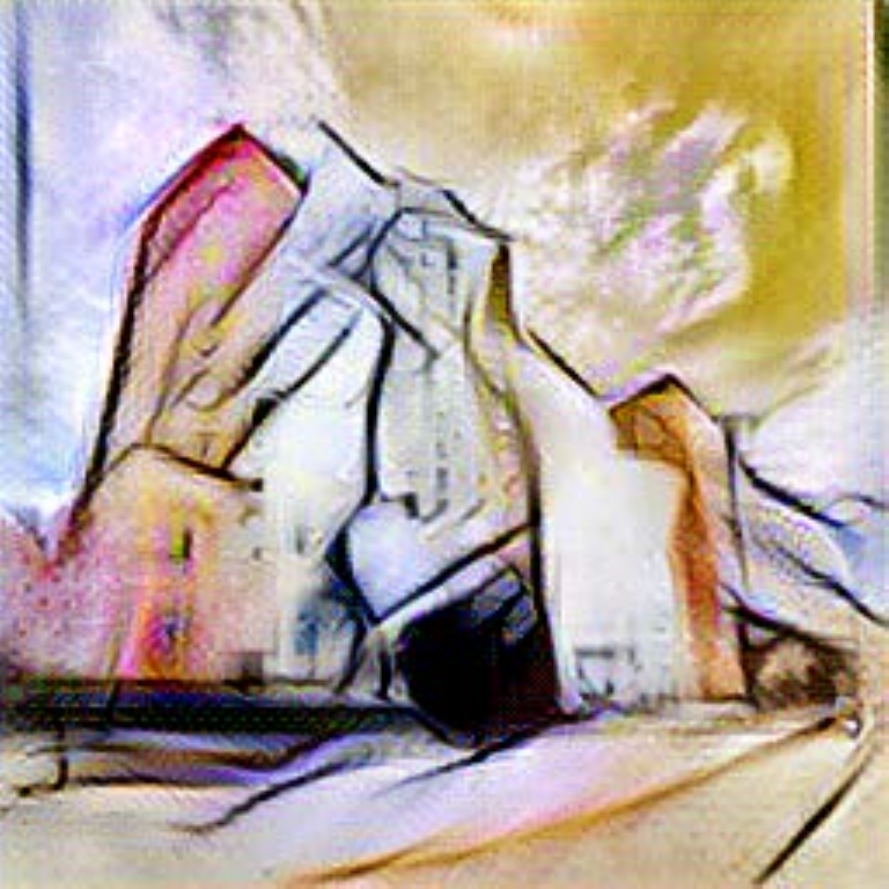}
\end{minipage}%
}%
\vfill

\subfigure{
\begin{minipage}[t]{0.2\linewidth}
\centering
\includegraphics[width=\linewidth]{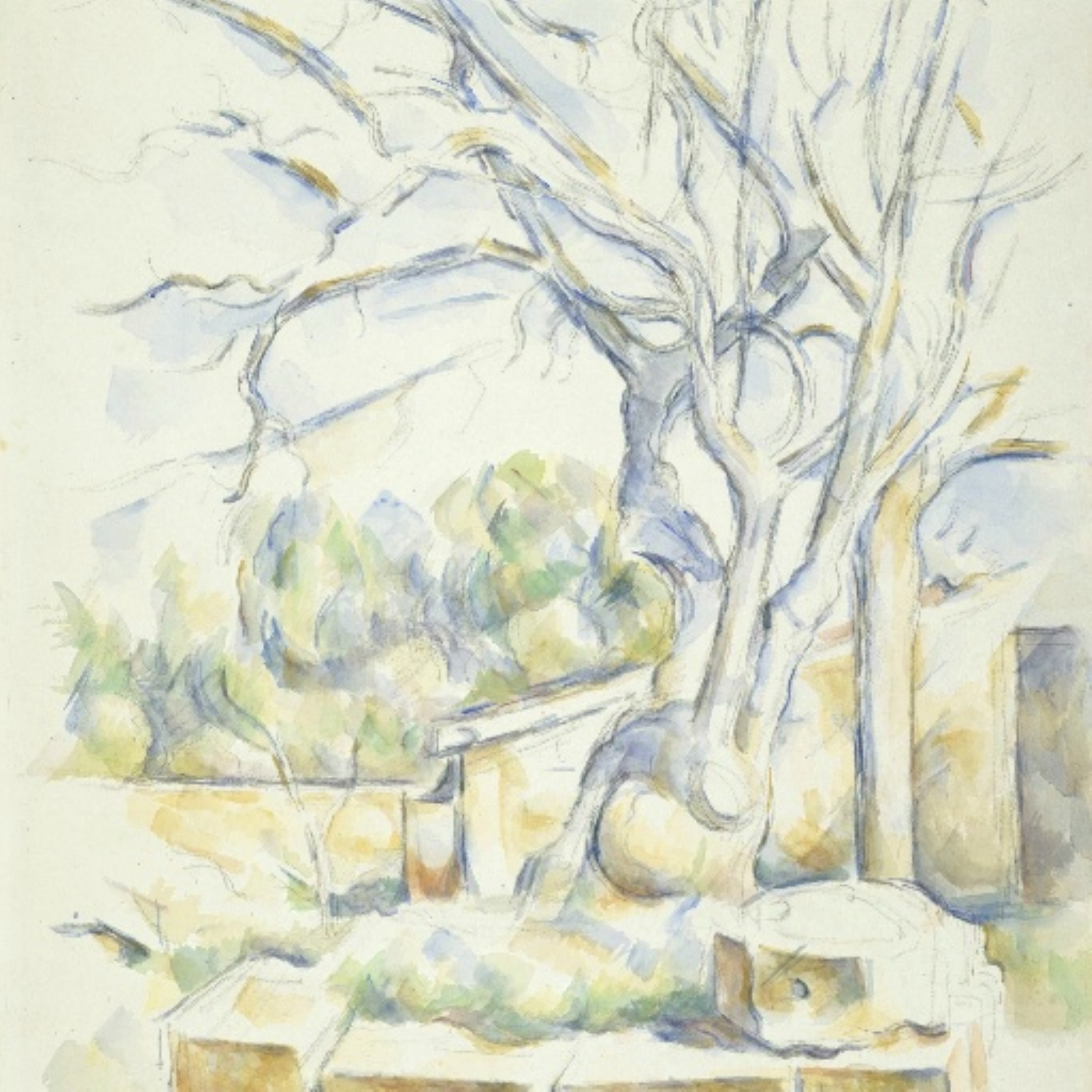}
\end{minipage}%
}%
\subfigure{
\begin{minipage}[t]{0.2\linewidth}
\centering
\includegraphics[width=\linewidth]{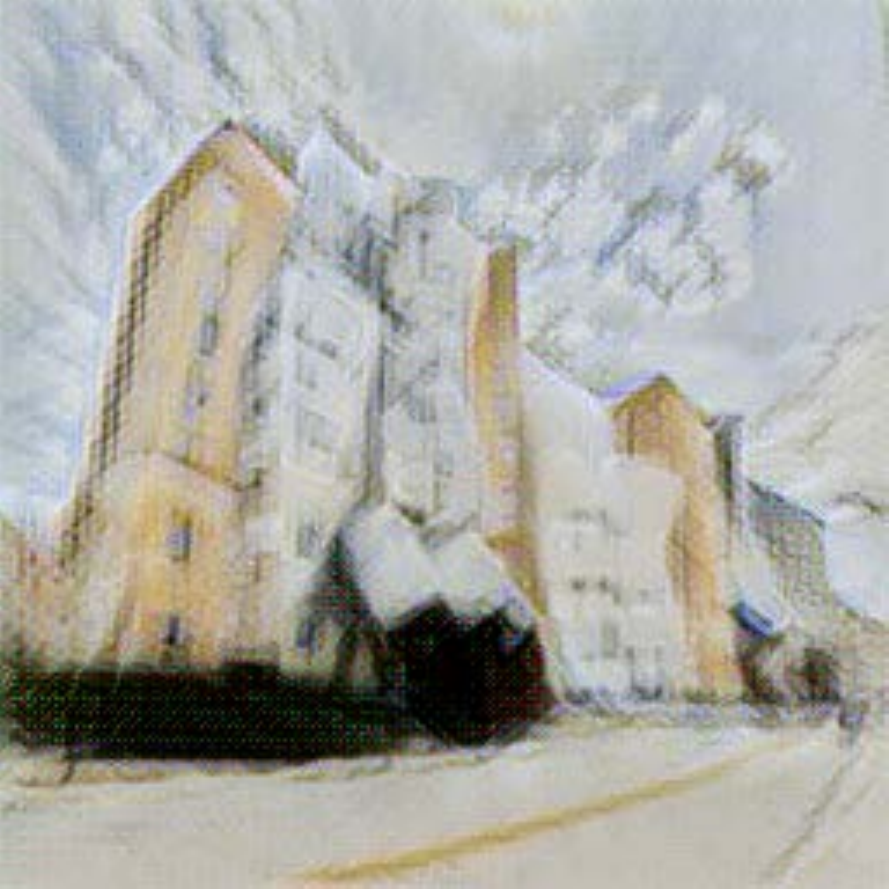}
\end{minipage}%
}%
\subfigure{
\begin{minipage}[t]{0.2\linewidth}
\centering
\includegraphics[width=\linewidth]{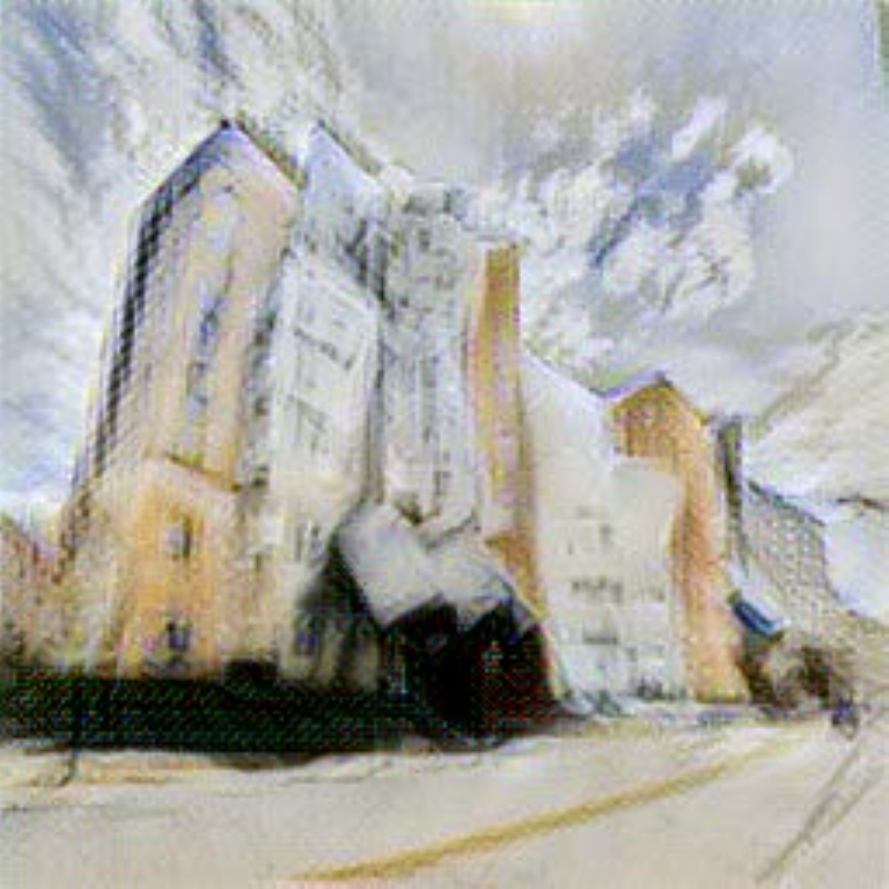}
\end{minipage}%
}%
\subfigure{
\begin{minipage}[t]{0.2\linewidth}
\centering
\includegraphics[width=\linewidth]{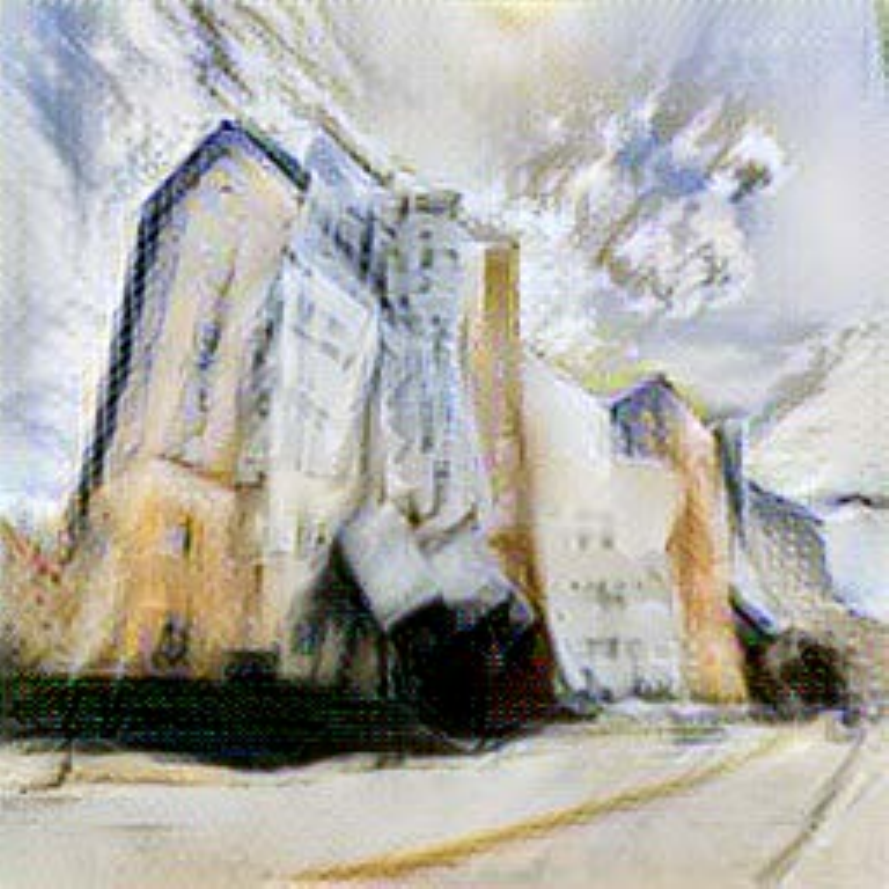}
\end{minipage}%
}%
\subfigure{
\begin{minipage}[t]{0.2\linewidth}
\centering
\includegraphics[width=\linewidth]{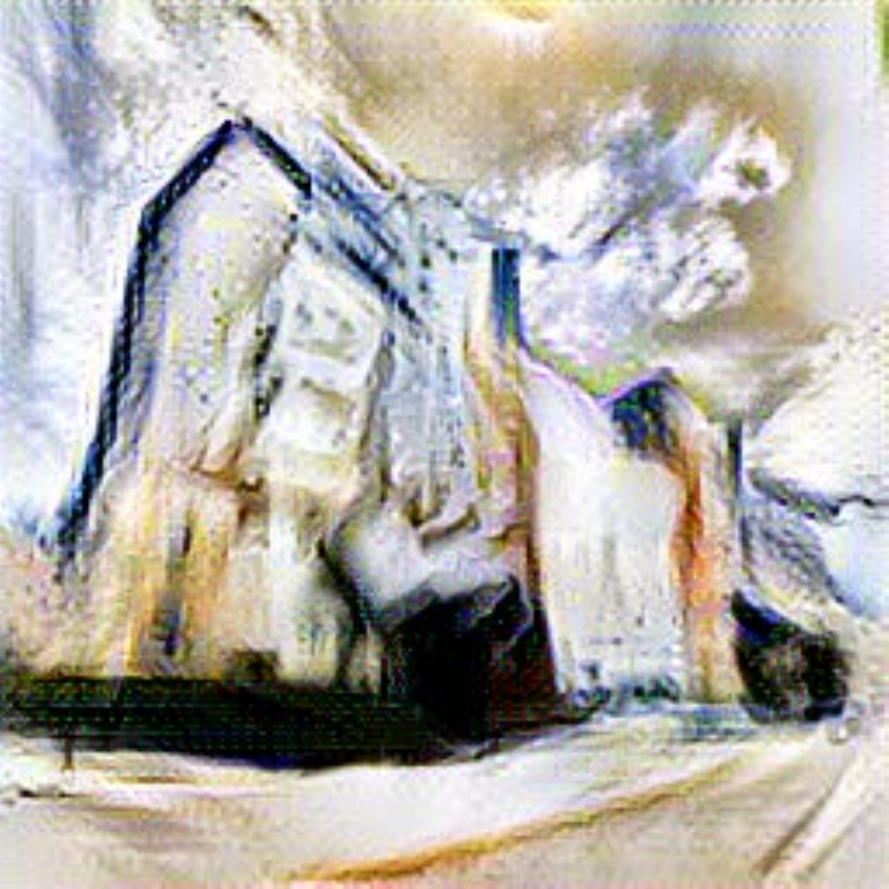}
\end{minipage}%
}%
\vfill

\subfigure{
\begin{minipage}[t]{0.2\linewidth}
\centering
\includegraphics[width=\linewidth]{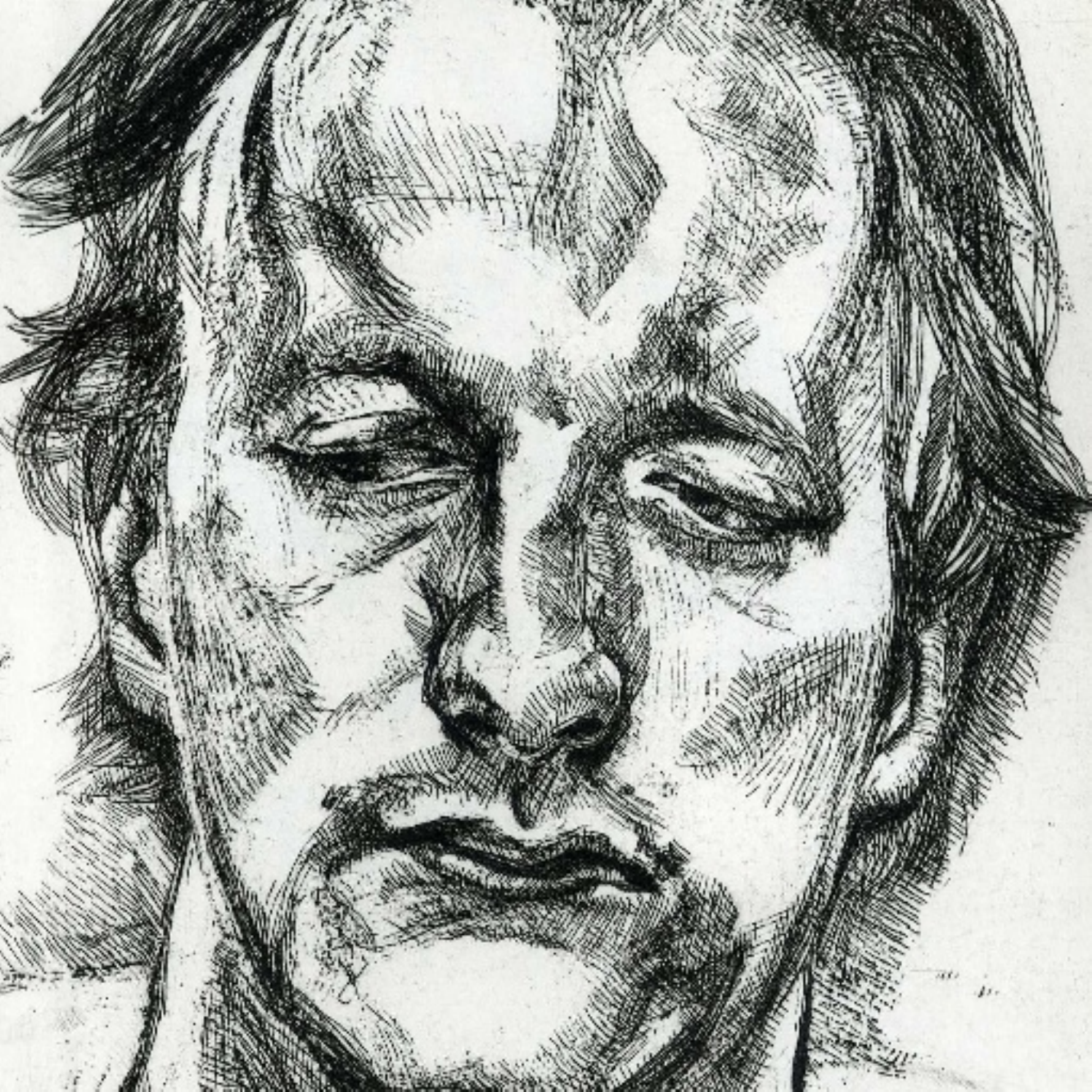}
\end{minipage}%
}%
\subfigure{
\begin{minipage}[t]{0.2\linewidth}
\centering
\includegraphics[width=\linewidth]{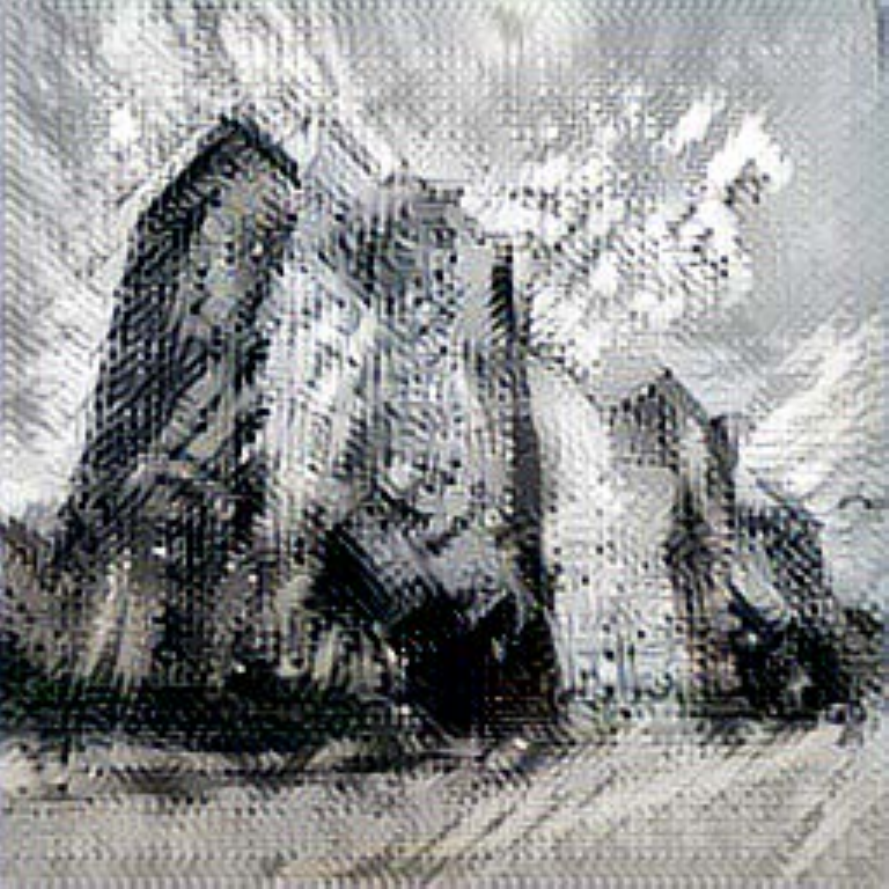}
\end{minipage}%
}%
\subfigure{
\begin{minipage}[t]{0.2\linewidth}
\centering
\includegraphics[width=\linewidth]{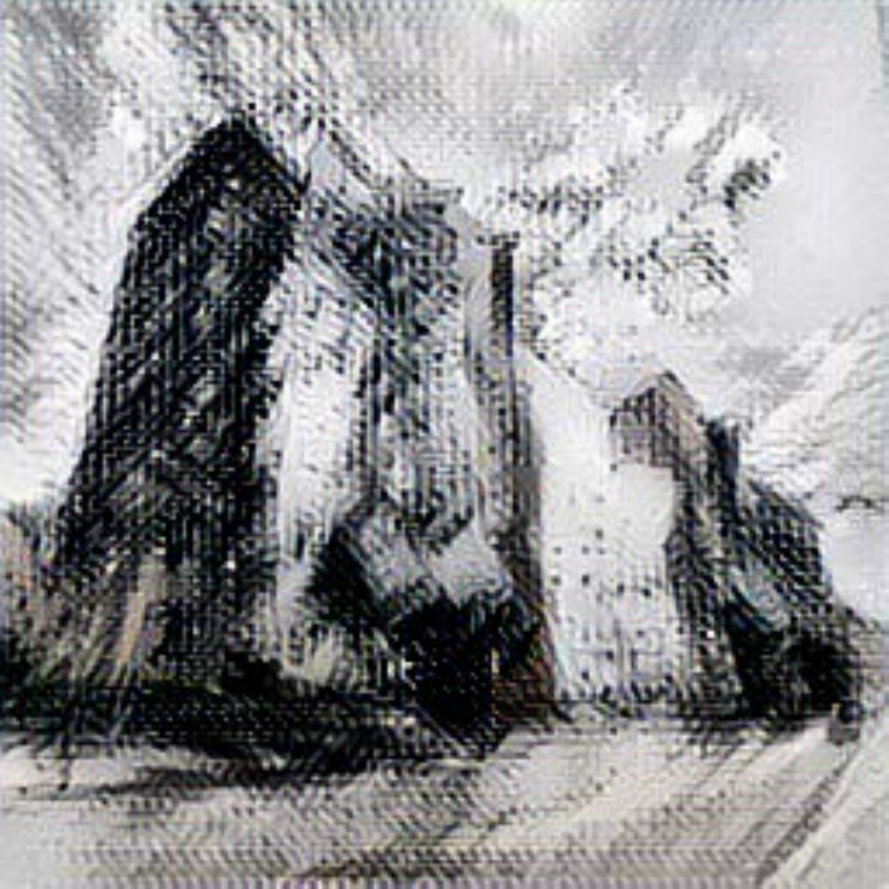}
\end{minipage}%
}%
\subfigure{
\begin{minipage}[t]{0.2\linewidth}
\centering
\includegraphics[width=\linewidth]{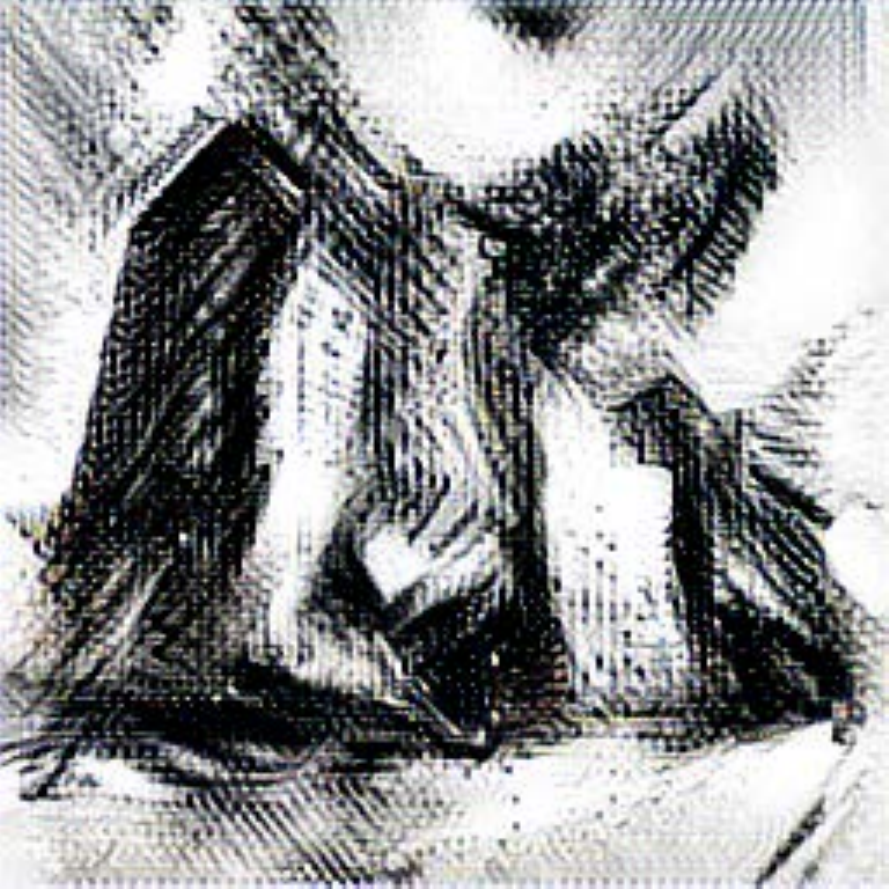}
\end{minipage}%
}%
\subfigure{
\begin{minipage}[t]{0.2\linewidth}
\centering
\includegraphics[width=\linewidth]{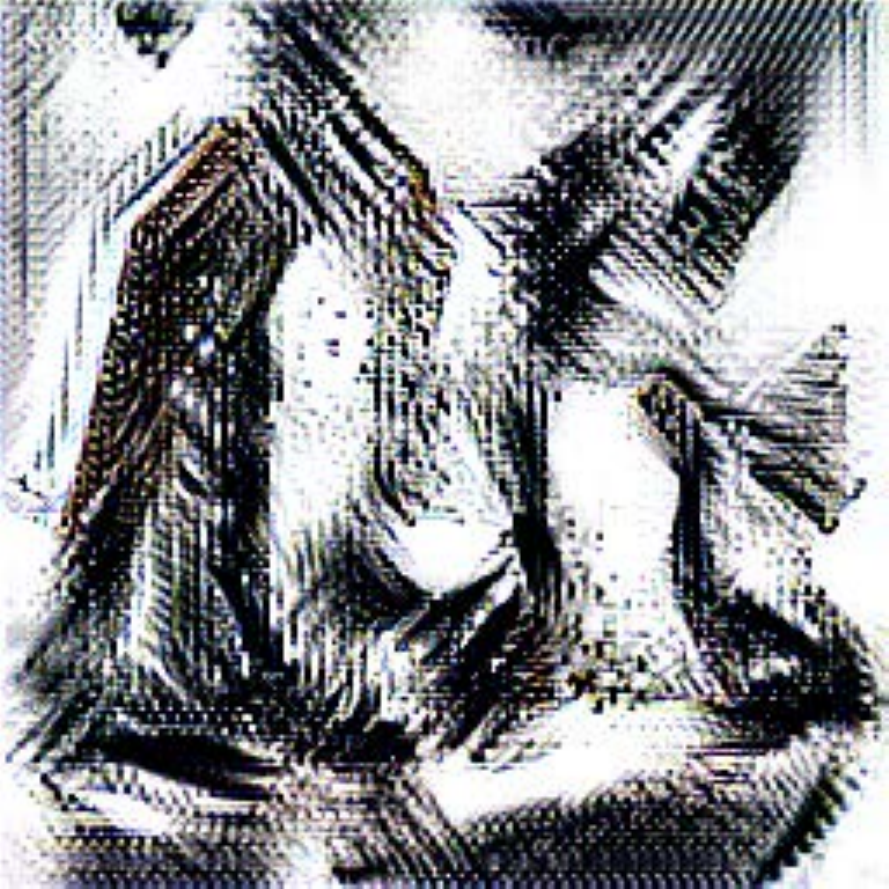}
\end{minipage}%
}%
\centering
\caption{The styled image with stroke basis intervened using spectrum based methods. The left most of each row shows the style images (From top to bottom: The Great Wave off Kanagawa - Katsushika Hokusai; Composition - Alberto Magnelli; Dancer - Ernst Ludwig Kirchner; Pistachio Tree in the Courtyard of the Chateau Noir - Paul Cezanne; Potrait - Lucian Freud). From left to right of each row, the effect of stroke is increasingly amplified.}
\label{spectrumIntervention-1}
\end{figure*}

\begin{figure*}[htb]
\centering
\subfigure{
\begin{minipage}[t]{0.2\linewidth}
\centering
\includegraphics[width=\linewidth]{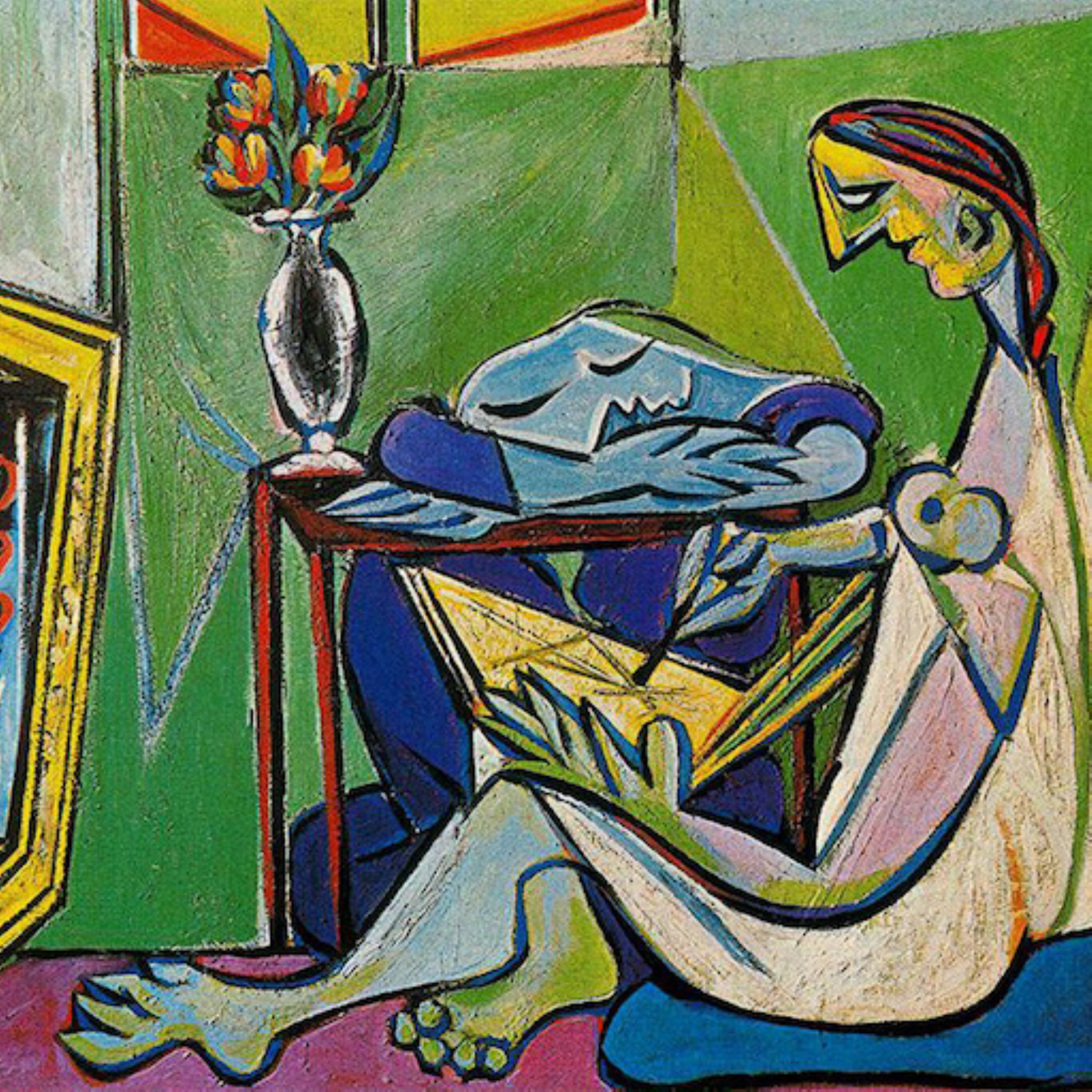}
\end{minipage}%
}%
\subfigure{
\begin{minipage}[t]{0.2\linewidth}
\centering
\includegraphics[width=\linewidth]{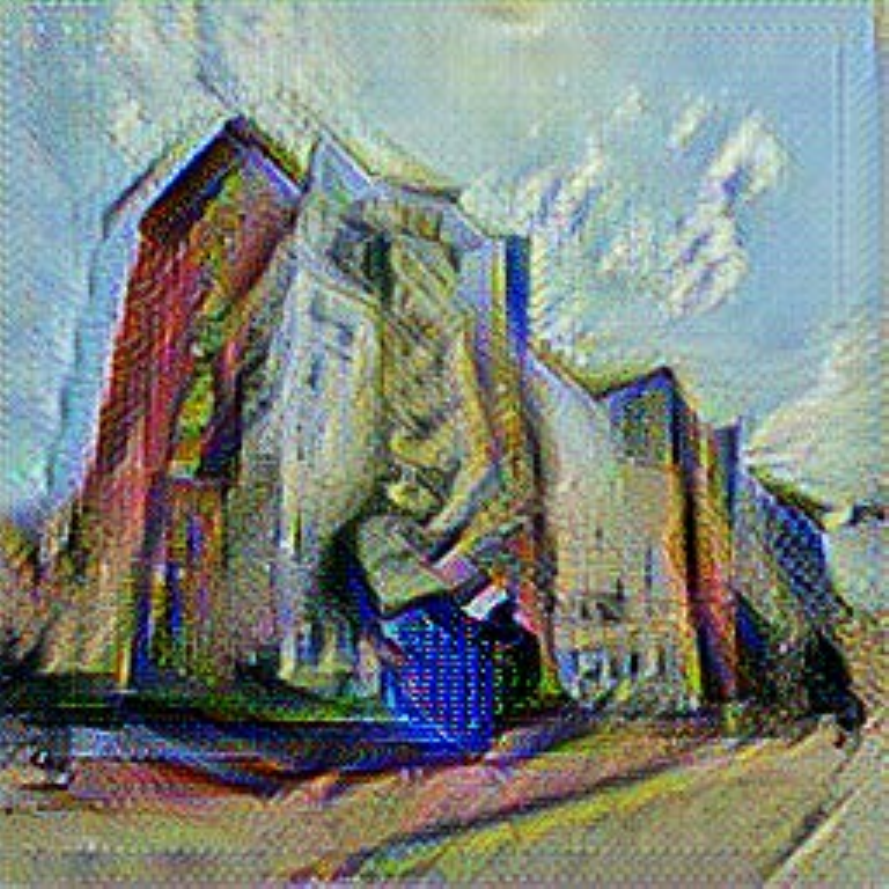}
\end{minipage}%
}%
\subfigure{
\begin{minipage}[t]{0.2\linewidth}
\centering
\includegraphics[width=\linewidth]{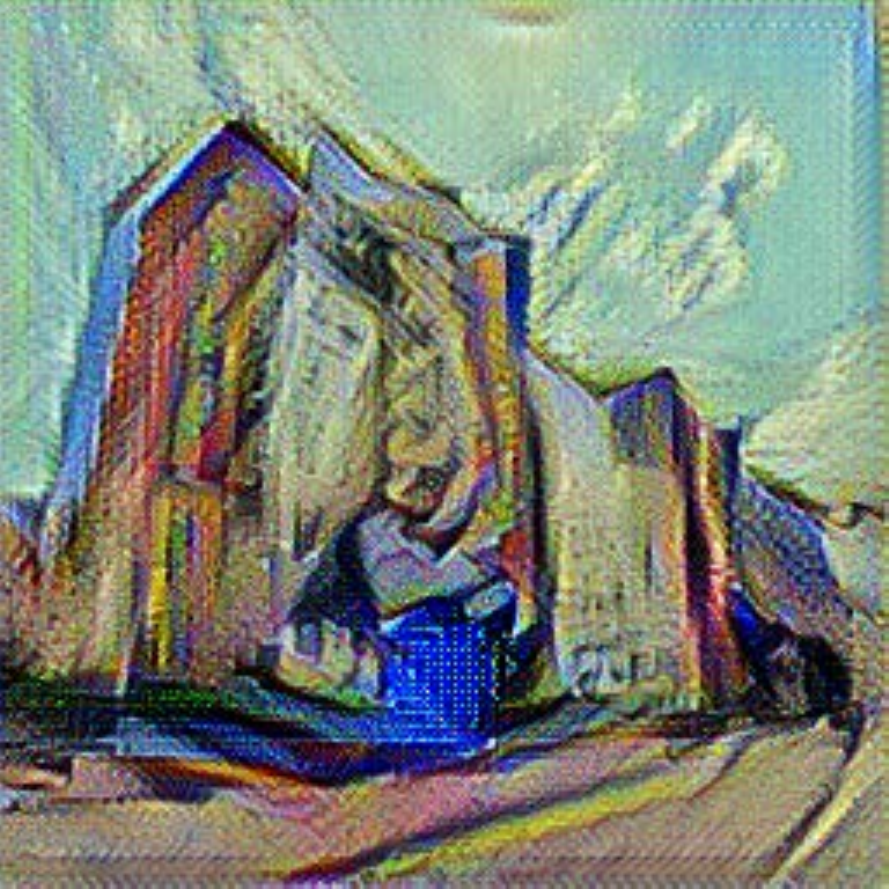}
\end{minipage}%
}%
\subfigure{
\begin{minipage}[t]{0.2\linewidth}
\centering
\includegraphics[width=\linewidth]{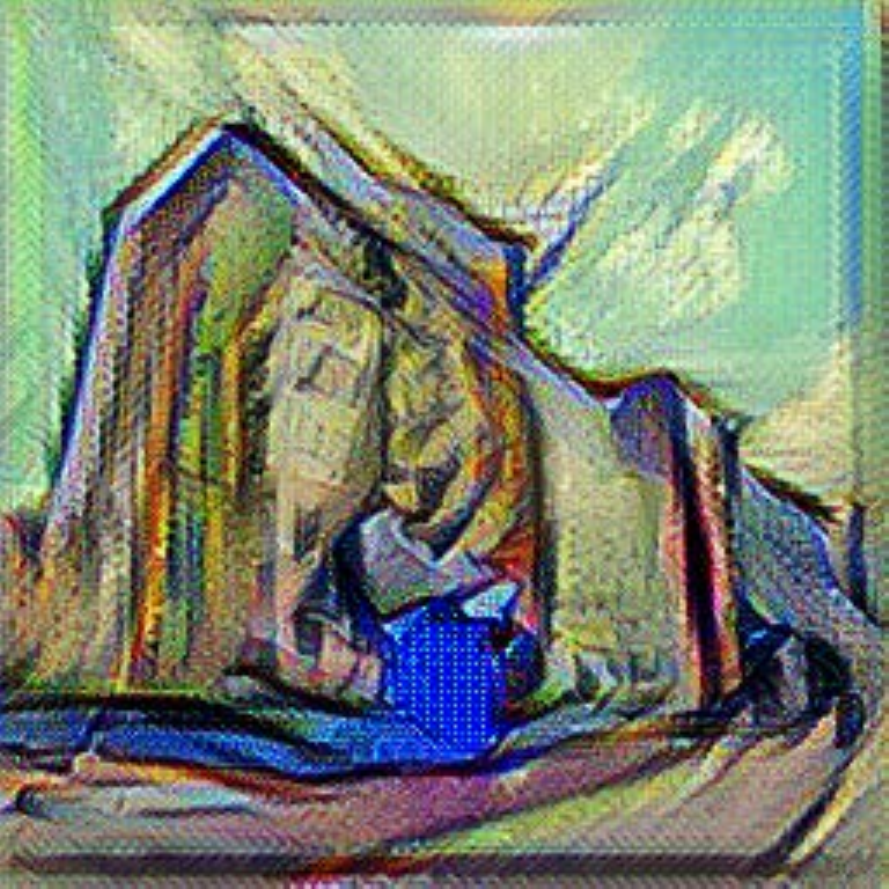}
\end{minipage}%
}%
\subfigure{
\begin{minipage}[t]{0.2\linewidth}
\centering
\includegraphics[width=\linewidth]{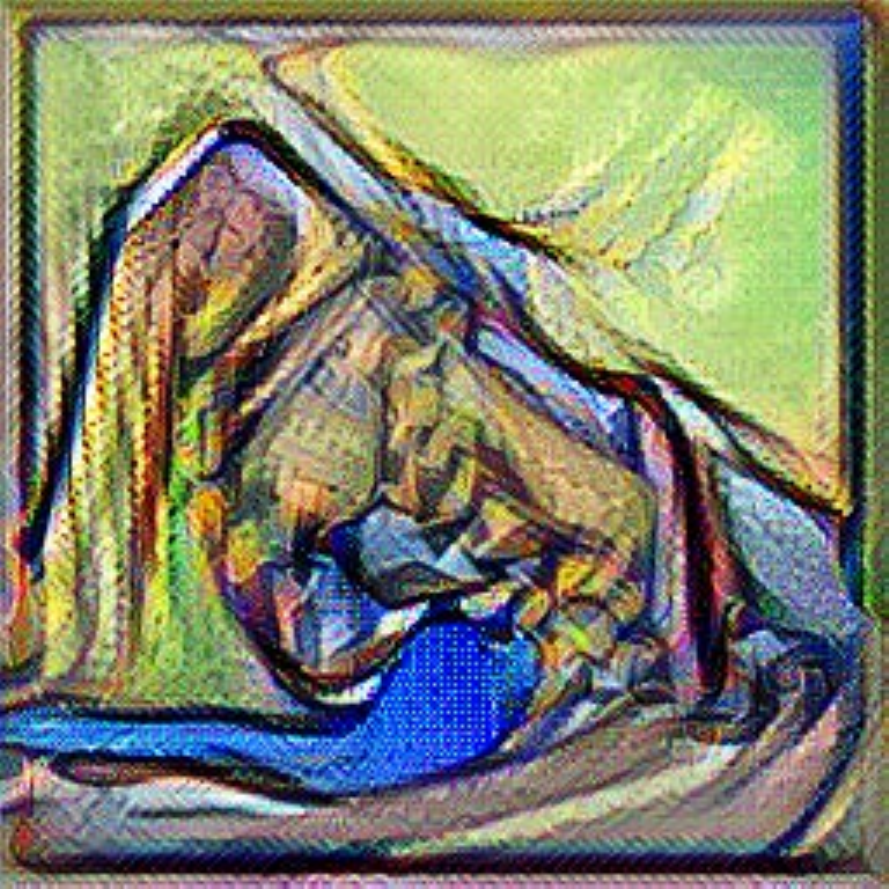}
\end{minipage}%
}%
\vfill

\subfigure{
\begin{minipage}[t]{0.2\linewidth}
\centering
\includegraphics[width=\linewidth]{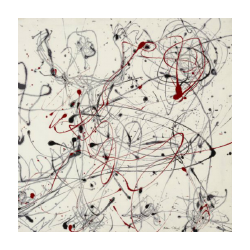}
\end{minipage}%
}%
\subfigure{
\begin{minipage}[t]{0.2\linewidth}
\centering
\includegraphics[width=\linewidth]{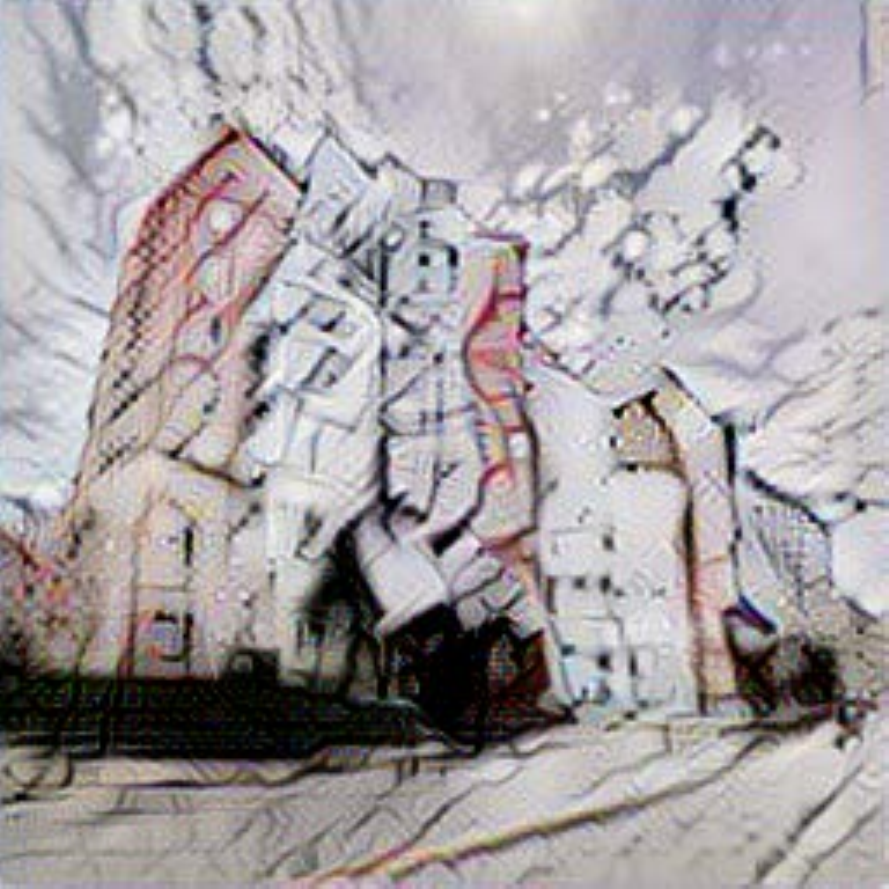}
\end{minipage}%
}%
\subfigure{
\begin{minipage}[t]{0.2\linewidth}
\centering
\includegraphics[width=\linewidth]{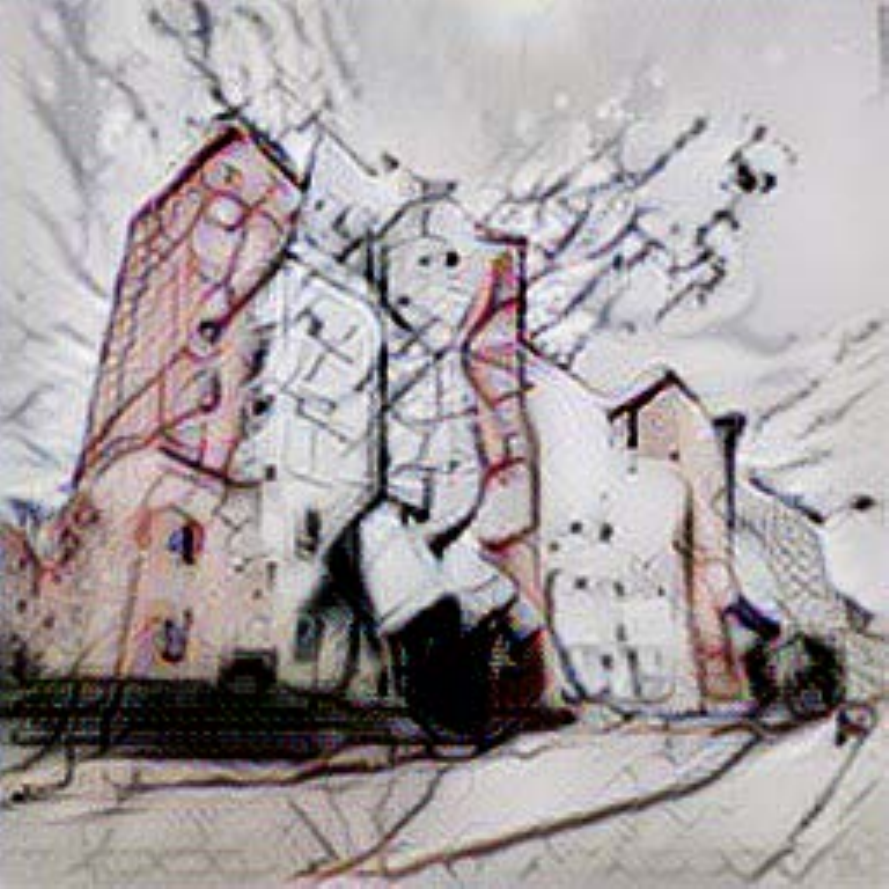}
\end{minipage}%
}%
\subfigure{
\begin{minipage}[t]{0.2\linewidth}
\centering
\includegraphics[width=\linewidth]{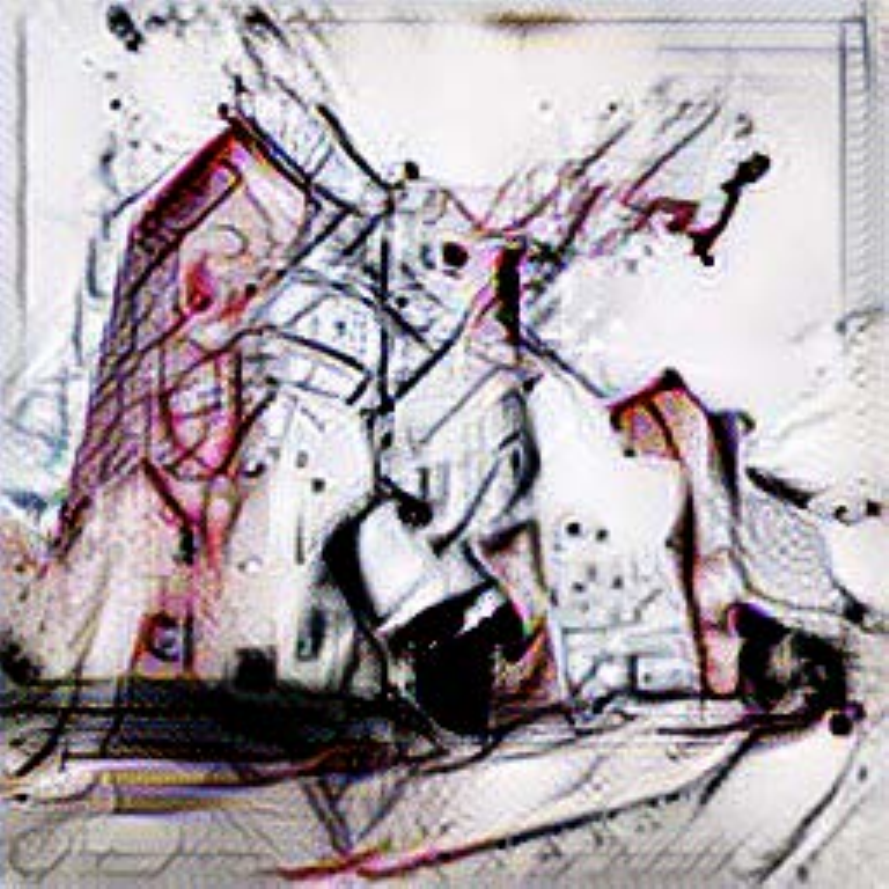}
\end{minipage}%
}%
\subfigure{
\begin{minipage}[t]{0.2\linewidth}
\centering
\includegraphics[width=\linewidth]{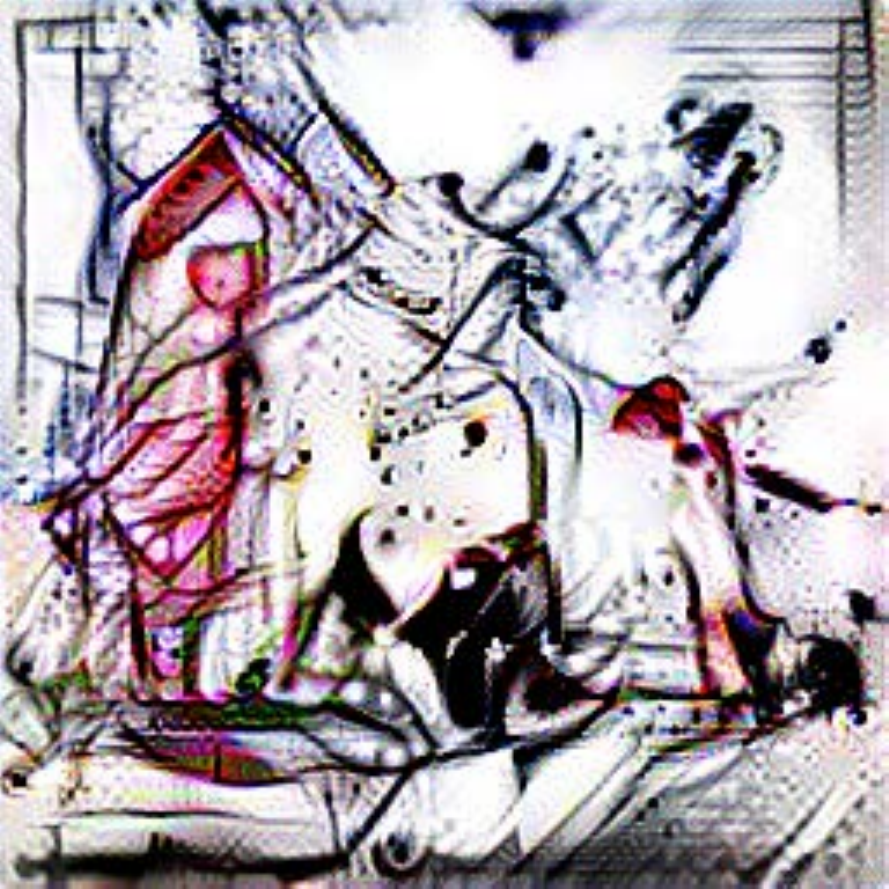}
\end{minipage}%
}%
\vfill

\subfigure{
\begin{minipage}[t]{0.2\linewidth}
\centering
\includegraphics[width=\linewidth]{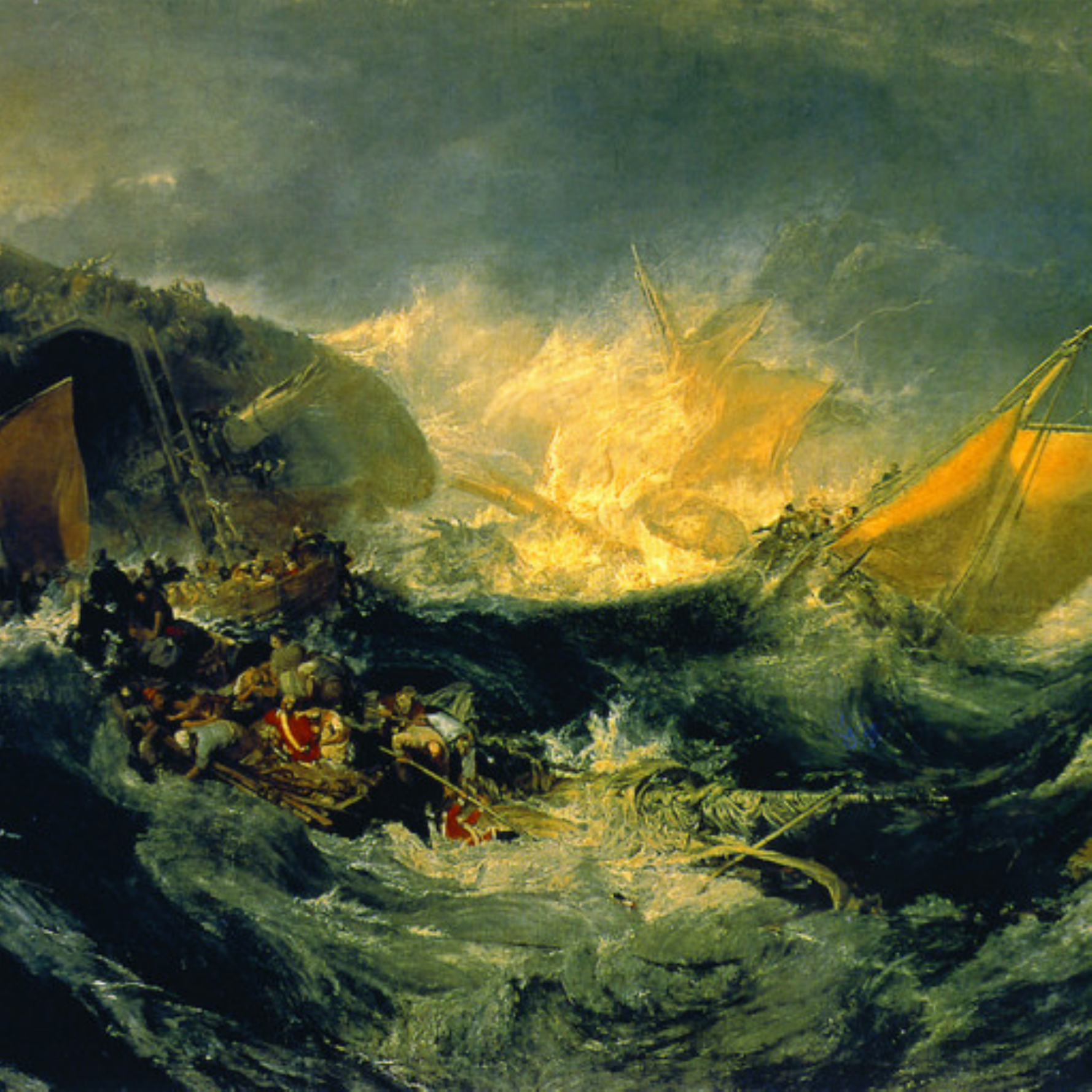}
\end{minipage}%
}%
\subfigure{
\begin{minipage}[t]{0.2\linewidth}
\centering
\includegraphics[width=\linewidth]{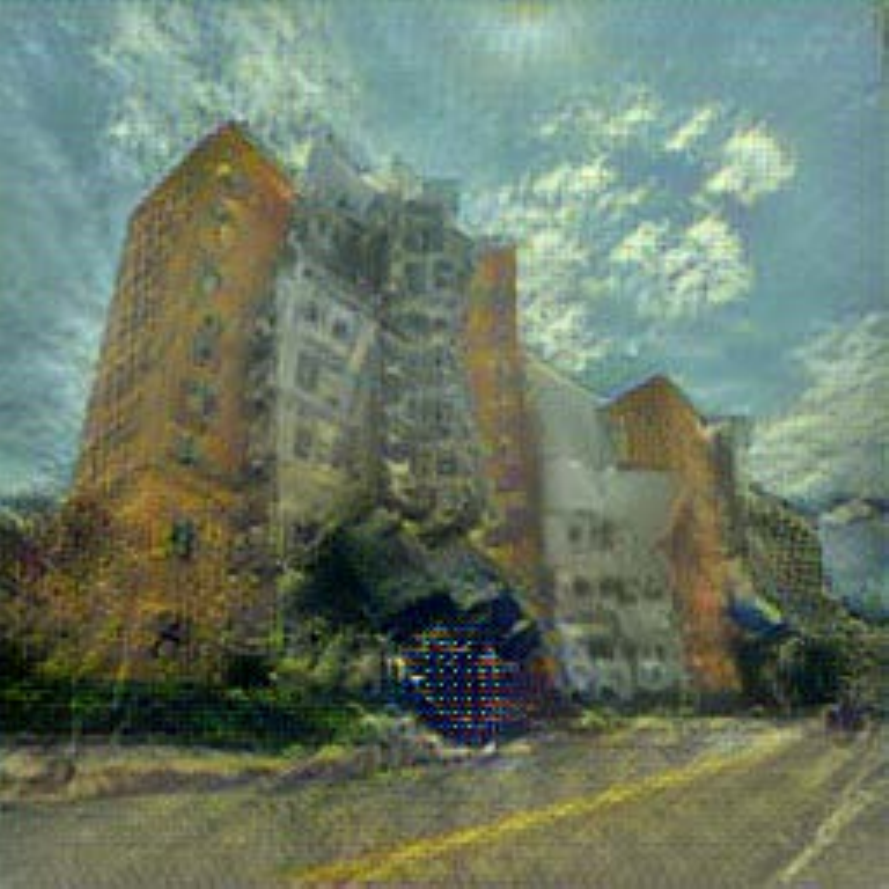}
\end{minipage}%
}%
\subfigure{
\begin{minipage}[t]{0.2\linewidth}
\centering
\includegraphics[width=\linewidth]{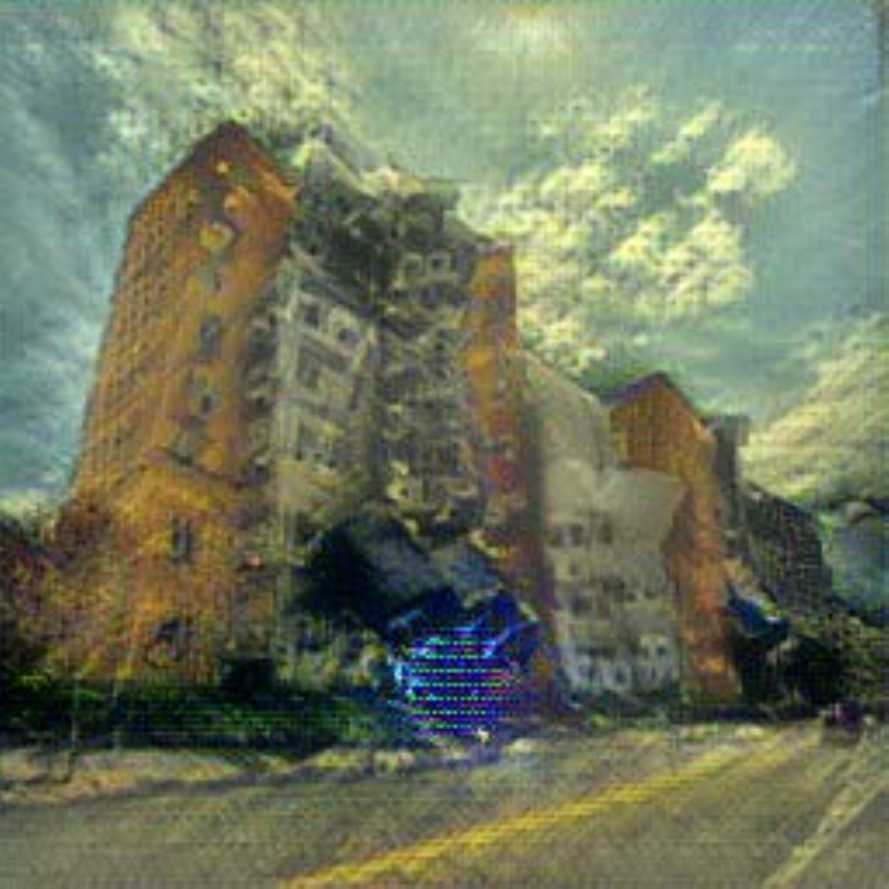}
\end{minipage}%
}%
\subfigure{
\begin{minipage}[t]{0.2\linewidth}
\centering
\includegraphics[width=\linewidth]{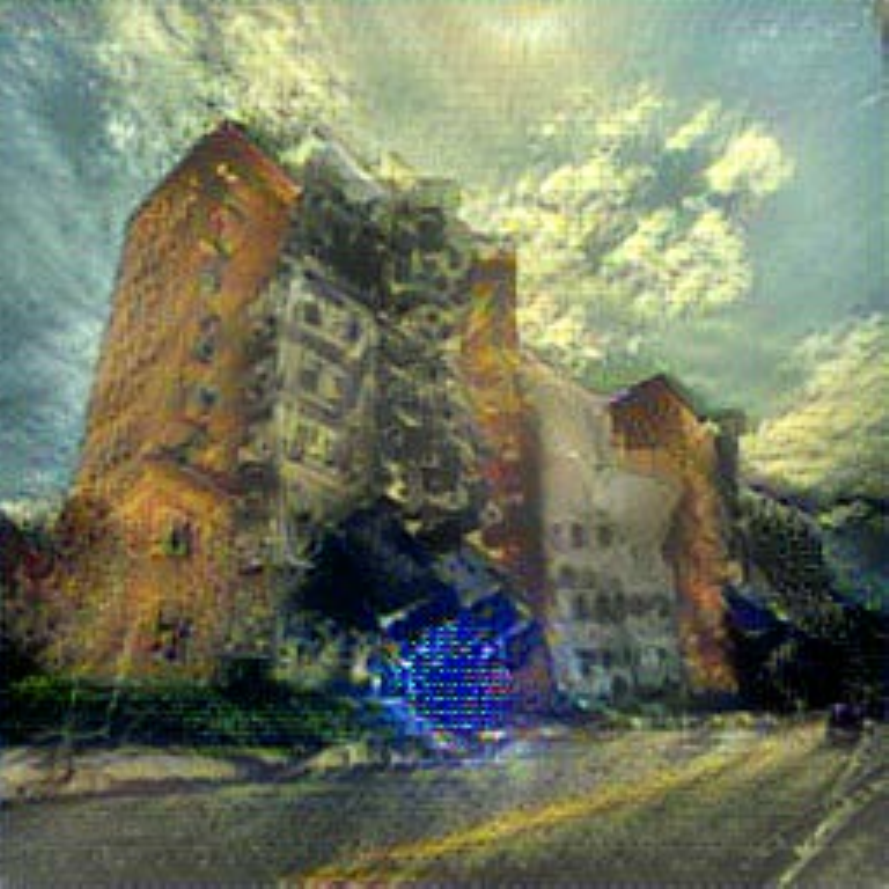}
\end{minipage}%
}%
\subfigure{
\begin{minipage}[t]{0.2\linewidth}
\centering
\includegraphics[width=\linewidth]{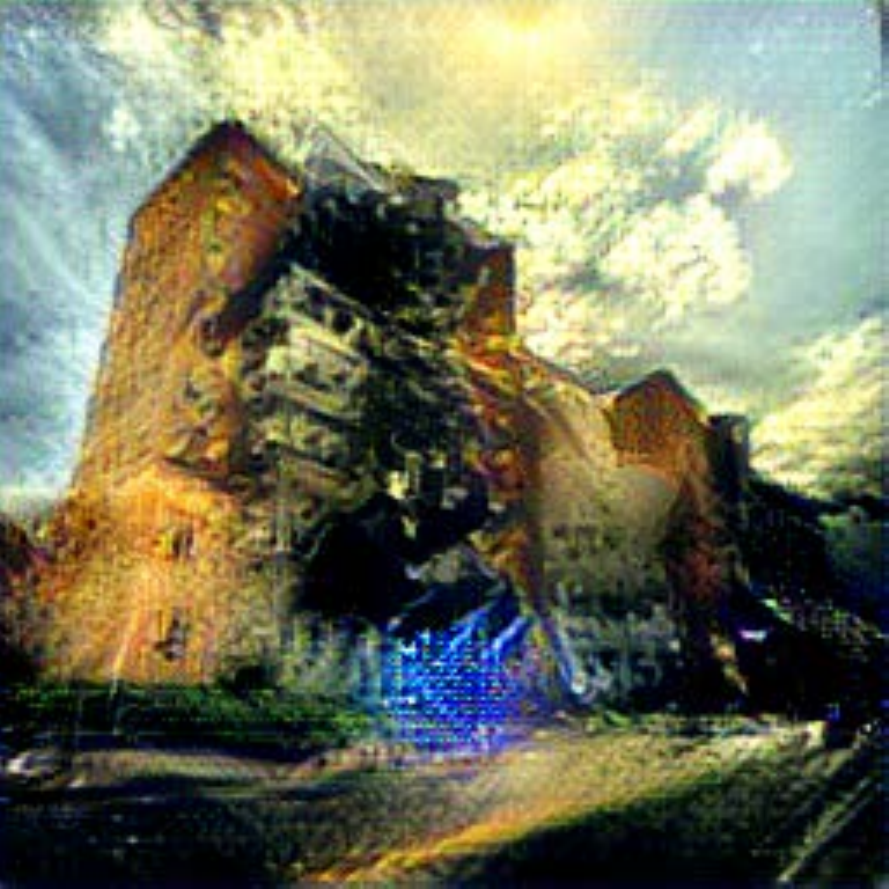}
\end{minipage}%
}%
\vfill

\subfigure{
\begin{minipage}[t]{0.2\linewidth}
\centering
\includegraphics[width=\linewidth]{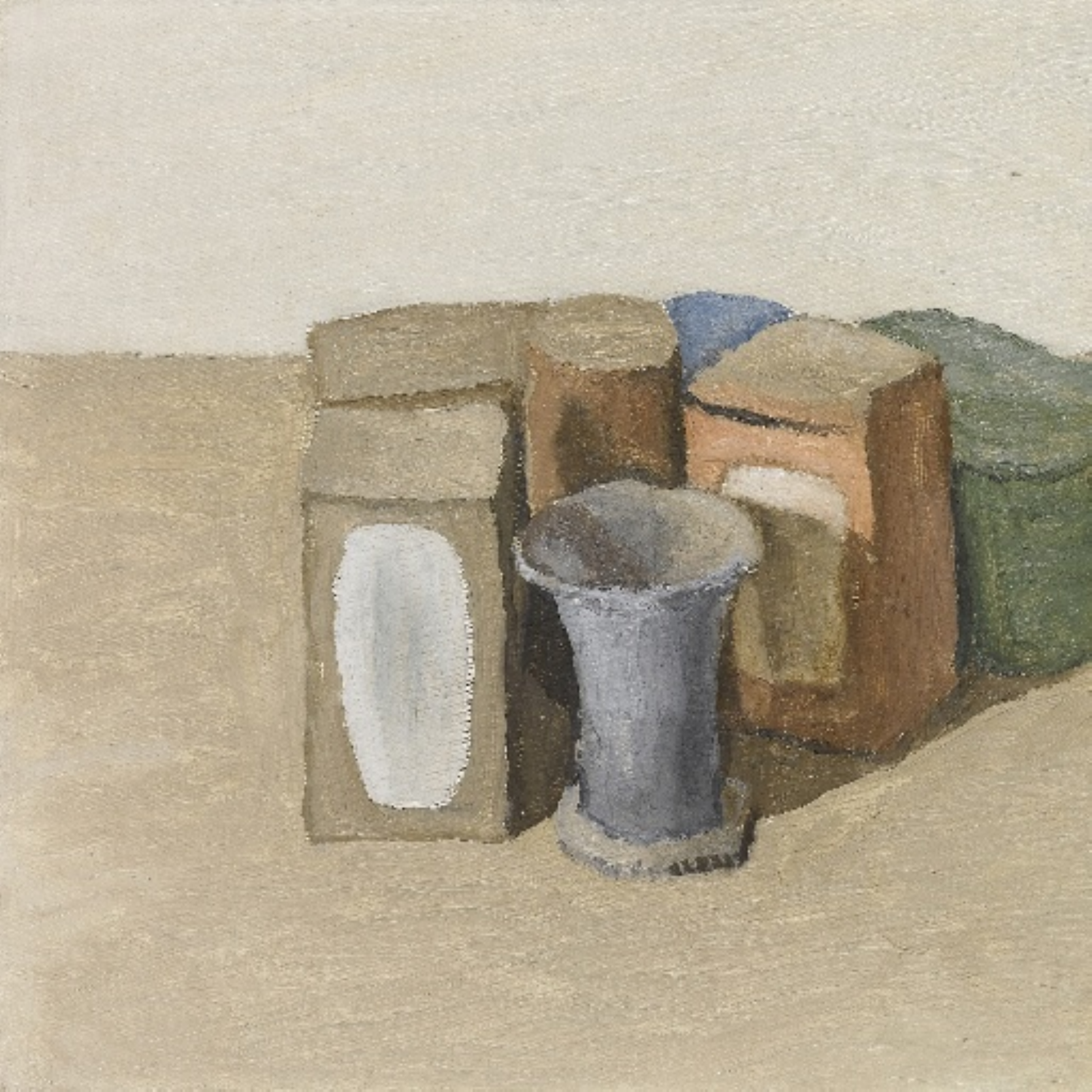}
\end{minipage}%
}%
\subfigure{
\begin{minipage}[t]{0.2\linewidth}
\centering
\includegraphics[width=\linewidth]{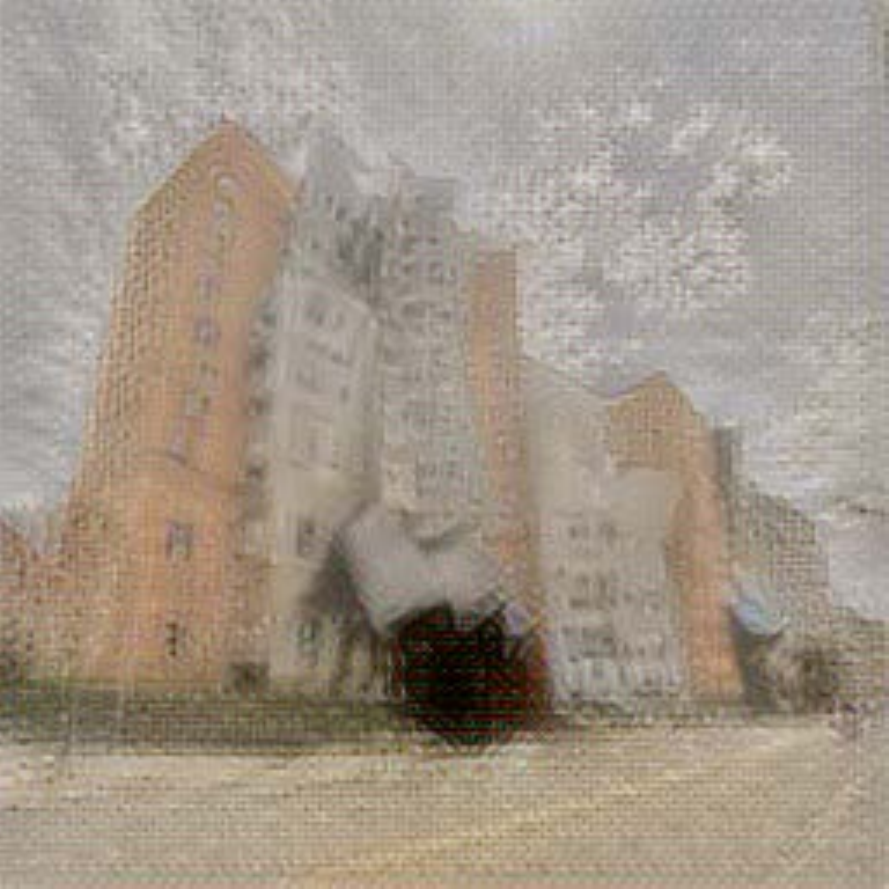}
\end{minipage}%
}%
\subfigure{
\begin{minipage}[t]{0.2\linewidth}
\centering
\includegraphics[width=\linewidth]{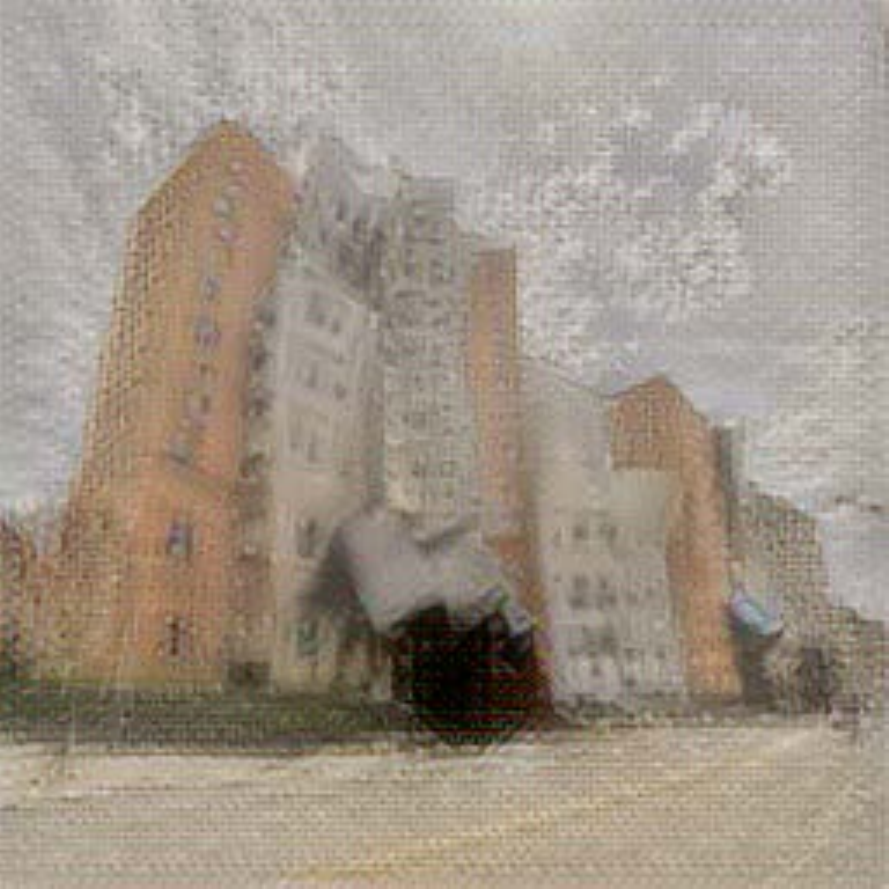}
\end{minipage}%
}%
\subfigure{
\begin{minipage}[t]{0.2\linewidth}
\centering
\includegraphics[width=\linewidth]{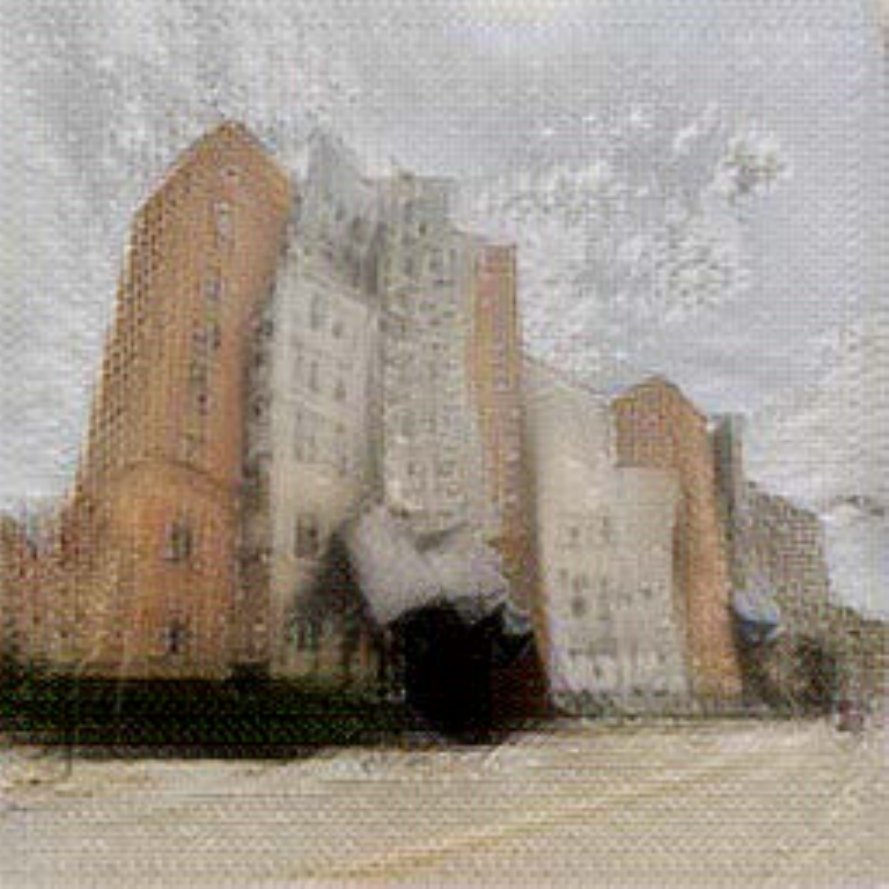}
\end{minipage}%
}%
\subfigure{
\begin{minipage}[t]{0.2\linewidth}
\centering
\includegraphics[width=\linewidth]{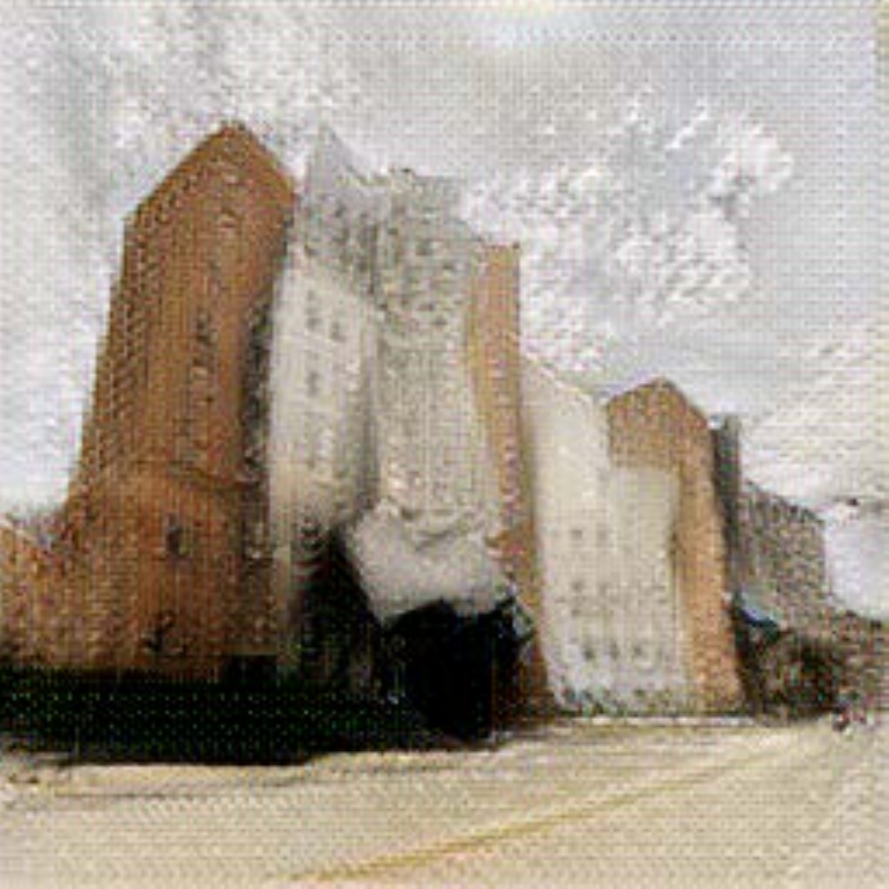}
\end{minipage}%
}%
\vfill

\subfigure{
\begin{minipage}[t]{0.2\linewidth}
\centering
\includegraphics[width=\linewidth]{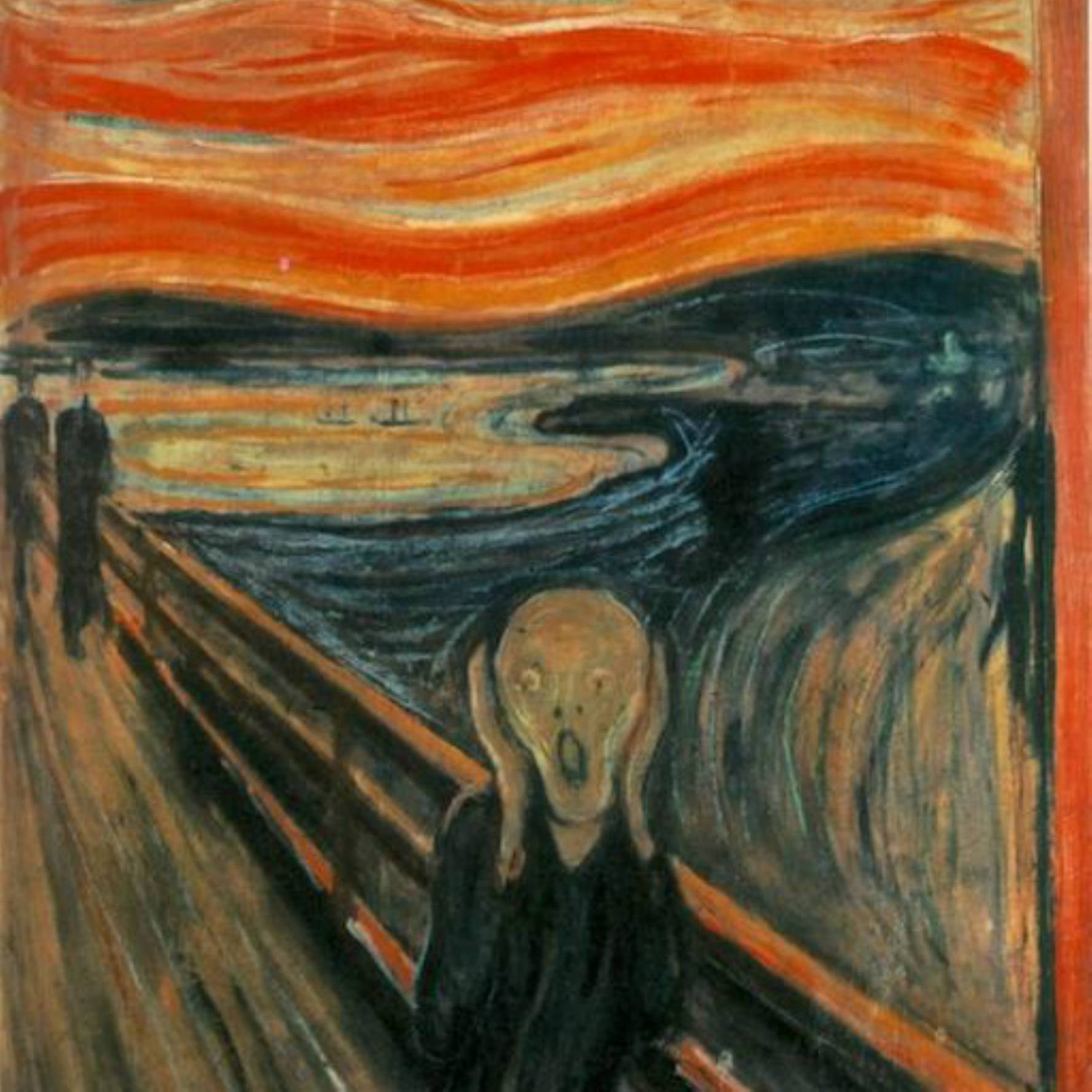}
\end{minipage}%
}%
\subfigure{
\begin{minipage}[t]{0.2\linewidth}
\centering
\includegraphics[width=\linewidth]{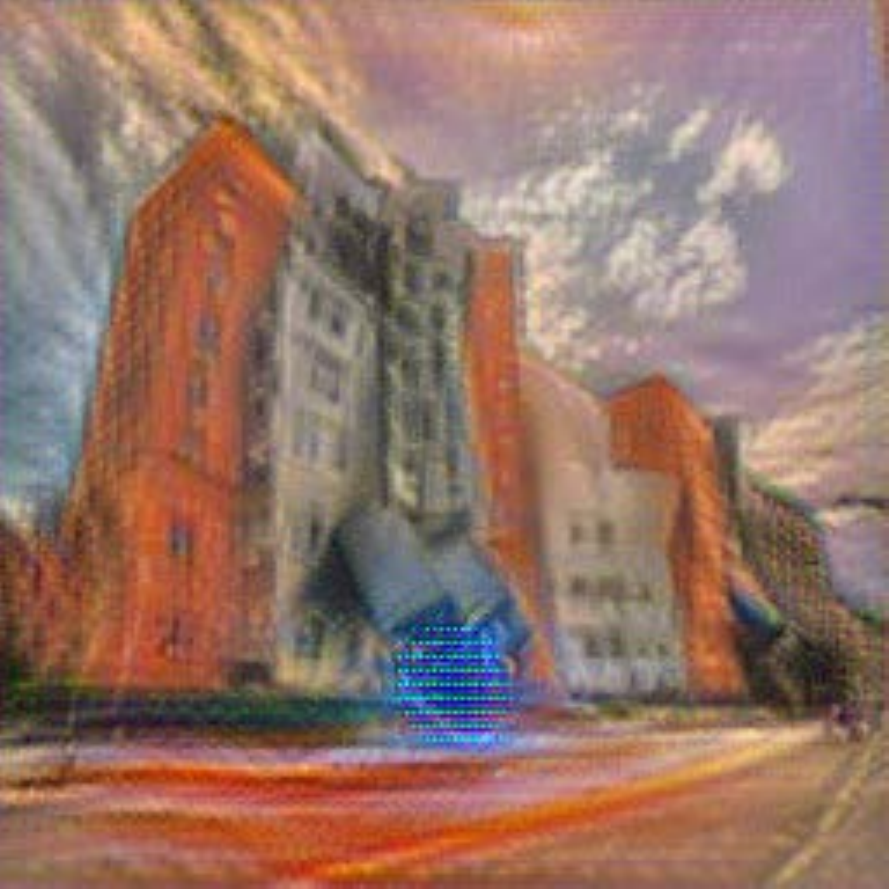}
\end{minipage}%
}%
\subfigure{
\begin{minipage}[t]{0.2\linewidth}
\centering
\includegraphics[width=\linewidth]{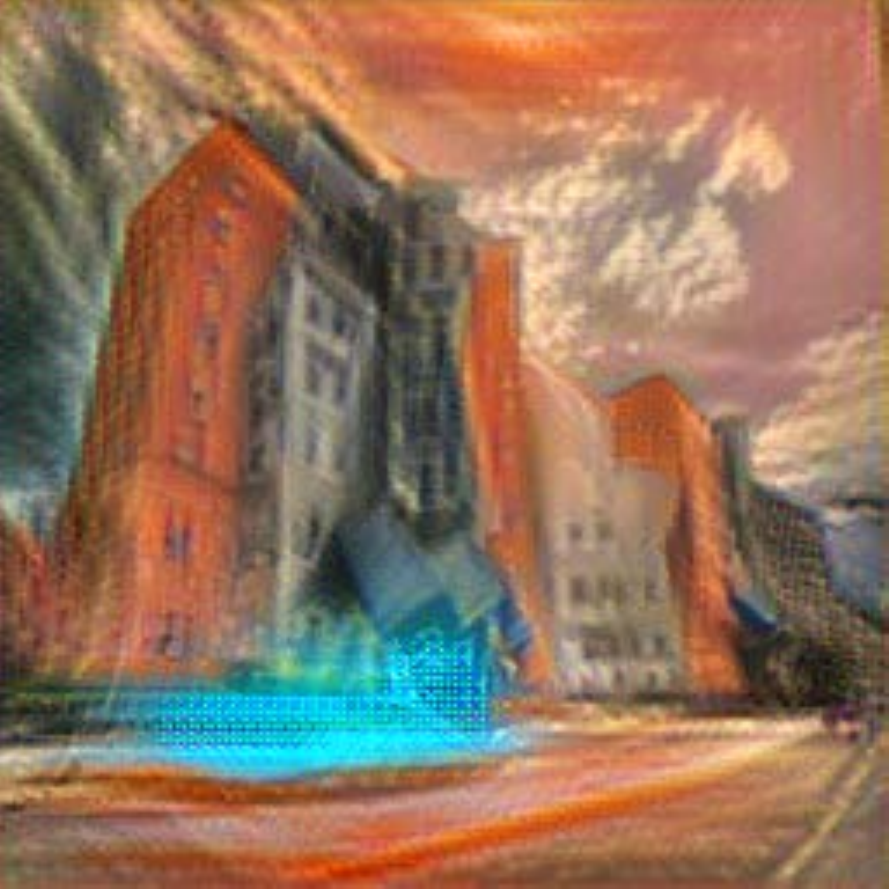}
\end{minipage}%
}%
\subfigure{
\begin{minipage}[t]{0.2\linewidth}
\centering
\includegraphics[width=\linewidth]{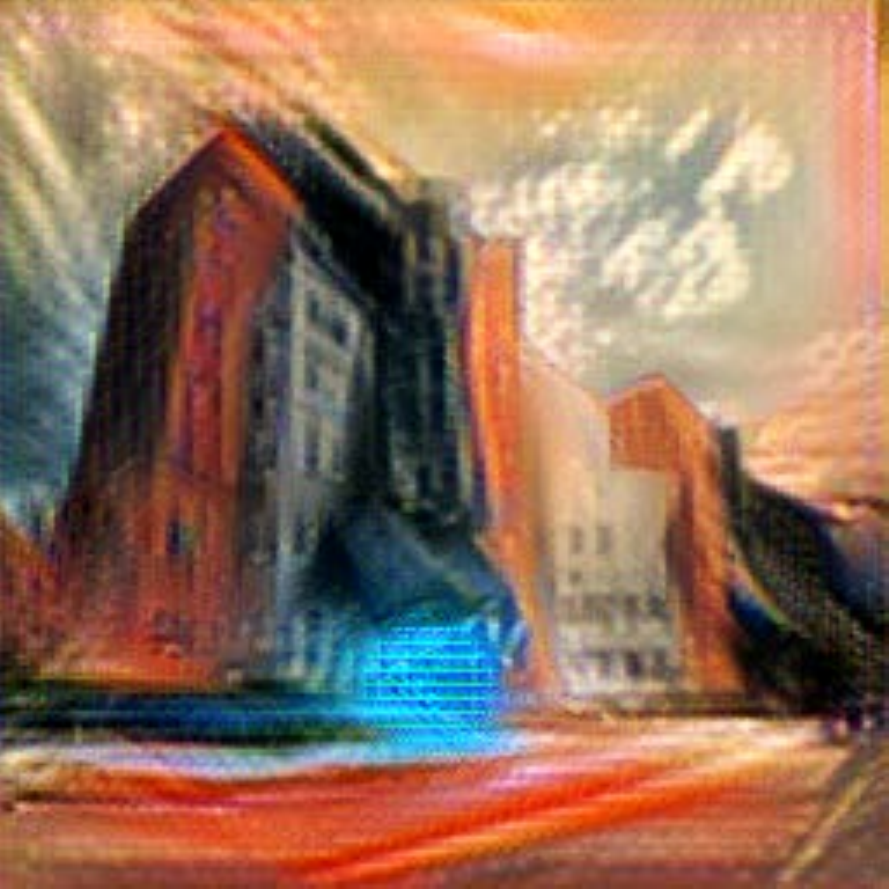}
\end{minipage}%
}%
\subfigure{
\begin{minipage}[t]{0.2\linewidth}
\centering
\includegraphics[width=\linewidth]{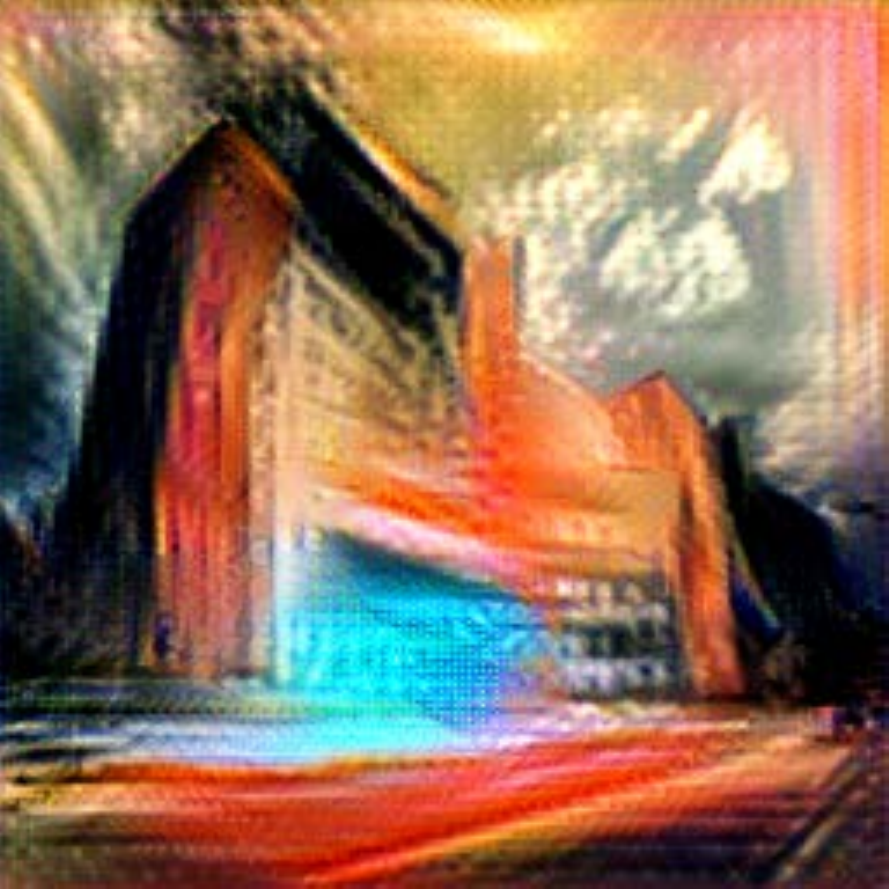}
\end{minipage}%
}%
\centering
\caption{The styled image with stroke basis intervened using spectrum based methods. The left most of each row shows the style images (From top to bottom: A Muse (La Muse) - Pablo Picasso; Number 4 (Gray and Red) - Jackson Pollock; Shipwreck - J.M.W. Turner; Natura Morta - Giorgio Morandi; The Scream - Edvard Munch). From left to right of each row, the effect of stroke is increasingly amplified.}
\label{spectrumIntervention-2}
\end{figure*}

\begin{figure*}[htb]
\centering
\subfigure{
\begin{minipage}[t]{0.2\linewidth}
\centering
\includegraphics[width=\linewidth]{new/wave.pdf}
\end{minipage}%
}%
\subfigure{
\begin{minipage}[t]{0.2\linewidth}
\centering
\includegraphics[width=\linewidth]{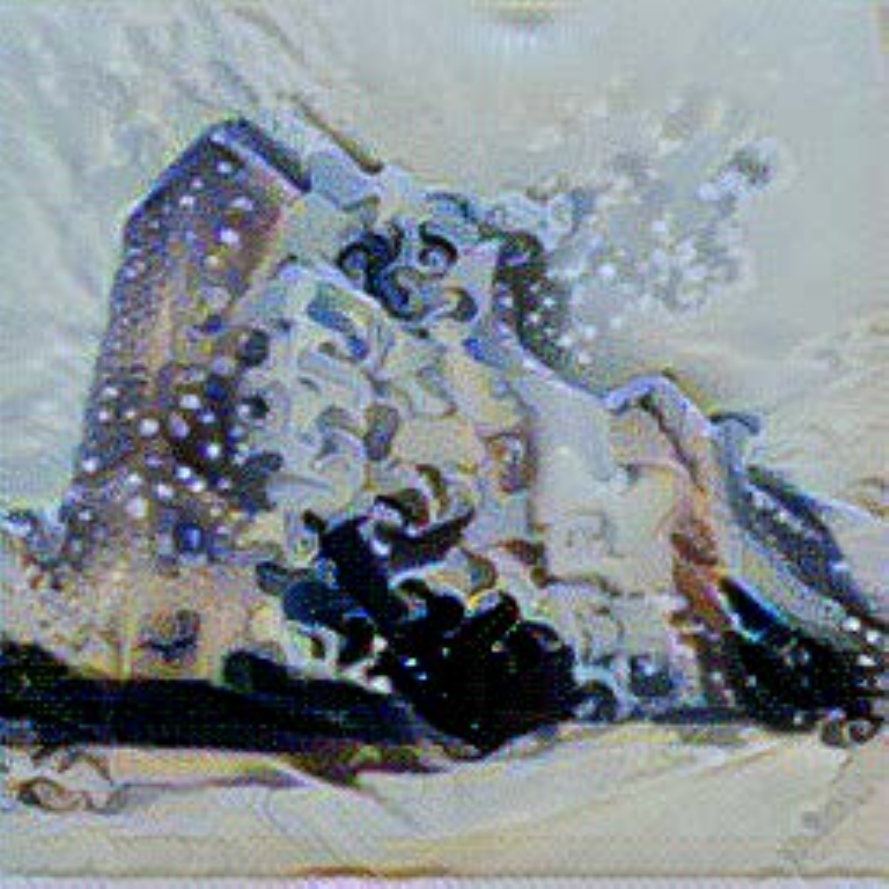}
\end{minipage}%
}%
\subfigure{
\begin{minipage}[t]{0.2\linewidth}
\centering
\includegraphics[width=\linewidth]{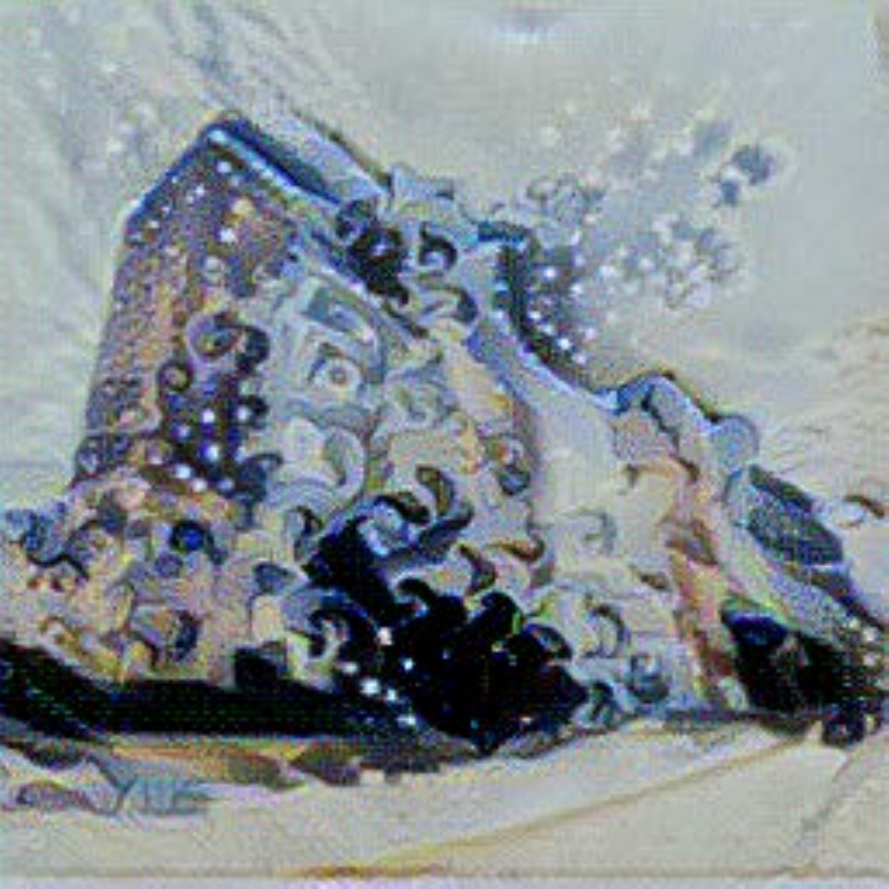}
\end{minipage}%
}%
\subfigure{
\begin{minipage}[t]{0.2\linewidth}
\centering
\includegraphics[width=\linewidth]{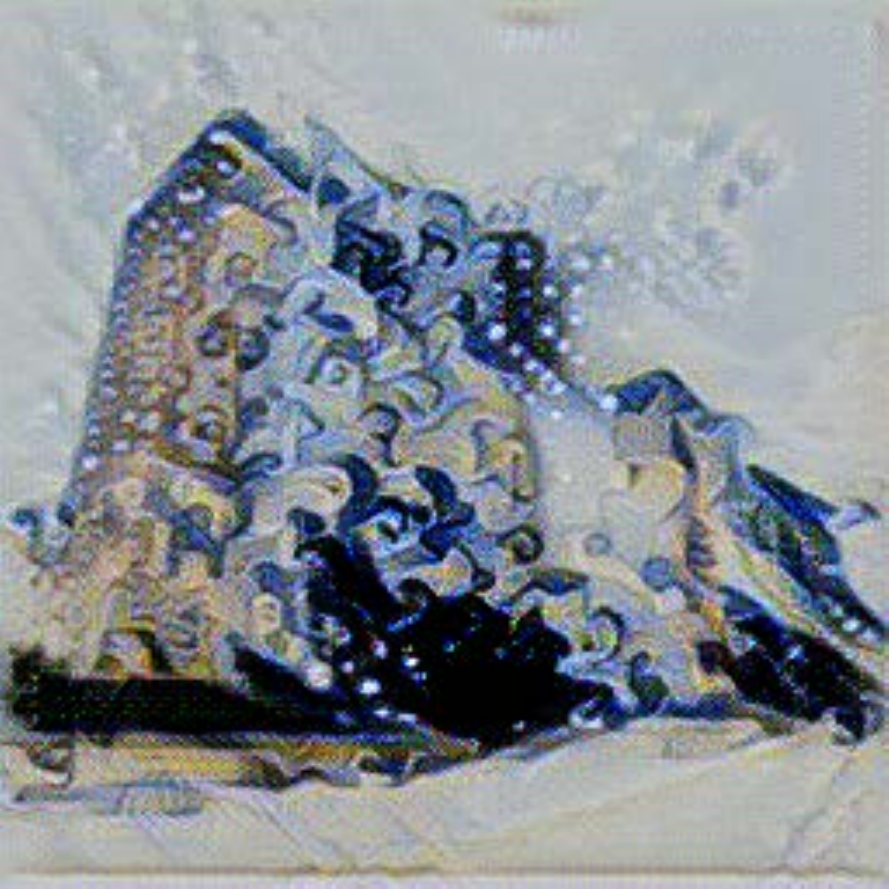}
\end{minipage}%
}%
\subfigure{
\begin{minipage}[t]{0.2\linewidth}
\centering
\includegraphics[width=\linewidth]{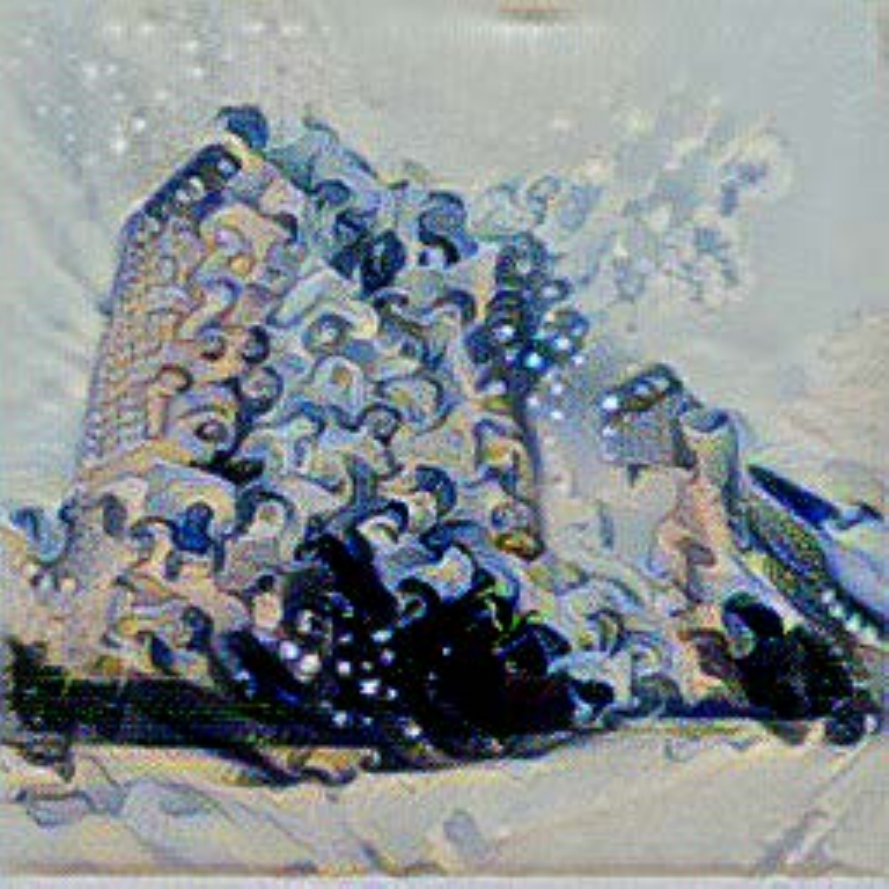}
\end{minipage}%
}%
\vfill

\subfigure{
\begin{minipage}[t]{0.2\linewidth}
\centering
\includegraphics[width=\linewidth]{new/composition.pdf}
\end{minipage}%
}%
\subfigure{
\begin{minipage}[t]{0.2\linewidth}
\centering
\includegraphics[width=\linewidth]{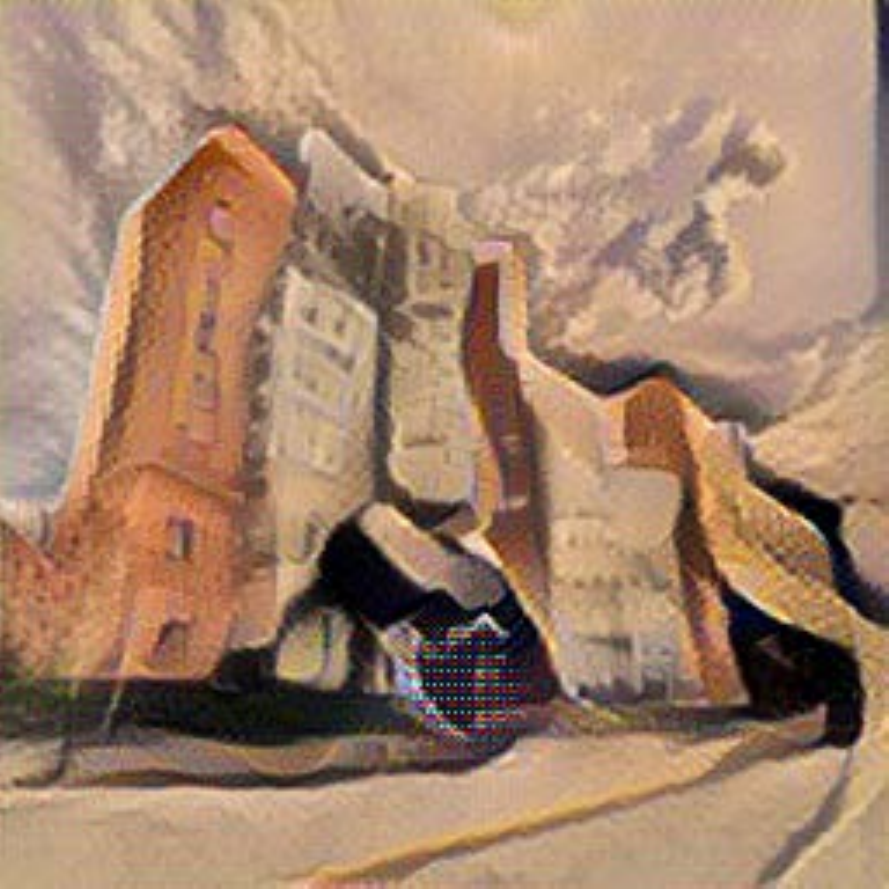}
\end{minipage}%
}%
\subfigure{
\begin{minipage}[t]{0.2\linewidth}
\centering
\includegraphics[width=\linewidth]{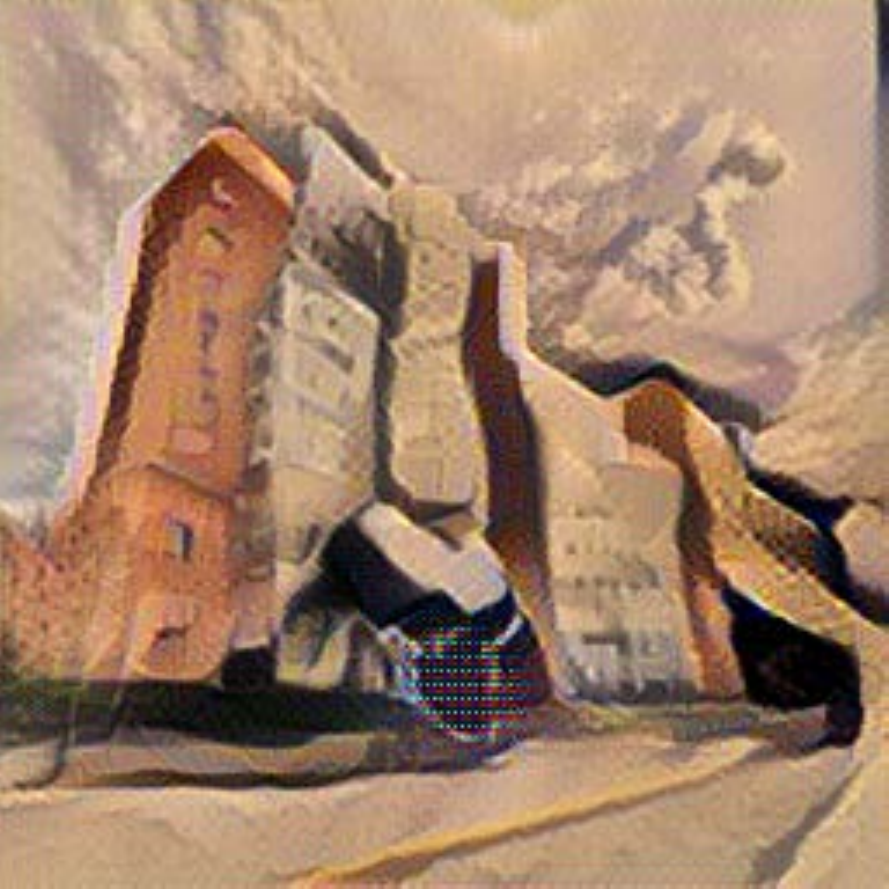}
\end{minipage}%
}%
\subfigure{
\begin{minipage}[t]{0.2\linewidth}
\centering
\includegraphics[width=\linewidth]{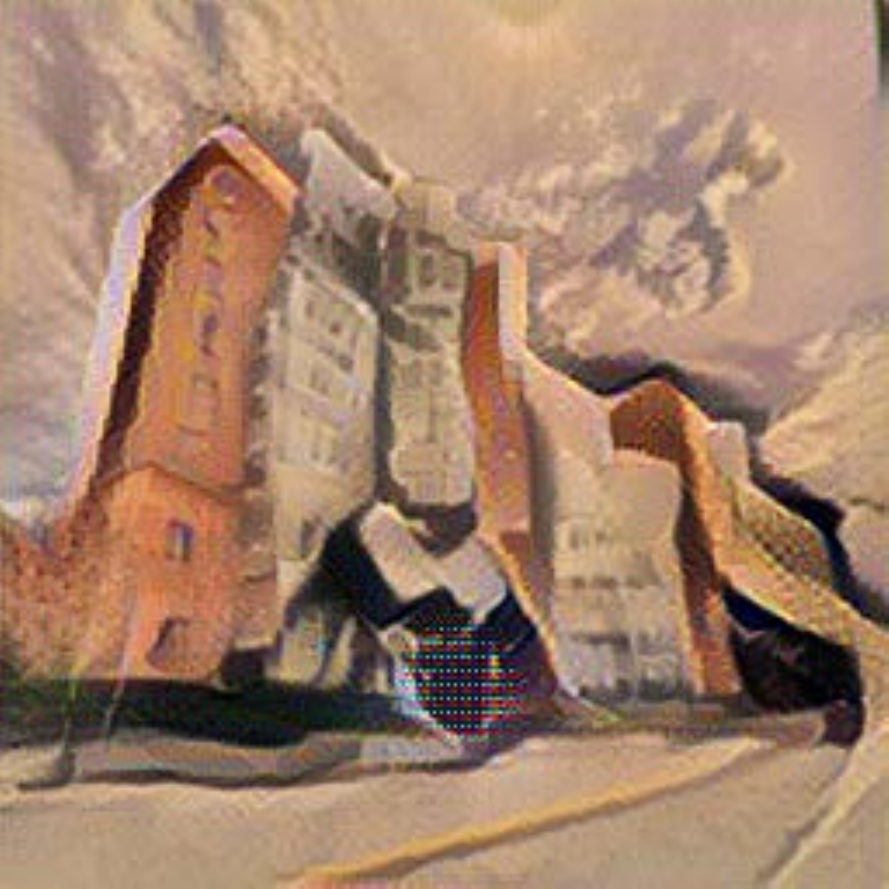}
\end{minipage}%
}%
\subfigure{
\begin{minipage}[t]{0.2\linewidth}
\centering
\includegraphics[width=\linewidth]{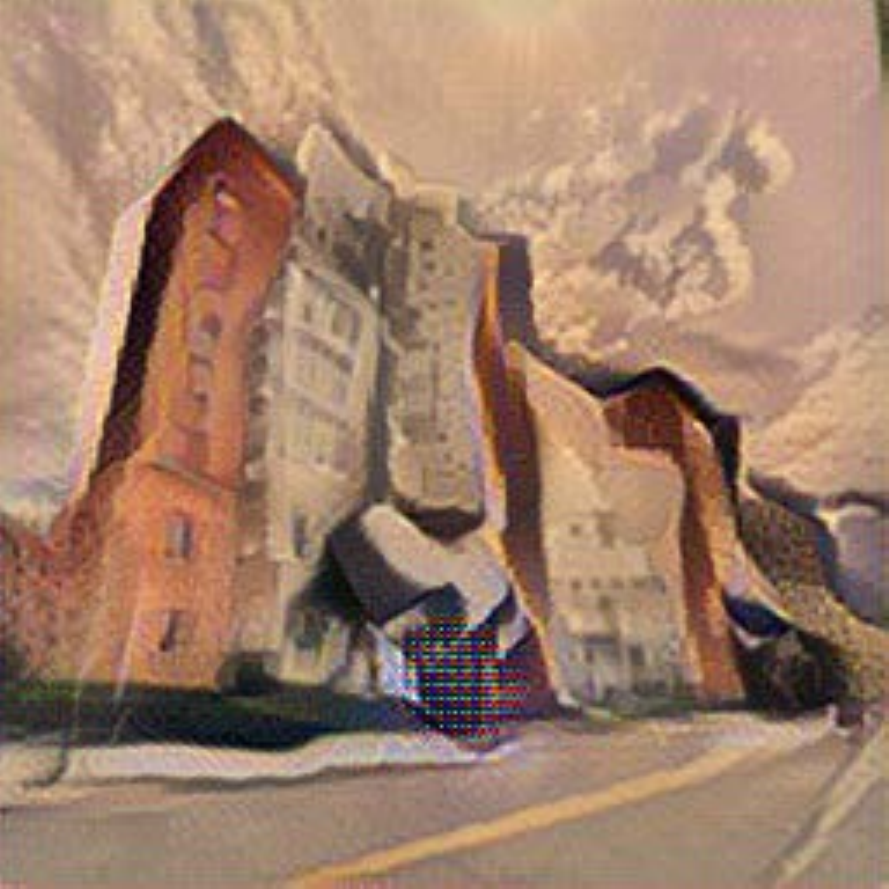}
\end{minipage}%
}%
\vfill

\subfigure{
\begin{minipage}[t]{0.2\linewidth}
\centering
\includegraphics[width=\linewidth]{new/dancer.pdf}
\end{minipage}%
}%
\subfigure{
\begin{minipage}[t]{0.2\linewidth}
\centering
\includegraphics[width=\linewidth]{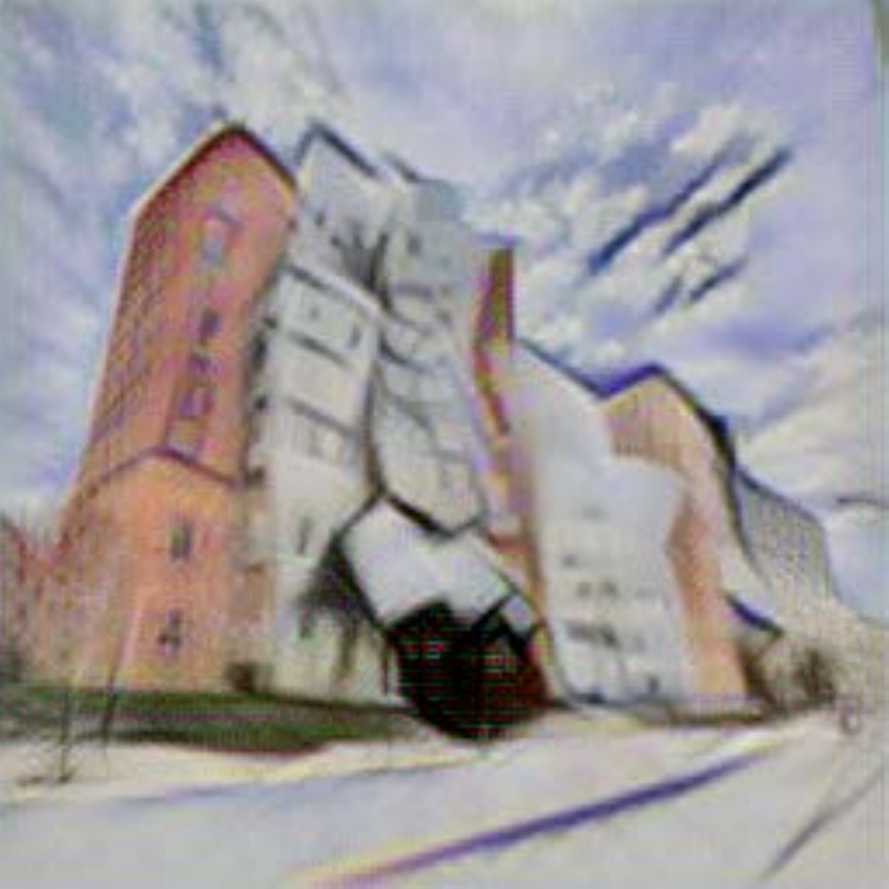}
\end{minipage}%
}%
\subfigure{
\begin{minipage}[t]{0.2\linewidth}
\centering
\includegraphics[width=\linewidth]{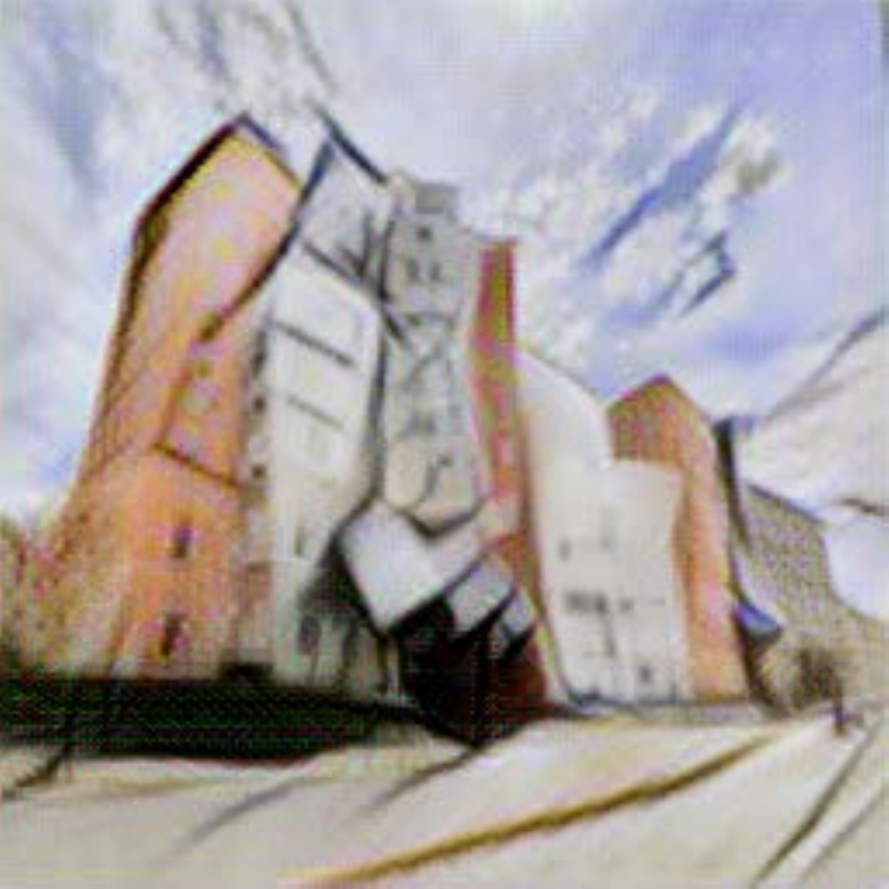}
\end{minipage}%
}%
\subfigure{
\begin{minipage}[t]{0.2\linewidth}
\centering
\includegraphics[width=\linewidth]{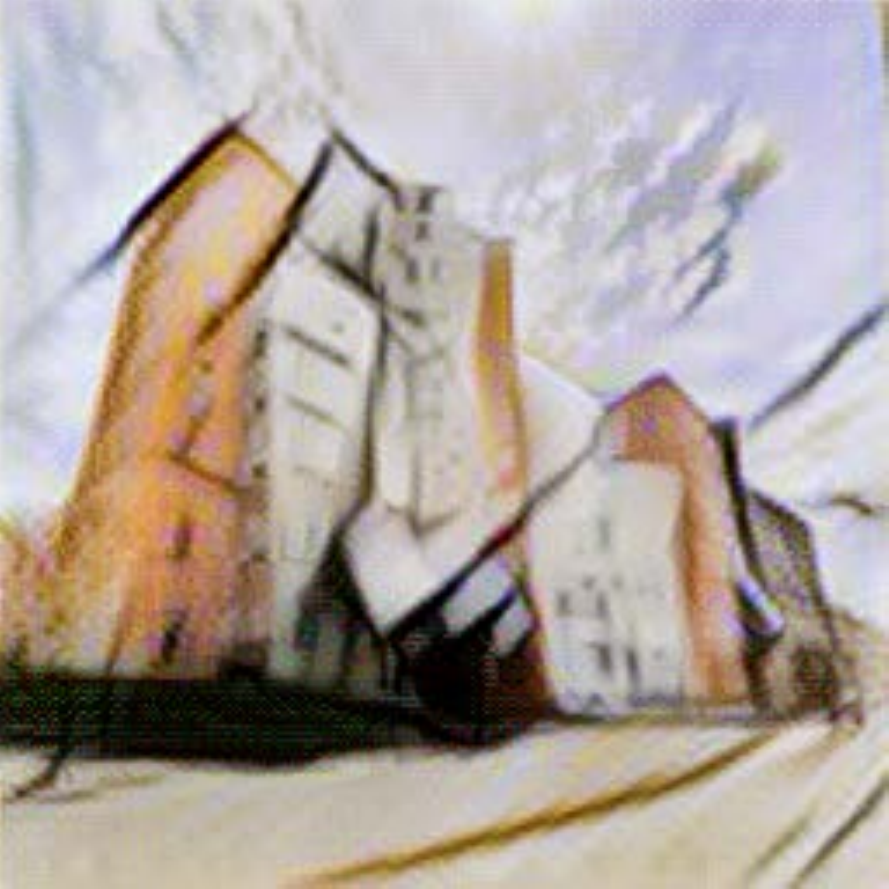}
\end{minipage}%
}%
\subfigure{
\begin{minipage}[t]{0.2\linewidth}
\centering
\includegraphics[width=\linewidth]{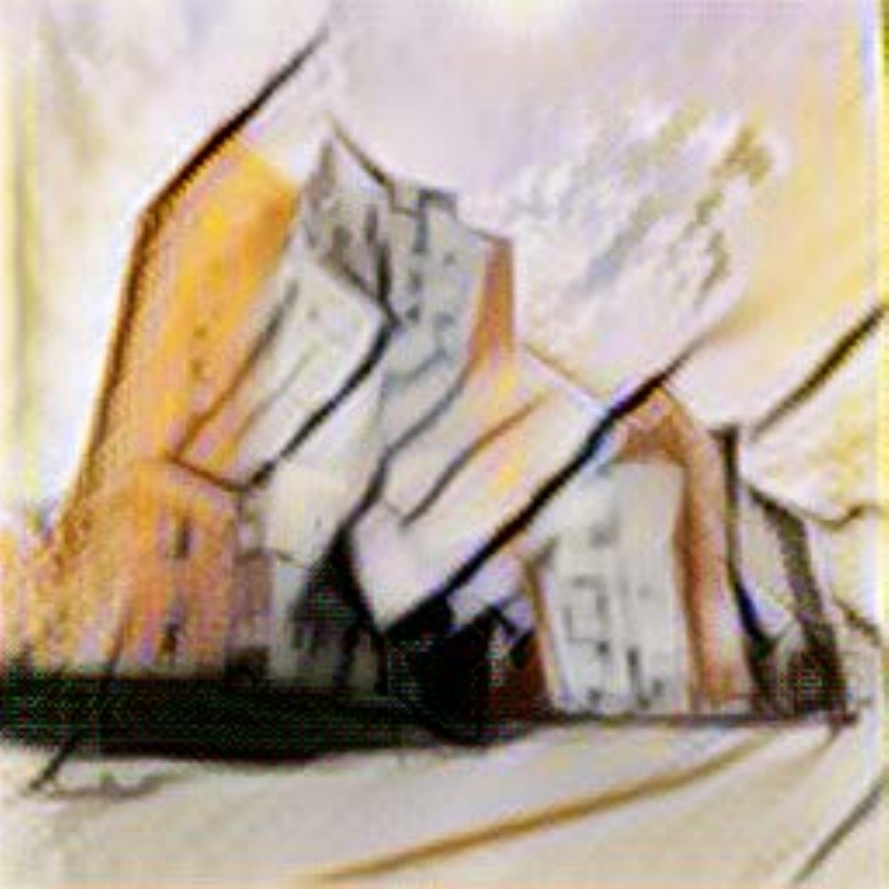}
\end{minipage}%
}%
\vfill

\subfigure{
\begin{minipage}[t]{0.2\linewidth}
\centering
\includegraphics[width=\linewidth]{new/draft.pdf}
\end{minipage}%
}%
\subfigure{
\begin{minipage}[t]{0.2\linewidth}
\centering
\includegraphics[width=\linewidth]{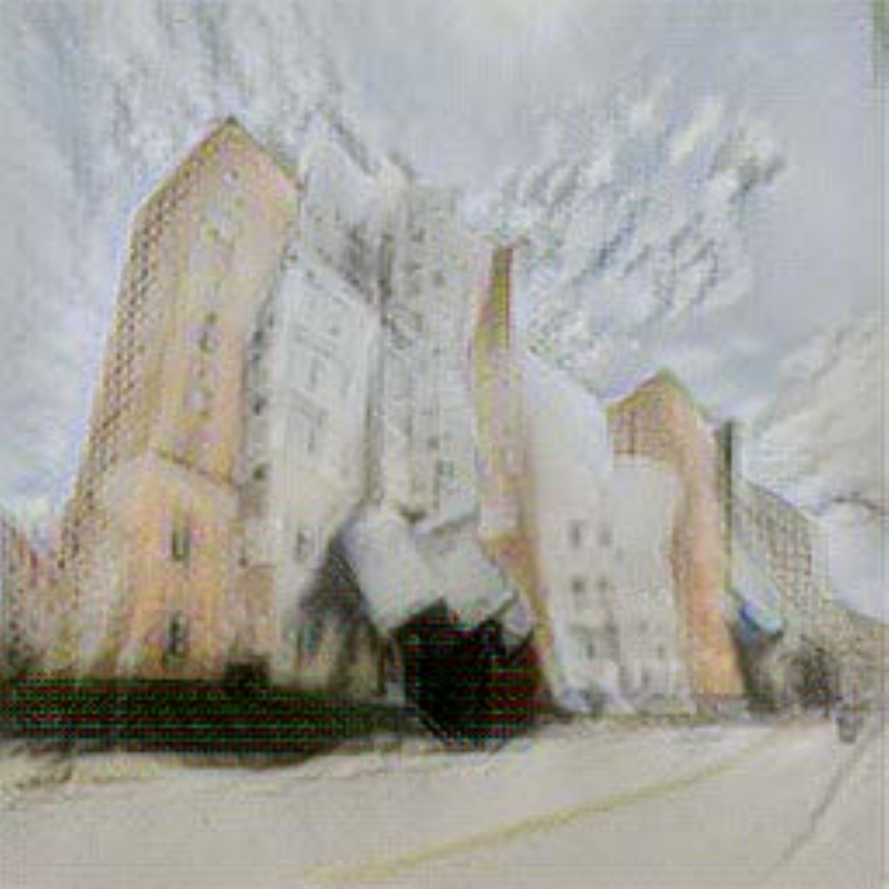}
\end{minipage}%
}%
\subfigure{
\begin{minipage}[t]{0.2\linewidth}
\centering
\includegraphics[width=\linewidth]{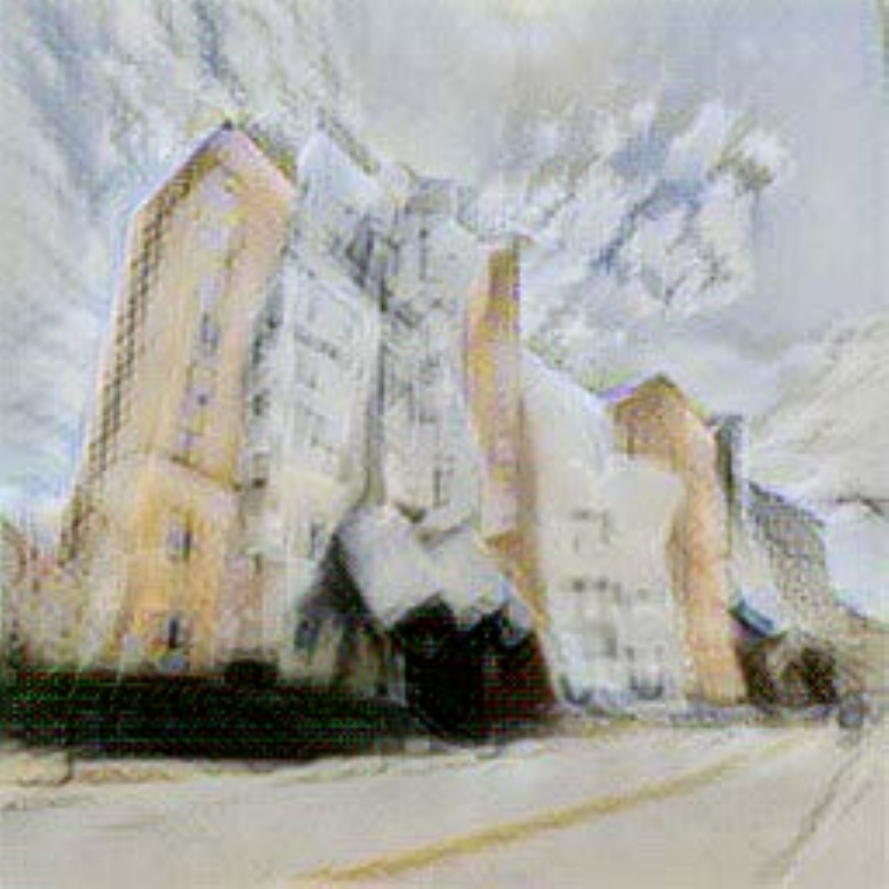}
\end{minipage}%
}%
\subfigure{
\begin{minipage}[t]{0.2\linewidth}
\centering
\includegraphics[width=\linewidth]{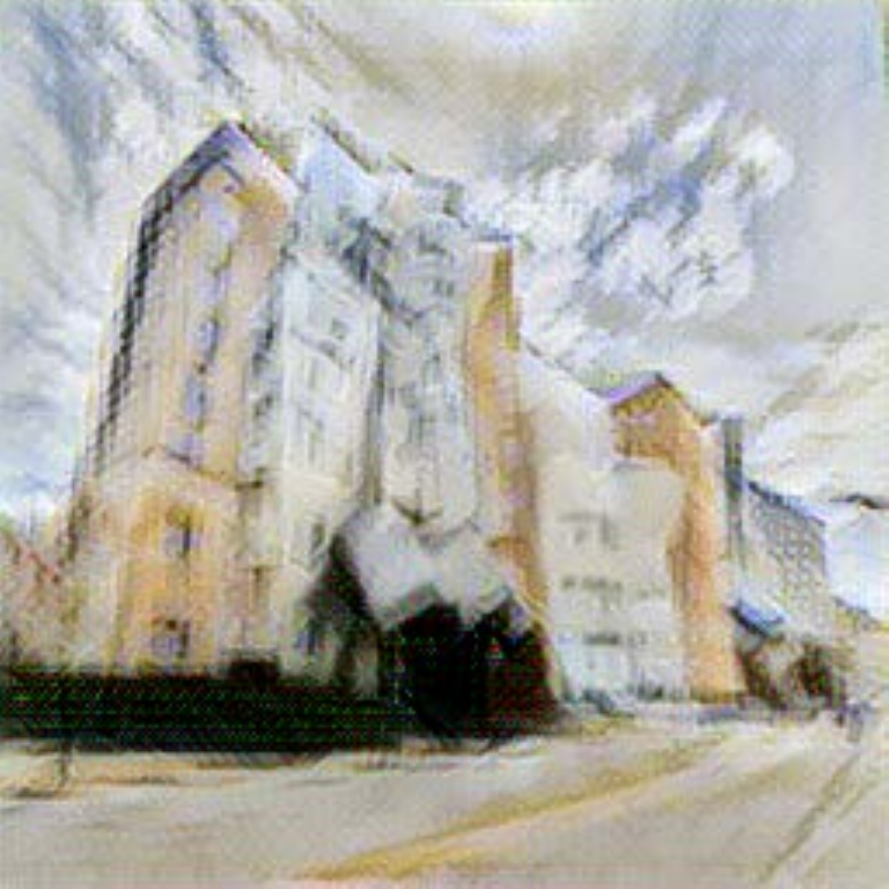}
\end{minipage}%
}%
\subfigure{
\begin{minipage}[t]{0.2\linewidth}
\centering
\includegraphics[width=\linewidth]{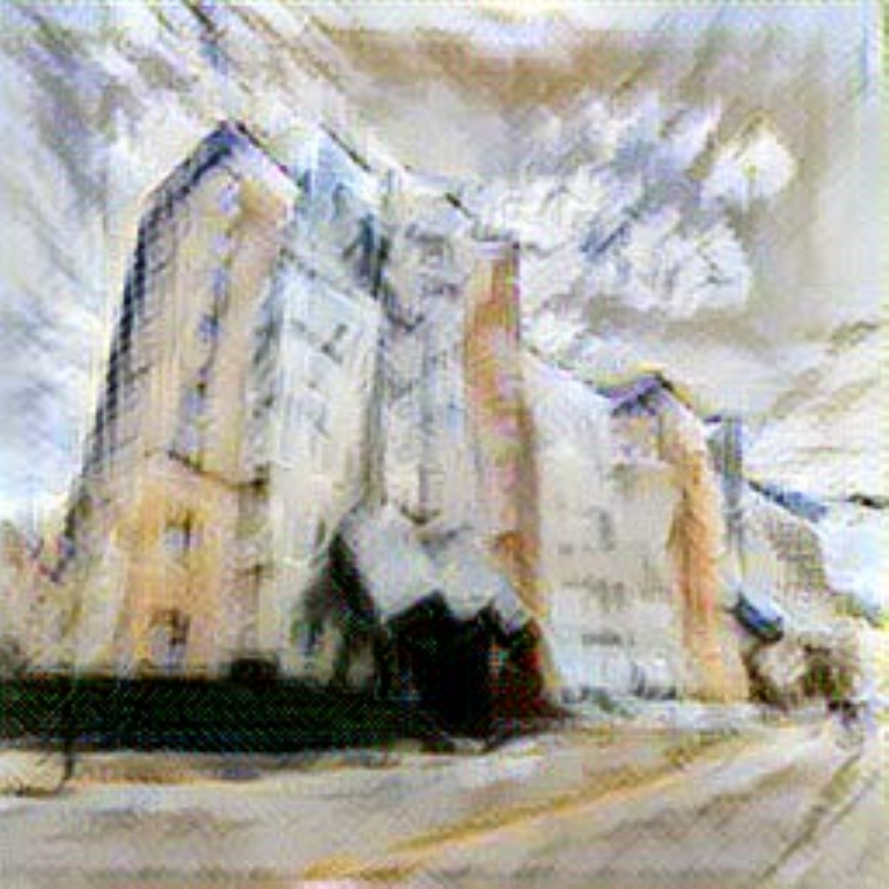}
\end{minipage}%
}%
\vfill

\subfigure{
\begin{minipage}[t]{0.2\linewidth}
\centering
\includegraphics[width=\linewidth]{new/face.pdf}
\end{minipage}%
}%
\subfigure{
\begin{minipage}[t]{0.2\linewidth}
\centering
\includegraphics[width=\linewidth]{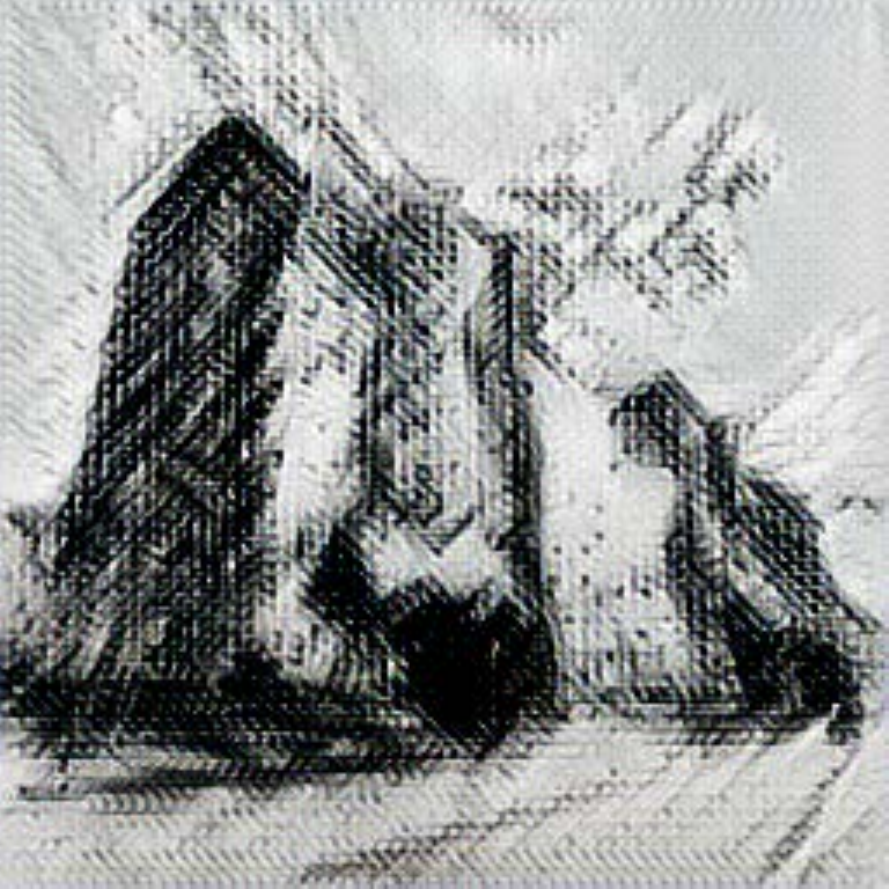}
\end{minipage}%
}%
\subfigure{
\begin{minipage}[t]{0.2\linewidth}
\centering
\includegraphics[width=\linewidth]{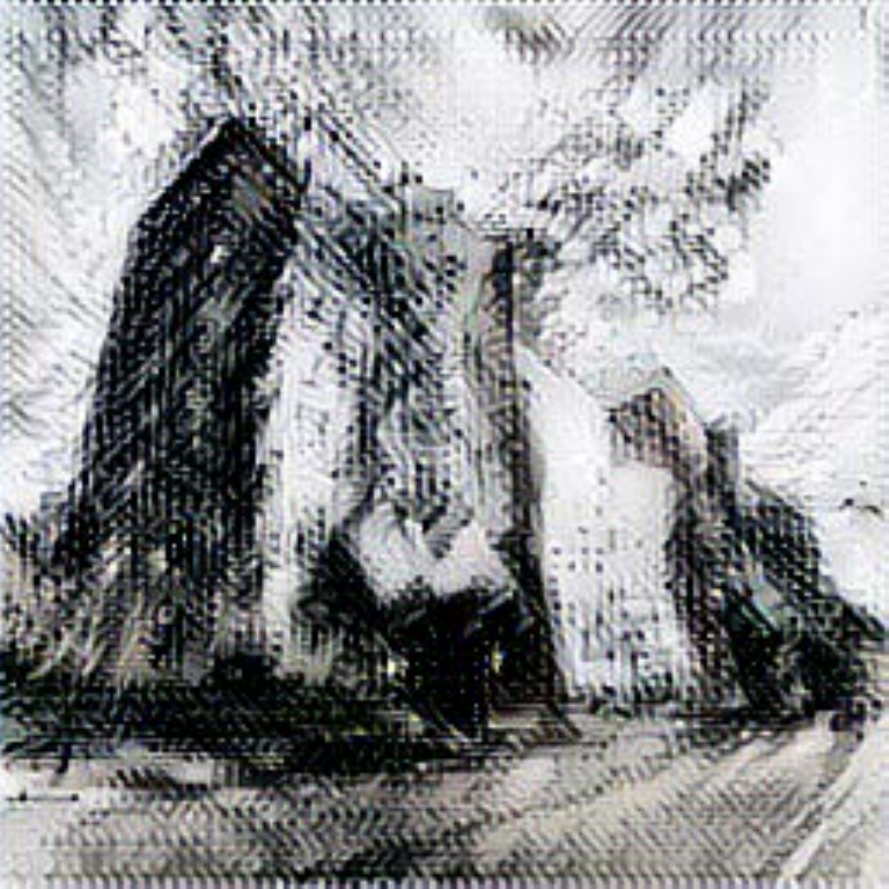}
\end{minipage}%
}%
\subfigure{
\begin{minipage}[t]{0.2\linewidth}
\centering
\includegraphics[width=\linewidth]{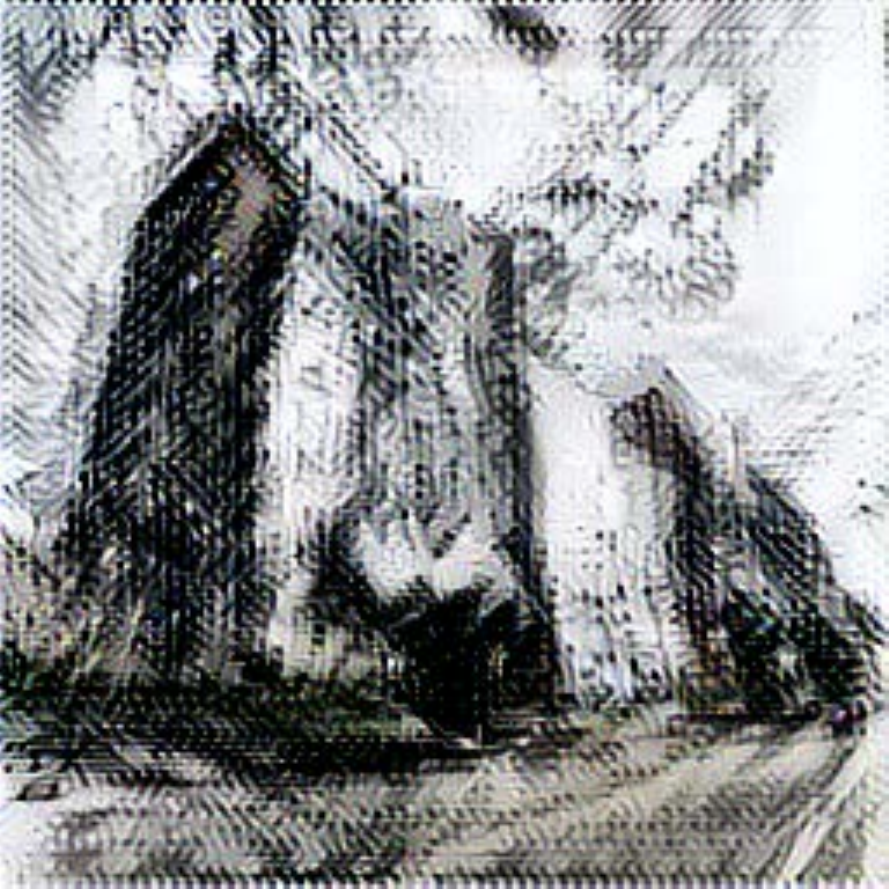}
\end{minipage}%
}%
\subfigure{
\begin{minipage}[t]{0.2\linewidth}
\centering
\includegraphics[width=\linewidth]{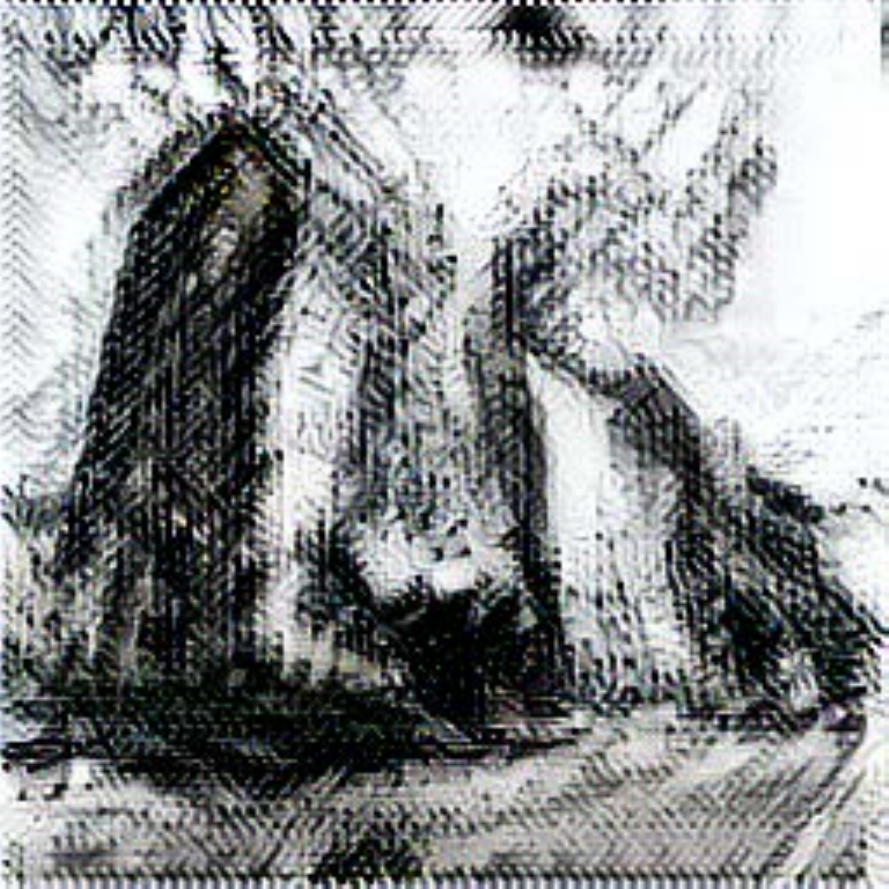}
\end{minipage}%
}%
\centering
\caption{The styled image with stroke basis intervened using ICA. The left most of each row shows the style images. From left to right of each row, the effect of stroke is increasingly amplified.}
\label{icaIntervention-1}
\end{figure*}

\begin{figure*}[htb]
\centering
\subfigure{
\begin{minipage}[t]{0.2\linewidth}
\centering
\includegraphics[width=\linewidth]{new/la_muse.pdf}
\end{minipage}%
}%
\subfigure{
\begin{minipage}[t]{0.2\linewidth}
\centering
\includegraphics[width=\linewidth]{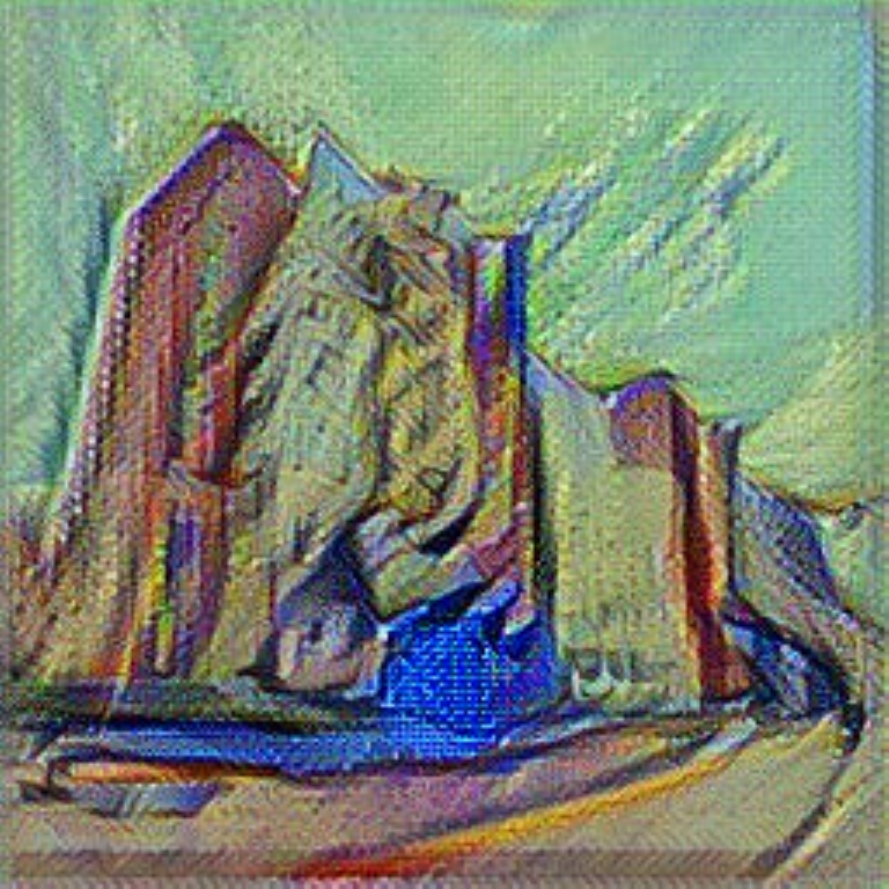}
\end{minipage}%
}%
\subfigure{
\begin{minipage}[t]{0.2\linewidth}
\centering
\includegraphics[width=\linewidth]{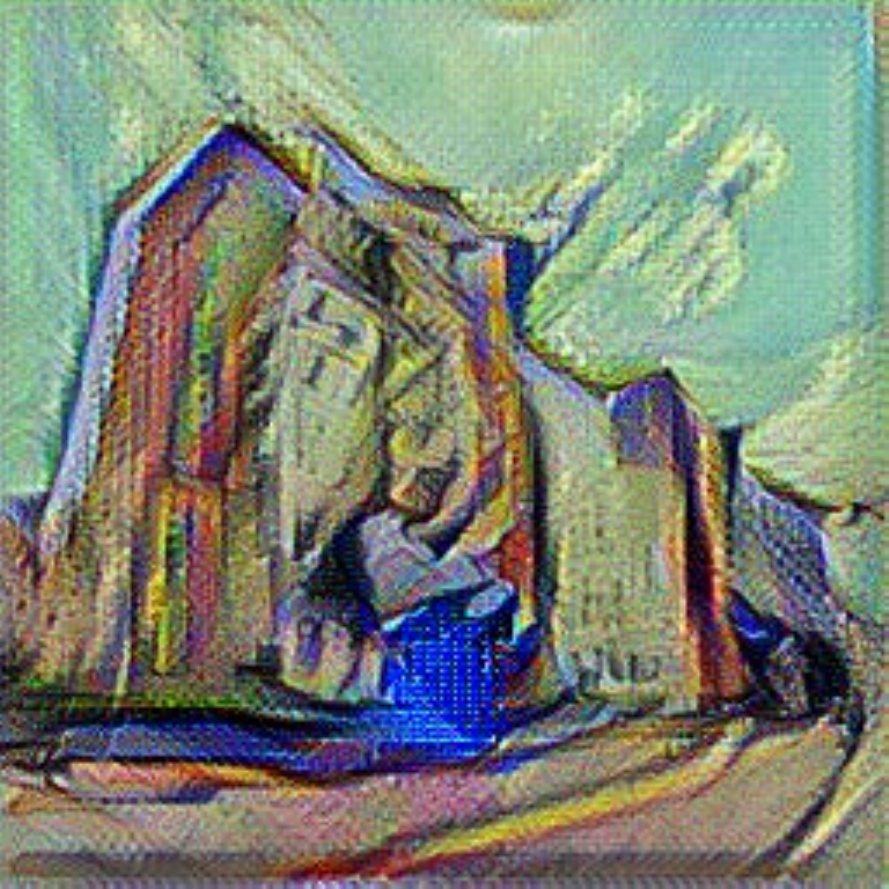}
\end{minipage}%
}%
\subfigure{
\begin{minipage}[t]{0.2\linewidth}
\centering
\includegraphics[width=\linewidth]{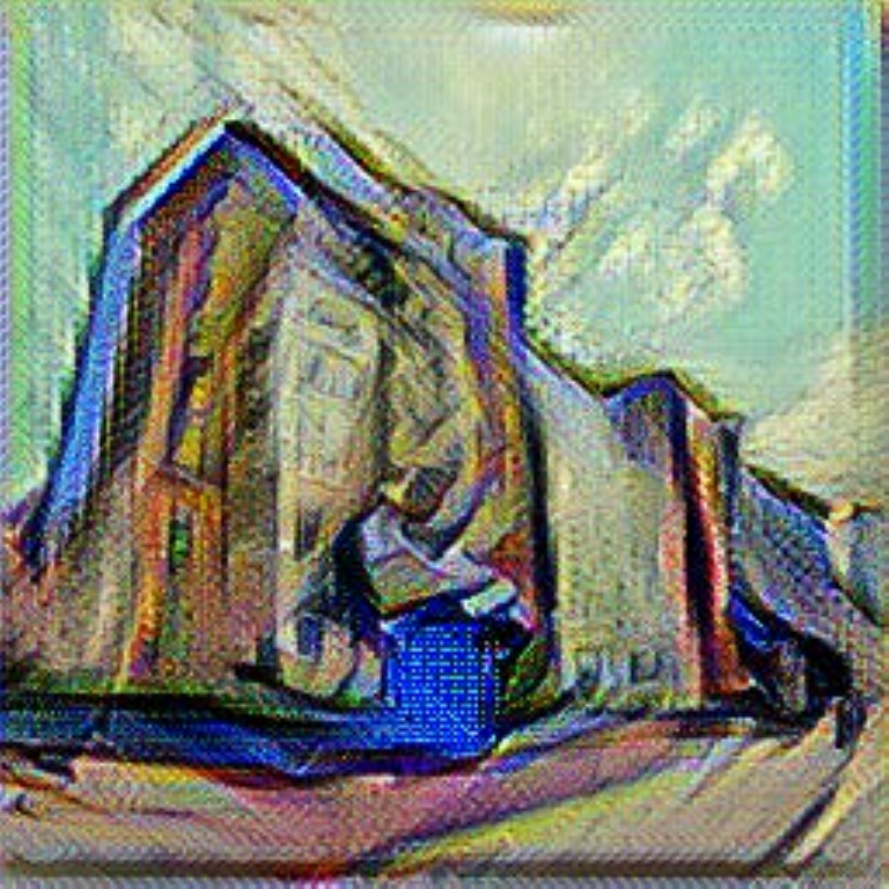}
\end{minipage}%
}%
\subfigure{
\begin{minipage}[t]{0.2\linewidth}
\centering
\includegraphics[width=\linewidth]{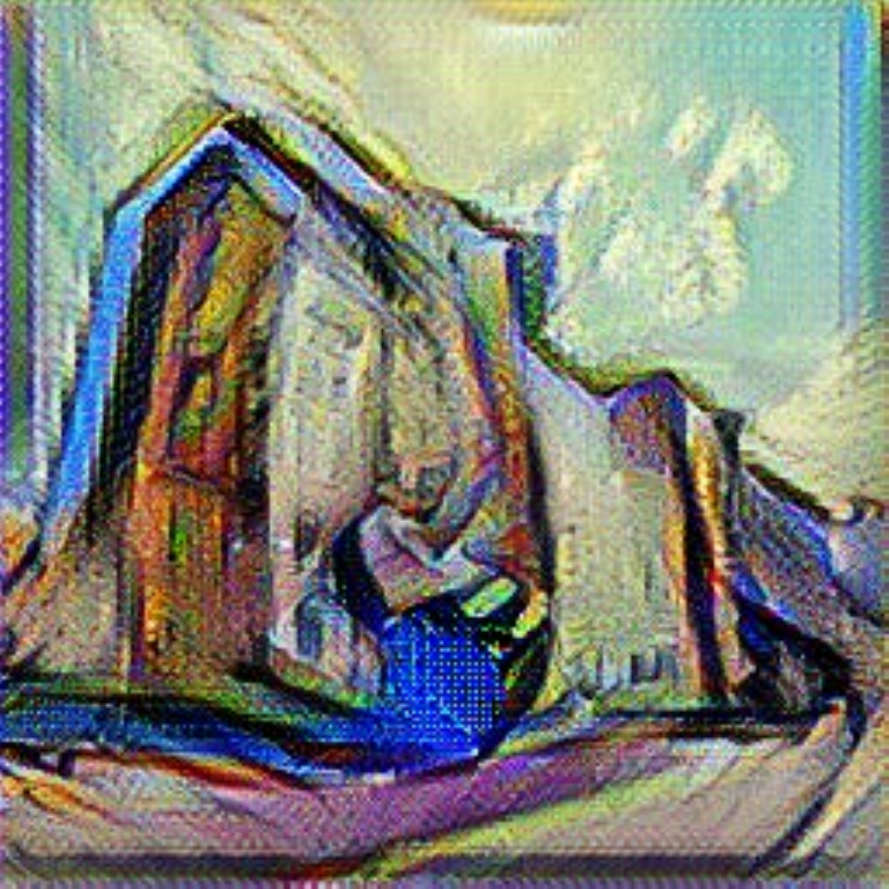}
\end{minipage}%
}%
\vfill

\subfigure{
\begin{minipage}[t]{0.2\linewidth}
\centering
\includegraphics[width=\linewidth]{new/gray_red.pdf}
\end{minipage}%
}%
\subfigure{
\begin{minipage}[t]{0.2\linewidth}
\centering
\includegraphics[width=\linewidth]{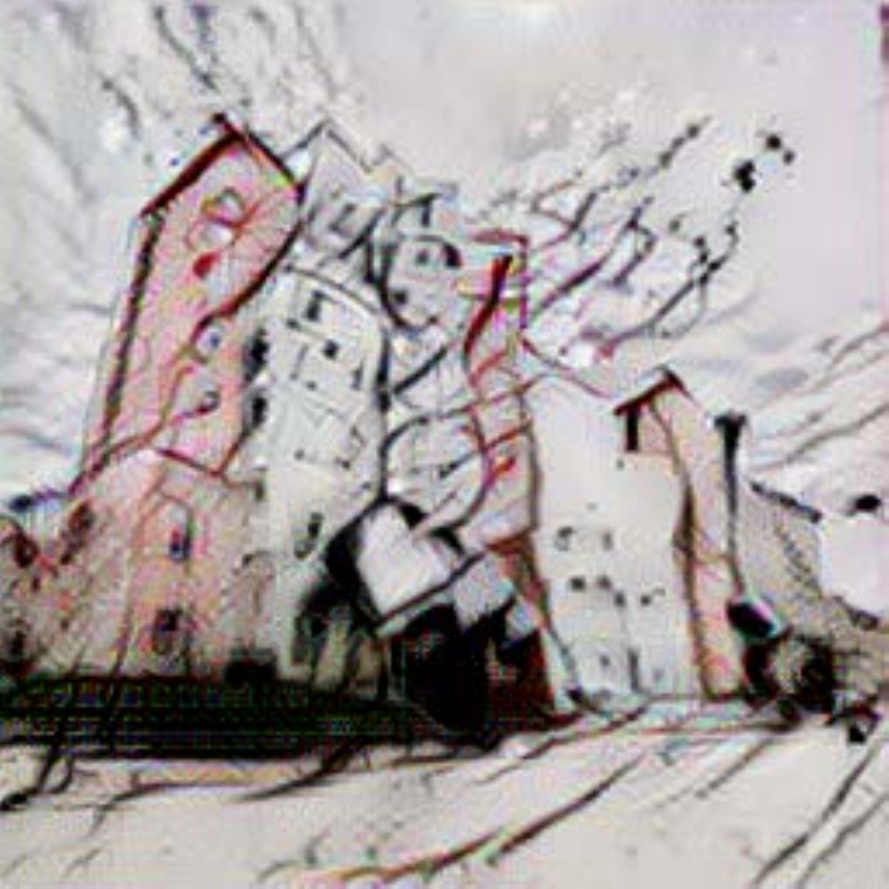}
\end{minipage}%
}%
\subfigure{
\begin{minipage}[t]{0.2\linewidth}
\centering
\includegraphics[width=\linewidth]{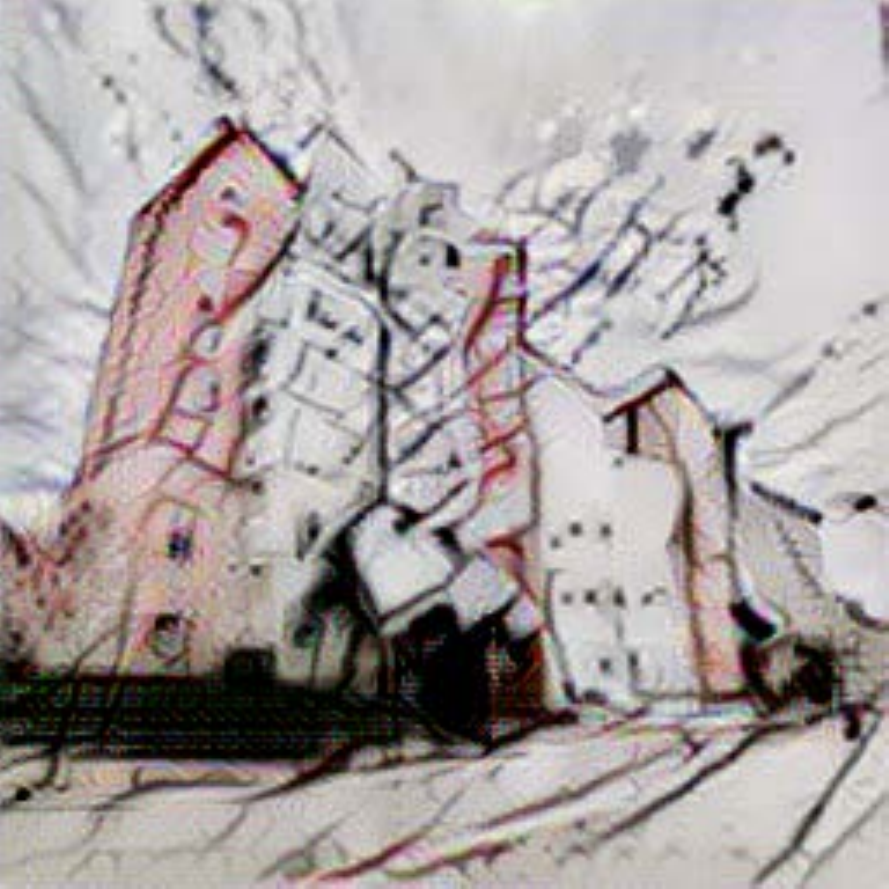}
\end{minipage}%
}%
\subfigure{
\begin{minipage}[t]{0.2\linewidth}
\centering
\includegraphics[width=\linewidth]{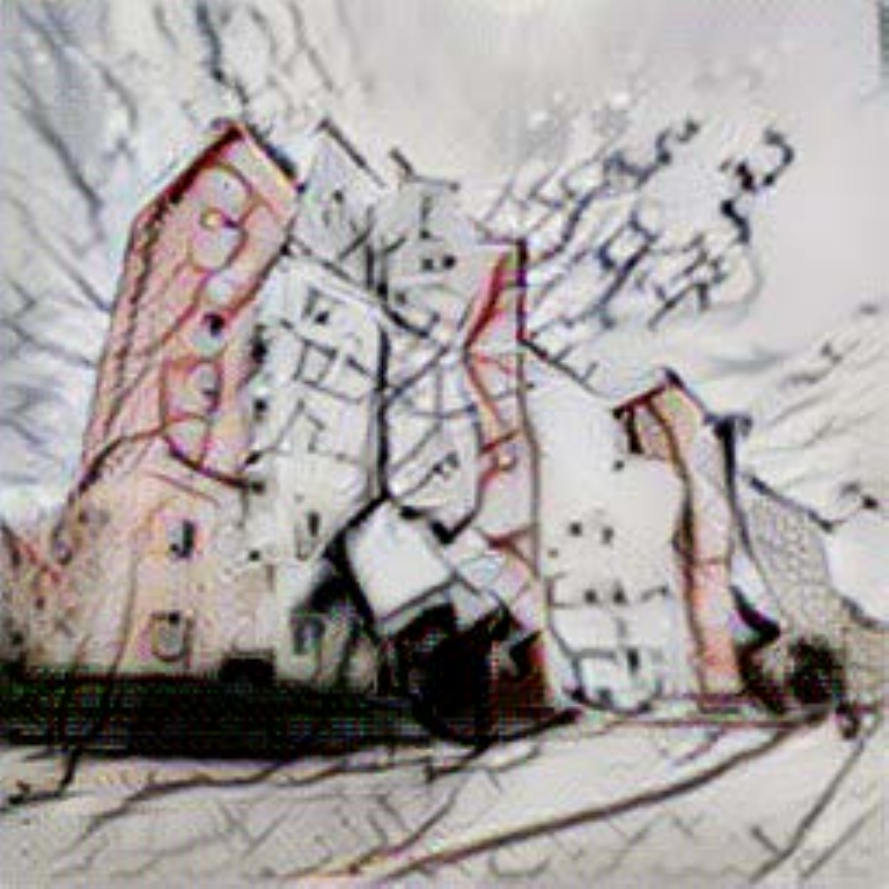}
\end{minipage}%
}%
\subfigure{
\begin{minipage}[t]{0.2\linewidth}
\centering
\includegraphics[width=\linewidth]{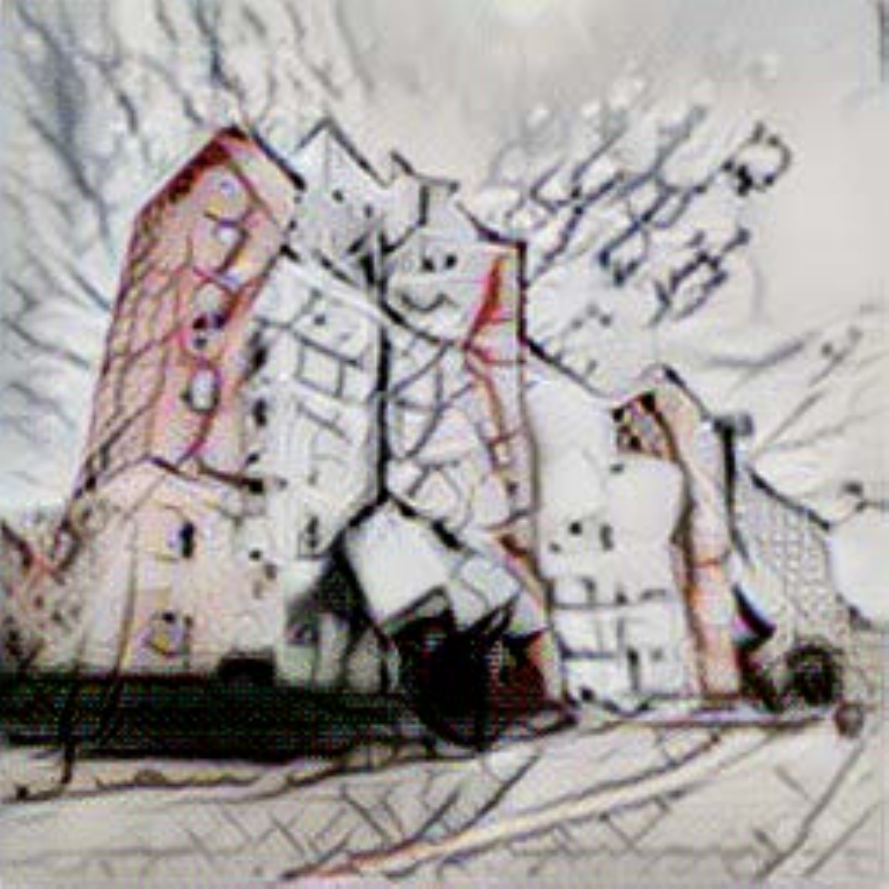}
\end{minipage}%
}%
\vfill

\subfigure{
\begin{minipage}[t]{0.2\linewidth}
\centering
\includegraphics[width=\linewidth]{new/ship.pdf}
\end{minipage}%
}%
\subfigure{
\begin{minipage}[t]{0.2\linewidth}
\centering
\includegraphics[width=\linewidth]{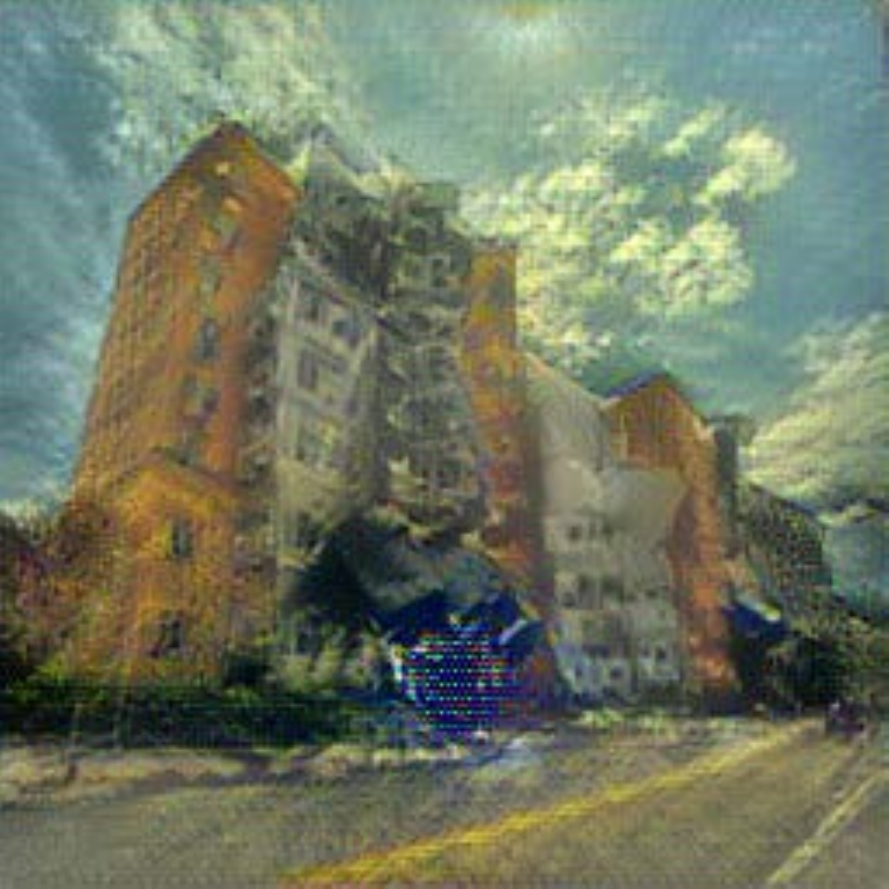}
\end{minipage}%
}%
\subfigure{
\begin{minipage}[t]{0.2\linewidth}
\centering
\includegraphics[width=\linewidth]{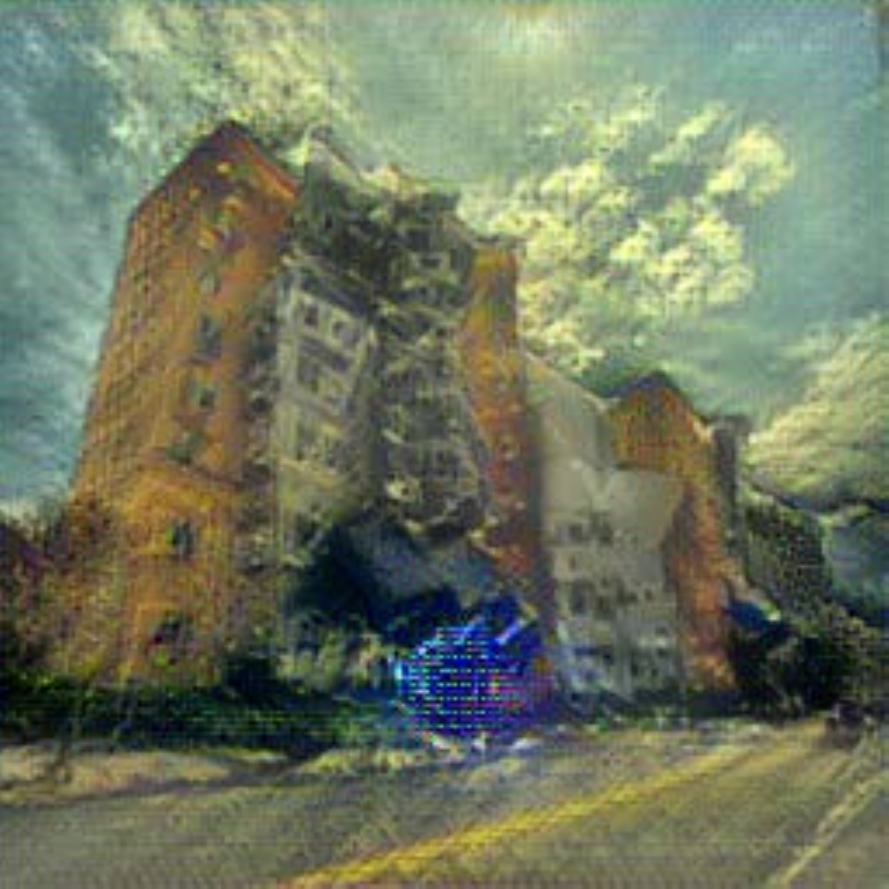}
\end{minipage}%
}%
\subfigure{
\begin{minipage}[t]{0.2\linewidth}
\centering
\includegraphics[width=\linewidth]{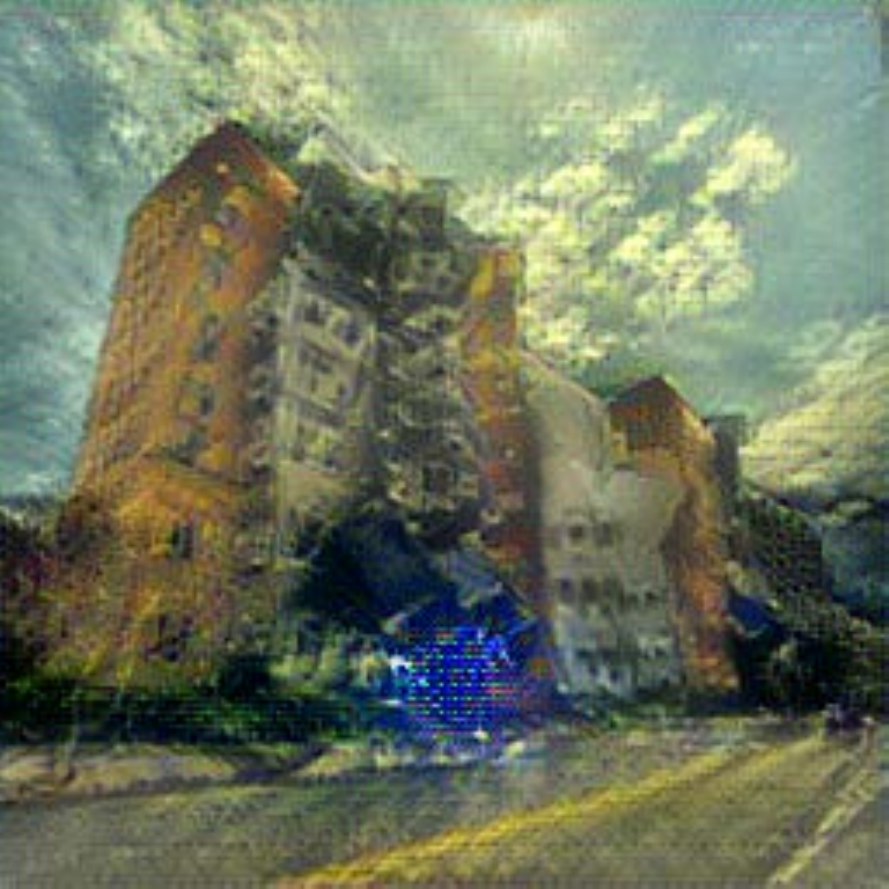}
\end{minipage}%
}%
\subfigure{
\begin{minipage}[t]{0.2\linewidth}
\centering
\includegraphics[width=\linewidth]{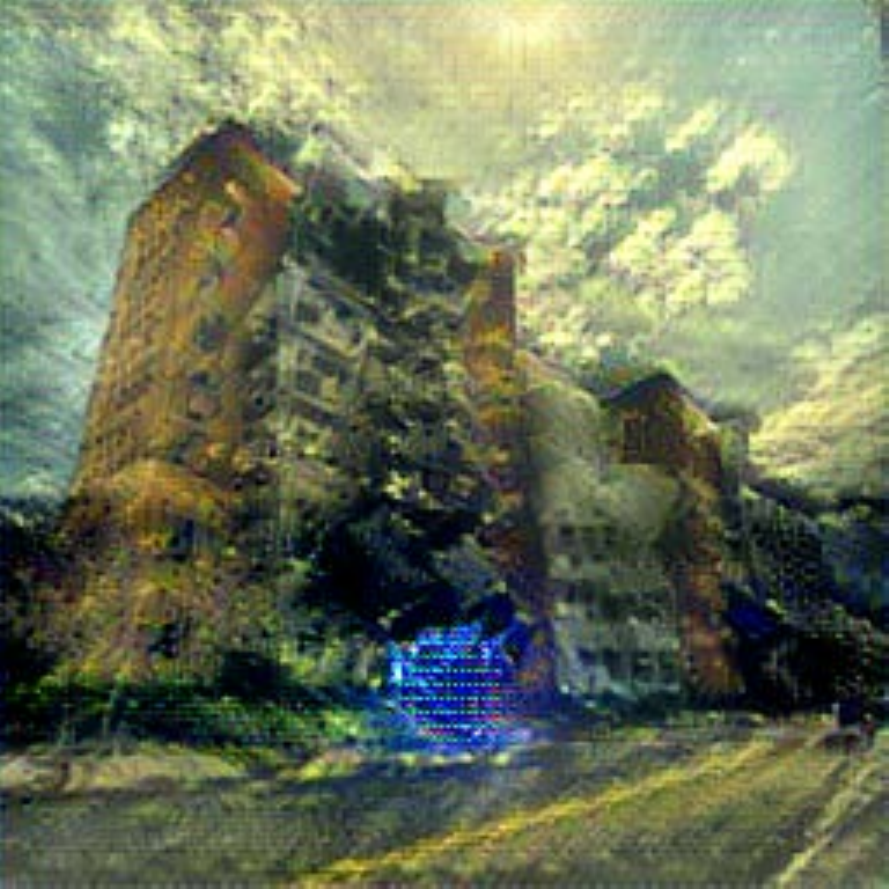}
\end{minipage}%
}%
\vfill

\subfigure{
\begin{minipage}[t]{0.2\linewidth}
\centering
\includegraphics[width=\linewidth]{new/static.pdf}
\end{minipage}%
}%
\subfigure{
\begin{minipage}[t]{0.2\linewidth}
\centering
\includegraphics[width=\linewidth]{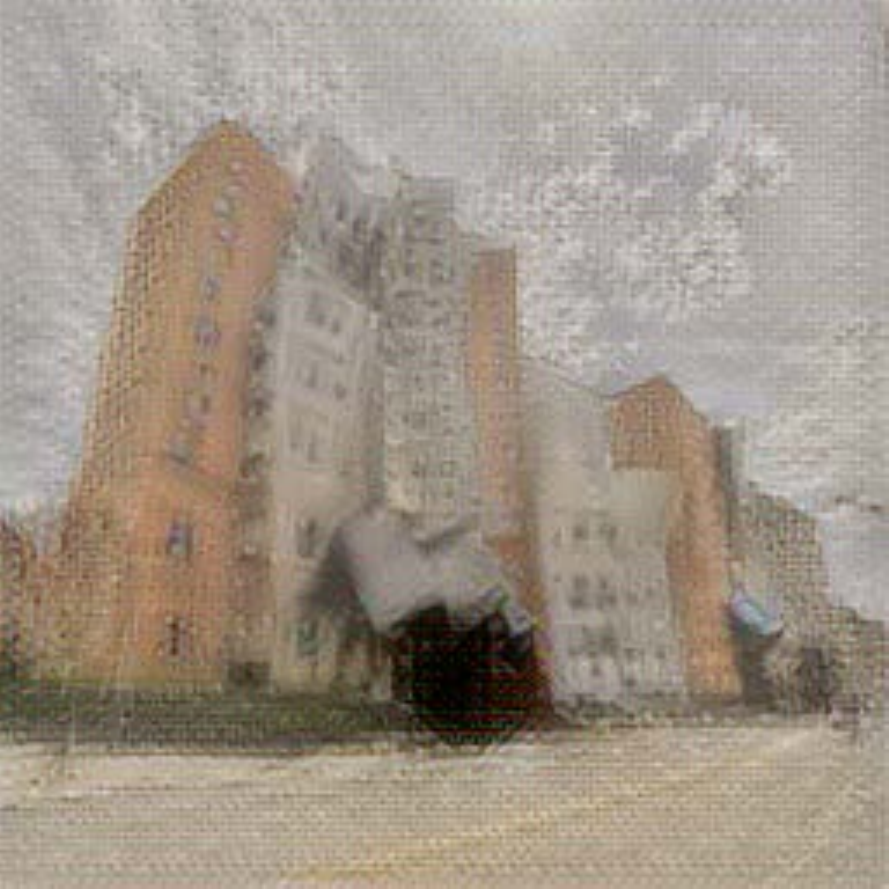}
\end{minipage}%
}%
\subfigure{
\begin{minipage}[t]{0.2\linewidth}
\centering
\includegraphics[width=\linewidth]{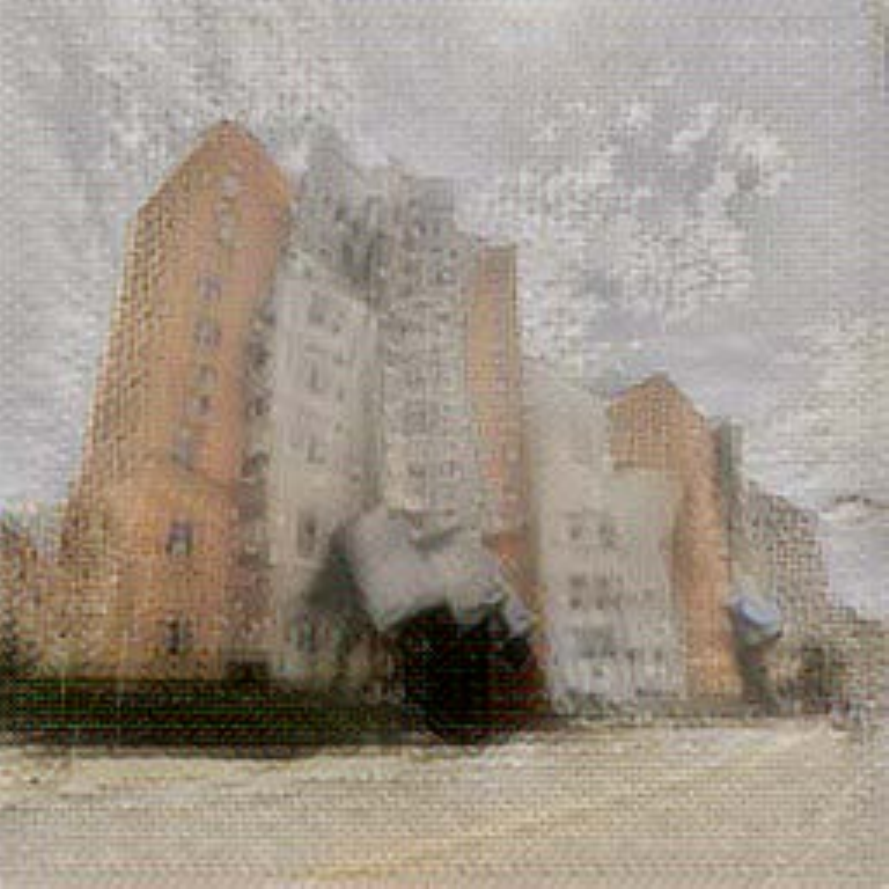}
\end{minipage}%
}%
\subfigure{
\begin{minipage}[t]{0.2\linewidth}
\centering
\includegraphics[width=\linewidth]{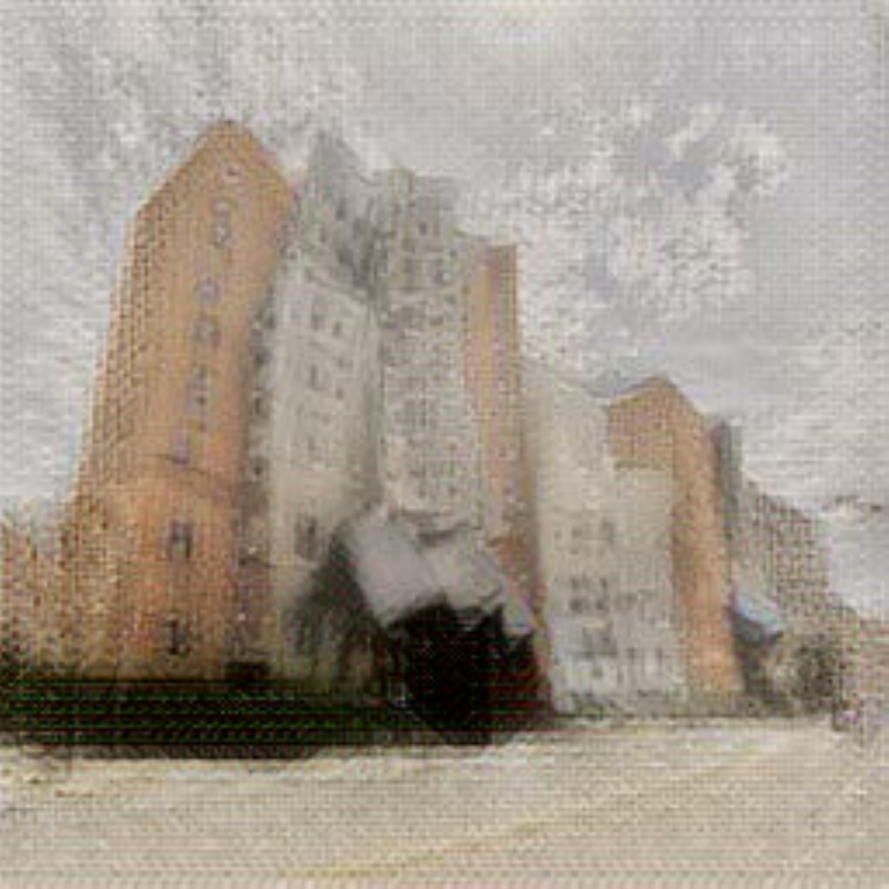}
\end{minipage}%
}%
\subfigure{
\begin{minipage}[t]{0.2\linewidth}
\centering
\includegraphics[width=\linewidth]{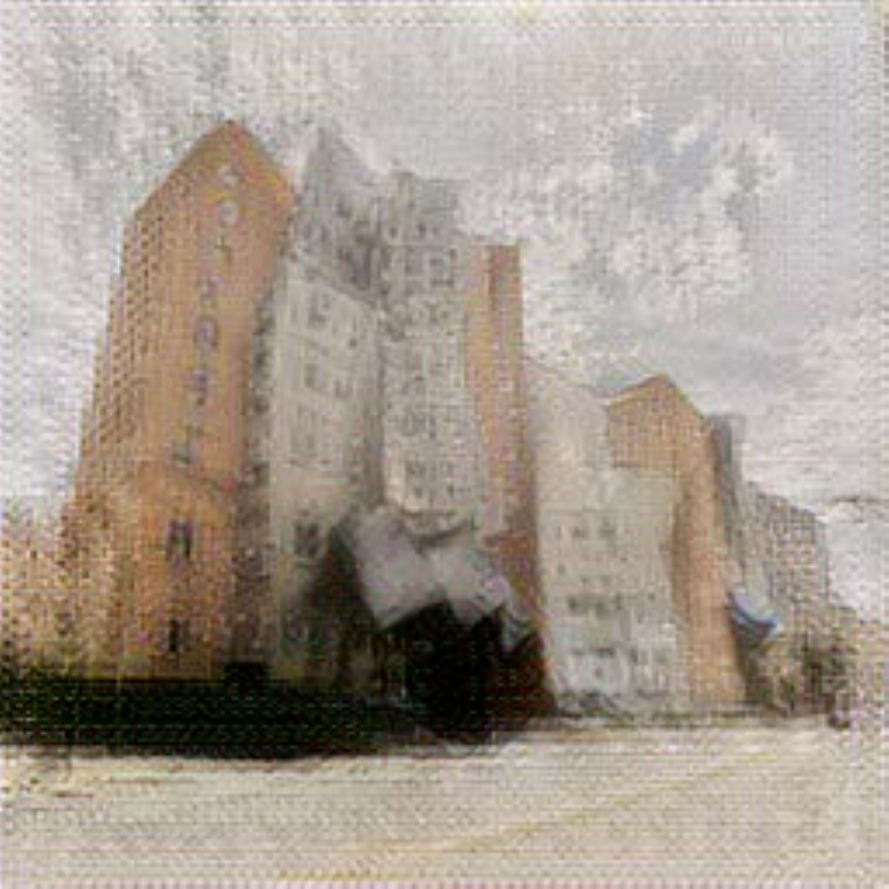}
\end{minipage}%
}%
\vfill

\subfigure{
\begin{minipage}[t]{0.2\linewidth}
\centering
\includegraphics[width=\linewidth]{new/the_scream.pdf}
\end{minipage}%
}%
\subfigure{
\begin{minipage}[t]{0.2\linewidth}
\centering
\includegraphics[width=\linewidth]{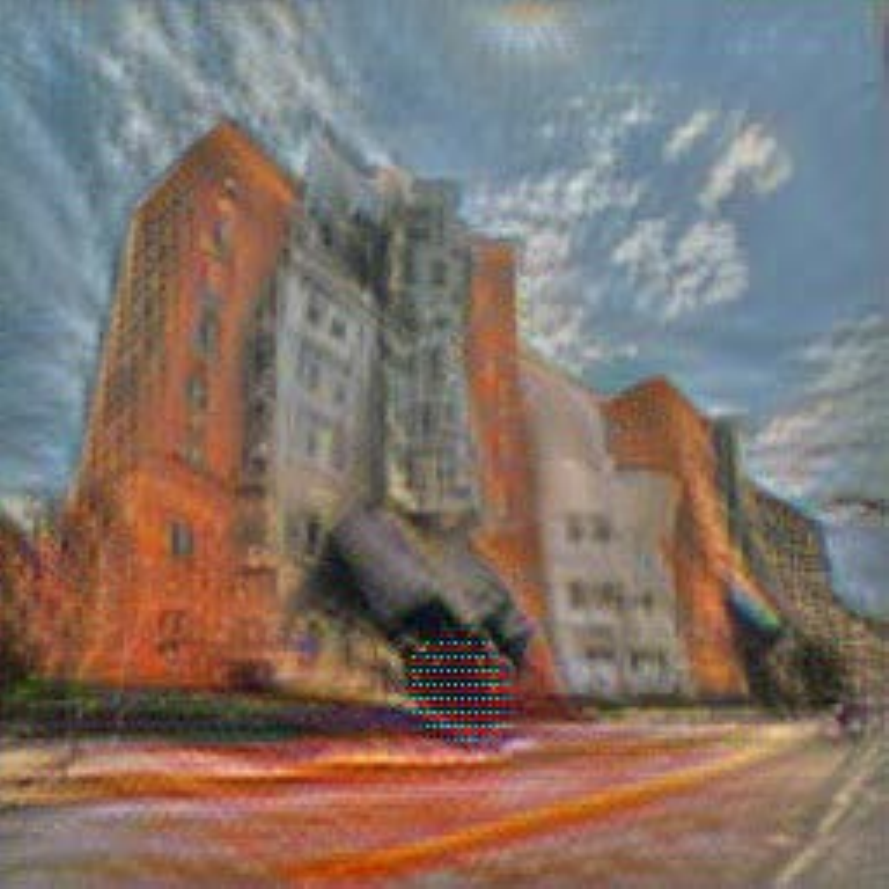}
\end{minipage}%
}%
\subfigure{
\begin{minipage}[t]{0.2\linewidth}
\centering
\includegraphics[width=\linewidth]{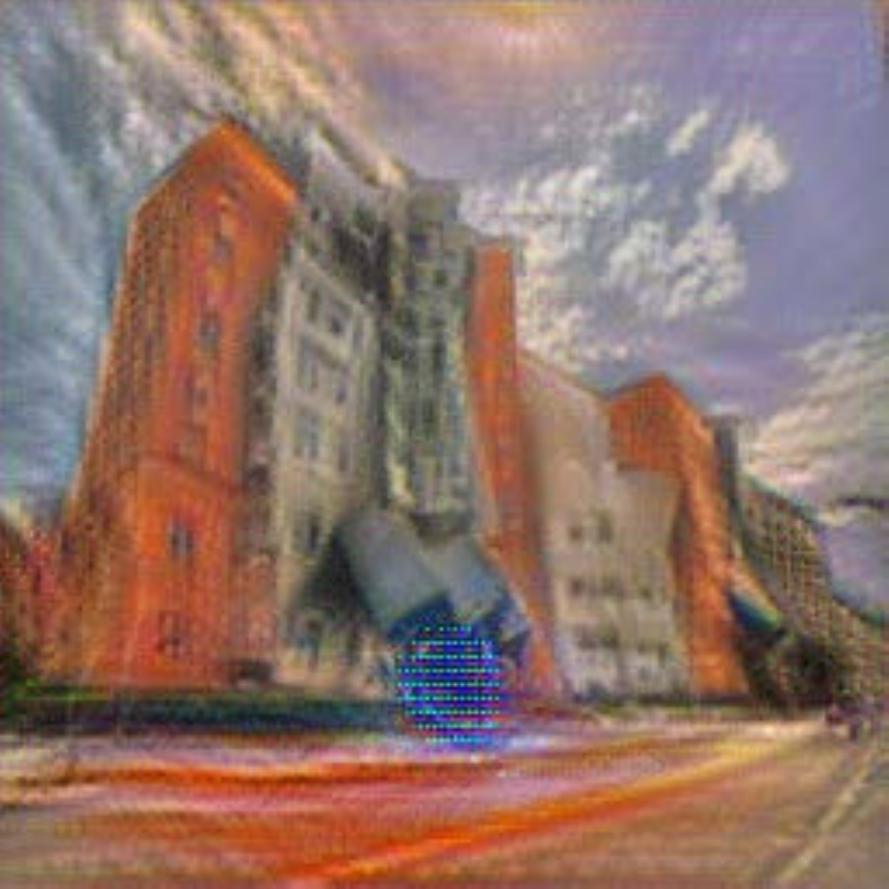}
\end{minipage}%
}%
\subfigure{
\begin{minipage}[t]{0.2\linewidth}
\centering
\includegraphics[width=\linewidth]{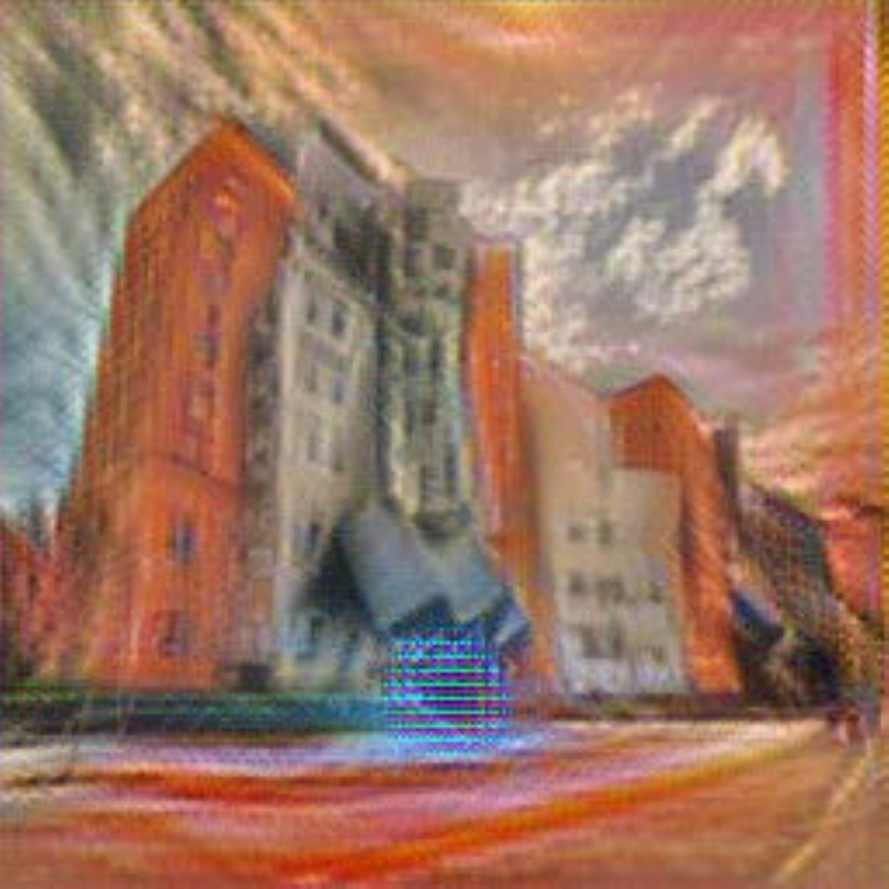}
\end{minipage}%
}%
\subfigure{
\begin{minipage}[t]{0.2\linewidth}
\centering
\includegraphics[width=\linewidth]{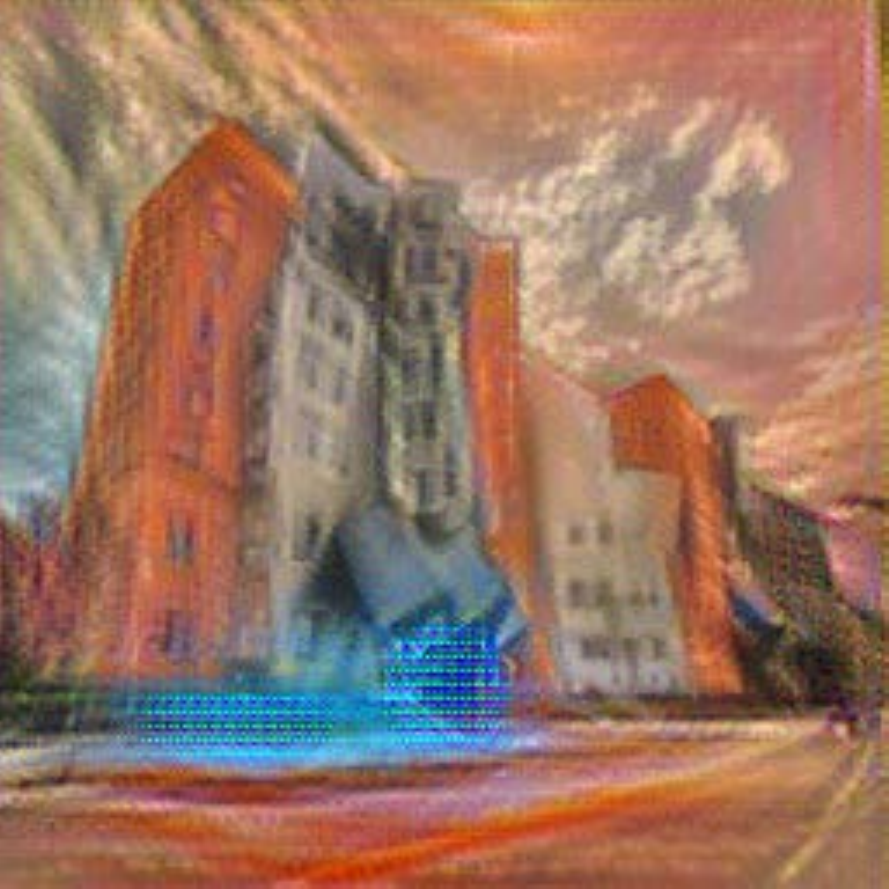}
\end{minipage}%
}%
\centering
\caption{The styled image with stroke basis intervened using ICA. The left most of each row shows the style images. From left to right of each row, the effect of stroke is increasingly amplified.}
\label{icaIntervention-2}
\end{figure*}

We demonstrate more styled images with stroke basis intervened using spectrum based method (Figure \ref{spectrumIntervention-1} and Figure \ref{spectrumIntervention-2}) and ICA (Figure \ref{icaIntervention-1} and Figure \ref{icaIntervention-2}) respectively.

\subsection*{D. Style Mixing}
We demonstrate more styled images transferred with compound style generated by mixing the color basis and stroke basis of two different styles. The results of both spectrum based method and ICA method are shown in Figure \ref{mixing} with comparison with traditional mixing method - interpolation.

\begin{figure*}[ht]
\centering
\subfigure{
\begin{minipage}[t]{0.2\linewidth}
\centering
\includegraphics[width=\linewidth]{new/la_muse.pdf}
\end{minipage}%
}%
\subfigure{
\begin{minipage}[t]{0.2\linewidth}
\centering
\includegraphics[width=\linewidth]{new/wave.pdf}
\end{minipage}%
}%
\subfigure{
\begin{minipage}[t]{0.2\linewidth}
\centering
\includegraphics[width=\linewidth]{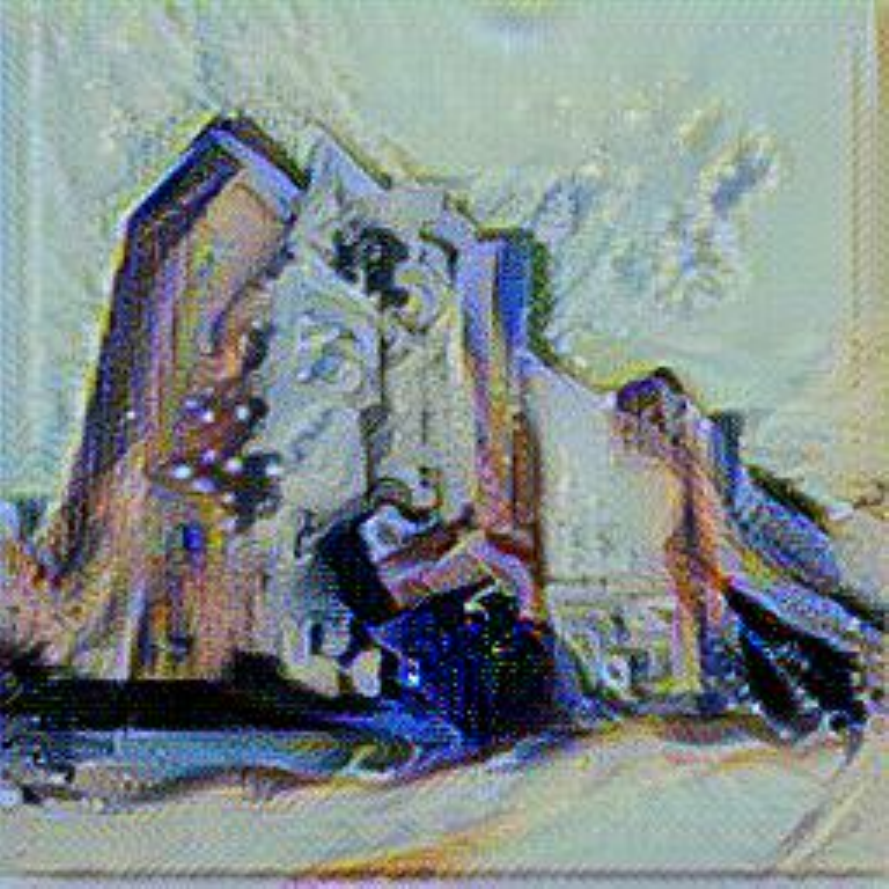}
\end{minipage}%
}%
\subfigure{
\begin{minipage}[t]{0.2\linewidth}
\centering
\includegraphics[width=\linewidth]{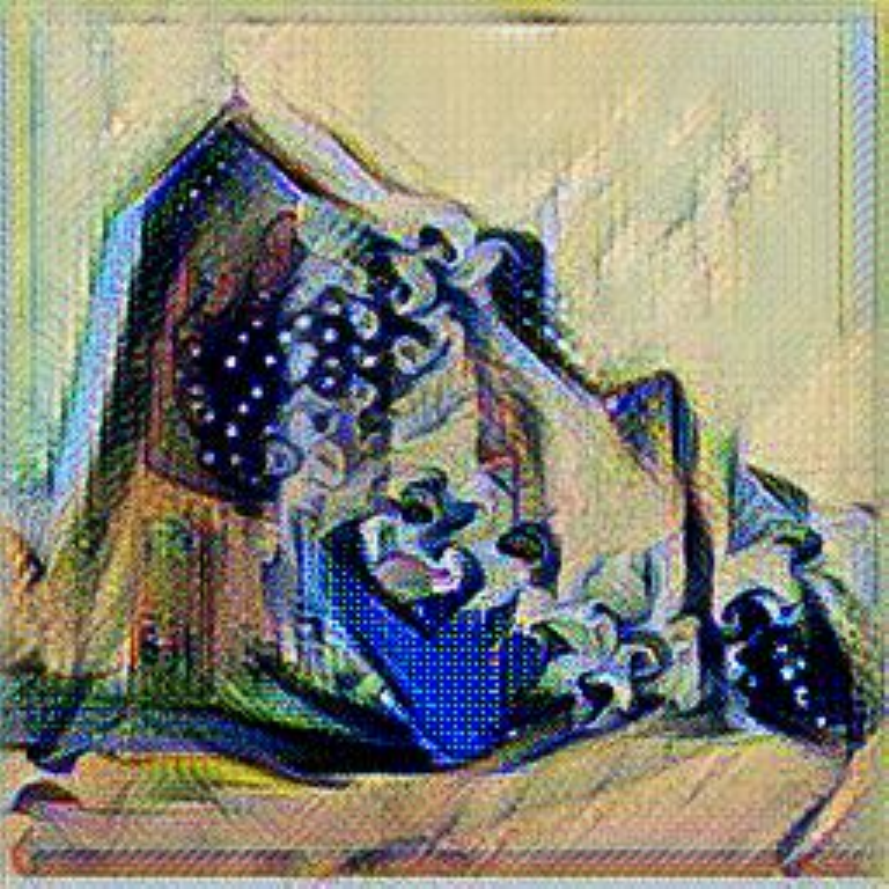}
\end{minipage}%
}%
\subfigure{
\begin{minipage}[t]{0.2\linewidth}
\centering
\includegraphics[width=\linewidth]{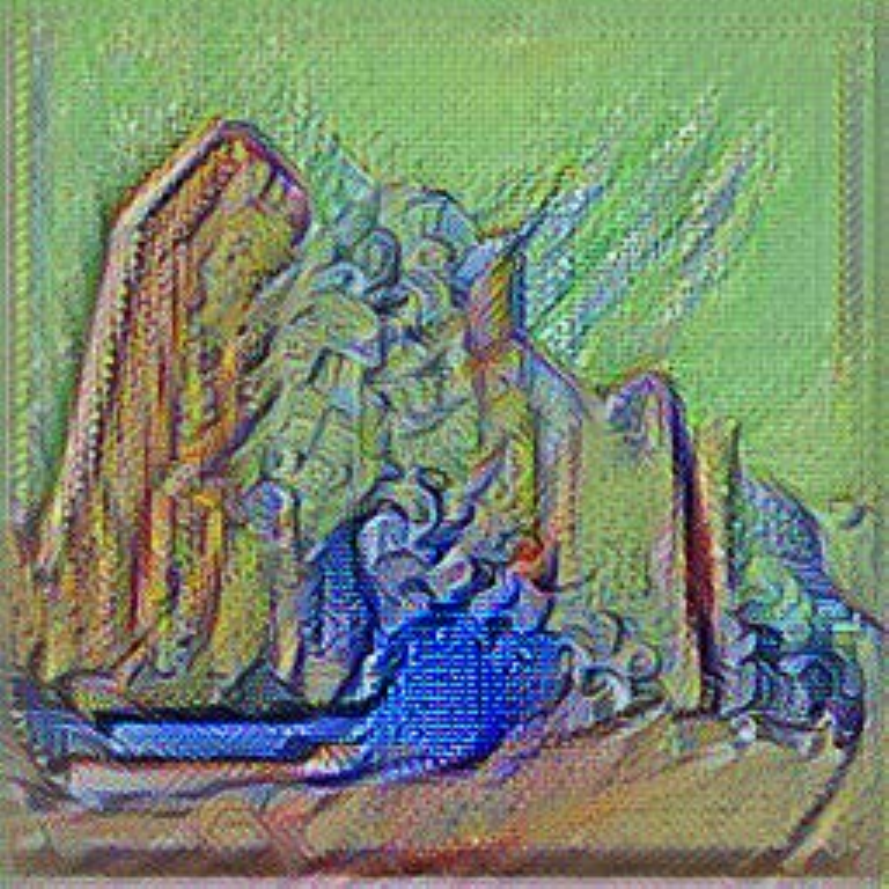}
\end{minipage}%
}%
\vfill

\subfigure{
\begin{minipage}[t]{0.2\linewidth}
\centering
\includegraphics[width=\linewidth]{new/the_scream.pdf}
\end{minipage}%
}%
\subfigure{
\begin{minipage}[t]{0.2\linewidth}
\centering
\includegraphics[width=\linewidth]{new/ship.pdf}
\end{minipage}%
}%
\subfigure{
\begin{minipage}[t]{0.2\linewidth}
\centering
\includegraphics[width=\linewidth]{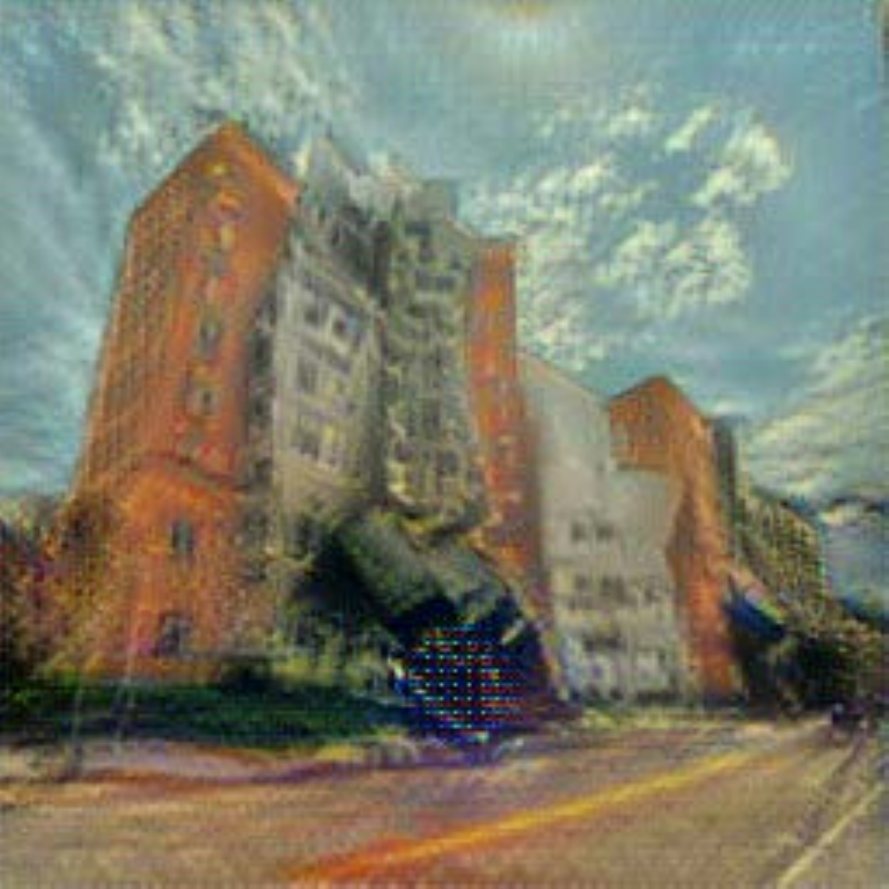}
\end{minipage}%
}%
\subfigure{
\begin{minipage}[t]{0.2\linewidth}
\centering
\includegraphics[width=\linewidth]{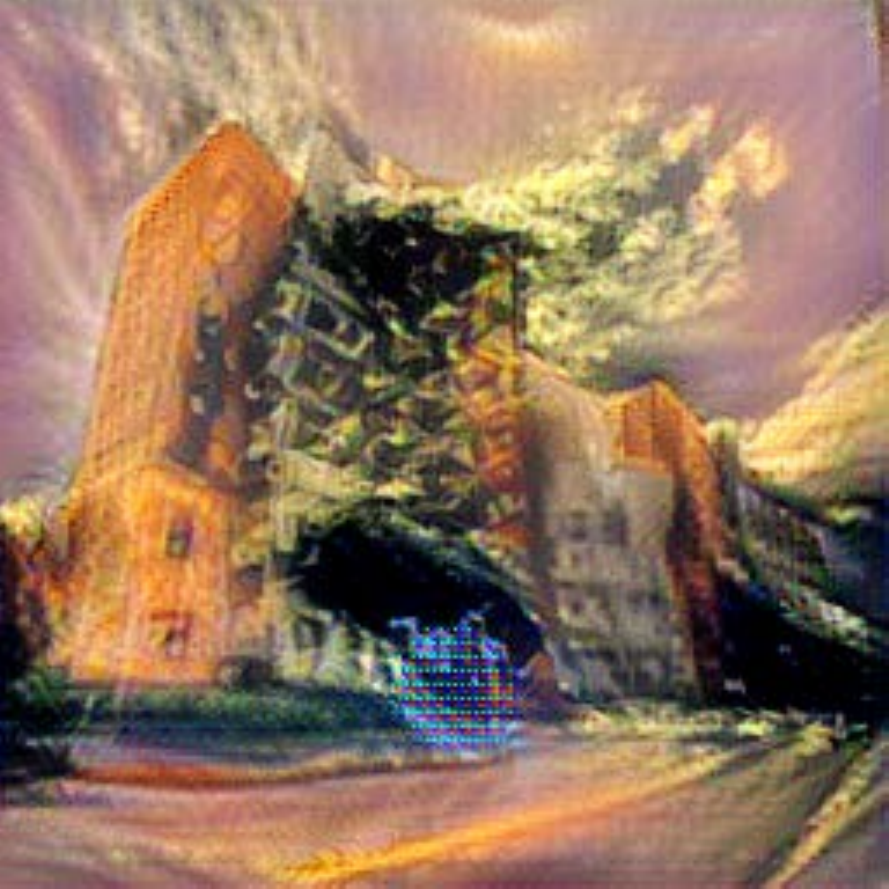}
\end{minipage}%
}%
\subfigure{
\begin{minipage}[t]{0.2\linewidth}
\centering
\includegraphics[width=\linewidth]{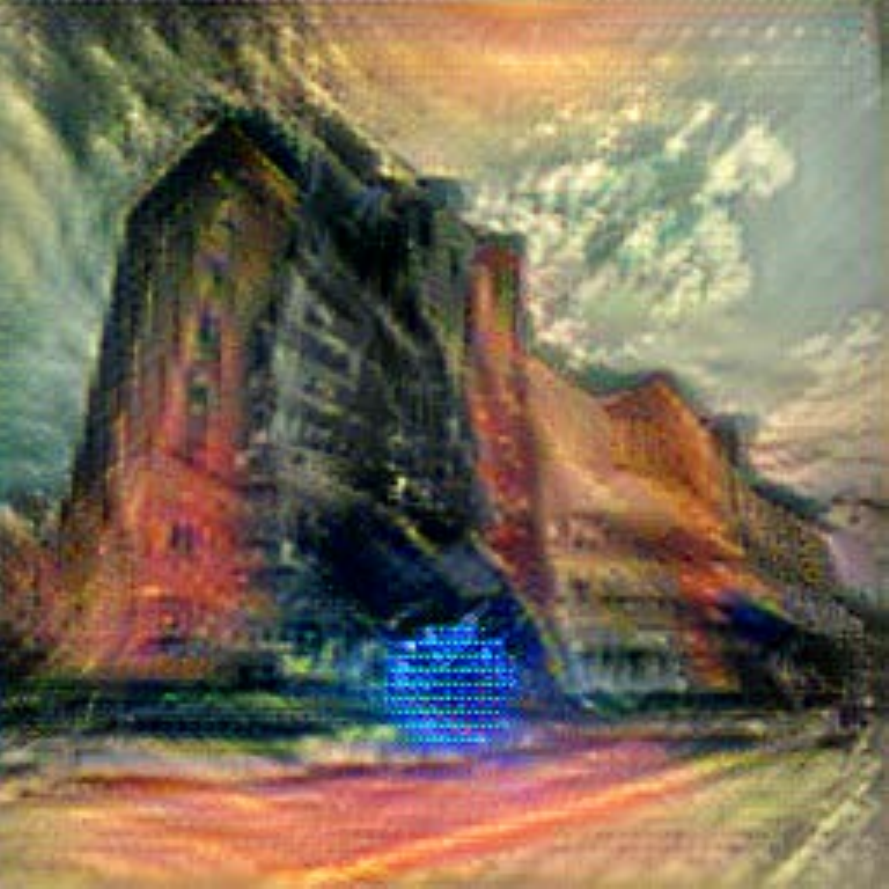}
\end{minipage}%
}%
\vfill

\subfigure{
\begin{minipage}[t]{0.2\linewidth}
\centering
\includegraphics[width=\linewidth]{new/draft.pdf}
\end{minipage}%
}%
\subfigure{
\begin{minipage}[t]{0.2\linewidth}
\centering
\includegraphics[width=\linewidth]{new/gray_red.pdf}
\end{minipage}%
}%
\subfigure{
\begin{minipage}[t]{0.2\linewidth}
\centering
\includegraphics[width=\linewidth]{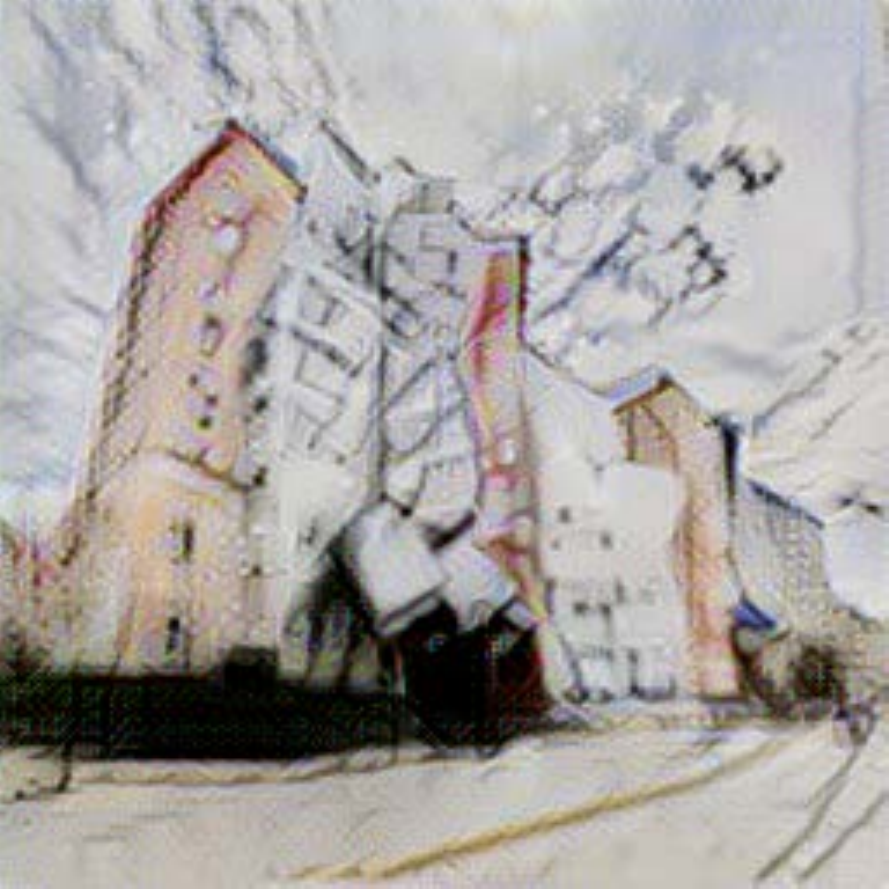}
\end{minipage}%
}%
\subfigure{
\begin{minipage}[t]{0.2\linewidth}
\centering
\includegraphics[width=\linewidth]{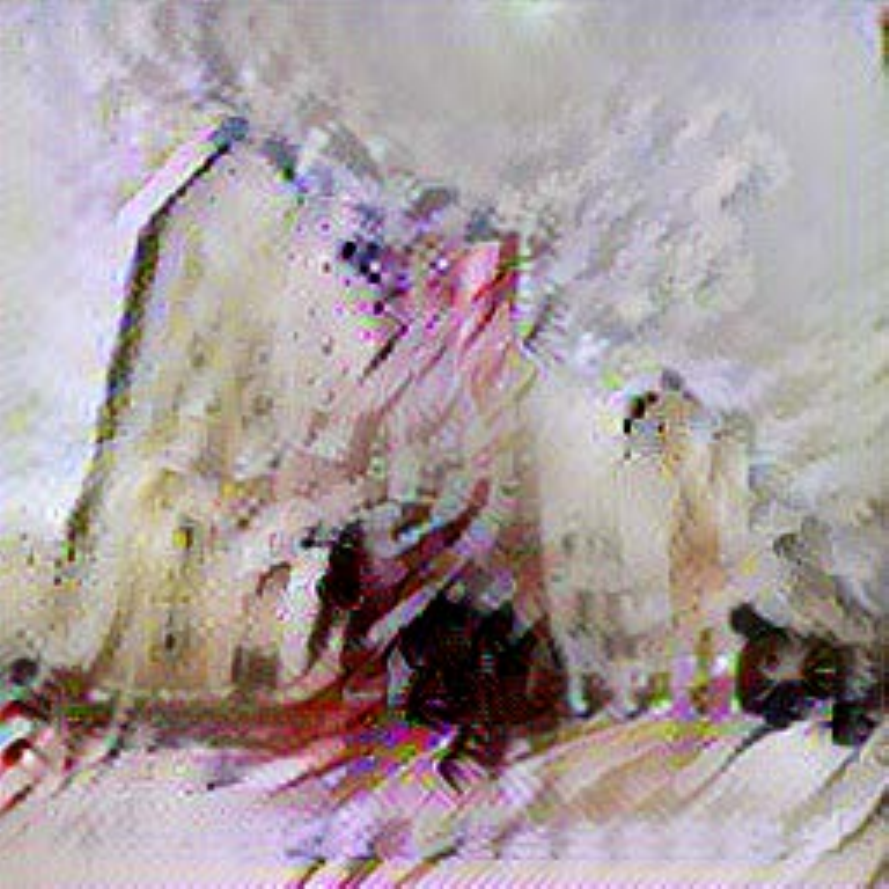}
\end{minipage}%
}%
\subfigure{
\begin{minipage}[t]{0.2\linewidth}
\centering
\includegraphics[width=\linewidth]{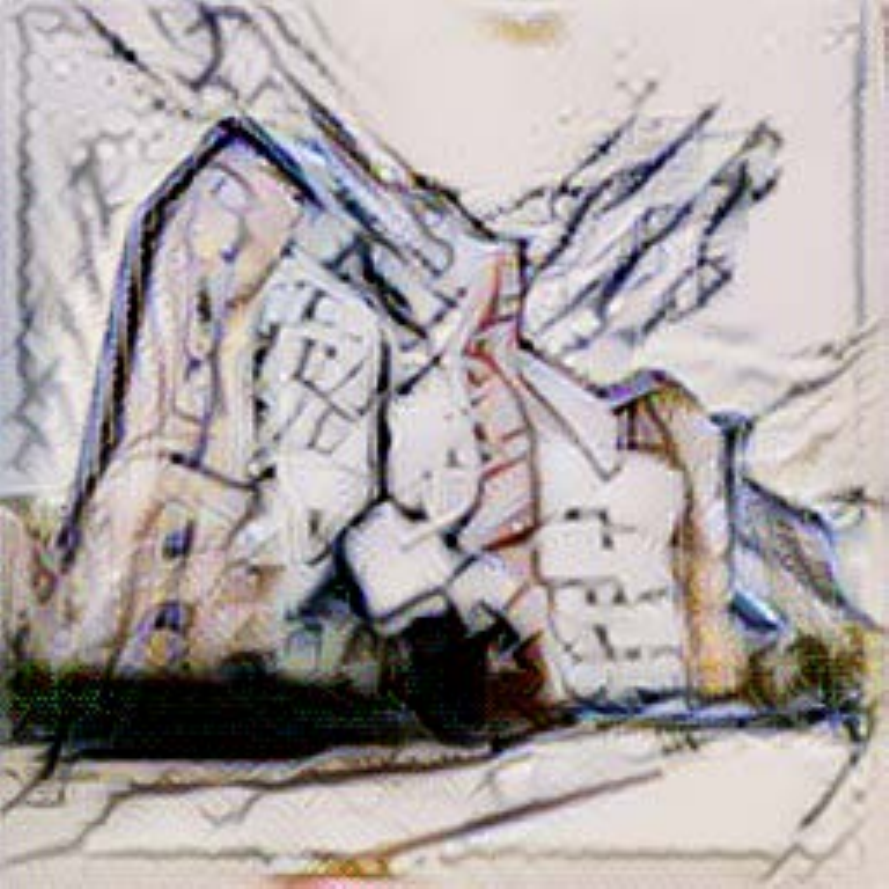}
\end{minipage}%
}%
\vfill

\subfigure{
\begin{minipage}[t]{0.2\linewidth}
\centering
\includegraphics[width=\linewidth]{new/static.pdf}
\end{minipage}%
}%
\subfigure{
\begin{minipage}[t]{0.2\linewidth}
\centering
\includegraphics[width=\linewidth]{new/face.pdf}
\end{minipage}%
}%
\subfigure{
\begin{minipage}[t]{0.2\linewidth}
\centering
\includegraphics[width=\linewidth]{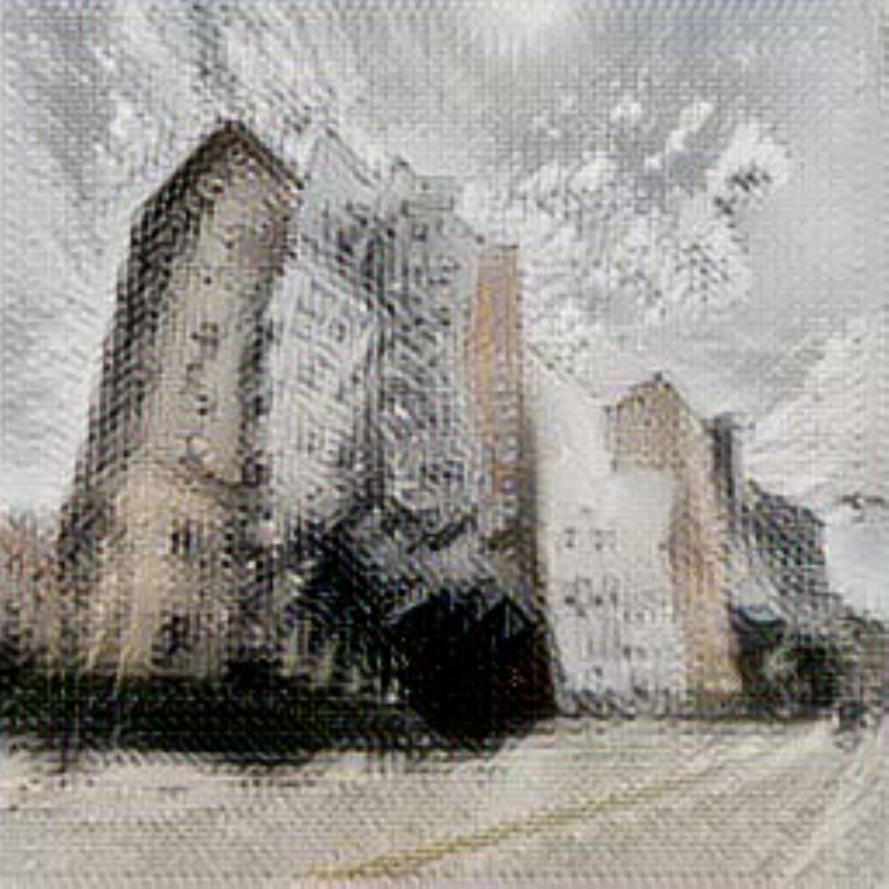}
\end{minipage}%
}%
\subfigure{
\begin{minipage}[t]{0.2\linewidth}
\centering
\includegraphics[width=\linewidth]{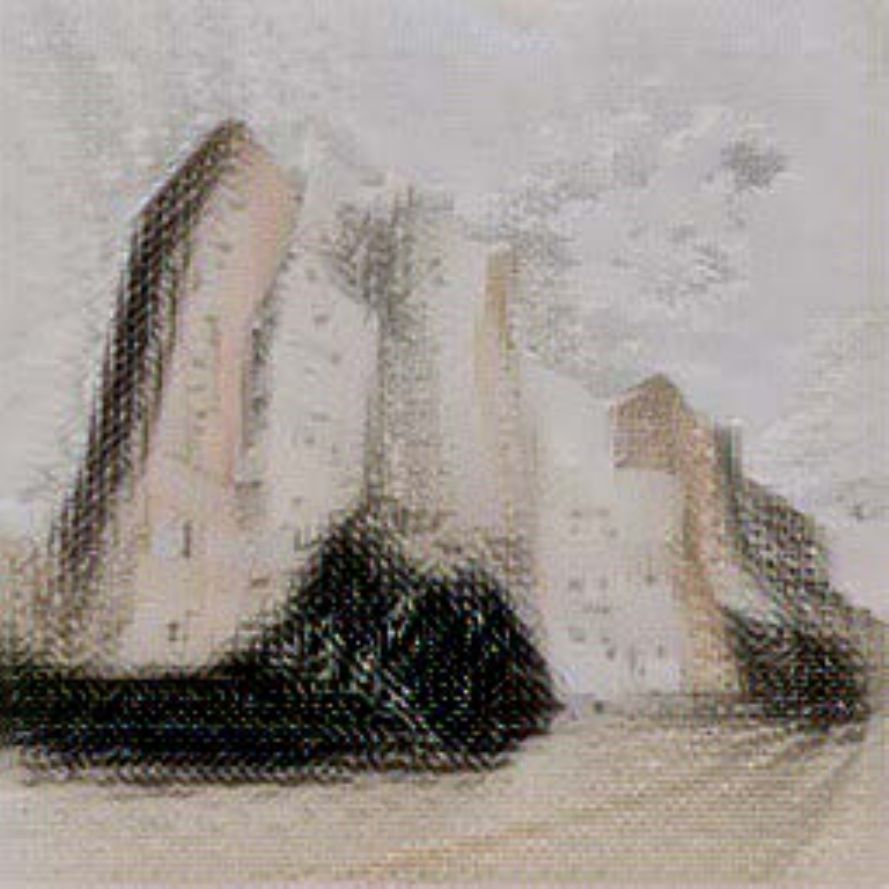}
\end{minipage}%
}%
\subfigure{
\begin{minipage}[t]{0.2\linewidth}
\centering
\includegraphics[width=\linewidth]{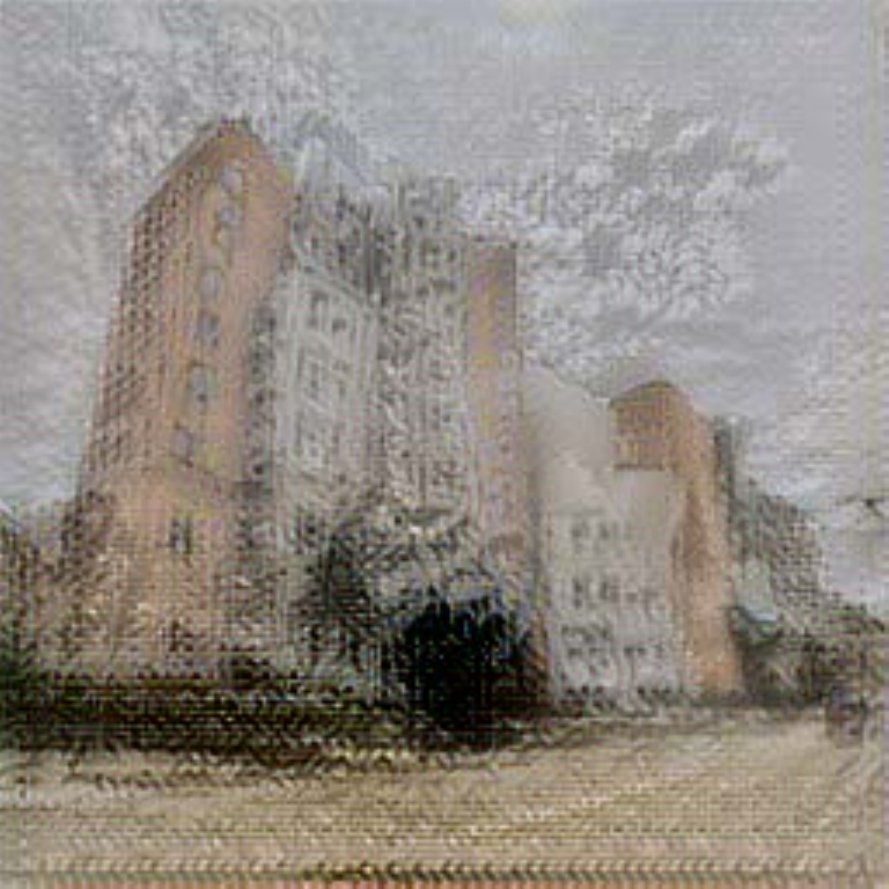}
\end{minipage}%
}%
\vfill

\subfigure{
\begin{minipage}[t]{0.2\linewidth}
\centering
\includegraphics[width=\linewidth]{new/dancer.pdf}
\end{minipage}%
}%
\subfigure{
\begin{minipage}[t]{0.2\linewidth}
\centering
\includegraphics[width=\linewidth]{new/composition.pdf}
\end{minipage}%
}%
\subfigure{
\begin{minipage}[t]{0.2\linewidth}
\centering
\includegraphics[width=\linewidth]{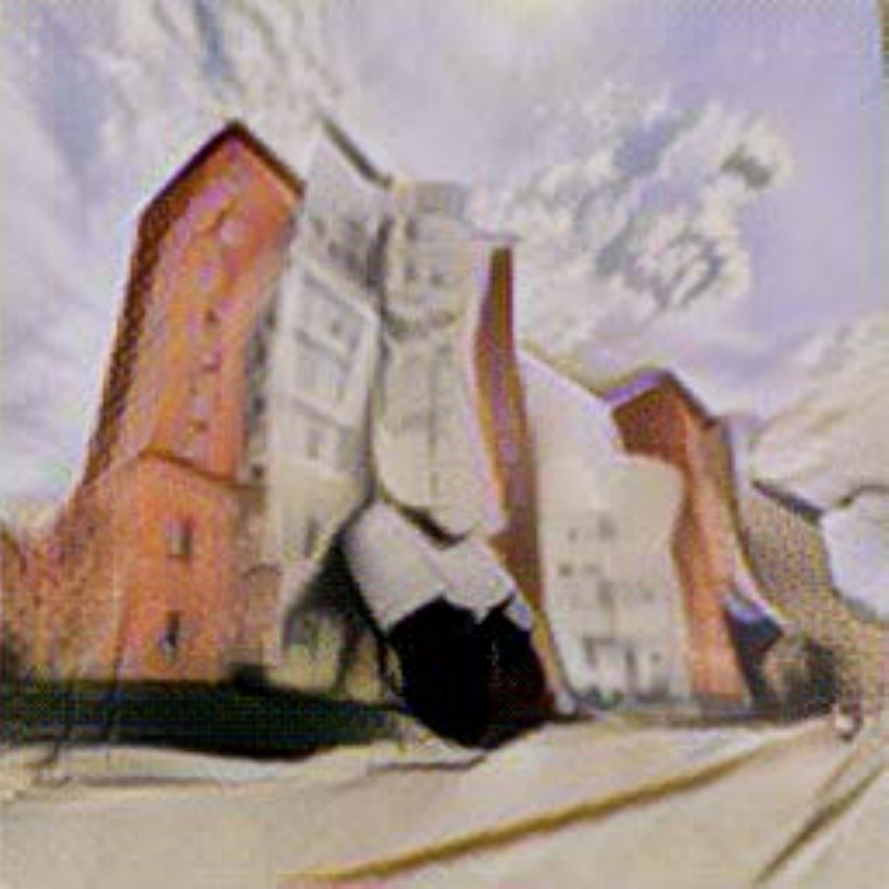}
\end{minipage}%
}%
\subfigure{
\begin{minipage}[t]{0.2\linewidth}
\centering
\includegraphics[width=\linewidth]{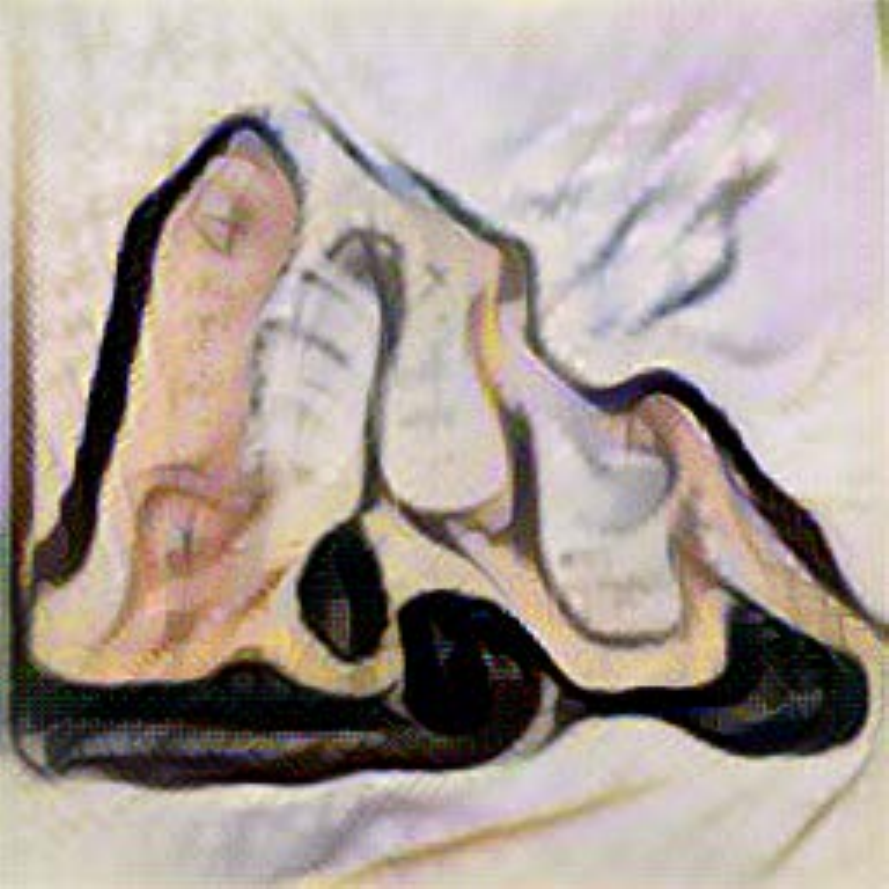}
\end{minipage}%
}%
\subfigure{
\begin{minipage}[t]{0.2\linewidth}
\centering
\includegraphics[width=\linewidth]{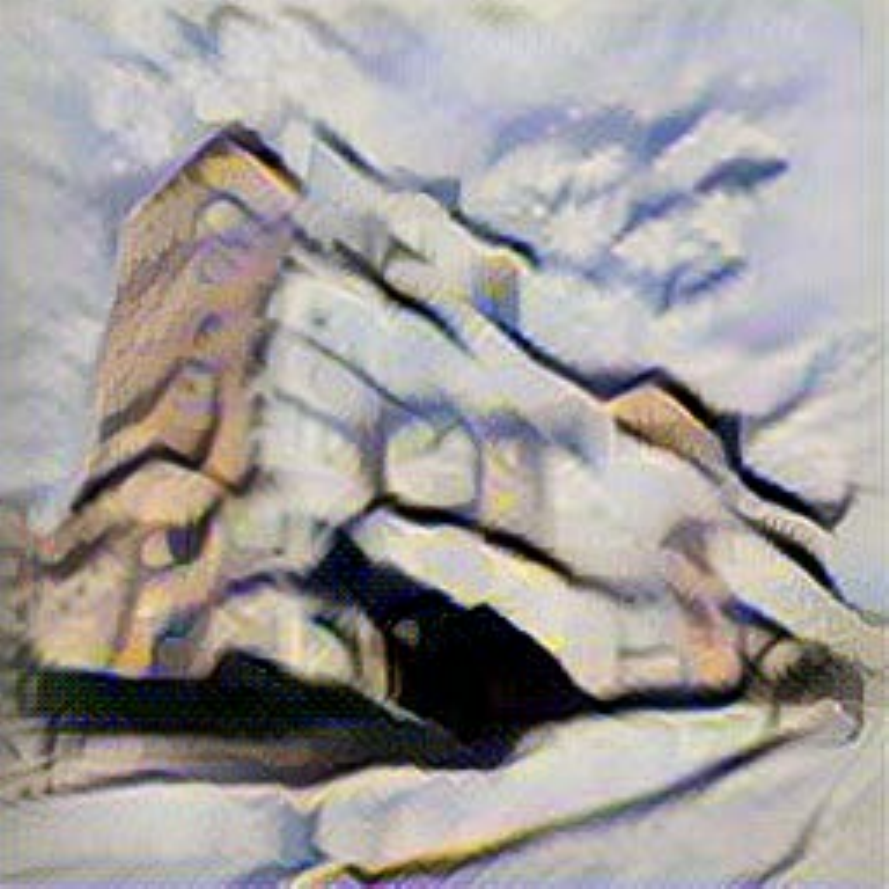}
\end{minipage}%
}%

\centering
\caption{The left two columns are the style images used for mixing. Specifically, we mix the color of the most left one with the stroke of the second left one. The third left column shows the styled images with traditional interpolation method. The second right column shows the styled images using spectrum mixing method. The most right column shows the styled images using ICA mixing method.}
\label{mixing}
\end{figure*}

\end{document}